\documentclass[a4paper,11pt,oneside,titlepage]{book}
\usepackage[T1]{fontenc}		% Codifica dei font; T1 codifica di output dell'italiano e di lingue occidentali
\usepackage[utf8]{inputenc}		% Codifica degli input; abilita l'utilizzo di caratteri accentati
\usepackage[english]{babel}		% Set the languages; the last one is the main language
\usepackage{geometry}			% Allow to change the margin
\usepackage{setspace}			% Allow to change the interline with \begin{onehalfspaceng} ecc.
\usepackage{fancyhdr}			% Fancy style for page layout
\usepackage{afterpage} 			% Allow to load blank pages with the command '\afterpage{\null\thispagestyle{empty}\clearpage}' 

\usepackage{url}
\usepackage{color}			% Allow to use colors
\usepackage{xcolor}			% manage colors
\usepackage{enumerate}			% Numbered lists
\usepackage{enumitem}			% Allow to manage the style of the list
\usepackage{graphicx}			% Images
\usepackage{epstopdf}			% allows to convert eps images to pdf for use with pdflatex
\usepackage{pstool}			% Allows to use psfrag with pdflatex (no latex compiler needed)
\usepackage{psfrag}
\usepackage{array}			% Tables
\usepackage{tabularx}			% allow to set the width of the whole table
\usepackage[bottom]{footmisc}		% To attach footnote at the end of the page
\usepackage{booktabs}			% is a must for professional-looking layout
\usepackage{longtable}			% is very popular for multi-page tables
\usepackage{caption}			% Captions
\usepackage{float}			% Manage float objects (image, table, ecc.)
\usepackage{rotating}			% permette di ruotare immagini, tabelle, ecc. di 90° o di 270°
\usepackage{rotfloat}			% costruisce un ponte tra i pacchetti float e rotating
\usepackage{amsmath} 			% Equations
\usepackage{amssymb}			% Mathematical Symbols
\usepackage{cancel}                     % Simplification Cancellation
\usepackage{mleftright}
\usepackage{listings}  
\usepackage[swapnames,norules,nouppercase]{frontespizio}
\usepackage[intoc]{nomencl}
\usepackage[acronym,toc,nomain,nopostdot,nonumberlist]{glossaries}
\newglossarystyle{mylong}{%
  \setglossarystyle{long}%
}
\setglossarystyle{mylong}
\usepackage[square, numbers, comma, sort&compress]{natbib}
\usepackage{verbatim}                   % Needed for the "comment" environment to make LaTeX comments
\usepackage{wrapfig}                 % Per inserire le figure contornate da testo
\usepackage{blindtext}
\fancypagestyle{plain}{%                                        % modifica dello stile predefinito plain
                        \fancyhf{}                              % cancella tutti i campi di  intestazione e pie di pagina
                        \fancyfoot[R]{\thepage}                 % mette al centro il numero di pagina
                        \renewcommand{\headrulewidth}{0pt}      % spessore della linea di separazione in alto (0 per eliminare la linea)
                        \renewcommand{\footrulewidth}{0.4pt}    % spessore della linea di separazione in basso (0 per eliminare la linea)
                        }                                       % modifica dello stile predefinito plain
\definecolor{sapienza}{RGB}{130,36,51} % example \definecolor{name}{model}{color-spec}
\definecolor{cust1}{RGB}{85,85,85}
\definecolor{cust2}{RGB}{212,212,212}
\makenomenclature % Generate the nomenclature
\makeglossaries % Generate the glossary
\newenvironment{dedication}
{
  \phantom{.}
  \vspace{13cm}
  \begin{quote} \begin{flushright}}
{\end{flushright} \end{quote}}
\usepackage{calligra}

\usepackage{glossaries}
\usepackage{mathtools}
\mathtoolsset{showonlyrefs}
\usepackage{tikz}
\usetikzlibrary{arrows.meta, positioning, calc}
\usepackage{pgfplots}
\pgfplotsset{compat=1.18}
\usepgfplotslibrary{colormaps}
\usepackage[linktocpage]{hyperref}			% Crea collegamenti ipertestuali rendendo cliccabili i riferimenti
\hypersetup{colorlinks=true, linkcolor=red, citecolor=red, filecolor=black, urlcolor=black, menucolor=black}
\usepackage{mathrsfs} 
\usepackage{adjustbox}
\usepackage{optidef}
\usepackage{algorithm}
\usepackage[noend]{algpseudocode}
\usepackage{subcaption}
\usepackage[absolute,overlay]{textpos}

\nomenclature{$\co$}{Coarse image}
\nomenclature{$\x$}{Original image}
\nomenclature{$\s$}{Semantic Segmentation Map}
\nomenclature{$\br$}{Residual}
\nomenclature{$\m$}{Mask}
\nomenclature{$\Loss$}{Loss function}
\newignoredglossary{ignored}
\newacronym{sc}{SemC}{Semantic Communication}
\newacronym{goc}{GOC}{Goal-Oriented Communication}
\newacronym{sgoc}{SemGOC}{Semantic-Goal-Oriented Communication}
\newacronym{bpp}{BPP}{Bits Per Pixel}
\newacronym{bppf}{BPPF}{Bits Per Pixel per Frame}
\newacronym{bpg}{BPG}{Better Portable Graphics}
\newacronym{flif}{FLIF}{Free Lossless Image Format}
\newacronym[type=ignored]{jpeg}{JPEG}{Joint Photographic Experts Group}
\newacronym[type=ignored]{jpeg2000}{JPEG2000}{Joint Photographic Experts Group 2000}

\newacronym[type=ignored]{l1}{$\ell_1$}{}
\newacronym[type=ignored]{l2}{$\ell_2$}{}
\newacronym{fid}{FID}{Frechet Inception Distance}
\newacronym{mse}{MSE}{Mean Squared Error}
\newacronym{mae}{MAE}{Mean Absolute Error}
\newacronym{nmse}{NMSE}{Normalized Mean Squared Error}
\newacronym{psnr}{PSNR}{Peak Signal-to-Noise Ratio}
\newacronym{miou}{mIoU}{mean Intersection over Union}
\newacronym{ssim}{SSIM}{Structural Similarity Index}
\newacronym{lpips}{LPIPS}{Learned Perceptual Image Patch Similarity}
\newacronym{msssim}{MS-SSIM}{Multi-Scale Structural Similarity}

\newacronym{ddpm}{DDPM}{Denoising Diffusion Probabilistic Model}
\newacronym{ddim}{DDIM}{Denoising Diffusion Implicit Model}
\newacronym{nn}{NN}{Neural Network}
\newacronym{ai}{AI}{Artificial Intelligence}
\newacronym{dnn}{DNN}{Deep Neural Network}
\newacronym{cnn}{CNN}{Convolutional Neural Network}
\newacronym{ml}{ML}{Machine Learning}
\newacronym[type=ignored]{elbo}{ELBO}{Evidence Lower Bound}
\newacronym{gan}{GAN}{Generative Adversarial Networks}
\newacronym{vqgan}{VQ-GAN}{Vector Quantized GAN}
\newacronym{sota}{SOTA}{state-of-the-art}
\newacronym{srr}{SRR}{Semantic Relevant Residual}
\newacronym{ssm}{SSM}{Semantic Segmentation Map}
\newacronym{ssmodel}{SS-Model}{Semantic Segmentation Model}
\newacronym{sseg}{SSeg}{Semantic Segmentation}
\newacronym{vae}{VAE}{Variational Autoencoder}
\newacronym{vqvae}{VQ-VAE}{Vector Quantized VAE}
\newacronym{maskvqvae}{MQ-VAE}{Masked VQ-VAE}
\newacronym{ae}{AE}{Autoencoder}
\newacronym[type=ignored]{unet}{U-Net}{}
\newacronym[type=ignored]{resblock}{ResBlock}{Residual Block}
\newacronym[type=ignored]{resblockdown}{ResBlock-Down}{}
\newacronym[type=ignored]{resblockup}{ResBlock-Up}{}
\newacronym{fc}{FC}{Fully Connected}
\newacronym{mlp}{MLP}{Multilayer Pereptron}
\newacronym{cfg}{CFG}{Classifier-Free Guidance}
\newacronym{pe}{PE}{Positional Embedding}

\newacronym{awgn}{AWGN}{additive white Gaussian noise}
\newacronym{sr}{SR}{Super-Resolution}
\newacronym{lr}{LR}{Low-Resolution}
\newacronym{hr}{HR}{High-Resolution}
\newacronym{roi}{ROI}{Region-Of-Interest}
\newacronym{pca}{PCA}{Principal Component Analysis}
\newacronym{gmm}{GMM}{Gaussian Mixture Model}
\newacronym{rbm}{RBM}{Restricted Boltzmann Machine}

\newacronym[type=ignored]{dsslic}{DSSLIC}{}%{Deep Semantic Segmentation based Learned Image Compression}

\newacronym{spic}{SPIC}{Semantic-Preserving Image Coding}
\newacronym{cspic}{C-SPIC}{Class specific SPIC}
\newacronym{semcore}{SemCoRe}{Semantic-Conditioned Super-Resolution Diffusion Model}
\newacronym{spade}{SPADE}{Spatially-Adaptive Normalization}
\newacronym{spe}{SemPE}{Semantic Positional Embedding}
\newacronym{amm}{AMM}{Adaptive Mask Module}
\newacronym{samm}{SAMM}{Semantic conditioned Adaptive Mask Module}
\newacronym{adm}{ADM}{Adaptive De-Mask Module}
\newacronym{pi}{PI}{Positional Index}
\newacronym{sqgan}{SQ-GAN}{Semantic masked VQ-GAN}

\newacronym{dpi}{DPI}{Data Processing Inequality} 
\newacronym{it}{IT}{Information Theory}
\newacronym{iot}{IoT}{Internet of Things}
\newacronym{ib}{IB}{Information Bottleneck}
\newacronym{gib}{GIB}{Gaussian Information Bottleneck}

\newacronym{en}{EN}{Edge Network}
\newacronym{ed}{ED}{Edge Device}
\newacronym{es}{ES}{Edge Server}
\newacronym{flops}{FLOPS}{FLoating point Operations Per Second}
\newacronym{flopc}{FLOPC}{Floating Point Operations Per Cycle}

\newacronym{kb}{KB}{Knowledge Base}

\newcommand{\x}{\mathbf{x}}
\newcommand{\s}{\mathbf{s}}
\newcommand{\co}{\mathbf{c}}
\newcommand{\br}{\mathbf{r}}
\newcommand{\m}{\mathbf{m}}
\newcommand{\Loss}{\mathcal{L}}
\newcommand{\z}{\mathbf{z}}
\newcommand{\n}{\mathbf{n}}
\newcommand{\y}{\mathbf{y}}
\newcommand{\h}{\mathbf{h}}
\newcommand{\w}{\mathbf{w}}
\newcommand{\ob}{\mathbf{o}}
\newcommand{\bepsilon}{\boldsymbol{\epsilon}}
\newcommand{\N}{\mathcal{N}}
\newcommand{\bmu}{\boldsymbol{\mu}}
\newcommand{\bSigma}{\boldsymbol{\Sigma}}
\newcommand{\I}{\mathbb{\mathbf{I}}}
\newcommand{\e}{\mathbf{e}}
\newcommand{\C}{\mathcal{C}}

\newcommand{\bB}{\mathbf{B}}
\newcommand{\bM}{\mathbf{M}}

\newcommand{\eref}[1]{Eq.~\eqref{#1}}
\newcommand{\tref}[1]{Table~\ref{#1}}
\newcommand{\fref}[1]{Fig.~\ref{#1}}
\newcommand{\sref}[1]{Section~\ref{#1}}
\newcommand{\cref}[1]{Chapter~\ref{#1}}
\newcommand{\aref}[1]{App.~\ref{#1}}

\newcommand{\circv}{\tikz[baseline=(char.base)]{
    \node[shape=circle,draw,inner sep=1pt] (char) {v};}}

\begin{document}
%   FRONT MATTER                                            %
\frontmatter 	   % Begin Roman style (i, ii, iii, iv...) page numbering
\pagestyle{empty}  % No headers or footers for the following pages
%   TITLE PAGE
% Titlepage.tex
\newcommand{\fronttitlefont}{\fontsize{24pt}{24pt}\selectfont\bfseries}
\newcommand{\frontsubtitlefont}{\fontsize{17pt}{21pt}\selectfont\bfseries}
\begin{titlepage}
  \begin{center}
    % Create a table to align logos and university names
    \begin{tabular}{@{\hspace{-0.5cm}}c@{\hspace{1.5cm}}c@{}}
      % Left University
      \begin{minipage}[t][\totalheight][t]{0.48\textwidth}
        \centering
        \includegraphics[height=2.8cm]{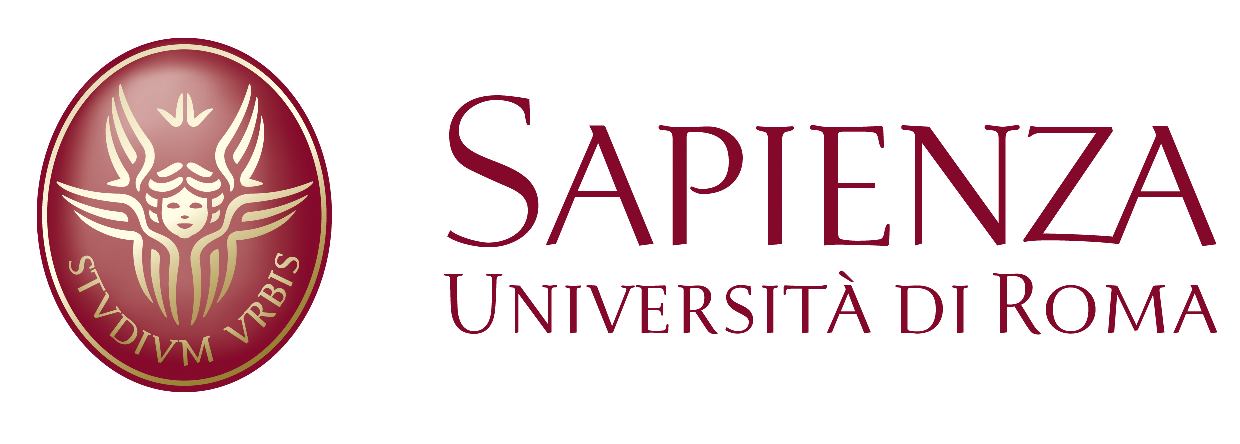}\\[0.5cm]
        {\LARGE \textbf{Sapienza University of Rome}}\\[0.45cm]
        {\Large \textit{Department of Computer, Control, and Management Engineering}}\\[0.5cm]
        {\Large PhD in Data Science}
      \end{minipage}
      &
      % Right University
      \begin{minipage}[t][\totalheight][t]{0.48\textwidth}
        \centering
        \includegraphics[height=2.8cm,trim=0 15 0 15, clip]{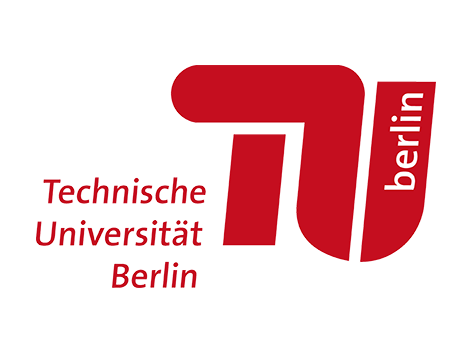}\\[0.5cm]
        {\LARGE \textbf{Technische Universität Berlin}}\\[0.45cm]
        {\Large \textit{Faculty 4 - Electrical Engineering and Computer Science}}\\[0.5cm]
        {\Large PhD in Engineering}
      \end{minipage}
    \end{tabular}

    \vfill

    % Thesis title
    {\fronttitlefont
      \begin{center}
        {\Huge Semantic Communication based on\\[0.3cm]
        Generative AI:}
        {\huge A New Approach \\[0.3cm] to Image Compression and Edge Optimization}
      \end{center}
    }

    \vfill

    % Advisors and candidate with increased font size
    {\Large
    \begin{minipage}[t]{0.45\textwidth}
      \raggedright
      \textbf{Thesis Advisors:} \\
      Prof. Sergio Barbarossa \\
      Prof. Giuseppe Caire \\
    \end{minipage}
    \hfill
    \begin{minipage}[t]{0.45\textwidth}
      \raggedleft
      \textbf{Candidate:} \\
      Francesco Pezone \\
    \end{minipage}
    }

    \vfill

    % Footer
    {\large 31st of October 2024}

  \end{center}
\end{titlepage}
\clearpage
\thispagestyle{empty}

\vspace*{1cm}

\noindent Thesis defended on 22 January 2025 \\
\noindent in front of a Board of Examiners composed by:
\vspace{-0.2cm}
\begin{tabbing}
    \hspace{0.5cm} \= Prof. Paolo Di Lorenzo (Chairman) \\
    \> Prof. Sergio Barbarossa \\
    \> Prof. Giuseppe Caire \\
    \> Prof. Marios Kountouris \\
    \> Prof. Mehdi Bennis \\
\end{tabbing}

\vspace{13cm}

% \noindent \rule{8cm}{0.2pt} % Horizontal line 5 cm wide and 0.4pt thick
\hrule

\vspace{0.2cm}

\noindent \textbf{Semantic Communication based on Generative AI: A New Approach to Image Compression and Edge Optimization}\\
\noindent \textit{Ph.D. thesis}, Cotutelle \textit{Sapienza University of Rome} and \textit{Technische Universität Berlin}

\vspace{0.5cm}

\noindent \copyright \;2025 Francesco Pezone. All rights reserved.

\vspace{0.5cm}

\noindent This thesis has been typeset by \LaTeX\ and the Sapthesis class.

\vspace{0.5cm}

\noindent Author's email: \href{mailto:francesco.pezone.ds@gmail.com}{\textcolor{blue}{francesco.pezone.ds@gmail.com}}

\clearpage
%   QUOTE (OPTIONAL)
\begin{dedication}

% {\Huge
%     % {\fontfamily{calligra}\selectfont
%     % {\Huge
% \textit{\textbf{D}reams\\\textbf{A}lways\\\textbf{J}ustify\\\textbf{E}fforts}

% }
%%%%%%%%%%%%%%%%%%%
\vspace*{-10cm} % Adjust this to control vertical position

\begin{flushright} % This keeps the text on the right side
    \begin{tabular}{l} % Left-aligned tabular column
        {\Huge \textit{\textbf{D}reams}} \\
        {\Huge \textit{\textbf{A}lways}} \\
        {\Huge \textit{\textbf{J}ustify}} \\
        {\Huge \textit{\textbf{E}fforts}}
    \end{tabular}
\end{flushright}
%%%%%%%%%%%%%%%%%%%
% \vspace*{2cm} % Adjust this to control vertical position

% \begin{flushright} % This keeps the text on the right side
%     \begin{tabular}{l} % Left-aligned tabular column
%         {\Huge \textit{Dreams}} \\
%         {\Huge \textit{Always}} \\
%         {\Huge \textit{Justify}} \\
%         {\Huge \textit{Efforts}}
%     \end{tabular}
% \end{flushright}
\end{dedication}

\newpage
%   ABSTRACT
\afterpage{\null\thispagestyle{empty}\clearpage} % Blank page
\thispagestyle{plain}			% Supress header 
\setlength{\parskip}{0pt plus 1.0pt}
\section*{Abstract}
As digital technologies continue to advance, modern communication networks face unprecedented challenges in handling the vast amounts of data produced daily by connected intelligent devices. Autonomous vehicles, smart sensors, IoT systems etc., are gaining more and more interest and new communication paradigms are needed. This thesis addresses these challenges by combining semantic communication with generative models to optimize image compression and resource allocation in edge networks. Unlike traditional bit-centric communication systems, semantic communication prioritizes the transmission of meaningful data specifically selected to convey the meaning rather than obtain a faithful representation of the original data. The communication infrastructure can benefit of the focus solely on the relevant parts of the data due to significant improvements in  bandwidth efficiency and latency reduction.

Central to this work is the design of semantic-preserving image compression algorithms, utilizing advanced generative models such as Generative Adversarial Networks and Denoising Diffusion Probabilistic Models. These algorithms compress images by encoding only semantically relevant features and exploiting the generative power at the receiver side. This allows for the accurate reconstruction of high-quality images with minimal data transmission. The thesis also introduces a Goal-Oriented edge network optimization framework based on the Information Bottleneck problem and stochastic optimization, ensuring that communication resources are dynamically allocated to maximize efficiency and task performance.

By integrating semantic communication into edge networks, the proposed system achieves a balance between computational efficiency and communication effectiveness, making it particularly suited for real-time applications. The thesis compares the performance of these semantic communication models with conventional image compression techniques, using both classical and semantic-aware evaluation metrics. The results demonstrate the potential of combining generative AI and semantic communication to create more efficient semantic-goal-oriented communication networks that meet the demands of modern data-driven applications.

\vfill
Keywords: Semantic Communication, Generative AI, Goal-Oriented Communication, Edge Network Optimization, Image Compression.

\thispagestyle{empty}
\mbox{}
\newpage
%   ZUSAMMENFASSUNG
\afterpage{\null\thispagestyle{empty}\clearpage} % Blank page
\thispagestyle{plain}			% Supress header 
\setlength{\parskip}{0pt plus 1.0pt}
\section*{Zusammenfassung}
Mit der ständigen Weiterentwicklung digitaler Technologien stehen moderne Kommunikationsnetzwerke vor der beispiellosen Herausforderung, die riesigen Datenmengen zu bewältigen, die täglich von vernetzten intelligenten Geräten erzeugt werden. Autonome Fahrzeuge, intelligente Sensoren, IoT-Systeme usw. gewinnen zunehmend an Interesse und es werden neue Kommunikationsparadigmen benötigt. Diese Arbeit befasst sich mit diesen Problemen, indem ein neues Kommunikationsparadigma eingesetzt wird, das semantische Kommunikation mit generativen Modellen kombiniert, um die Bildkompression und Ressourcenallokation in Edge-Netzwerken zu optimieren. Im Gegensatz zu traditionellen bitzentrierten Kommunikationssystemen priorisiert die semantische Kommunikation die Übertragung von bedeutungsvollen Daten, die speziell ausgewählt werden, um den Inhalt zu vermitteln, anstatt eine getreue Darstellung der Originaldaten zu erreichen. Die Kommunikationsinfrastruktur profitiert von dem Fokus auf relevante Datenanteile durch signifikante Verbesserungen in der Bandbreiteneffizienz und Reduzierung der Latenz.

Zentraler Bestandteil dieser Arbeit ist die Entwicklung semantik-erhaltender Bildkompressionsalgorithmen unter Verwendung fortschrittlicher generativer Modelle wie Generative Adversarial Networks und Denoising Diffusion Probabilistic Models. Diese Algorithmen komprimieren Bilder, indem sie nur semantisch relevante Merkmale kodieren und die generative Leistung auf der Empfängerseite nutzen. Dies ermöglicht die genaue Rekonstruktion von qualitativ hochwertigen Bildern bei minimaler Datenübertragung. Die Arbeit stellt außerdem ein zielorientiertes Optimierungsframework für Edge-Netzwerke vor, das auf dem Information-Bottleneck-Problem und stochastischer Optimierung basiert und sicherstellt, dass Kommunikationsressourcen dynamisch zugewiesen werden, um Effizienz und Aufgabenleistung zu maximieren.

Durch die Integration semantischer Kommunikation in Edge-Netzwerke erreicht das vorgeschlagene System ein Gleichgewicht zwischen rechnerischer Effizienz und Kommunikationseffektivität, was es besonders für Echtzeitanwendungen geeignet macht. Die Arbeit vergleicht die Leistung dieser semantischen Kommunikationsmodelle mit herkömmlichen Bildkompressionstechniken, wobei sowohl klassische als auch semantikbewusste Bewertungsmetriken verwendet werden. Die Ergebnisse zeigen das Potenzial der Kombination von generativer KI und semantischer Kommunikation, um effizientere, semantisch zielorientierte Kommunikationsnetzwerke zu schaffen, die den Anforderungen moderner, datengetriebener Anwendungen gerecht werden.

\thispagestyle{empty}
\mbox{}
\newpage
\begin{singlespace}
 \tableofcontents
 \addcontentsline{toc}{chapter}{\listfigurename}
 \listoffigures
 \addcontentsline{toc}{chapter}{\listtablename}
 \listoftables
% \renewcommand{\lstlistlistingname}{List of Code}
% \addcontentsline{toc}{chapter}{\lstlistlistingname}
% \lstlistoflistings

%   NOMENCLATURE AND ACRONYMS
%  \printnomenclature
% Define a new glossary for displayed acronyms
 \printglossaries
\end{singlespace}

%   MAIN MATTER
\mainmatter	  % Begin normal, numeric (1,2,3...) page numbering
\clearpage
% ------ set page style fancy with the follow
\pagestyle{fancy}
\renewcommand{\chaptermark}[1]{\markright{\chaptername\ \thechapter.\ #1}{}}
\renewcommand{\sectionmark}[1]{\markright{\thesection.\ #1}}
\lhead{}
\chead{}
\rhead{\slshape \rightmark}
\lfoot{Francesco Pezone}
\cfoot{}
\rfoot{\thepage}
\renewcommand{\headrulewidth}{0.4pt}
\renewcommand{\footrulewidth}{0.4pt}

% %   CHAPTERS
\chapter{\textcolor{black}{Introduction}}
\thispagestyle{plain}
The rapid advancement of digital technology has fundamentally transformed the way information is generated, transmitted, and consumed. From high-resolution multimedia content to the proliferation of the \gls{iot}, autonomous vehicles, and smart cities, the modern world is producing data at an unprecedented rate \cite{Aliyu2017Towards, Balaji2023Machine}. Traditional communication systems, initially designed for human-to-human interaction and optimized for transmitting raw data, are struggling to keep pace with the vast amount of information generated on a daily basis \cite{Jaafreh2018Multimodal, Mordacchini2020HumanCentric}. Bandwidth limitations, latency constraints and numerous other issues represent significant challenges in modern network designs \cite{Tassi2017Modeling}.

At the core of these issues lies a fundamental mismatch between the growth of data generation and the expansion of communication infrastructure capabilities. Conventional communication paradigms, based on Shannon's information theory \cite{Shannon1948Communication}, focus on the accurate compression and delivery of bits and symbols, regardless of their contextual significance. While this bit-centric approach is effective in scenarios where preserving every detail of the original data is crucial (such as transmitting ultra-high-definition images e.g. to be posted on social networks), it becomes increasingly inefficient in contexts where only a fraction of the transmitted data is relevant to the end-user or application \cite{Strinati20216G, Gunduz20246G}. Considering all symbols as equally important can lead to suboptimal use of bandwidth and computational resources, increasing the pressure on communication networks.

One possible scenario where the bit-centric communication framework could increase the pressure on the communication network is in the so-called machine-to-machine communication. The absence of humans in this type of communication relies on the fact that machines can autonomously collect, process, and send data from a transmitter to a receiver that will further process them to take some actions. In this context, transmitting the raw data in its entirety can be both redundant and counterproductive. Machines are, in fact, designed to process the data via some algorithms. These can either be a simple rule-based algorithm or a more complex and advanced \gls{nn}. In both cases the model will focus mainly on particular features that most influence the decision. All the other features will be discarded as they are non-relevant \cite{Tishby2015DNNIB}. This implies that the same action or decision can often be executed without the need for the original raw data but instead just by using a transformation that preserves information about the relevant parts. This process will guarantee the same level of performance, eliminating the necessity for bit-by-bit transmission \cite{Luo2022Semantic}. An example where it is beneficial to consider only relevant features is the so-called human-robot interaction. In this framework a human is interacting with machines that communicate with each other. This can happen, for example, in an industrial scenario where the human works in close proximity to robotic arms. These arms will have to move loads and perform some actions, but more importantly, avoid any potentially dangerous contact with the human. This task requires constant and real-time monitoring and prediction of the human movements'. This task could be performed by recording \gls{hr} images of the human and transmitting them to another machine that will process and aggregate multiple images to decide whether to stop the movement of the arm or not. Another better option would be to pre-process the data locally. To this scope, it is possible to locally extract the graph representation of the human's pose, transmit it to the receiver which will further process the information considering possible transmission errors \cite{Sampieri2022Pose, Testa2024Stability}. By transmitting only this distilled information, the communication load can be drastically reduced without compromising the final goal.

\begin{wrapfigure}[12]{r}{0.45\textwidth}
    \vspace{-10pt} % Adjust as needed
    \centering
    \includegraphics[width=0.4\textwidth]{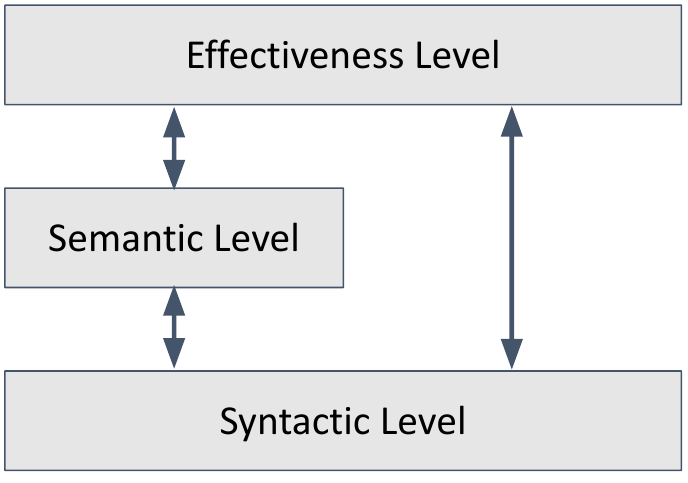}
    \caption[Scheme of three levels of communication]{Diagram illustrating the three levels of communication and their interconnection as proposed in \cite{WARREN1953semantic}}
    \label{fig: INTRO semantic_communication}
\end{wrapfigure}

This idea of going beyond the classical bit-by-bit reconstruction was proposed by Weaver in 1953 when he introduced the three different levels of communication \cite{WARREN1953semantic}: (i) the syntactic level, focusing on the accurate transmission of symbols; (ii) the semantic level, focusing on the conveyance of meaning; and (iii) the effectiveness level, focusing on the impact of the communication on the desired outcome or action. The syntactic level represents the fundamental level of communication. For this level, Shannon provided a rigorous and formal solution in what is known as \gls{it} \cite{Shannon1948Communication}. This level involves the technical aspects of transmitting and reconstructing symbols through physical channels. The semantic level builds upon this by addressing the interpretation and extraction of the \textit{semantic meaning} embedded within the data. Instead of considering the data as a series of symbols, the semantic level tries to focus on the information conveyed by the data. At the top is the effectiveness level, which ensures that the communication achieves its intended purpose. This level is designed to orchestrate and interact with both previous communication levels and the environment resources to fulfill specific goals, embodying the essence of what is referred to as \gls{goc}. 
As shown in \fref{fig: INTRO semantic_communication}, these levels are intrinsically interconnected, forming a hierarchical structure where each higher level builds upon and controls the lower ones.

The correct implementation of these higher-level communication paradigms has interested researchers for their potential in improving communications \cite{Strinati20216G, Han2022SemanticPreserved, Gunduz2023Beyond, Qin2021Semantic, Lu2021Rethinking}. \gls{sc} addresses the inefficiencies of traditional paradigms by focusing on the extraction and transmission of the actual information that the transmitter intends to convey, rather than the raw data itself. For example, in the context of autonomous vehicles, it is important to avoid any collision between the vehicle and any other object. The classical way to perform this task is by continuously sending \gls{hr} images to a base station that will process the images and select the best action to avoid a car accident. However, the vehicle could pre-process these images locally to identify and extract critical information about the position and shape of other objects, such as pedestrians, other vehicles, traffic signals, and more \cite{Grigorescu2020survey}. All of this information can be represented via the so-called \gls{ssm}, a representation that can be compressed far more than the original image while preserving all the essential semantic information necessary for navigation and obstacle avoidance. Sometimes, to safely navigate a vehicle it might be necessary to have more information than the position and shape of the surrounding objects. For example, only knowing the shape of a traffic sign might not be enough. In this case, the effectiveness level may identify that the information conveyed by the \gls{ssm} is not enough and instruct the transmitter to send additional information, such as a low-resolution version of the original image. This adaptive approach significantly reduces the amount of data that needs to be transmitted and adapts it to the specific semantic content and to the goals of the application.

In contexts where the semantic and the goal of communication are crucial, the integration of \gls{ai}, particularly generative models, provides valuable benefits. Combined with classical optimization techniques, these approaches offer promising ways to improve communication efficiency and effectiveness \cite{Xie2021DeepLearningEnabled}.

Generative models, such as \glspl{gan} \cite{goodfellow2014generative}, \glspl{vae} \cite{Kingma2014VAE}, and \glspl{ddpm} \cite{Ho2020ddpm}, have demonstrated incredible capabilities in learning complex data distributions and generating high-quality synthetic data \cite{Dhariwal2021DDPM_beat_GAN}. By leveraging these models, it becomes possible to reconstruct or generate data at the receiver side based on few compressed semantic representations of the original data \cite{He2022Robust}. For example, when a transmitter processes an image of a street scene and sends only the \gls{ssm}, the receiver can use that information as input of a generative model to reconstruct an image that, while visually different in terms of colors and textures, maintains all objects in their correct positions. This ensures that the essential semantic information required for tasks like navigation or scene understanding are preserved while drastically reducing the amount of bits to be transmitted thus decreasing latency which is critical for applications requiring real-time responses.

However, the practical implementation of \gls{sc} and \gls{goc} using generative models in real-world applications present significant challenges. One of the most critical issues is related to resource constraints, including limited computational power and energy availability, especially on devices \cite{Zhou2019Edge, Chis2016PerformanceEnergy, Wang2019An}. Generative models, while capable of producing high-quality images, are computationally intensive and typically require substantial processing power and memory. Running these models on servers or cloud infrastructures is relatively feasible. Instead, deploying them on devices like smartphones, autonomous vehicles, or \gls{iot} sensors introduces constraints due to limited hardware capabilities and the necessity for energy efficiency. Moreover, the communication channels themselves may have limited bandwidth and are susceptible to latency and reliability issues. Therefore, trade-offs between computational and communication resources, compression efficiency, reconstruction quality, and semantic preservation must be meticulously balanced to meet application-specific requirements. Pre-processing the data too much could decrease the autonomy of the device battery, while insufficient pre-processing could excessively load the communication infrastructure.

A special and quite useful case where it is possible to monitor and control every single aspect of the communication is the edge computing in the \gls{en}. Edge computing has emerged as a pivotal component in modern communication networks, offering a paradigm where computation and data storage are brought closer to the data sources \cite{Shi2016edge}. This proximity significantly reduces latency and bandwidth usage, enabling real-time applications like augmented reality, autonomous driving, and \gls{iot}. 
Research has focused on developing techniques for optimizing resources within the context of \glspl{en} to achieve a favorable trade-off between computational demands and communication efficiency \cite{Binucci2023goaloriented, Hu2024Semantic}.

By integrating generative models at the edge, it becomes possible to perform complex data processing tasks locally, reducing the need for extensive data transmission to centralized servers. For instance, an \gls{ed} can observe an image, and preprocess it via some classic compression algorithms or \gls{nn} model before transmitting only the essential semantic information. It will be at the \gls{es} that the power of the generative models is used to reconstruct an image similar to the original one in terms of semantic information preservation. Additionally, the use of \gls{goc} protocols ensures that the transmitted data aligns with the specific objectives of the application, further optimizing resource utilization.\\

This thesis will focus  on the interconnection between \gls{sc}, \gls{goc} and edge computing optimization in the context of image compression. Multiple \gls{sc} image compression algorithms will be proposed in parallel with resource allocation strategies to enhance communication efficiency and effectiveness. The structure of the semantic image compression and reconstruction algorithm will be based on the use of generative models. The reconstruction of the image will be mainly, but not only, based on the information contained in the \gls{ssm}. The performance of the proposed models will be compared with classical image compression algorithms like \gls{bpg} or \gls{jpeg2000} in terms of classical and semantic metrics. Moreover, these models will be designed to be integrated into an \gls{en} framework. A resource allocation strategy based on the \gls{ib} problem \cite{Tishby1999IB,Tishby2015DNNIB} will be proposed to enhance communication efficiency and the resources will be optimized via stochastic optimization techniques to provide an overall insight into the performance of the \gls{en}. 
\section{\textcolor{black}{Contribution and Thesis Outline}}

In this section, the structure of the thesis is introduced and a brief description of the main contributions is provided.

\begin{itemize}[label={}]
    \item {\textbf{Chapter \ref{ch: SEMCOM}) Foundations of Semantic and Goal-Oriented Communication:}} This chapter provides an overview of the three levels of communication proposed by Weaver in 1953, focusing on the \gls{sc} and \gls{goc} levels. The differences and the relations between these levels are discussed, highlighting the importance of semantic information in communication systems. Additionally, in this chapter the advantages of employing generative models to achieve \gls{sc} and how the \gls{ib} problem can be useful for \gls{goc} are suggested.
    \item {\textbf{Chapter \ref{ch: generative_models}) Foundations of Generative Models and Evaluation Metrics:}} This chapter goes over the fundamental concepts of generative models, including \glspl{vqgan}, \glspl{vqvae}, and \glspl{ddpm}, providing all the necessary background on the most important architectures that will be used in the following chapters. Furthermore, various classical and semantic evaluation metrics used to assess the performance of the proposed image compression models will be discussed. After a general overview of these metrics, the last part of this chapter will be devoted to the introduction of a new specifically designed metric. This is expressed in terms of traffic sign classification accuracy and involves a specific process for identification and classification of traffic signs.
    \item {\textbf{Chapter \ref{ch: SPIC}) Semantic-Preserving Image Coding based on Conditional Diffusion Models:}} This chapter introduces a novel semantic image coding scheme designed to preserve the semantic content of an image  while ensuring a good trade-off between coding rate and image quality preservation. The proposed Semantic-Preserving Image Coding framework (\acrshort{spic}) is based on a modular approach where the transmitter encodes the \gls{ssm} and a low-resolution version of the original image. The receiver then employs the proposed Semantic-Conditioned Super-Resolution Diffusion Model (\acrshort{semcore}) to reconstruct the \gls{hr} image based on the information contained in the \gls{ssm} and the low-resolution image. 
    The modular approach allows the framework to be easily updated, and to be more flexible in case of architectural changes without requiring any further fine-tuning or re-training. Thanks to this design choice, other variations are proposed to address multiple scenarios where different elements are considered semantically relevant and computational and/or communication resources are limited.
    \item {\textbf{Chapter \ref{ch: SQGAN}) Semantic Image Coding Using
Masked Vector Quantization:}} This chapter introduces the Semantic Masked VQ-GAN (\acrshort{sqgan}), a novel approach that integrates \gls{vqgan} and \gls{sc} principles to optimize image compression and transmission. This new model is designed to selectively encode and transmit only semantically relevant parts of the image, effectively reducing redundancy and enhancing communication efficiency. The task is performed by a designed module, the Semantic Conditioned Adaptive Mask Module (\acrshort{samm}), that learns how to identify and rank semantically relevant features. This chapter also introduces a data augmentation technique designed to target specific semantically relevant classes to allow better reconstruction. Moreover, a variation of the discriminator network designed to focus only on some specific semantically relevant classes is introduced in this chapter.
    The proposed model outperforms traditional compression algorithms such as \gls{bpg} and \gls{jpeg2000} across multiple metrics and works at extremely low values of \gls{bpp}. Moreover, it is designed to be easily implemented in an \gls{en} framework due to the presence of two parameters that can be dynamically adjusted in a \gls{goc} fashion and directly influence the level of compression.
    \item {\textbf{Chapter \ref{ch: Goal_oriented}) Goal-Oriented Resource Allocation in Edge Networks:}} This chapter introduces the resource allocation of the \gls{en}. The goal is to minimize the average power consumption while meeting constraints on average service delay and evaluation metrics of the learning task. Stochastic optimization methods are used to dynamically adjust all the computational, communication, and compression resources in a Goal-Oriented fashion. This chapter begins with combining the \gls{ib} problem and stochastic optimization to achieve \gls{goc}. The \gls{ib} principle is used to design the encoder in order to find an optimal balance between representation complexity and relevance of the transmitted data with respect to the goal. Additionally, the same stochastic optimization framework is used to integrate in the \gls{en} the proposed \acrshort{sqgan} introduced in \cref{ch: SQGAN}. This will provide an overview of what is defined as \gls{sgoc}.
\end{itemize}

\section{Related Publications}
The main contributions of this thesis will be introduced in \cref{ch: SPIC}, \ref{ch: SQGAN} and \ref{ch: Goal_oriented} and are based on the following publications:
\begin{quotation}
\noindent \textit{\textbf{\large Goal-Oriented Communication for Edge Learning based on the Information Bottleneck}}\\
\textit{Francesco Pezone, Sergio Barbarossa, Paolo Di Lorenzo}\\
Proceedings of 2022 IEEE International Conference on Acoustics, Speech and Signal Processing (ICASSP), 2022, pp. 8832-8836

\vspace{0.2cm}

\noindent \textit{\textbf{\large Semantic-Preserving Image Coding based on Conditional Diffusion Models}}\\
\textit{Francesco Pezone, Osman Musa, Giuseppe Caire, Sergio Barbarossa}\\
Proceedings of 2024 IEEE International Conference on Acoustics, Speech and Signal Processing (ICASSP), 2024, pp. 13501-13505

\vspace{0.2cm}

\noindent \textit{\textbf{\large C-SPIC: Class-Specific Semantic-Preserving Image Coding with Residual Enhancement for Accurate Object Recovery}}\\
\textit{Francesco Pezone, Osman Musa, Giuseppe Caire, Sergio Barbarossa}\\
Manuscript in Preparation 

\vspace{0.2cm}

\noindent \textit{\textbf{\large SQ-GAN: Semantic Image Coding Using
Masked Vector Quantization}}\\
\textit{Francesco Pezone, Sergio Barbarossa, Giuseppe Caire}\\
Manuscript in Preparation
\end{quotation}
\noindent Other contributions more related to the overall idea and advantages of generative models for \gls{sc} and the \gls{ib} problem for \gls{goc} are presented in \cref{ch: SEMCOM} and are based on the following publications:
\begin{quotation}
\noindent \textit{\textbf{\large Semantic and Goal-Oriented Communications}}\\
\textit{Sergio Barbarossa, Francesco Pezone}\\
\textit{6G Wireless Systems: Enabling Technologies}, edited by M. Chiani, S. Buzzi, L. Sanguinetti, and U. Spagnolini, CNIT Tech Report, 2022

\vspace{0.2cm}

\noindent \textit{\textbf{\large Semantic Communications based on Adaptive Generative Models and Information Bottleneck}}\\
\textit{Sergio Barbarossa, Danilo Comminiello, Eleonora Grassucci, Francesco Pezone, Stefania Sardellitti, Paolo Di Lorenzo}\\
IEEE Communications Magazine, vol. 61, no. 11, 2023, pp. 36-41
\end{quotation}
\noindent Additionally, a publication leveraging the syntactic level of communication to perform radio frequency denoising could have been included. However, it has been left out for sake of exposition and is reported here for reference:
\begin{quotation}
\noindent \textit{\textbf{\large Demucs for Data-Driven RF Signal Denoising}}\\
\textit{Çağkan Yapar, Fabian Jaensch, Jan C. Hauffen, Francesco Pezone, Peter Jung, Saeid K. Dehkordi, Giuseppe Caire}\\
Proceedings of 2024 IEEE International Conference on Acoustics, Speech, and Signal Processing Workshops (ICASSPW), 2024, pp. 95-96
\end{quotation}

\section{Remark on Notation}
This thesis extensively utilizes concepts from \gls{ml}, for this reason it is important to introduce some notation that will be used throughout the following chapters. \\
The original image will be referred to as $\x$ and the original \gls{ssm} as $\s$. They are both represented by tensors in three dimensions. In the context of tensors, the term "shape" will be used to refer to the dimensions of the tensor, that is, how many elements exist in each dimension. The term "size" will be used to refer to the total number of elements present in the entire tensor.

The original image $\x$ is a tensor of shape $3 \times H \times W $, where $3$ refers to the RGB channels, $H$ is the height and $W$ is the width of the frame. The \gls{ssm} $\s$ is a tensor of shape $n_c \times H \times W $, where $n_c$ refers to the number of semantic classes, $H$ is the height and $W$ is the width of the frame. In some contexts, the \gls{ssm} will be represented by a tensor of shape $3 \times H \times W $ for the RGB representation of $1 \times H \times W $ for the grayscale representation, they can all be considered interchangeable. However, the first representation will be the one used as input to any \gls{nn} model.

The proposed models in this thesis are designed to compress $\x$ and $\s$ at the transmitter side and reconstruct an image $\hat{\x}$ and a \gls{ssm} $\hat{\s}$ at the receiver. The term "reconstructed" will be used when referring to the output of the proposed models, i.e. $\hat{\x}$ and  $\hat{\s}$, while the term "generated" refers to the \gls{ssm} obtained from an image, either $\x$ or $\hat{\x}$, via some out-of-the-shelf pre-trained \gls{sota} \gls{ssmodel}. In this context, the term \textit{\gls{ssm} preservation} or \textit{\gls{ssm} retention} will refer to the property of the reconstructed image $\hat{\x}$ to generate a \gls{ssm} that is similar to the original $\s$ associated to $\x$. 

Any intermediate output of the model is referred to as the "latent representation," the "latent tensor," or the "features tensor".  

Moreover, a generic part $P$ of the model is understood to depend on some trainable parameters $\theta$. If 
$P(\cdot)$ is used instead, it denotes the function implemented by this part, mapping a given input to its corresponding output based on the current parameters.

Other specific notations are defined along the thesis when necessary.

\chapter{\textcolor{black}{Foundations of Semantic and Goal-Oriented Communication}}
\label{ch: SEMCOM}
\thispagestyle{plain}
The content of this chapter is based on the current state of the art and on the contributions at the core of \sref{sec: SEMCOM sem_gen}, \sref{sec: SEMCOM sem_go} and the implementation of the \gls{ib} principle in the context of \gls{goc}. These contributions are based on the following publications:
\begin{quotation}
\noindent \textit{\textbf{\large Goal-Oriented Communication for Edge Learning based on the Information Bottleneck}}\\
\textit{Francesco Pezone, Sergio Barbarossa, Paolo Di Lorenzo}

\vspace{0.1cm}
\noindent \textit{\textbf{\large Semantic and Goal-Oriented Communications}}\\
\textit{Sergio Barbarossa, Francesco Pezone}

\vspace{0.1cm}

\noindent \textit{\textbf{\large Semantic Communications based on Adaptive Generative Models and Information Bottleneck}}\\
\textit{Sergio Barbarossa, Danilo Comminiello, Eleonora Grassucci, Francesco Pezone, Stefania Sardellitti, Paolo Di Lorenzo}
\end{quotation}

\section{Introduction}
In 1983, during an interview, the famous physicist Richard Feynman was asked why two magnets repel each other. Faced with this apparently simple question, one of the greatest physicists who ever lived took the opportunity to illustrate an important lesson: answering a "why" question is not easy at all! One of the first assumptions is that the questioner and the respondent share some common knowledge. It is useless if the questioner is a 5-year-old and the answer involves concepts of quantum mechanics. Another assumption is that there must be a point at which the question is considered answered. It is always possible to respond with another "why" question; it's the classic game that kids love to play—keep asking "why?". For this reason, Richard Feynman ultimately told the interviewer, \textit{"I'm not going to be able to give you an answer to why magnets attract or repel, except to tell you that they do."}\\

This anecdote is a simple example of how communication can sometimes be difficult. Two interlocutors can be in the same room and talk for hours, but if what they say is not understood by the other, or if it does not satisfy the other's curiosity, then the communication is not effective.\\
These concepts are not new in the field of \gls{it}.  It was in 1953 when Weaver suggested that the broad subject of communication can be divided into three main levels \cite{WARREN1953semantic}:
\begin{itemize}[{label={--}}]
    \item \textbf{Syntactic level}: \textit{How accurately can the symbols of communication be transmitted? (The technical problem.)}
    \item \textbf{Semantic level}: \textit{How precisely do the transmitted symbols convey the desired meaning? (The semantic problem.)}
    \item \textbf{Effectiveness level}: \textit{How effectively does the received meaning affect conduct in the desired way? (The effectiveness problem.)}
\end{itemize}

\section{Syntactic Level}
The syntactic level is one of the most studied in the field of communication. It refers to the technical problem of how to transmit symbols to guarantee a correct reconstruction at the receiver end. This level is based on the work proposed by Shannon in 1948 \cite{Shannon1948Communication}. At the time, there was a lack of a mathematical framework to understand and optimize the communication process. Telephone networks and radio transmissions were becoming more popular, and engineers were struggling with issues related to signal noise, bandwidth, and the capacity of communication channels. Shannon proposed a mathematical model to describe the communication process, and his contribution completely changed the way communication is approached.

The theory proposed by Shannon is still the basis of many modern communication systems. Building on his work, researchers have developed a vast number of communication strategies, such as sophisticated forms of error detection and correction \cite{Mercier2010errorcorrection}, multiple-input multiple-output (MIMO) communications \cite{Jensen2016MIMO}, mitigation of multi-user interference \cite{Yang2022interference}, etc.

At the same time, new network and communication infrastructures are being developed at an incredible pace. Technologies like 4G and 5G are now part of the daily life of billions of people, and 6G is on the horizon \cite{Saad2019Vision6G}. Unfortunately, the rate of improvement of physical devices and communication infrastructures is subject to physical limitations. All the players in telecommunications spend billions of dollars every year to access finite resources like bandwidth. The management of these resources is a complex task, and the optimization of the communication process is a never-ending challenge. For this reason, the syntactic level alone might not be sufficient anymore.

The subtle problem is that, from theory, it is known that even with the best possible compression algorithm, there is a limit to the number of bits at which a piece of data can be compressed. This limit depends on its entropy, also referred to as Shannon entropy. It is not possible to perform better than this. This means that if the idea is to reconstruct the exact sequence of symbols, the best that can be achieved in terms of compression is given by its entropy. However, if the idea is to convey the meaning of a piece of data regardless of the form, this can be potentially achieved at values lower than the entropy of the original data. This is the idea behind the semantic level of communication.

\section{Semantic Communication (Semantic Level)}\label{sec: SEMCOM sem}
On the semantic level, the way the message is reconstructed is not relevant as long as the \textit{semantic information} is preserved. The term semantic information refers to the information that is conveyed by the data and is relevant to the receiver, allowing the receiver to understand the message without reconstructing it symbol by symbol.

Multiple works have proposed formal theories concerning \gls{sc} \cite{Bao2011SemEntropy, Gunduz2024SemTheory, Carnap1954SemCommTheory}. This thesis will present \gls{sc} a more intuitive and high level way. It is in fact possible to consider any given piece of data $\x$ as composed of two parts: 
\begin{itemize}[{label={}}]
    \item \textbf{Syntactic component} $V$: This quantity refers to the subset $V=\{v_i\}$ of the symbols $v_i \in \mathcal{V}$ used to represent the data $\x$ in its original domain. Here, $\mathcal{V}$ denotes the alphabet of the symbols.
    \item \textbf{Semantic component} $S$: This quantity refers to the subset $S=\{s_i\}$ of the semantic information/meanings $s_i \in \mathcal{S}$ selected from the semantic alphabet $\mathcal{S}$ and associated with the data $\x$. 
\end{itemize}
Only by having access to both components is it possible to fully describe the data $\x$. In fact, without the semantic component $S$, the data $\x$ is just a sequence of symbols $V$ that can be processed only at a symbolic level. Without the syntactic component $V$, the data $\x$ itself will not exist. However, to fully describe the data $\x$ in a semantic way, the semantic component $S$ is the most important part. Once $S$ is defined, it is possible to associate it with multiple symbolic components $V$.

To clarify the concept, it is possible to consider $\x$ as an image of an urban environment. The symbolic component $V$ will be composed of the sequence of RGB pixel values $v_i$ of the image. The semantic component $S$ will instead be associated with the more abstract semantic information $s_i$ contained in the image. Some examples might be accessing if there is a pedestrian crossing the street, if a car is driving too close, or if the right window on the fifth floor is open or not.

From $S$, it is possible to derive multiple symbolic components. For example if $S$ represents only the semantic meaning \textit{"the traffic light is red"}, there are countless possible configurations $V$ of the RGB values $v_i$ that represent an image with the same semantic meaning.

\begin{figure}
    \centering
    \includegraphics[width=\textwidth]{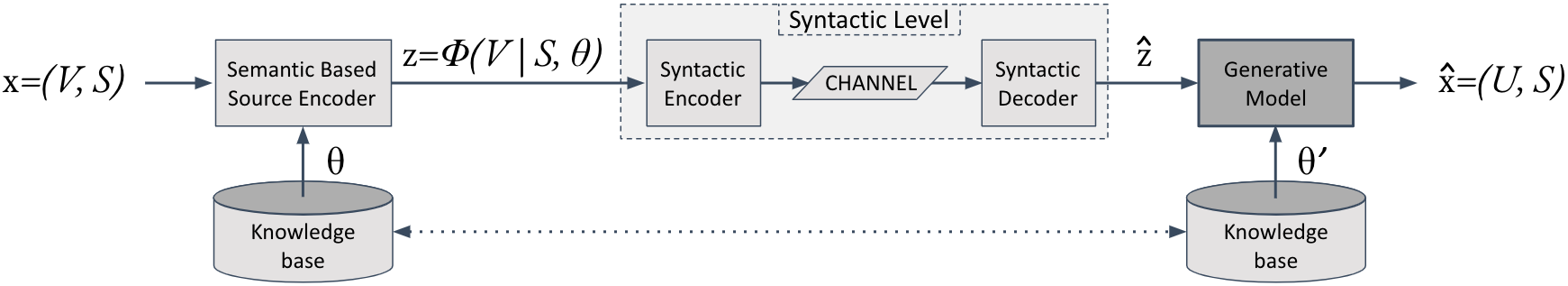}
    \caption[Semantic Communication Scheme]{Semantic Communication scheme.}
    \label{fig: SEMCOM semantic_comm_scheme}
\end{figure}

In fact, in the context of \gls{sc}, only the semantic information matters. Two pieces of data $\x=(V,S)$ and $\hat{\x}=(U,T)$ are defined as \textit{semantically equivalent}, represented with $\x \longleftrightarrow \hat{\x}$, if and only if $S=T$. This means that as long as the semantic components are the same, the data can be considered the same on a semantic level. No assumption is made about the syntactic components $V$ and $U$, which might be completely different from one another.

This idea is shown in \fref{fig: SEMCOM semantic_comm_scheme}, where the data $\x=(V,S)$ is compressed and transmitted to the receiver. The syntactic component $V$ is transformed via a Semantic-Based Source Encoder to produce a representation defined as $\z=\Phi(V|S,\theta)$. This representation is composed of semantic symbols $z_i\in \mathcal{Z}$ selected from an alphabet $\mathcal{Z}$ to convey the semantic information $s_i$. Each $z_i$ is selected on the basis of $v_i$, the semantic information $s_i$, and the \gls{kb} $\theta$.

The \gls{kb} $\theta$ refers to a set of facts, rules, constraints, etc., specific to the semantic context at hand. It is used to connect the syntactic symbols $v_i$ to the semantic symbols $z_i$ associated with the semantic information $s_i$ and vice versa. This is a fundamental part of the \gls{sc} framework since only if the \gls{kb} is shared between the transmitter and the receiver can \gls{sc} happen. Ideally, $\theta$ and $\theta'$ should be the same at both ends, even if they could potentially differ within some margins.

The semantic symbols $\z$ can now be forwarded to the syntactic level. In fact, \gls{sc} relies on the syntactic communication to encode, transmit, and decode every semantic symbol $z_i$. Without the syntactic level and the use of conventional compression techniques, \gls{sc} would not be possible.

At the receiver end, the received transformation $\hat{\z}$ can be used to reconstruct a semantically equivalent representation of the data $\x$. The reconstruction is influenced by the \gls{kb} $\theta'$ and is performed by a specialized semantic decoder (represented by a generative model in \fref{fig: SEMCOM semantic_comm_scheme}) to reconstruct $\hat{\x}=(U,S)$.

To clarify this concept, consider again the example of the image of the street environment. Assume that the semantic component $S$ is composed only of the semantic information $s_1$ that the traffic light is red. Also, consider that the \gls{kb} $\theta$ knows that the data refers to images of a street environment, and the \gls{kb} $\theta'$ knows that the images refer to a street environment on a cloudy day. In this case, the Semantic-Based Source Encoder will process all the RGB pixels $v_i$, and the output transformation might be represented as a single sentence $\z=$ \textit{"There is a red traffic light"} composed of the semantic symbols, characters, $z_i$ associated with $s_1$. This string is encoded, transmitted, and at the receiver decoded by the syntactic level to obtain the string $\hat{\z}=$ \textit{"There ys a red trafgic licht."} The transmission introduced some errors that the syntactic level was not able to remove.

The string $\hat{\z}$ can now be used as input to a text-to-image generative model. This model will process $\hat{\z}$, and the \gls{kb} $\theta'$ will be used to give more context about the fact that the images are from a street environment on a cloudy day. By exploiting this additional context, the model will be able to correct the syntactic errors in $\hat{\z}$. The output of the model will be an image $\hat{\x}=(U,S)$ that is coherent with the semantic information contained in $S$.

Of course, the syntactic component $U$ of the image $\hat{\x}$ will not be the same as the $V$ of the original image $\x$, since there are countless combinations of pixel values that can represent the same semantic information. Nonetheless, $\x$ and $\hat{\x}$ can still be considered semantically equivalent, thus the communication happend in a semantically lossless way.

As introduced in the example, it is possible at the syntactic level to have the introduction of some errors on the semantic symbols $z_i$. In general the correction of such errors is demanded to the syntactic level. However, while this process can fail from time to time, the additional semantic decoder at the receiver can help correct these errors on a semantic level. The assumption is that the errors introduced by the syntactic level might not impact the preservation of the semantic information $s_i$. In the previous example, the two sentences \textit{"There is a red traffic light"} and \textit{"There ys a red trafgic licht"} present multiple errors that the text-to-image model employed at the receiver is fortunately able to correct. In fact, differently to other works that employ end-to-end architecture to incorporate the channel in the training phase \cite{Gunduz2019DeepJSCC, Felix2018OFDM}, this will consider a modular approach. All the three levels of communication will be considered separately and the possible errors introduced at the syntactic level will be corrected by the generative model at the semantic level. 

As shown in \fref{fig: SEMCOM semantic_comm_scheme}, in this thesis the \gls{sc} framework is designed to be based on the presence of generative models at the receiver. In fact, generative models are useful tools to extract and work with semantic information from the data and use them to produce semantically equivalent reconstructions.

\subsection{Semantic Communication Based on Generative Models}\label{sec: SEMCOM sem_gen}
Generative models have become increasingly popular in the field of machine learning due to their exceptional ability to model complex data distributions and generate new data samples that retain the essence of the original content. These models have achieved remarkable results in various tasks such as image \gls{sr} \cite{Ledig2017Photo}, denoising \cite{Vincent2010Stacked}, image-to-audio translation \cite{Zhou2017Visual}, 3D synthesis \cite{Wu2016Learning} and more.

The fundamental principle of generative modeling involves designing a model that can learn the underlying distribution of the data $\x$. This distribution is then used to produce representative samples similar to the data contained in the training dataset. In learning the distribution of the data, the model captures the semantic information $s_i \in S$ contained in it. For instance, when a generative model is trained to generate human faces, it starts by learning how to identify simple patterns (features), like horizontal and vertical lines, rounded objects, or abrupt changes in colors. Progressively, as the model processes the input data through its hidden layers, more complex semantic features will be identified and considered. Instead of simple patterns, the model will focus on structures like eyes, nose, mouth, etc. and progressively incorporate more complex features \cite{Karras2019Style, Zeiler2014Visualizing}.

This way, the model is able to obtain a hidden representation $\z$ of the data that contains all the semantic information $s_i$ about faces. By using these hidden representations, the model can generate new samples that are coherent with the original data.

One important advantage of employing generative models is their flexibility. These models are able to reconstruct a plausible $\hat{\x}$ in an incremental way. If the received semantic information increases, then the model will be able to reconstruct data that are more semantically close to the original one, with higher \textit{semantic similarity}. By semantic similarity is intended any metric that is able to capture the differences between the semantic content of data. This will be discussed in detail in \sref{sec: GM evaluation metrics} where various semantic metrics will be introduced.

\begin{figure}
    \centering
    \includegraphics[width=\textwidth]{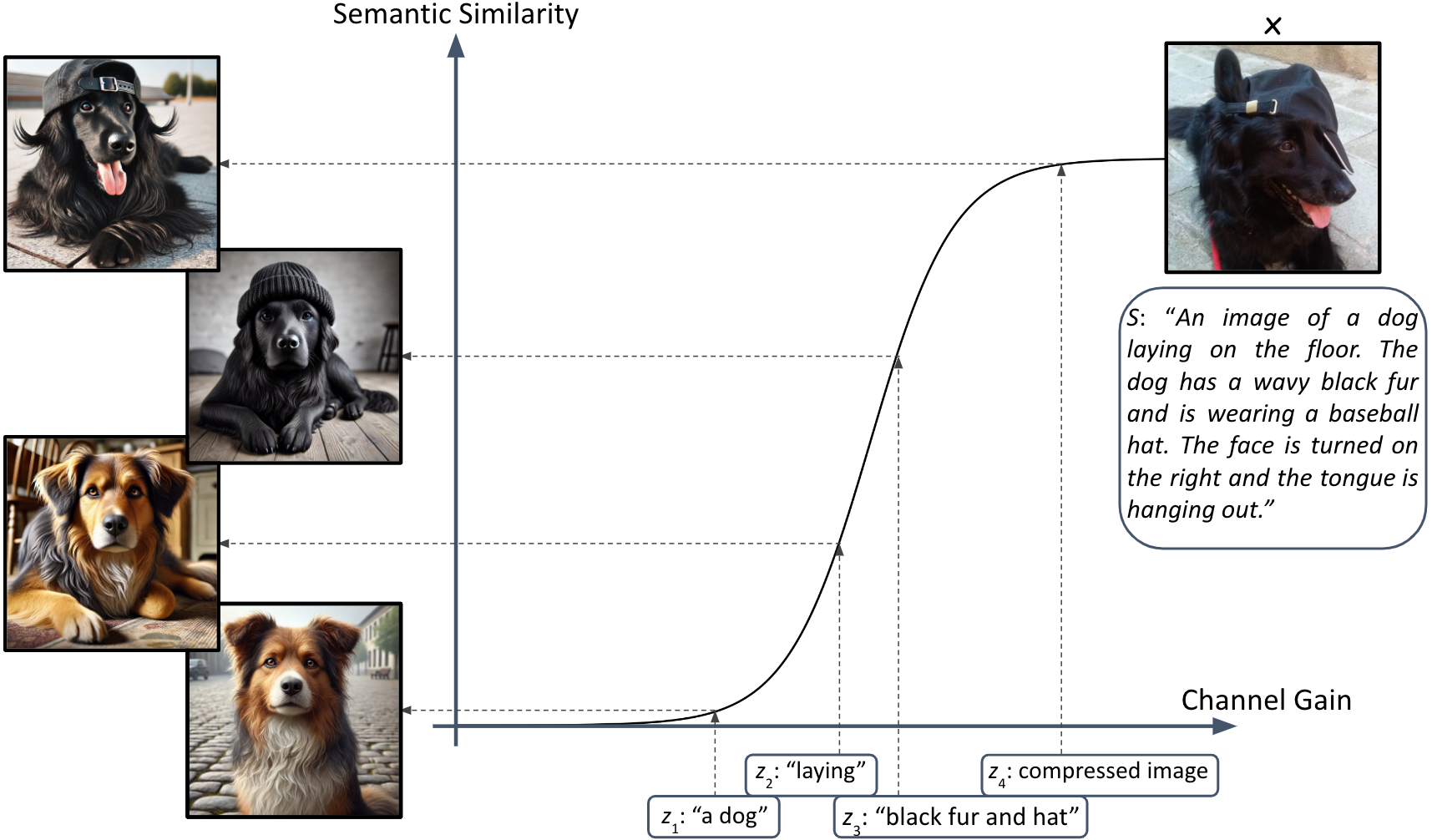}
    \caption[Semantic reconstruction performances as channel quality varies]{Impact of varying channel conditions on semantic reconstruction and how the generative model adapt to these changes. As the channel quality improves, the generative model is able to receive more semantic symbols $\z_i$ and eventually reach semantically equivalence between $\x$ and $\hat{\x}$.}
    \label{fig: SEMCOM generative_model_channel}
\end{figure}

This concept of flexibility provided by generative models is depicted in \fref{fig: SEMCOM generative_model_channel}.

Suppose that the original data $\x$ is the picture of the dog in the top-right corner \cite{Lexthehead2024Alex}. Associated with this picture is the semantic meaning $S$ produced with the  LLaVA-v1.5-13B model \cite{Liu_2024_CVPR} and reported under $\x$. In a real scenario the transmission might not always be possible. In fact, the channel might be noisy, the bandwidth might be limited or the transmission might be too expensive.

In these cases, it is important to have a communication system that is able to adapt to the channel conditions. To this end, suppose that the syntactic level is based on a successive refinement approach \cite{Tian2008SuccRefinement}. In this way, the receiver will be able to decode more semantic symbols $z_i$ as the channel conditions improve. If the channel performances are very poor, it might be possible to transmit only one semantic symbol. This $z_1$=\textit{"a dog"} can be inserted as input in the SDXL generative model \cite{Podell2023SDXL} to produce the bottom-left image in \fref{fig: SEMCOM generative_model_channel}. This will already be enough to generate a realistic image of a dog, but the semantic similarity with the original one might be very low.

As soon as the channel conditions improve, the generative model will be able to receive more semantic symbols $z_i$ and generate images that are increasingly semantically similar to the original one. If the channel conditions are very good, then it is possible to transmit all the semantic symbols $z_i$, and the generative model will be able to generate an image that is semantically equivalent to the original one.

However, the realization of this adaptive scenario is not trivial. The selection of the semantic symbols $z_i$ to be transmitted is a complex task related to the effectiveness level of communication.

\section{Goal-Oriented Communication (Effectiveness Level)} \label{sec: SEMCOM go}
The last level of communication proposed by Weaver is the effectiveness level. This level is related to the effect of the received message on the receiver end. In other words, it is related to the ability of the message to induce the desired behavior in the receiver. In this context, the idea is to transmit not all the information, but only those that are strictly relevant to the fulfillment of a certain goal. For this reason, the communication is also referred to \gls{goc}.

This level of communication can be considered as a higher level that can directly orchestrate the behavior of lower (semantic and syntactic) levels and how the communication infrastructure process the data and control the different connected components \cite{Zhang2022goaloriented, Kountouris2024goal}.

\begin{figure}
    \centering
    \includegraphics[width=\textwidth]{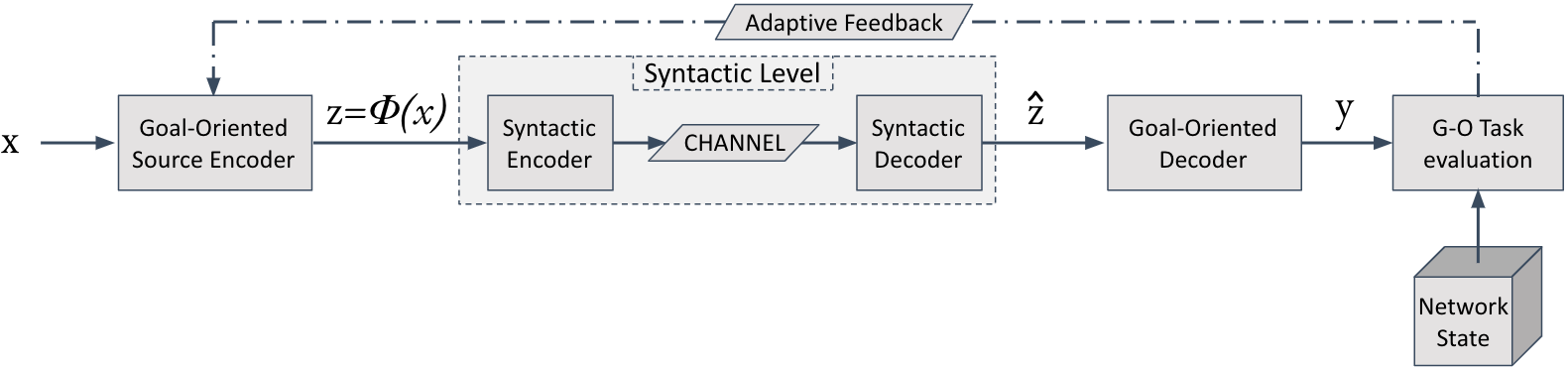}
    \caption[Goal-Oriented Communication Scheme]{Goal-Oriented Communication Scheme.}
    \label{fig: SEMCOM go_comm_scheme}
\end{figure}

The scheme of the \gls{goc} is depicted in \fref{fig: SEMCOM go_comm_scheme}.

Consider a scenario where the transmitter can observe some data $\x$, and the communication is happening to perform a certain task at the receiver end. The nature of this task can range from classification to parameter estimation. It can be constrained by power consumption or bandwidth limitations. In this context consider $\y$ as the data that contains the necessary information to perform a certain task at the receiver.

The idea of \gls{goc} is to compress and transform $\x$ to obtain a representation $\z=\Phi(\x)$ that is able to preserve all the information that $\x$ has on $\y$.

Additionally, the level of compression can be dynamically adapted to external factors that depend on the communication infrastructure. An example is in the context of \glspl{en} where multiple \glspl{ed} communicate to the same \gls{es}. It might happen that the communication infrastructure is at capacity but an additional \gls{ed} has to be inserted. In this case, the \gls{goc} paradigm is required to optimize the resource in the \gls{en}. This might cause some \glspl{ed} to compress the data more to reduce their impact on the network and allow all the \glspl{ed} to communicate with the \gls{es}.

This is just one of the possible examples of how \gls{goc} can be used to adapt to external factors. The dynamic adaptation of the network will be discussed in \cref{ch: Goal_oriented}, where the \gls{goc} framework will be introduced in the context of \glspl{en} and the resource optimization process presented.

In this section, the focus will be posed on one of the possible approaches to obtain a representation $\z$ that is effective.

\subsection{Goal-Oriented Communication Based on Information Bottleneck} \label{sec: SEMCOM ib}

In \gls{goc}, the idea is to identify the transformation $\z=\Phi(\x)$ so that $\z$ contains the same level of relevant information that $\x$ has on $\y$, and at the same time, $\z$ is maximally compressed.

When a transformation $\z$ satisfies these two conditions, it is said to be a \textit{minimal sufficient statistic} of $\x$ with respect to $\y$ \cite{Cover2006IT}. This concept can be expressed in terms of the mutual information between the terms as:
\begin{equation}
    I(\x; \z) = \min_{\w: I(\w; \y) = I(\x; \y)} I(\x; \w).
    \label{eq: SEMCOM min_suff_stat}
\end{equation}
Minimizing the mutual information $I(\x; \z)$ ensures that the transformation $\z$ is as compressed as possible. Simultaneously, the constraint $I(\z; \y) = I(\x; \y)$ guarantees that $\z$ preserves all the information that $\x$ originally had regarding $\y$.
The advantage of transmitting this transformation $\z$ is that the receiver will be able to perform the task as well as if $\x$ was transmitted but with a potentially high advantage in terms of transmitted bits.

Unfortunately, the process of identifying the minimal sufficient statistic is not an easy task. For this reason in \cite{Strinati20216G} was proposed, and further extended in \cite{Shao2021learning},  the use of the \gls{ib} method \cite{Tishby1999IB} to perform \gls{goc}. The \gls{ib} is used to loosen the constraints described in \eref{eq: SEMCOM min_suff_stat} and is represented as follows:
\begin{equation}
    \min_{\Phi=p(\z|\x)} I(\x; \z) - \beta I(\z; \y).
    \label{eq: SEMCOM ib_problem}
\end{equation}
The new formulation is a trade-off between the compression capabilities of $\Phi$ and the information that $\z$ is able to retain about $\y$. In the \gls{ib} problem the transformation function $\Phi=p(\z|\x)$ is represented in a probabilistic way and not deterministic. The parameter $\beta$ is a non-negative parameter used to explore the trade-off between the two mutual information where $I(\z; \y)$ is referred to as \textit{relevance} while $I(\x; \z)$ as \textit{complexity}.

\begin{figure}
    \centering
    \includegraphics[width=0.85\textwidth]{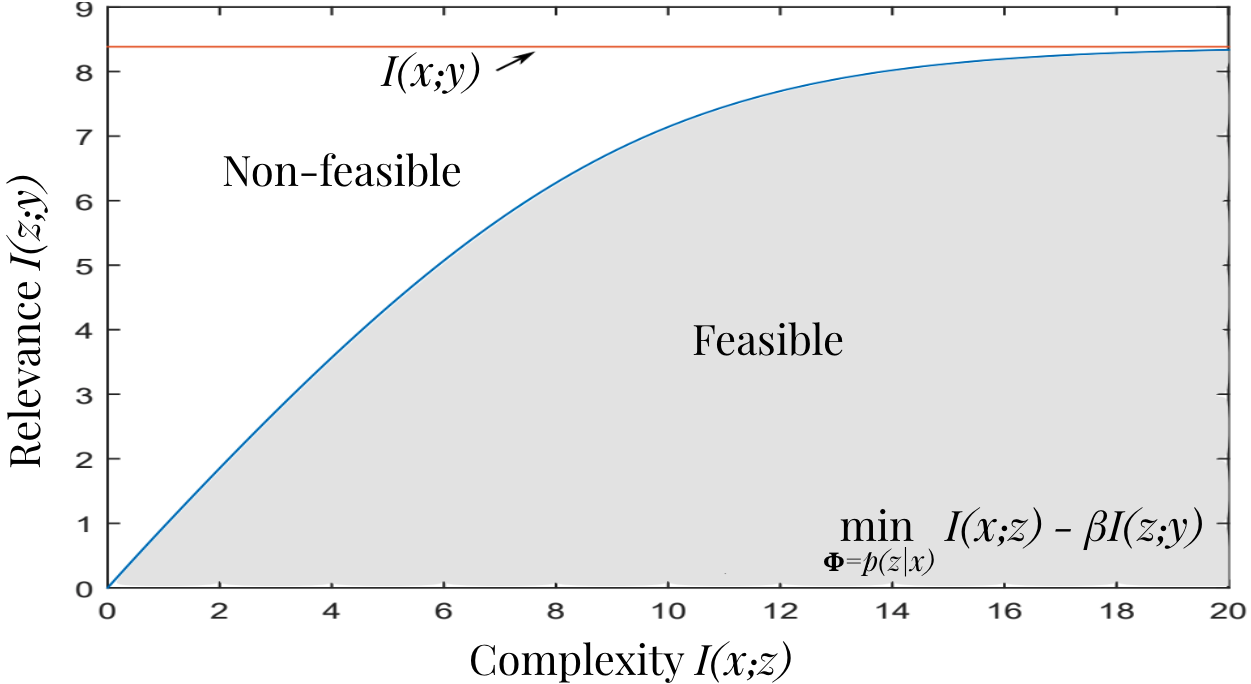}
    \caption[Information Bottleneck Trade-Off in the Relevance-Complexity plane]{Trade-off representation of the \acrshort{ib} problem in the Relevance-Complexity plane \cite{Zaidi2020IB}. The blue line represents the optimal solutions for different values of $\beta$ while the red line represents the limit of relevance given by all the information that $\x$ has on $\y$. The optimal solution is achieved in the case of  \acrshort{gib} where the projection matrix $\mathbf{A}$ is selected as in \eref{eq: SEMCOM Matrice_A}.}
    \label{fig: SEMCOM gib_tradeoff}
\end{figure}

For low values of $\beta$, the \gls{ib} problem will prefer compression over performance. This translates into a transformation $\z$ that has very low complexity but, unfortunately, also the performance will be negatively impacted, and the effectiveness of the communication will be lower. At high values of $\beta$, the transformation $\z$ will be obtained by preferring the maximization of the relevance. This will inevitably improve the effectiveness of the communication as well as the complexity of the transformation, resulting in more bits to be transmitted. The relationship between the two quantities is reported in \fref{fig: SEMCOM gib_tradeoff}.

In the plot, it is possible to identify a feasible region where the solution will be found. The optimal solutions for any given value of $\beta$ are represented by the points on the blue boundary between the two regions. As $\beta$ increases, the feasible region approaches the red line that represents the upper-bound of the relevance given by all the information that $\x$ has on $\y$, formally $I(\x;\y)$.

This problem is not trivial to solve in general. However, there are two cases in which it is possible to be solved: (i) the case of discrete random variables that admits a solution achievable via an iterative algorithm based on the Blahut-Arimoto algorithm \cite{Tishby1999IB}, and (ii) the case where $\x$ and $\y$ are jointly Gaussian \cite{Chechik2004GIB}.\\

In this work is presented the application of this second case, also referred to as the \gls{gib}, in the context of \gls{goc}. In fact, this case is helpful to better understand how the \gls{ib} works but also represents one of the cases where the solution is optimal. For this reason in this section will be recalled the main results of the \gls{gib} while in \sref{sec: EN_ib} it will be presented the integration in the \gls{en}.

Denote by $\x\sim \mathcal{N}(\mathbf{0}, \Sigma_{X})$ and $\y\sim \mathcal{N}(\mathbf{0}, \Sigma_{Y})$ two centered multivariate jointly Gaussian vectors of dimension $d_\x$ and $d_\y$, respectively. Also, let $\Sigma_{XY}$ represent the cross-covariance between $\x$ and $\y$. Under these conditions, the optimal encoding rule $\Phi$ is a linear transformation \cite{Chechik2004GIB} expressed as:
\begin{equation}
    \z = \Phi(\x)  = \mathbf{A} \x + \boldsymbol{\xi},
\end{equation}
where $\boldsymbol{\xi} \sim  \mathcal{N}(\mathbf{0}, \Sigma_{\xi})$ is a Gaussian noise, statistically independent of $(\x, \y)$.

In this configuration, Chechik et al. \cite{Chechik2004GIB} were able to evaluate the optimal transformation matrix $\mathbf{A}$ as a function of the trade-off parameter $\beta$. The structure of $\mathbf{A}$ is given by:
\begin{equation}
\mathbf{A} = \left\{\begin{matrix}
            [\mathbf{0}^T;...;\mathbf{0}^T] & 0 \leq \beta \leq \beta_1^c\\
            [\alpha_1\mathbf{v}_1^T; \mathbf{0}^T;...;\mathbf{0}^T] & \beta_1^c < \beta \leq \beta_2^c\\
            [\alpha_1 \mathbf{v}_1^T;\alpha_2\mathbf{v}_2^T;\mathbf{0}^T;\ldots;\mathbf{0}^T] & \beta_2^c < \beta \leq \beta_3^c\\
            \vdots\\
            [\alpha_1 \mathbf{v}_1^T;\alpha_2\mathbf{v}_2^T;\ldots;\alpha_{n_{\beta}}\mathbf{v}_{n_{\beta}}^T] & \beta_{n_{\beta}}^c < \beta 
            \end{matrix}\right.
\label{eq: SEMCOM Matrice_A}
\end{equation}
where $\mathbf{v}_i$ represent the left eigenvectors of the matrix $\bSigma_{X/Y}\,\bSigma_X^{-1}$ sorted by their corresponding ascending eigenvalues $\lambda_i$, for all $i=1, \ldots, n_{\beta}$; also, $\beta_i^c = \frac{1}{1-\lambda_i}$ denote the critical values of $\beta$ at which the number of components varies, $\alpha_i=\sqrt{\frac{\beta(1-\lambda_i)-1}{\lambda_i r_i}}$, and $r_i = \mathbf{v}_i^T\Sigma_{X} \mathbf{v}_i$,  for all $i=1, \ldots, n_{\beta}$. The value $n_{\beta}$ represent the highest index $i$ such that $\lambda_i \leq \frac{\beta-1}{\beta}$. This condition guarantees that the solution found is a global optimum.

The structure of the matrix $\mathbf{A}$ is very peculiar. For certain values of the trade-off parameter $\beta < \beta_1^c$, the preference for compression over performance is so high that the transformation $\z$ is the null vector and no information is sent to the receiver. As $\beta$ starts to increase, more importance is given to the performance, and $\z$ starts to populate with elements.

Because of the design of the \gls{gib}, it is possible to express the mutual information $I(\x; \z)$ and $I(\z; \y)$ in closed form as a function of $\beta$ as follows:
\begin{align}
    &I(\x; \z)=\frac{1}{2} \sum_{i=1}^{n_{\beta}} \log_2\left((\beta-1)\frac{1-\lambda_i}{\lambda_i}\right),\\
    &I(\z; \y)=I(x; z)-\frac{1}{2} \sum_{i=1}^{n_{\beta}}\log_2\left(\beta(1-\lambda_i)\right).
\end{align}
Since the solution of the \gls{gib} problem presented in \cite{Chechik2004GIB} is optimal, by representing these values on the relevance-complexity plane in \fref{fig: SEMCOM gib_tradeoff}, the trade-off will lie on the blue boundary between the feasible regions.

The interesting advantage of using the \gls{gib}, or more generally the \gls{ib}, is that the performance can be controlled by the trade-off parameter $\beta$. As will be discussed in \cref{ch: Goal_oriented}, this parameter allows communication to be adjusted based on external factors that might influence it. When multiple \glspl{ed} are connected to the same \gls{es}, a feedback mechanism from the \gls{es} to each \gls{ed} can influence and force it to adjust the value of $\beta$. By modifying the complexity of the transformation and its relevance, this approach becomes a powerful tool for facilitating the optimization of communication.

\section{Semantic-Goal-Oriented Communication} \label{sec: SEMCOM sem_go}
\begin{figure}
    \centering
    \includegraphics[width=\textwidth]{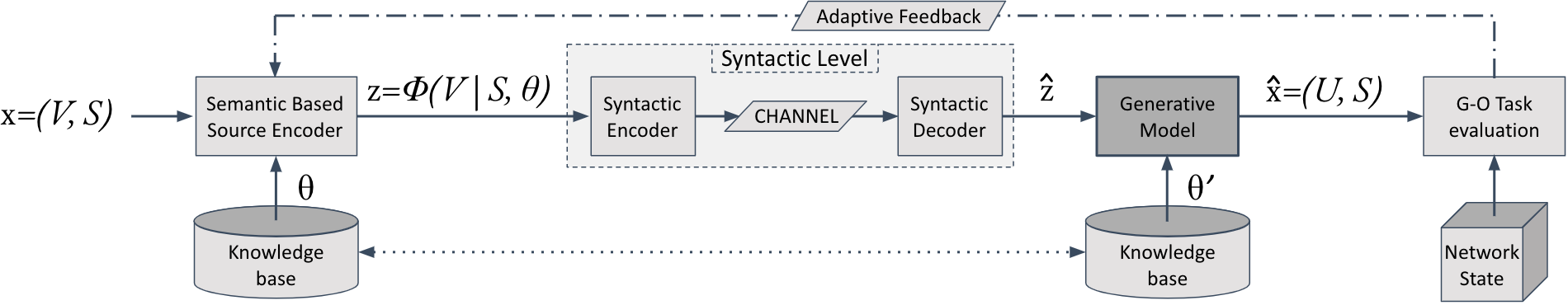}
    \caption[Semantic-Goal-Oriented Communication Scheme]{Semantic-Goal-Oriented Communication Scheme.}
    \label{fig: SEMCOM sem_go_comm_scheme}
\end{figure}

After discussing the three levels of communication, it is interesting to consider merging them all. As already discussed in the previous section and depicted in \fref{fig: INTRO semantic_communication}, the effectiveness level can work in synergy with the other levels. In this section, the structure of the so-called \gls{sgoc} will be discussed, as illustrated in \fref{fig: SEMCOM sem_go_comm_scheme}.

This type of communication is based on the premise that the semantic information $S$ is the most critical component of the data $\x$, and its preservation, along with power optimization, is the ultimate goal. Recalling \fref{fig: SEMCOM generative_model_channel}, even after the semantic level has successfully produced the semantic symbols $z_i$, the channel conditions may not be adequate to transmit all these symbols.

In such cases, \gls{goc} interacts with the semantic and syntactic levels, as well as the network transmitting the semantic symbols, to orchestrate the communication process. This orchestration involves selecting an appropriate subset of semantic symbols $z_i$ that are sufficient for the specific task and current network conditions.

By adaptively choosing which semantic symbols to transmit, the system ensures that the most relevant information is conveyed. This approach maintains the effectiveness of the communication by focusing on transmitting the semantic content that is most crucial for the receiver's task.

The practical implementation of this concept will be presented in \sref{sec: EN_nn}, where the model introduced in \cref{ch: SQGAN} will be integrated into a \gls{goc} framework.

% %   CHAPTERS
\chapter{\textcolor{black}{Foundations of Generative Models and Evaluation metrics}}\label{ch: generative_models}
\thispagestyle{plain}
The content of this chapter is based almost entirely on the current state of the art. The contribution of the author of this thesis is the introduction of the Traffic signs classification accuracy metric. This is presented at the end of the chapter and proposed in:
\begin{quotation}
\noindent \textit{\textbf{\large C-SPIC: Class-Specific Semantic-Preserving Image Coding with Residual Enhancement for Accurate Object Recovery}}\\
\textit{Francesco Pezone, Osman Musa, Giuseppe Caire, Sergio Barbarossa}
\end{quotation}

\section{Introduction}
Generative models represent a powerful class of \gls{ml} algorithms. Their capability to learn the underlying data distribution and to reconstruct or generate new data samples has made them some of the most utilized tools in \gls{ml}.

Since the introduction of early statistical methods like \textit{Restricted Boltzmann Machine} and \textit{Gaussian Mixture Model} \cite{Salakhutdinov2015surveyGenMod}, the field of generative models has undergone significant transformations. As task complexity increased, these initial approaches became inadequate for capturing the complexity of modern data distributions. While early models were effective in generating convincing images on simple datasets such as MNIST \cite{Lecun2010mnist}, which contains handwritten digits and the NORB dataset \cite{LeCun2004NORBdataset}, featuring small black-and-white toy images, more complex datasets like CIFAR10 and CIFAR100 \cite{Krizhevsky2009CIFAR} or ImageNet \cite{Russakovsky2015ImageNet} highlighted their limitations.

The first important breakthrough in the field of \gls{ml} came in 2012 with the introduction of AlexNet \cite{Krizhevsky2012Alexnet}. This architecture completely revolutionized the field by showing the incredible power of \glspl{dnn}, and more specifically of deep \gls{cnn}, in capturing complex details in the data distribution. AlexNet also popularized GPU acceleration for model training, the ReLU activation function \cite{Nair2010ReLU} and dropout for regularization \cite{Hinton2012dropout}, fundamentally changing the approach to \gls{ml}.

Shortly afterwards, architectures like the \gls{ae} \cite{Rumelhart1986Autoencoder} and the \gls{unet} \cite{Ronneberger2015Unet} were developed. While AlexNet was designed for classification tasks, these new architectures were proposed to work towards domains more closely related to generative models. The \gls{ae} was designed to capture a simplified latent representation of images and reconstruct them, while the \gls{unet} was specifically designed for \gls{sseg} tasks.

Building upon these architectures, a wide variety of new models have been developed in recent years. The \gls{ae} inspired the development of \glspl{vae} \cite{Kingma2014VAE, Kingma2019VAE}, incorporating probabilistic frameworks into the \gls{ae} structure. Then the introduction of \glspl{gan} \cite{goodfellow2014generative} introduced a new training paradigm based on two adversarial networks being optimized in a game theory scenario. Further advancements led to vector-quantized versions, such as the \gls{vqvae} \cite{Oord2017VQ-VAE} and the \gls{vqgan} \cite{Esser2O21Taming}. More recently, the family of \gls{ddpm} \cite{Ho2020ddpm} has gained significant attention for their innovative multistep approach in generating new data samples. These models have dramatically improved data generation quality,  enabling the generation of highly detailed images from simple text descriptions.

This chapter explores these models in parallel to some of the most important building blocks commonly used across many different architectures.  The \gls{ae}, \gls{unet}, \gls{resblock}, \gls{vae} and the \textit{Attention Mechanism} are fundamental components in a majority of the architectures discussed in this thesis. Understanding them is crucial before moving on to more complex architectures.

In \sref{sec: GM gan} the \gls{gan} architecture is discussed and in \sref{sec: GM vq-vae} the vector-quantized versions, such as the \gls{vqvae} and \gls{vqgan}, are introduced highlighting the advantages of vector quantization. Finally, the \gls{ddpm} models are analyzed, covering the techniques and concepts most relevant to this work.

In addition to exploring these generative models, this chapter also introduces the evaluation metrics and loss functions employed throughout this work. Selecting appropriate metrics and losses is crucial for designing, training, and assessing the performance of the models. The last section provides a comprehensive overview of both classic evaluation metrics, commonly used in image compression and reconstruction tasks, and semantic-relevant metrics, which evaluate the preservation of semantic information in the data. These metrics are particularly important in the context of \gls{sc} frameworks.

\section{The Building Blocks}\label{sec: GM unet}
This section is devoted to some of the most fundamental building blocks in the design of many generative models: the \gls{ae}, the \gls{unet}, the \gls{vae} and the \textit{Attention Mechanism}. The first three refer to complete architectures that serve as the starting points for the more complex models used in this thesis. They provide the necessary structures for encoding, decoding, and transforming data across different latent representation. The last one, the Attention Mechanism, is instead an important component widely used in modern \gls{ml}. This component is used to improve the performance of other models by allowing them to focus on specific parts of the data. Unlike classical convolutional layers, the attention mechanism can connect parts of the data that are spatially far apart.

\subsection{Autoencoder}

Proposed in \cite{Rumelhart1986Autoencoder}, the \acrlong{ae} is a type of neural network designed to reconstruct its input by learning an efficient representation of the data in an unsupervised way.\\
\glspl{ae} can vary greatly in complexity, ranging from \gls{ae} based on convolution operations to \gls{fc} layers. The one constant is the underlying structure composed of two main parts:

\begin{itemize}
    \item \textbf{Encoder:} The encoder network is the interface between the input data $\x$ and the latent representation $\z$. Its goal is to reduce the complexity of the initial data by preserving only the most relevant features.  

    \item \textbf{Decoder:} The decoder network is responsible for reconstructing the input data $\x$ from the latent representation $\z$. This network typically mirrors the encoder in structure but reverses the dimensionality reduction layers to progressively up-scale the latent tensor back to the original input size. The decoder's goal is to produce an output that is as close as possible to the original input by minimizing the reconstruction error.
\end{itemize}

By altering the type of encoder and decoder, different results can be achieved by an \gls{ae}. For example, if the encoder and decoder are composed of a series of \gls{fc} layers without nonlinear layers, then the \gls{ae} achieves the same latent representation as the \textit{Principal Component Analysis}, as shown in \cite{Plaut2018FromPS}.\\
Other powerful ways to design the \gls{ae} involve the use of \glspl{dnn} and convolutional layers. These more advanced architectures allow the capture of more complex patterns in the input data and provide the necessary power to handle input domains of increased complexity.\\
In general, \glspl{ae} are trained using some form of reconstruction loss to ensure that the reconstructed output resembles the input data. Examples of these reconstruction losses include \gls{l2} or \gls{l1}. However, for specific needs and advanced tasks, other forms of loss can be used to better approximate non-trivial data distributions.

\subsection{U-Net Architecture}\label{sec: GM unet}
\begin{figure}[!t]
    \centering
    \includegraphics[width=0.5\textwidth]{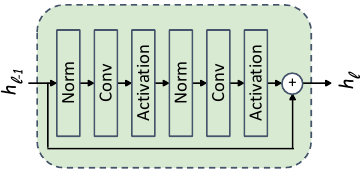}
    \captionsetup{width=.8\linewidth}
    \caption[\acrshort{resblock} architecture scheme]{Architectural diagram of the \gls{resblock}. The residual connection is represented by the line that goes from $h_{l-1}$ to the sum symbol.}
    \label{fig: GM resblock}
\end{figure}
The increasing complexity of both tasks and \gls{nn} depth presented various challenges to the classical \gls{ae}. \gls{sseg} tasks, for instance, highlighted the limitations of the bottleneck structure of the \gls{ae}, which often caused the loss of fine detail. In fact, the down-scaled latent space was insufficient for guaranteeing a good reconstruction of an output data with the same resolution as the input, since it did not effectively preserve spatial information. Another challenge was related to adapting the training process to the increased depth of \glspl{nn}. Since these networks are usually trained through backpropagation, the deeper the network, the longer the gradient has to "travel" backward. This results in the so-called problem of vanishing gradient \cite{Pascanu2013vanishGrad}, where the gradient becomes too small for effective training. 

To overcome these problems, the \gls{unet} was introduced in \cite{Ronneberger2015Unet}. Its structure is based on the \gls{ae} architecture, with the same encoder-decoder mirrored structure, and with some additional layers in between the encoder and decoder. These additional layers are called bottleneck layers and are used to further process the latent representation before it is handled by the decoder. However, the most important innovations in the \gls{unet} architecture consist in the extensive use of \glspl{resblock} and the introduction of the \textit{skip connections}. 

\begin{itemize}[label={}]
    \item{\textbf{ResBlock}:} Introduced by He et al. in \cite{He2016ResBlock}, not specifically for \glspl{unet}, this block was designed to help overcome the problem of the vanishing gradient. Before its introduction, the main approach was to use a long series of alternating convolutional layers, some nonlinear activation functions, and normalization, with a noticeable example being AlexNet \cite{Krizhevsky2012Alexnet}. He et al. proposed a new way of designing a network. Instead of a long uninterrupted sequence of convolution, activation, and normalization layers, the idea was to create a block, the \gls{resblock}, with a particular characteristic. 
    Every block is composed of a series of convolution, activation, and normalization layers that are repeated a given number of times, generally twice. The main difference lies in the introduction of the \textit{residual connection}. Every time a hidden layer $h_i$ encounters a \gls{resblock}, it is processed by the layers of the \gls{resblock} and at the end, the input is summed with the output, as illustrated in \fref{fig: GM resblock}.\\
    The role of the residual connection is to reduce the risk of degradation by providing shortcuts for the propagation of critical information. At the same time, during backpropagation their role is to split the gradient: one part is directed inside the block, allowing the parameters to be updated, while the other is copied at the other end, mitigating the vanishing gradient problem.
    
    \item{\textbf{Skip Connections}:} The innovation introduced by the \gls{unet} is the concept of skip connections. In fact, while the residual connection is adopted locally between a few layers, the skip connection works on a completely different scale. 
    \begin{figure}[!t]
        \centering
        \includegraphics[width=0.8\textwidth]{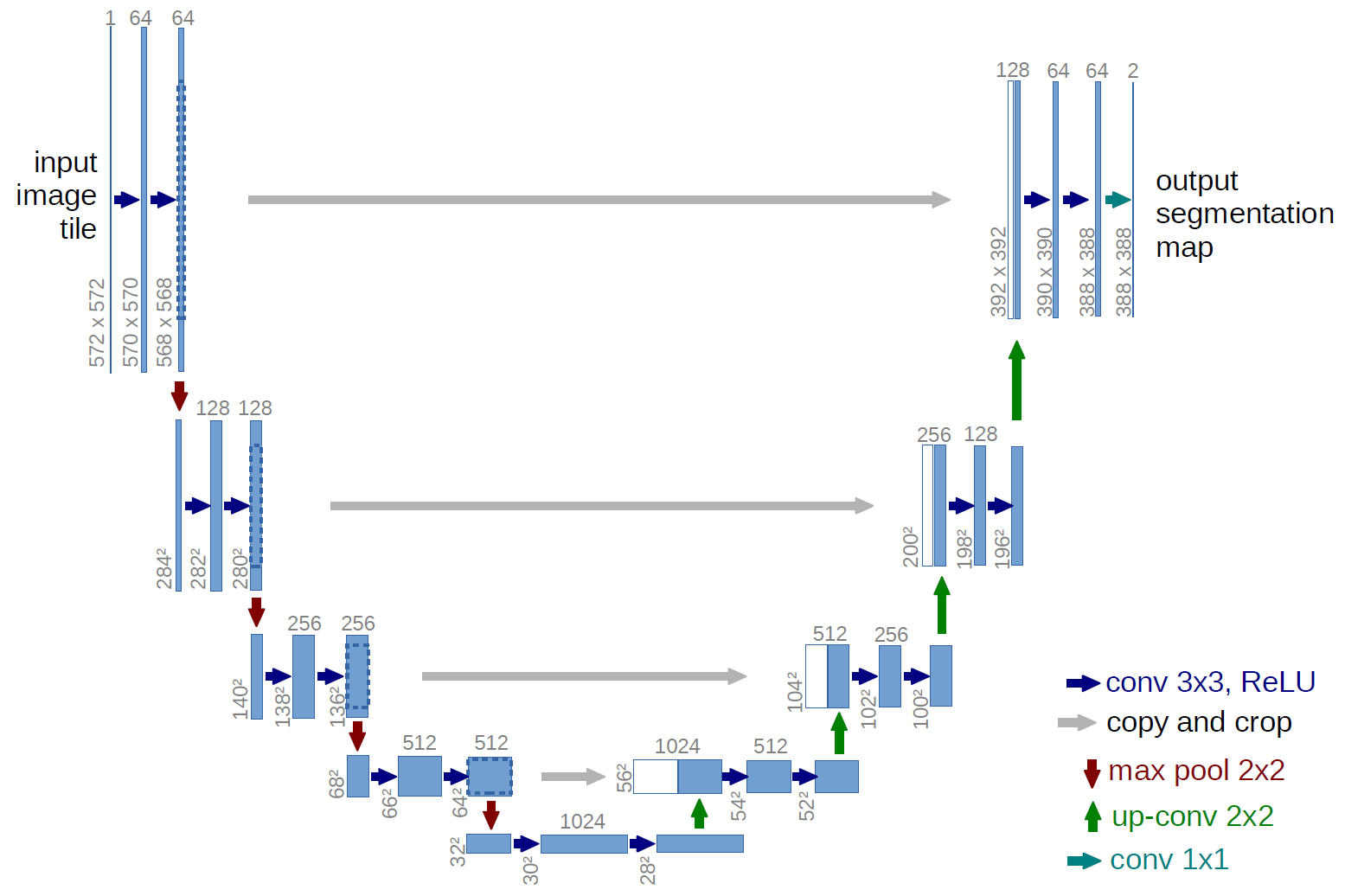}
        \captionsetup{width=.8\linewidth}
        \caption[\acrshort{unet} architecture scheme]{Architectural diagram of the \gls{unet} as proposed in \cite{Ronneberger2015Unet}. The encoder and decoder are linked via the skip connections represented in gray.}
        \label{fig: GM unet}
    \end{figure}
    As illustrated in \fref{fig: GM unet}, the skip connections link corresponding layers between the encoder and decoder. This has a double advantage: (i) during backpropagation, they allow the gradient to flow from the decoder to the encoder unchanged, avoiding the vanishing gradient, and (ii) during the forward process, they help preserve details. \\
    With skip connections, the model can reuse features extracted in the encoder at different resolutions even at the decoder. This enhances the ability of the \gls{unet} to generate detailed and accurate outputs. The decoder can access features processed both through the bottleneck but also features that are at the same level of detail. This is particularly important in tasks where preserving spatial information is critical, such as \gls{sseg} and \gls{sr}.
\end{itemize}

The main factor distinguishing the \gls{unet} from the \gls{ae} is the presence or absence of skip connections.\\
Even though skip connections encourage the preservation of detail, this does not automatically mean that the \gls{unet} is always better than the \gls{ae}. One of the advantages of using a \gls{unet} is its ability to preserve fine details, while one of its drawbacks is its shortcomings in data compression. In fact, while an \gls{ae} only requires the latent representation $\z$ to reconstruct the output, this would not be sufficient for a \gls{unet}, as all the intermediate skip connections would also have to be used. The choice of an \gls{ae} over a \gls{unet} is therefore strongly dependent on the task.\\

\begin{figure}[t]
    \centering
    \includegraphics[width=0.5\textwidth]{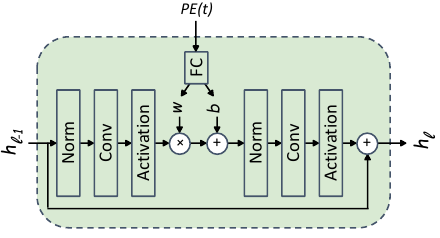}
    \caption[\acrshort{resblock} architecture scheme with time conditioning]{Architectural diagram of the \gls{resblock} conditioned on the time $t$. The value $t$ is mapped to an higher dimensional space via the \gls{pe} before being inserted in the \gls{resblock}.}
    \label{fig: GM resblock with time}
\end{figure}

While skip connections are specific to \glspl{unet}, \glspl{resblock} are a tool widely used in \gls{ml}. They are not specifically designed for \glspl{unet}; an \gls{ae} can be designed to include \glspl{resblock} as well. The advantages of \glspl{resblock} go beyond countering the vanishing gradient. They also offer a practical and simpler way to design a model. In fact, a model based on \glspl{resblock} can be easily modified to achieve better control over the output.\\
The scheme illustrated before in \fref{fig: GM resblock} represents only the base structure of a \gls{resblock}. However, by modifying its structure it is possible to condition its behavior on external factors. For example, in some cases it might be useful to introduce some form of time dependency. This can be easily achieved by modifying the \gls{resblock} architecture, as depicted in \fref{fig: GM resblock with time}, where the time conditioning is introduced by a linear application.
The first step involves mapping the time $t$ into a high-dimensional space that can be easily interpreted by the \gls{nn}. This is done via the so-called \gls{pe}, a technique developed for the Transformer architecture \cite{Vaswani2017AttentionIsAllYouNeed}. The idea is to map the value $t$ into a fixed or learnable higher-dimensional space, the $PE(t)$ vector, that can then be further transformed, for example via a \gls{fc} layer. This final transformation is now inserted in the middle of the \gls{resblock} and this conditioning will allow the model to learn the effects of time and reflect them in the result.\\
Another interesting example is the case where the performances are influenced by the \gls{ssm}. In this case, it is possible to modify the structure of the \gls{resblock} by substituting the classical normalization layer with the \gls{spade} layer introduced by Park et al. in \cite{Park2019SPADE}. They proposed a variation of the classical Batch Normalization layer \cite{Ioffe2015Batchnorm} that uses the information extracted from the \gls{ssm} to condition the normalization.

\begin{figure}[!t]
    \centering
    \includegraphics[width=0.4\textwidth]{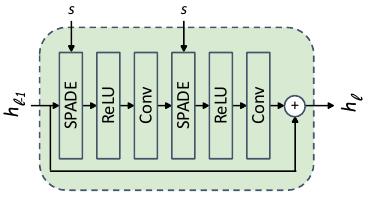}
    \hfill
    \raisebox{0\height}{\includegraphics[width=0.59\textwidth]{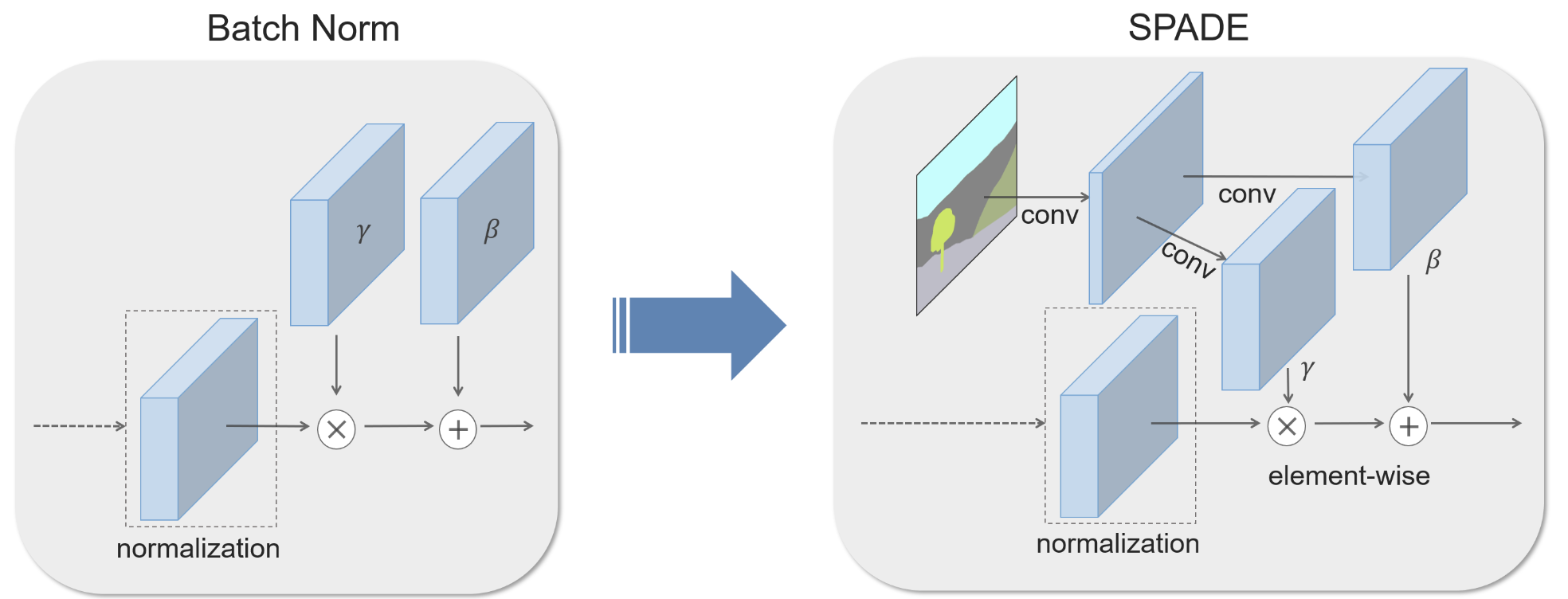}}
    \caption[\acrshort{spade} layer and \acrshort{resblock} architecture scheme]{On the left, the architectural diagram of the \acrshort{resblock} proposed in \cite{Park2019SPADE} that introduces \acrshort{spade}, a normalization technique able to condition on the \acrshort{ssm}. On the right, the \acrshort{spade} architecture is compared with batch normalization. The parameters $\gamma$ and $\beta$ are now learned as functions of the input \gls{ssm}.}
    \label{fig: GM resblock spade}
\end{figure}

More specifically, the main idea of their new \gls{spade} layer is depicted in the right scheme in \fref{fig: GM resblock spade}. In Batch Normalization, the re-scaling is achieved through two learnable parameters, $\gamma$ and $\beta$. These parameters are used to perform the linear transformation $BatchNorm(\h) = \gamma \h + \beta$ of the hidden variable $\h$. The intuition behind the \gls{spade} layer is to learn these parameters as a function of the \gls{ssm}. This is done as follows:
\begin{equation}
    SPADE(\h,\s) = f(\s) \h + g(\s),
\end{equation}
where $f$ and $g$ represent the \glspl{cnn} responsible for transforming the \gls{ssm} $\s$. The new $\gamma=f(\s)$ and $\beta=g(\s)$, while being used to normalize are also able to enforce the desired structure of the \gls{ssm} on the output. The structure of the new \gls{ssm} conditioned \gls{resblock} is shown on the left side of \fref{fig: GM resblock spade}. 

An important consideration has to be made on the effects of these conditioning. In fact, it is fundamental to introduce these modified \glspl{resblock} only in those parts of the \gls{unet} where their impact is maximized. To condition the feature extraction is good practice to condition the encoder blocks of the \gls{unet}. On the other hand, if the final purpose is to influence the output result the best option is to condition in the bottleneck and the decoder. For example, when the \gls{ssm} needs to be enforced on the output the modified \gls{resblock} in \fref{fig: GM resblock spade} should be placed in the bottleneck and decoder. Instead, in some other cases the conditioning variable is relevant throughout the whole process and all the \glspl{resblock} would be influenced. This is the case with the time conditioning.
 
\subsection{Variational Autoencoders}\label{sec: GM vae}
\begin{figure}[!h]
    \centering
    \includegraphics[width=0.7\textwidth]{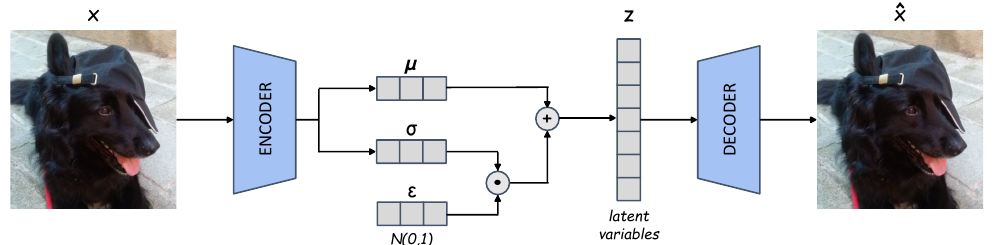}
    \caption[\acrshort{vae} architecture scheme]{Overview of the \acrshort{vae} architecture. The latent tensor is obtained as $\z=\boldsymbol{\mu} + \boldsymbol{\sigma} \odot \boldsymbol{\epsilon}$ via the reparameterization trick by selecting a random value $\bepsilon \sim \N(0, I)$.}
    \label{fig: GM schema_vae}
\end{figure}
A \gls{vae} \cite{Kingma2014VAE, Kingma2019VAE} extends the concept of classical \glspl{ae} by introducing a probabilistic framework to the encoding and decoding processes, \fref{fig: GM schema_vae}. Unlike the \glspl{ae}, that uses a deterministic approach, the  \glspl{vae} consider the encoding process as probabilistic. The data $\x$ is associated to a probability distribution $q(\x)$ that allow better generating capabilities. However, the direct use of $q(\x)$ to generate data is unfeasible because of its complexity in being directly estimated. To overcome this issue the \gls{vae} is designed to work with two simpler probability distributions: $q(\x|\z)$ and $q(\z|\x)$. They represent the core processes of encoding the data $\x$ to a latent representation $\z$ and then retrieving the original data back. Unfortunately, even if simpler than the $q(\x)$, these two distributions are still very complex and impossible to know exactly. This problem can be tackled by approximating them with the help of \glspl{dnn}.

Instead of dealing with $q(\z|\x)$ and $q(\x|\z)$, it is possible to consider the two approximating distributions $q_\phi(\z|\x)$ and $p_\theta(\x|\z)$ parameterized by two \gls{dnn} with parameters $\phi$ and $\theta$, respectively. Moreover, to further reduce the complexity, these approximations are usually forced to be Gaussian distributions.  This choice drastically reduces the complexity by shifting the problem from estimating an unknown probability distribution to estimating the mean and variance of a Gaussian distribution.\\

The key objective of a \gls{vae} is to maximize the likelihood of the observed data $q(\x) = \int q(\x|\z) q(\z) d\z$. However, directly maximizing this term is intractable due to the complexity of computing this value. To overcome this, \gls{vae} optimize the \gls{elbo}, which provides a tractable lower bound on the likelihood:

\begin{equation} \log q(\x) \geq ELBO(\x) =  \underbrace{\mathbb{E}_{q_{\phi}(\z|\x)} \left[ \log p_\theta(\x|\z) \right]}_{\text{Reconstruction Term}} - \underbrace{\mathcal{D}_{\text{KL}}(q_\phi(\z|\x) \parallel q(\z))}_{\text{Regularization Term}}. \label{eq: GM vae elbo_summarized} \end{equation}

The ELBO consists of two terms:

\begin{itemize} 
    \item \textbf{Reconstruction Term:} Encourages the decoder $p_\theta(\x|\z)$ to accurately reconstruct the input data from the latent representation. 
    \item \textbf{Regularization Term:} Encourages the encoder $q_\phi(\z|\x)$ to be close to the prior distribution $q(\z)$, typically chosen as a standard normal distribution as well. 
\end{itemize}

The training idea is to use the encoding network $q_\phi(\z|\x)$ to estimate the mean $\boldsymbol{\mu}$ and standard deviation $\boldsymbol{\sigma}$ of the true $q(\z|\x)$ and use them to sample the latent variable $\z$. This value is then used to reconstruct the data $\x$ with the decoding network $p_\theta(\x|\z)$. However, during backpropagation it would be better if the gradient could flow backward through the network and update the weights. Unfortunately, the sampling operation of $\z$ introduces a non-differentiability, which prevents backpropagation from updating the encoder weights.

To solve this problem, Kingma and Welling introduced the \textit{Reparameterization Trick} in the context of \gls{vae} \cite{Kingma2014VAE}. Instead of sampling $\z$ directly from $q_\phi(\z|\x)$, it is sampled with the help of an auxiliary variable $\boldsymbol{\epsilon} \sim \mathcal{N}(0, \mathbf{I})$ obtaining: 
\begin{equation}
    \z = \boldsymbol{\mu} + \boldsymbol{\sigma} \odot \boldsymbol{\epsilon}, 
\label{eq: GM vae z_reparametrized_summarized} 
\end{equation}

where $\boldsymbol{\mu}$ and $\boldsymbol{\sigma}$ are the mean and standard deviation output by the encoder network, and $\odot$ denotes element-wise multiplication.

The advantage is that the sampling process is now differentiable in $\boldsymbol{\mu}$ and $\boldsymbol{\sigma}$ and the gradient can flow back through the encoder to train the model. All the stochasticity is confined and depends on $\bepsilon$. It is important to notice that the system is still not differentiable with respect to $\bepsilon$, but this is not relevant in the optimization of the network.

The loss function used to train the \gls{vae} is derived from the negative ELBO as follows:
\begin{align}
    - \text{ELBO}(\x_i) \approx \Loss(\x_i ) &=  \frac{1}{L} \sum_{l=1}^{L} \frac{1}{2\sigma^2} \left( \left\| \x_i - \hat{\x}_{i,l} \right\|^2 \right) + \mathcal{D}_{\text{KL}}(q_\phi(\z|\x_i) \parallel q(\z))  \label{eq: GM vae loss_function_extended}\\
    &= \Loss_{\text{rec}}(\x_i) + \Loss_{\text{KL}}(\x_i ) \equiv \Loss_{\text{rec}} + \Loss_{\text{KL}} ,
\end{align}
where $\x_i$ is the $i$-th input data and $\hat{\x}_{i,l} = D_\theta(\z_{i,l})$ is the output of the decoder network.\\

Even if this loss function can already allow the \gls{vae} to generate good samples, it is good practice, as introduced in \cite{Higgins2016betaVAE}, to re-scale the different terms by a factor $\beta$. Also known as $\beta$-\gls{vae}, this new approach uses the following loss function:
\begin{equation}
    \Loss = \Loss_{\text{rec}} + \beta \Loss_{\text{KL}}.
\label{eq: GM vae beta_vae}
\end{equation}

The advantages are numerous, in fact the $\beta$-\gls{vae} is able to control the trade-off between the reconstruction and the regularization term, allowing the user to decide which one to prioritize. Different values of $\beta$ can lead to different outputs:
\begin{itemize}
    \item $\beta=1$: the classical \gls{vae} discussed until now
    \item $\beta>1$: prioritizes the regularization term, encouraging the latent representations to be more disentangled, meaning that individual dimensions of the latent space capture more distinct and interpretable factors. For example one component might focus on the shape of the eyes and one on the color. However, this can come at the cost of reconstruction accuracy.
    \item $\beta<1$: prioritizes the reconstruction term, leading to better reconstructions but poorer disentanglement of the latent factors. In this case any component might influence more than one factor.
\end{itemize}

By adjusting $\beta$, the trade-off between reconstruction fidelity and latent space regularization can be tweaked according to specific requirements. This approach is beneficial for applications where the interpretability of the latent space is crucial, or when the goal is to enhance the generative capabilities of the model.

\subsection{Attention Mechanism}\label{sec: GM attention}

The attention mechanism is a fundamental component in modern machine learning models, particularly in tasks involving sequence data, such as natural language processing and time series analysis. Introduced in \cite{Bahdanau2014Attention} and popularized by the Transformer architecture proposed in \cite{Vaswani2017AttentionIsAllYouNeed}, the attention mechanism allows the model to focus on specific parts of the input data. 

The core idea of the attention mechanism is to compute a weighted sum of the input elements, where the weights are dynamically determined based on the relevance of each element to the current task. This approach enables the model to selectively pay attention, \textit{attend}, to important information while ignoring non-relevant parts, leading to improved performance in tasks requiring context awareness.\\
The attention mechanism works by assigning three main vector representations to each input element: the query ($\mathbf{Q}$), the key ($\mathbf{K}$), and the value ($\mathbf{V}$). Drawing from concepts in information retrieval, the query $\mathbf{Q}$ identifies the information being sought, while the key $\mathbf{K}$ encodes the relevant features of the data that could match the query. The query and keys are processed together through a function that measures their compatibility, generating a score that indicates how well they align. This attention score ($\text{A}_{\text{score}}$) highlights which keys are most relevant to the query. The value $\mathbf{V}$ contains the actual data associated with each key and is retrieved based on the attention score, with higher scores giving more importance to the corresponding values.

The simplest form of attention is the so-called \textit{scaled dot-product attention}. In this version, the attention score is computed as the dot product of the query and key vectors, scaled by the inverse of the square root of the dimensionality of the key vectors, $d_k$. Then, a non-linearity is applied via the softmax function and these weights are then used to compute a weighted sum of the value vectors. The formula is:

\begin{equation}
\text{Attention}(\mathbf{Q}, \mathbf{K}, \mathbf{V}) = \text{A}_{\text{score}} \mathbf{V} = \text{softmax}\left(\frac{\mathbf{Q} \mathbf{K}^T}{\sqrt{d_k}}\right) \mathbf{V}
\label{eq: GM attention single head}
\end{equation}

The use of the attention score allows the model to dynamically weigh the importance of different input elements. The weights are directly linked to the query and key and the attention score adapt better than classical linear transformations to complex tasks.\\
Similar to the information retrieval scenario, every attention layer correspond to a request of information to the data. A single question is in general not necessary to extract multiple information. For this reason, a more commonly used practice is to use the \textit{Multi-Head Attention mechanism}. This mechanism consists of multiple attention mechanisms (or heads) running in parallel. This is used to capture different types of relationships in the data, where each head has its own set of query, key, and value matrices. The outputs from each head are then concatenated and linearly transformed to produce the final output:
\begin{equation}
\text{multi-head}(\mathbf{Q}, \mathbf{K}, \mathbf{V}) = \text{Concat}(\text{head}_1, \text{head}_2, \dots, \text{head}_h) \mathbf{W}^O,
\label{eq: GM attention multi head}
\end{equation}
where each attention head is computed as:
\begin{equation}
\text{head}_i = \text{Attention}(\mathbf{Q} \mathbf{W}^Q_i, \mathbf{K} \mathbf{W}^K_i, \mathbf{V} \mathbf{W}^V_i),
\end{equation}
with \(\mathbf{W}^Q_i\), \(\mathbf{W}^K_i\), \(\mathbf{W}^V_i\), and \(\mathbf{W}^O\) as learnable weight matrices specific to each attention head and the output linear transformation.\\

This mechanism can be used in many fields, from computer vision to natural language processing. By changing the input data and the query, key, and value structure, the attention mechanism can be adapted to specific tasks, resulting in different types of attention. One example is \textit{Self-Attention}, where the query, key, and value vectors are all derived from the same input. This allows each element to attend to all other elements. If the query is derived from a different input than the key and value, the attention mechanism is called \textit{Cross-Attention}. This is another possible technique used to condition the reconstruction on an external input. In Stable Diffusion \cite{rombach2022high}, cross-attention is widely used to condition the output on text input. In this case, the query is derived from the latent representation, while the key and value are obtained from the conditioning text. This allows the model to focus on relevant aspects of the text when generating the image, guiding the reconstruction process based on the given input.

\section{Generative Adversarial Networks (GAN)} \label{sec: GM gan}
\begin{figure}[!h]
    \centering
    \includegraphics[width=0.7\textwidth]{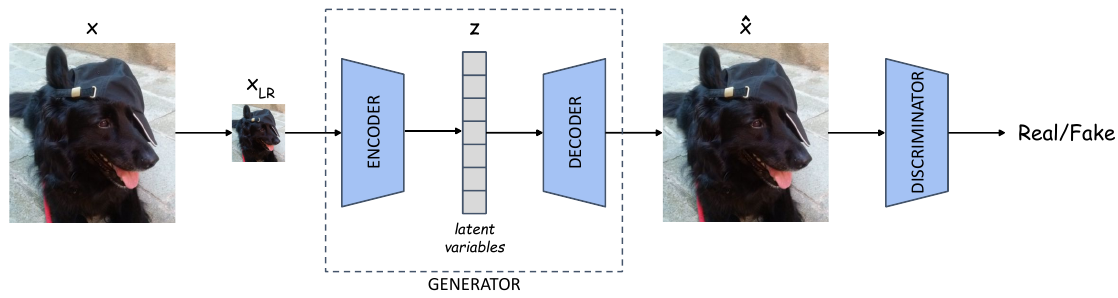}
    \caption[\acrshort{gan} architecture scheme]{Overview of the \acrshort{gan} architecture. The example reported involves a \acrshort{sr} task where the Generator tries to up-scale the $\x_{LR}$ such that the reconstructed $\hat{\x}$ can has enough details to fool the Discriminator.}
    \label{fig: GM schema_gan}
\end{figure}

In 2014, Goodfellow et al. introduced the  \gls{gan} \cite{goodfellow2014generative}. Unlike previous approaches, such as the \gls{vae}, the \gls{gan} is based on a \textit{game-theoretic} framework, where two neural networks are trained simultaneously in a competitive \textit{zero-sum minimax} game. The idea is that these two networks, the \textbf{generator} $G$ and the \textbf{discriminator} $D_{disc}$, compete against each other, trying to increase their own score while reducing the score of the other network. The generator tries to produce realistic samples such that it is hard for the discriminator to understand if they are real or fake. At the same time the discriminator tries to distinguish between real and fake samples, penalizing the generator if the classification is correct. This adversarial process results in a Nash equilibrium where the generator produces increasingly realistic samples that the discriminator cannot distinguish between real and fake any better than random guessing. For this reason in the long run the generator will reach an optimal generating strategy and changing this strategy will favor the discriminator. On the other hand in the long run the discriminator will have no incentive in changing its strategy either since it cannot discriminate between true and fake images any better than random guessing.\\

\subsection{GAN Objective}
In the original work introducing the Vanilla \gls{gan} \cite{goodfellow2014generative}, the generator $G$ and discriminator $D$ were designed to produce realistic images using a random noise vector as input. The process begins with generating a noise vector $\mathbf{z}$, which is then passed through the generator $G$ to produce a synthetic image $\hat{\mathbf{x}} = G(\mathbf{z})$. This generated image $\hat{\mathbf{x}}$ is fed to the discriminator $D$, which tries to distinguish between the generated image and a real image $\mathbf{x}$ sampled from the dataset. The discriminator outputs a probability that the input image is real or fake. Meanwhile, the generator is updated to improve its ability to generate images that can fool the discriminator.

The problem with this approach is that the vanilla \gls{gan} is designed to generate good-looking images, but without any specific application. The generated image will be every time a random image similar to those in the dataset used to train the model. This leaves no control over the output, making it impossible to generate a specific type of images.\\

This problem was partially solved with the introduction of the conditional \gls{gan} \cite{Mirza2014ConditionalGAN}. In this framework, alongside with the random noise, the generator take as input also a conditioning variable referring to the desired class of the generated object. In this way by using random noise and forcing the output to belong to a class, i.e. "dog", the generator was able to generate a random dog close to the one present in the dataset. 

However, this approach still presented too many degrees of freedom in the generation progress. Other than the class there was no other control over the generated image. 

Isola et. al in  \cite{isola2017image2image} tackled this problem proposing  a specific approach tailored for image-to-image applications.

Suppose that starting from a \gls{lr} image $\x_{LR}$ the idea is to generate its \gls{hr} version  $\hat{\x}$, as depicted in  \fref{fig: GM schema_gan}. The dataset will consist of pairs $(\x_{LR},\x)$ of \gls{lr} and real \gls{hr} images. 
The idea is to use an \gls{ae} architecture as generator that will takes as input $\x_{LR}$ and generates the \gls{hr} version $G(\x_{LR})=\hat{\x}$. The discriminator's goal will then be to understand if the image given as input is real or fake. This is done by allowing the discriminator to see at every iteration both the real $\x$ and the generated $\hat{\x} = G(\x_{LR})$ images and training it on both. In this way the discriminator will learn to classify images between real and fake by focusing on some details. This game can be formalized, as mentioned in the previous section, as a minmax problem with the following objective function:
\begin{equation}
    \min_G \max_{D_{disc}} \Loss_{\text{GAN}} = \underbrace{\mathbb{E}_{\x \sim  q(\x)} \left[ \log D_{disc}(\x) \right]}_{\text{Realism loss}} + \underbrace{\mathbb{E}_{\x_{LR} \sim q(\x_{LR})} \left[ \log \left( 1 - D_{disc}(G(\x_{LR})) \right) \right]}_{\text{Generator loss}},
\label{eq: GM gan objective_minmax}
\end{equation}
where $q(\x)$ is the distribution of the real \gls{hr} data and $q(\x_{LR})$ is the distribution of the \gls{lr} data.\\
The loss function $\Loss_{\text{GAN}}$ is composed of two terms:
\begin{itemize}
    \item The first term is the \textit{realism loss}, which encourages the discriminator to classify real \gls{hr} images as real. The discriminator is trained to maximize this term.
    \item The second term is the \textit{generator loss}, which encourages the generator to produce \gls{hr} images that the discriminator classifies as real. The generator is trained to minimize this term.
\end{itemize}

The loss function in  \eref{eq: GM gan objective_minmax} is sufficient to train the generator to produce images that can fool the discriminator. However, when the goal is to generate realistic images that are not only able to deceive the discriminator but are also visually appealing to humans, the $\Loss_{\text{GAN}}$ alone is not enough.\\
The work proposed by Pathak et al. in \cite{Pathak2016contextEncoders} showed how the loss function could be modified by adding terms to improve the quality of the generated images without affecting the discriminator network. This resulted in the introduction of the following composite loss :
\begin{equation}
    \Loss = \Loss_{\text{GAN}} +  \lambda_{\text{rec}} \Loss_{\text{rec}},
\label{eq: GM gan objective_complete}
\end{equation}
where $\Loss_{\text{rec}}$ is a term used to improve pixel-by-pixel similarity \cite{Pathak2016contextEncoders}, e.g. the \gls{l2} loss, and $\lambda_{\text{rec}}$ is a re-scaling factor to weight the effects of the reconstruction term. 

The is no strict rule about the choice of the $\Loss_{\text{rec}}$ that can differ depending on the task. For example, another simple and intuitive alternative could be the \gls{l1}, or a weighted combination of the two.

A parallel of this new loss can be drawn with the $\beta$-\gls{vae} discussed in  \eref{eq: GM vae beta_vae}. Even though the original \gls{vae} did not include the re-scaling factor $\beta$, overall performance could benefit from tuning this parameter. Similarly, in \glspl{gan}, adding another term to the loss function has shown improvements in results \cite{Pathak2016contextEncoders}. Another term often used in $\Loss_{\text{rec}}$ is the \textit{perceptual loss}. This term is introduced to push the generated image to have a better \textit{perceptual similarity} \cite{Bruna2016SuperResolution,Gatys2015textureSynthesis, Johnson2016PerceptualLoss} with the original image. This concept will be discussed in  \sref{sec: GM evaluation metrics}, when classical losses and evaluation metrics will be introduced alongside their semantic counterparts.

\section{Vector Quantized VAE and GAN}\label{sec: GM vq-vae}

\begin{figure}[!h]
    \centering
    \includegraphics[width=0.8\textwidth]{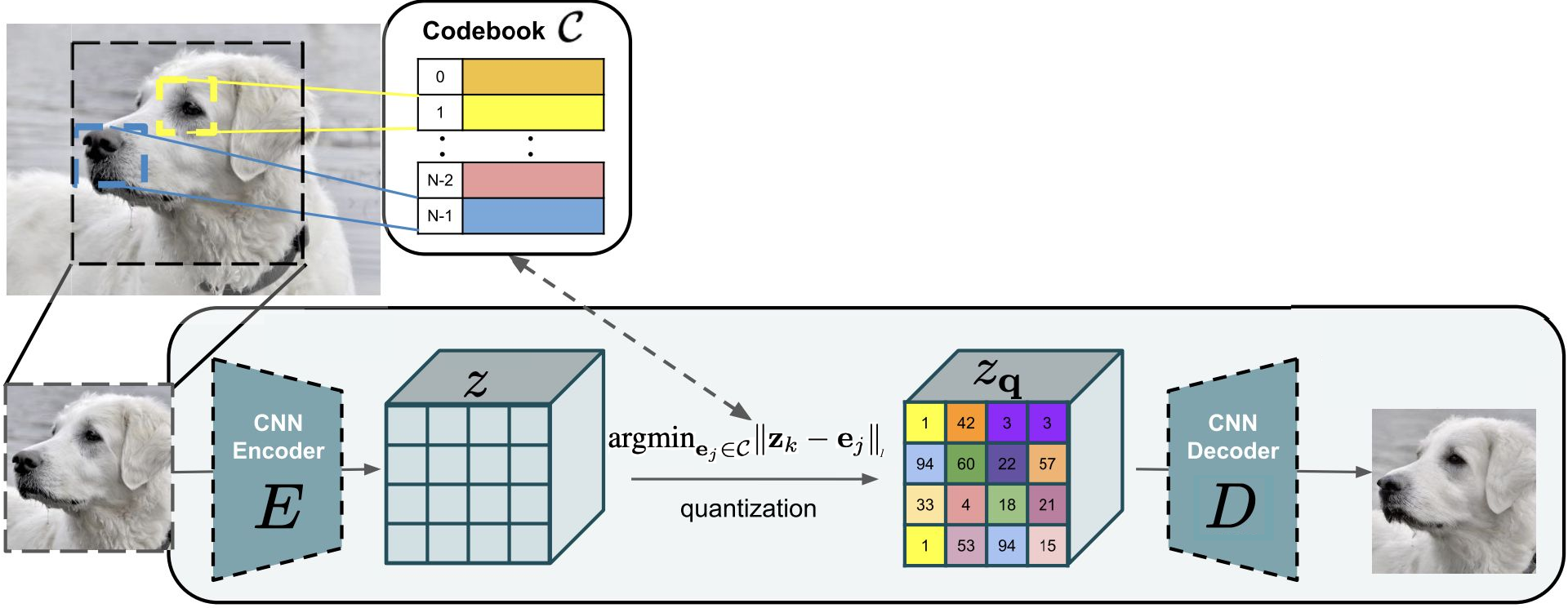}
    \caption[\acrshort{vqvae} architecture scheme]{Overview of the \acrshort{vqvae} architecture. After the encoding the latent vectors $\z_k$ are vector-quantized via the learnable codebook $\C$. The decoder then uses the quantized latent tensor $\z_q$ to generate the output instead of the original $\z$.}
    \label{fig: GM schema_vqvae}
\end{figure}

An interesting variant of the classical \gls{vae} or \gls{gan} are the respective vector quantization versions, namely \gls{vqvae} and \gls{vqgan}. \\

Proposed by Oord et al. in \cite{Oord2017VQ-VAE}, the idea behind these architectures is to replace the continuous latent representation $\z$ with a discrete vector-quantized version. By employing this discrete representation it is in fact possible to improve the quality of the final generated sample. 

In the classical versions discussed in the previous sections, the latent representation $\z$ is prone to contain some redundancy. This can happen because of the many ways to represent the same data in a continuous space and this might lead to inefficiencies. In the continue representation, small changes in the latent tensor $\z$ caused by some noise, could cause significant differences in the final output. These problems are partially or completely addressed by considering the vector quantization. 

In fact, by using a discrete representation, the model is forced to use a predefined set of vectors to represent the latent tensor $\z$. This will drastically reduce redundancy and improve latent space efficiency. 

A \gls{vqvae} uses the same structure as a classical \gls{ae}, with an encoder, decoder, and latent representation $\z$, as illustrated in \fref{fig: GM schema_vqvae}.  

Formally, the input data $\x$ is processed by the encoder and mapped to the continuous latent tensor $\z = E(\x)$. This latent tensor is composed of many vectors $\z_k$ of a predefined dimensionality $C$.  Each $\z_k$ is then quantized using a learnable codebook $\C = \{\e_j\}_{j=1}^{J}$, composed of $J$ vectors (codewords) $\e_j$. The vector quantization consist of selecting the codeword that is closer to the latent vector $\z_k$ according to the standard minimum \gls{l2} distance quantization rule: 
\begin{equation}
    \z_{q_k} = \e_i \quad \text{where} \quad i = \text{argmin}_{j\in{1,...,J}} \|\z_k - \e_j\|_2,
    \label{eq: GM vq-vae quantization}
\end{equation}
where $\z_{q_k}$ is the vector-quantized version of $\z_k$ and $\e_j$ is the $j$-th codeword in the codebook $\C$. The quantized vectors $\z_q^k$ are then reformatted into the quantized latent tensor $\z_q$. This tensor is then input to the decoder to produce the reconstructed image $\hat{\x} = D(\z_q)$.

%The generation is simple, but the problem is once again in the backpropagation. To find the code $\e_i$ that best approximates a given $\z_k$ in \eref{eq: GM vq-vae quantization} the argmin was used. Unfortunately the argmin is a non-continuous function that destroys the gradient. To overcome this issue, it is possible to consider that $\frac{\partial \Loss}{\partial \z} \approx \frac{\partial \Loss}{\partial \z_q}$. If the model has been properly trained the gradient with respect to the continuous latent $\z$ or with respect to the vector quantized $\z_q$ should be similar. By exploiting this fact it is possible to simply copy the gradient from one side of the projection to the other and continue with backpropagation, as if the projection never happened. This is solving the problem of non-differenciability without inficiating too much on the performances.\\
Because of the new learnable codebook $\C$ the loss function defined for \glspl{vae} in \eref{eq: GM vae loss_function_extended} cannot be applied anymore. The new loss function takes the following form:
\begin{align}
    \Loss_{\text{VQ-VAE}} &= \|\x - D_\theta(\z_q)\|_2^2 + \|\text{sg}[\z] - \z_q\|_2^2 + \lambda_{\text{commit}} \|\z - \text{sg}[\z_q]\|_2^2\\
    &= \Loss_{\text{rec}} + \Loss_{\text{vq}} + \lambda_{\text{commit}} \Loss_{\text{commit}}.
    \label{eq: GM vq-vae vq-loss}
\end{align}
where $\lambda_{\text{commit}}$ is a hyperparameter that controls the strength of the commitment loss, and $\text{sg}[\cdot]$ denotes the stop-gradient operator, which prevents gradients from updating the parameters during backpropagation.\\
The reconstruction loss $\Loss_{\text{rec}}$ is exactly the same as in \eref{eq: GM vae loss_function_extended} and is the part responsible that the reconstructed $\hat{\x}$ is close enough to $\x$. \\
The vector quantization loss $\Loss_{\text{vq}}$ is responsible for the training of the codebook. This term is used to allow the codebook to learn meaningful codewords. Thanks to $\text{sg}[\z]$, the gradient is prevented from flowing through the encoder, so only the codewords are updated.
The last term, the commitment loss $\Loss_{\text{commit}}$, acts as a regularizer to avoid overfitting. Since $\z$ has to be learned, it would be helpful if the encoder learned to produce latent vectors $\z_k$ close to the elements of the codebook. The commitment loss is the term that forces this process by training the encoder to output vectors $\z_k$ close to the codewords in the codebook. At the same time the elements of the codebook are prevented to be trained by the use of $\text{sg}[\z_q]$. This effect is then scaled by a factor $\lambda_{\text{commit}}$. If $\lambda_{\text{commit}}$ increases, the encoder will generate only latent vectors $\z_k$ that are very close to the elements of the codebook, at the expense of reconstruction performance. Conversely, if $\lambda_{\text{commit}}$ decreases, this will cause the codebook to learn less meaningful codewords. A value too high or too low will negatively impact reconstruction performance.

\begin{figure}[!h]
    \centering
    \includegraphics[width=0.8\textwidth]{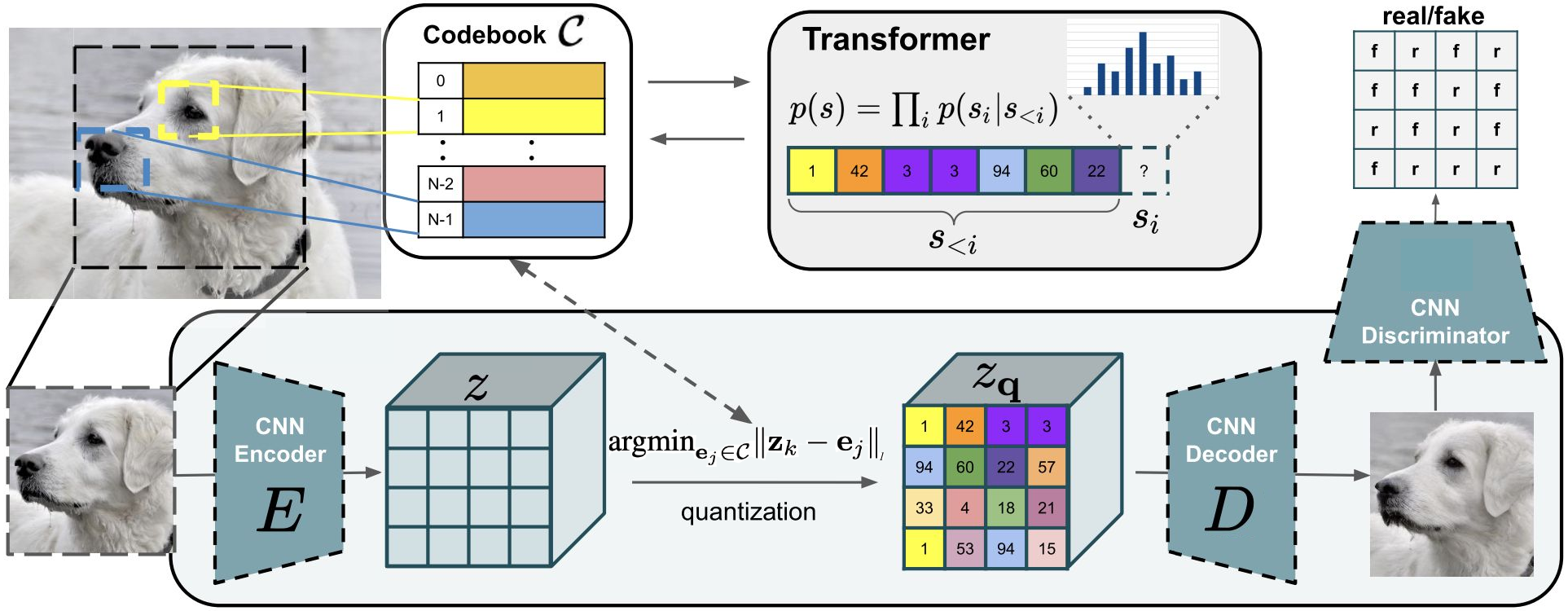}
    \caption[\acrshort{vqgan} architecture scheme]{Overview of the \acrshort{vqgan} architecture as proposed in \cite{Esser2O21Taming}. Similar to the \acrshort{vqvae} the latent vectors are vector-quantized and used in the decoding process. However, only the first $N$ codes from the top-left corner to the bottom-right are used while the other $K-N$ are predicted with the GPT2-based Transformer \cite{Radford2019GPT2}.}
    \label{fig: GM schema_vqgan}
\end{figure}

An advanced version of the \gls{vqvae} is represented by the \gls{vqgan}. These models combine the advantages of the characteristic adversarial training of the \gls{gan} with the vector quantization of the \gls{vqvae}. Popularized by Esser et al. in \cite{Esser2O21Taming}, the idea behind a \gls{vqgan} is the same as the \gls{vqvae} but with the introduction of the discriminator network. 

The loss function for this new architecture is a fusion between the loss function of \glspl{vqvae} and that of \glspl{gan}. It can be written as follows:
\begin{equation}
    \Loss_{\text{VQ-GAN}} = \lambda_{\text{GAN}} \Loss_{\text{GAN}} + \lambda_{\text{rec}} \Loss_{\text{rec}}  + \lambda_{\text{vq}} \Loss_{\text{vq}} + \lambda_{\text{commit}} \Loss_{\text{commit}},
\label{eq: GM vq-gan total_loss}
\end{equation}
where this new loss function is composed  of all the \gls{vqvae} loss terms plus the $\Loss_{\text{GAN}}$ introduced in \eref{eq: GM gan objective_minmax}. Here $\lambda_{\text{GAN}}$, $\lambda_{\text{rec}}$, $\lambda_{\text{vq}}$, and $\lambda_{\text{commit}}$ are the re-scaling factors responsible for  shifting the relative importance of each term.\\

An important observation introduced in \cite{Esser2O21Taming} is that there is no need to use all the vector-quantized $\z_q^k$ for reconstruction. In fact, these vectors are not statistically independent and knowing some of them can be enough to estimate the other. 

As shown in \fref{fig: GM schema_vqgan} the idea is to use only a smaller portion of the vector-quantized elements and predict the other with a GPT2-based transformer  architecture \cite{Radford2019GPT2}.

In \cite{Esser2O21Taming} it was proposed to select a fixed number $N$ of codewords. The selection rule was purely geometrical, obtained by sorting the vector-quantized $\z_{q_k}$ from the one located in the top-left corner to the one located in the bottom-right. Based on this order only the first $N$ vector-quantized elements were selected and the respective codeword index $e_j$ used for the next step that involved the use of a GPT2 transformer model as depicted in \fref{fig: GM schema_vqgan}. 

This model is used to predict the missing $K-N$ codewords based on the first ordered $N$. The performances of this new architecture are very impressing, by only selecting $70\%$ of the total $K$ vector-quantized elements is possible to obtain very good-looking reconstructed images.\\

This new idea is very interesting, but unfortunately it does not consider a fundamental aspect. Not all the vector-quantized vectors $\z_{q_k}$ have the same relevance. Ordering them via a geometrical pattern and taking the first $N$ is not necessarily the best option.

This is reminiscent of the standard approach in image coding, where only the most significant coefficients after a unitary transform (e.g., wavelet, discrete-cosine) are effectively quantized, while the "weak" coefficients are discarded. 
This issue was addressed by the introduction of the \gls{maskvqvae}.

\subsection{Masked VQ-VAE}\label{sec: GM mqvae}
\begin{figure}[!h]
    \centering
    \includegraphics[width=0.7\textwidth]{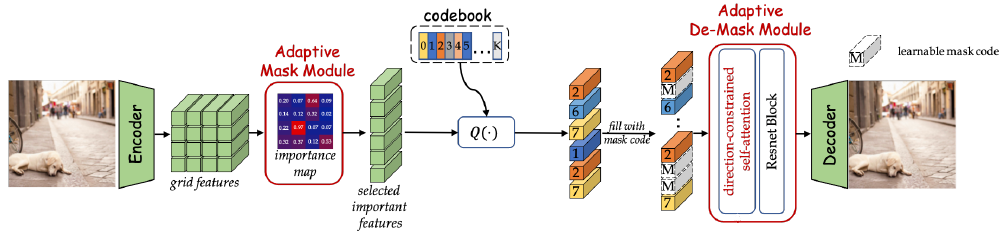}
    \caption[\acrshort{maskvqvae} architecture scheme]{Overview of the \acrshort{maskvqvae} architecture as proposed in \cite{Huang2023MaskedVQ-VAE}. Differently form the \acrshort{vqgan} now the introduction of the \acrlong{amm} select the relevant latent vectors $\z_k$ in order of importance instead that in order of appearance. The \acrlong{adm} is used to reconstruct the missing elements.}
    \label{fig: GM schema_masked_vqvae}
\end{figure}
To overcome the issue of selecting the vectors in order of relevance and not appearance Huang et al. proposed in \cite{Huang2023MaskedVQ-VAE} the \gls{maskvqvae}, depicted in \fref{fig: GM schema_masked_vqvae}. With this \gls{vqvae}-based architecture it is possible to select only the elements that are more relevant to the final goal.

The core idea behind the \gls{maskvqvae} is to use a 2-layer \gls{cnn} called \textit{\gls{amm}} which evaluates an importance score for different regions in the latent tensor $\z$ and selectively quantizes only those regions that are most critical for reconstruction. This selection is performed by evaluating for every latent vectors $\z_k$ a score $s_k \in  [0,1]$ referred to as \textit{relevance score}.  Only the $N$ latent vectors with the highest value will be selected. 

One problem with this approach is that the selection of the highest score introduces the non-differenciability of the sorting operation. For this reason the backpropagation cannot flow in the \gls{amm} and train the weights. This problem can be solved by multiply the normalized features vectors by the respective relevance score as:
\begin{equation}
    \z_k' = \text{LayerNorm}(\z_k) \times s_k .
    \label{eq: GM vq score times z}
\end{equation}
In doing so, the gradient can now flow in the scoring network and update the weights.

At this point the vector quantization is performed only of the $N$ most relevant re-scaled latent vectors $\z_k'$ via the use of a codebook $\C$ similarly to the classical \gls{vqvae}.

The last step before the use of the decoder is to reorganize these $N$ relevant vector-quantized elements in a tensor shape. Unfortunately, since not all the elements have been selected, some of them are empty. To solve this issue, Huang et al.  introduced a learnable placeholder codeword $\bM$ and the \gls{adm}. This additional learnable codeword has the same dimension of every other codeword but is only available at the receiver and placed in all the non-relevant positions. Since it has never seen the input data it conveys no information about the image $\x$. By using the codeword $\bM$ directly in the decoder this will interfere with the relevant vectors that contains a lot of information about the data $\x$. To avoid this the \gls{adm} block is introduced.

It is based on the usage of $L$ consecutive \textit{direction-constrained self-attention}, a modified self-attention mechanism that encourages the information flow from relevant quantized vectors to the non-relevant one while blocking the reverse. This mechanism is used to gradually add relevant information to the placeholder codeword $\bM$. 

The new attention score of every direction-constrained self-attention is structured as follows:
\begin{equation}
    \text{A}_{\text{score}} = \text{softmax}\left(\frac{\mathbf{Q} \mathbf{K}^T}{\sqrt{d_k}}\right) \odot \bB_l,
    \label{eq: GM vq adaptive_demask}
\end{equation}
where $\mathbf{Q}$ and $\mathbf{K}$ are the query and key as discussed in \sref{sec: GM attention}, and  $\bB_l$ is the matrix responsible for the desired properties of the information flow. 

Starting from the first of the $L$ blocks, the matrix $\bB_1$ is designed as follows:
\begin{equation}
    \bB_1 = 
    \m + (\mathbf{1}-\m)\times 0.02,
\end{equation}
where $\m$ is the masking matrix, which is $1$ if the element is relevant and $0$ otherwise. By defining $\bB_1$ in this way, when the codeword $\bM$ has no information about the original data, its influence on the relevant elements will be minimal. After every iteration of the direction-constrained self-attention the matrix $\bB_l$ is updated as $\bB_{l+1} = \sqrt{\bB_{l}}$ slowly increasing the flow of information in both directions as the non-relevant elements gain more information from the relevant elements. 

The result is a tensor $\hat{\z}$ that can be used by the decoder network to produce the final reconstructed image $\hat{\x}= D(\hat{\z})$.

\section{Denoising Diffusion Probabilistic Models (DDPM)}\label{sec: GM ddpm}
\begin{figure}[!h]
    \centering
    \includegraphics[width=0.7\textwidth]{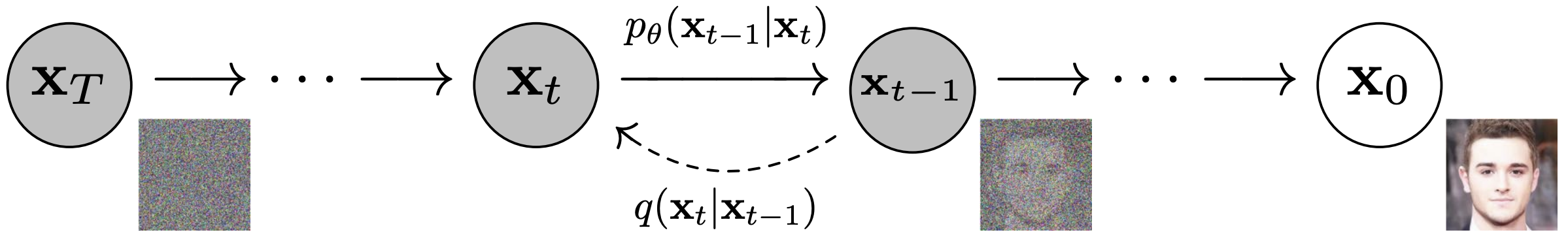}
    \caption[\acrshort{ddpm} visual interpretation of the forward and reverse process]{Visual representation of the \acrshort{ddpm} forward and reverse process \cite{Ho2020ddpm}. The forward process (from right to left) identified by $q(\x_t|\x_{t-1})$ is implemented by gradually adding Gaussian white noise to the original image. Starting from the $x_0$ at the end of $T$ iterations the result is a complete white noise signal $\x_T$. The reverse process (from left to right) uses a trained neural network, identified by $p_\theta(\x_{t-1}|\x_t)$, to iteratively denoise the image. From $\x_t$ it is possible to predict the less noisy version at step $\x_{t-1}$. After $T$ step the original image $\x_0$ can be reconstructed.}
    \label{fig: GM schema_ddpm_theory}
\end{figure}

The last family of generative models discussed in this chapter is the \glspl{ddpm}. Initially proposed by Ho et al. in \cite{Ho2020ddpm}, \glspl{ddpm} are designed to generate data from initial white noise, similar to the Vanilla \gls{vae} discussed in \sref{sec: GM vae}. However, unlike \glspl{vae}, \glspl{ddpm} implement this transformation over multiple iterations, gradually transitioning from the latent variable $\z=\x_T$ to the data $\x_0$.

The core idea behind \glspl{ddpm} is to approach the challenging task of transforming white noise into data samples by breaking it down into a sequence of simpler, gradual steps. This gradual approach is illustrated in \fref{fig: GM schema_ddpm_theory}. Instead of directly mapping a latent variable $\x_T$ to the data $\x_0$, \glspl{ddpm} learn to shift the distribution incrementally, transforming the latent vector into an image over several iterations.

In \glspl{ddpm}, the data generation process involves two key parts:

\begin{itemize}[label={}]
    \item {\textbf{Forward Process} $q(\x_t | \x_{t-1})$:} A fixed Markov chain that progressively adds Gaussian noise to the data, moving from $\x_0$ to $\x_T$. This is equivalent to the distribution $q(\z | \x)$ in the \gls{vae}, responsible for the transformation from the data to the latent space.
    
    \item {\textbf{Reverse Process} $q(\x_{t-1} | \x_t)$:} The process that progressively denoises the data, moving from $\x_T$ back to $\x_0$. This is equivalent to the distribution $q(\x | \z)$ in the \gls{vae}, responsible for the transformation from the latent space to the data.
\end{itemize}

The forward process is only used in the training phase and provides the intermediate steps to train the model. The Reverse process is instead the process that allows the generation of the data samples.

Formally, by considering an intermediate representation of the data $\x_t$ at time-step $t$, the  forward process is defined as:

\begin{equation} 
    q(\x_t | \x_{t-1}) = \N(\x_t; \sqrt{1-\beta_t} \x_{t-1}, \beta_t \mathbf{I}), 
    \label{eq: GM ddpm forward_process} 
\end{equation}
where $\beta_t$ is a variance schedule controlling the amount of noise added at each step. Typically, $\beta_t$ increases over time, ensuring that $\x_T$ approaches an isotropic Gaussian distribution.

To avoid the computational inefficiency of sequentially computing all intermediate $\x_t$, it is possible use the reparametrization trick, as for the \gls{vae}, to sample $\x_t$ directly from $\x_0$.  Using the property of Gaussian distributions the intermediate $\x_t$ can be sampled as:

\begin{equation} 
    \x_t = \sqrt{\bar{\alpha}_t} \x_0 + \sqrt{1 - \bar{\alpha}_t} \bepsilon_0, 
    \label{eq: GM ddpm forward_process_reparametrization} 
\end{equation}

where $\alpha_t = 1 - \beta_t$, $\bar{\alpha}_t = \prod_{s=1}^t \alpha_s$, and $\bepsilon_0 \sim \N(0, \mathbf{I})$. This reparameterization allows for an efficient computation of $\x_t$ at any time step $t$ without iterating through all previous steps.\\

However, the strength of the \glspl{ddpm} lies in inverting the forward process, in going from $\x_T$ back to the original $\x_0$ in what is referred to as the \textit{reverse process}.

In the reverse process, the image is reconstructed gradually removing the noise. The true unknown reverse process can be described by the distribution $q(\x_{t-1} | \x_t)$ that unfortunately is too complicated, and there is no way to know it exactly. Fortunately, Sohl-Dickste et al. in \cite{Sohl-Dickstein2015ddpm_x_0} came up with an idea. Instead of working with $q(\x_{t-1} | \x_t)$, they introduced $q(\x_{t-1} | \x_t, \x_0)$, the posterior distribution of the forward process obtained via Bayes' theorem as follows:
\begin{equation}
    q(\x_{t-1} | \x_t, \x_0) = \frac{q(\x_t | \x_{t-1})q(\x_{t-1} | \x_0)}{q(\x_t | \x_0)}.
\end{equation}
By doing so, it is possible to work with a simpler distribution defined as:
\begin{equation}
    q(\x_{t-1} | \x_t, \x_0) = \N(\x_{t-1}; \bmu_q(\x_t, \x_0), \bSigma_q(t)),
    \label{eq: GM ddpm reverse_process_conditioned_x_0}
\end{equation}
with:
\begin{align}
    \bmu_q(\x_t, \x_0) &= \frac{(1-\bar{\alpha}_{t-1})\sqrt{\alpha_t}}{1-\bar{\alpha}_t}\x_t + \frac{\beta_t \sqrt{\bar{\alpha}_{t-1}}}{1-\bar{\alpha}_t}\x_0 \label{eq: GM ddpm mean_ddpm}\\
    \bSigma_q(t) &= \frac{1-\bar{\alpha}_{t-1}}{1-\bar{\alpha}_t}\beta_t \I = \tilde{\beta}_t \I, \label{eq: GM ddpm var_ddpm}
\end{align}
where the mean $\bmu_q(\x_t, \x_0)$ is a function of $\x_t$, $\x_0$, and $t$, while $\bSigma_q(t)$ is a deterministic function of $t$. The subscript $_q$ is used to identify the true mean and variance of the distribution $q(\x_{t-1} | \x_t, \x_0)$ that are completely described as long as $\x_0$ and $t$ are known.

The drawback of \eref{eq: GM ddpm mean_ddpm} is that in practice, there is no direct way to access $\x_0$. Of course, during training, the original data is known, but the idea is to generate $\x_0$ from $\x_T$, not the opposite. 

For this reason the distribution $q(\x_{t-1} | \x_t, \x_0)$ can be approximated with another parametric distribution $p_\theta(\x_{t-1} | \x_t)$ that can be learned by a \gls{dnn}. In doing so, \eref{eq: GM ddpm reverse_process_conditioned_x_0} can be approximated as:
\begin{equation}
    p_\theta(\x_{t-1} | \x_t) = \N(\x_{t-1}; \bmu_\theta(\x_t), \bSigma_\theta(t)),
    \label{eq: GM ddpm reverse_process_approximated}
\end{equation}
where $\bmu_\theta(\x_t)$ and $\bSigma_\theta(t)$ are the mean and variance predicted by the \gls{dnn}\footnote{It will be used a compact notation where, whenever a model takes $\x_t$ as input, it is implied that the input also includes $t$. This means that $\bmu_\theta(\x_t)$ is the compact version of $\bmu_\theta(\x_t, t)$, while $\bSigma_\theta(t)$ takes only the time-step $t$ as input.}.

For sake of simplicity in the following discussions will be considered the approach when only the mean is learned and $\bSigma_\theta(t)= \tilde{\beta}_t \I$. \footnote{This is only to simplify the explanation. In reality the future implementation presented in this work will consider the covariance matrix as a learnable part of the model following the work introduced in \cite{Nichol2021Improved_DDPM}.}

Theoretically, what is needed now is to design a model $\bmu_\theta$ that can approximate the true mean $\bmu_q$. In practice, by using \eref{eq: GM ddpm mean_ddpm} the training process can be simplified even further. In fact, the true mean $\bmu_q(\x_t, \x_0)$ is expressed as a linear combination of $\x_t$ and $\x_0$. This means that at every step $t$, it is possible to estimate $\x_0$ from $\x_t$ and use this prediction to estimate the mean. In doing so it is possible to evaluate the error directly on the predicted and not on the mean.

To this scope, a new network $\hat{\x}_\theta$ called \textit{denoiser}, can be introduced. This network is responsible for predicting the initial $\x_0$ given any $\x_t$ and $t$. From this prediction the new approximated mean can be evaluated as follows:
\begin{equation}
    \bmu_\theta(\x_t) = \frac{(1-\bar{\alpha}_{t-1})\sqrt{\alpha_t}}{1-\bar{\alpha}_t}\x_t + \frac{\beta_t \sqrt{\bar{\alpha}_{t-1}}}{1-\bar{\alpha}_t}\hat{\x}_\theta(\x_t).
\label{eq: GM ddpm mu_theta_x_theta}
\end{equation}

Another similar but conceptually different approach is based on the relationship between $\x_0$ and the noise $\bepsilon_0$ introduced after the reparametrization trick in \eref{eq: GM ddpm forward_process_reparametrization}. It is in fact possible to rearrange the terms and express $\x_0$ as a function of the added noise $\bepsilon_0$ and $\x_t$. This noise can then be predicted by a network $\hat{\bepsilon}_\theta$  and by substituting it back into \eref{eq: GM ddpm mean_ddpm}. In this way the mean $\bmu_\theta$ can be also estimated as:
\begin{equation}
    \bmu_\theta(\x_t) = \dots = \frac{1}{\sqrt{\alpha_t}}\x_t - \frac{1-\alpha_t}{\sqrt{1-\bar{\alpha}_t}\sqrt{\alpha_t}}\hat{\bepsilon}_\theta(\x_t).
    \label{eq: GM ddpm mu_theta_epsilon_theta}
\end{equation}

Both these two approaches, using $\x_0$ or $\bepsilon_0$, allow to estimate the mean $\bmu_\theta$ in two complete different ways. For this reason it is important to clarify which are the differences and advantages between these two methods.
\begin{enumerate}
    \item \textbf{Direct prediction:} The idea is to use a model $\hat{\x}_\theta$ to directly predict the original $\x_0$. The model takes as input the noisy $\x_t$, along with $t$, and generates an estimate $\hat{\x}_\theta(\x_t)$ of the original data-point from which $\x_t$ was obtained. The loss to be minimized in this approach is the following:
    \begin{equation}
        \Loss(\theta) =  \sum_{t=1}^{T} \mathbb{E}_{q(\x_t|\x_0)} \left[ \frac{1}{2 \tilde{\beta}_t} \frac{(1-\alpha_t)^2 \bar{\alpha}_{t-1}}{(1-\bar{\alpha}_t)^2} \left\|\hat{\x}_\theta(\x_t) - \x_0 \right\|^2\right].
        \label{eq: GM ddpm loss_x_theta}
    \end{equation}
    One advantage of this method is its simplicity. The model directly estimates the target mage, making the approach intuitive and straightforward to implement. Additionally, the model is optimized for the final objective in an end-to-end manner, which can sometimes lead to superior performance. However, directly predicting $\x_0$ can be more challenging, especially when dealing with complex distributions, potentially causing difficulties in optimization and stability during training.
    
    \item \textbf{Noise prediction:} Alternatively, starting from $\x_t$ and $t$, the model $\hat{\bepsilon}_\theta$ can be used to predict the noise $\bepsilon_0$ that was added to the original $\x_0$ to obtain $\x_t$. The loss function associated to this version is the following:
    \begin{equation}
        \Loss(\theta) =  \sum_{t=1}^{T} \mathbb{E}_{q(\x_t|\x_0)} \left[ \frac{1}{2 \tilde{\beta}_t} \frac{(1-\alpha_t)^2 \bar{\alpha}_{t-1}}{(1-\bar{\alpha}_t)^2}  \left\|\hat{\bepsilon}_\theta(\x_t) - \bepsilon_0 \right\|^2\right].
        \label{eq: GM ddpm loss_epsilon_theta}
    \end{equation}
    This approach has the advantage of making the optimization process easier, as predicting the noise $\bepsilon_0$ is often simpler than directly predicting $\x_0$. This is particularly true in high-dimensional spaces when this approach is leading to better convergence properties. Additionally, this method is more flexible and can generalize better across different types of data and tasks, making it applicable in a wider range of scenarios. On the downside, the model is not directly optimized for the final objective of reconstructing $\x_0$ that has to be obtained by rearranging \eref{eq: GM ddpm forward_process_reparametrization}. This might result in suboptimal performance, like color shifting in certain cases.
\end{enumerate}

The choice between the two approaches depends on the specific task and the characteristics of the data. In general, the noise prediction method is more robust and easier to optimize, making it a popular choice in practice. However, the direct prediction approach can sometimes lead to better performance, especially when dealing with simpler data distributions.

In both cases, it is good practice to use a \gls{unet} architecture in designing the two models $\hat{\x}_\theta$ or $\hat{\bepsilon}_\theta$. The use of a \gls{unet} is not mandatory, but it is a common practice in the field since it has shown to be very effective in capturing the complex distribution of the data. Moreover, as discussed in \sref{sec: GM unet}, due to the time dependency of the $\x_t$, the \gls{unet} can be easily adapted to the task by employing the modified version of the \gls{resblock} depicted in \fref{fig: GM resblock with time}.
\subsection{Training and Inference}\label{sec: GM ddpm training_inference}
Before introducing some more advanced versions of \gls{ddpm}, it is useful to understand how the training and inference are performed, as it is not immediately intuitive.

As already discussed, both $\hat{\x}_\theta$ or $\hat{\bepsilon}_\theta$ are able to obtain an estimate of $\x_0$ in one single step only by knowing $\x_t$ and $t$.
However, this might cause some confusion since the whole purpose of the \gls{ddpm} was to use a multistep approach to gradually reconstruct $\x_0$.  Instead, now both $\hat{\x}_\theta$ and $\hat{\bepsilon}_\theta$ can perform this action in one single step. 

The fact is that yes, it is possible to perform the reverse process in only one step, but it is not recommended. In fact, performances are drastically influenced by the number of steps used in the transition. To clarify this better, it is possible to start by discussing the training process and then introduce the inference.

The following discussion will cover a single training and inference iteration for only one image $\x_0$. To perform the overall training and inference process, these single iteration steps should be repeated for all the images in the dataset for every epoch.
\begin{itemize}[label={}]
    \item{\textbf{Training}:} The training starts with the original image $\x_0$. Thanks to the forward process and the reparametrization trick in \eref{eq: GM ddpm forward_process_reparametrization}, it is possible to evaluate a random noisy version $\x_t$. This evaluation is done by selecting a random time-step $t$ and evaluating the sequence of $\beta_t$. The forward step can now be considered over and at this point the network $\hat{\x}_\theta$ is used to predict an estimate of $\x_0$. During training this is performed in one single step, allowing the reconstruction of the original $\x_0$ and concluding the reverse process. By knowing the original $\x_0$, the loss function in \eref{eq: GM ddpm loss_x_theta} can be used to train the model via backpropagation. \footnote{For simplicity, the direct prediction approach with $\hat{\x}_\theta$ has been considered, but the process is the same even when considering $\hat{\bepsilon}_\theta$.} The idea of every single training-step is to expose the network $\hat{\x}_\theta$ to as many $\x_t$ as possible, letting it learn the effect of $t$ on the original $\x_0$.
    \item{\textbf{Inference}:} While the training is straightforward, the inference involves more parts. Unfortunately, the same approach used for training cannot be applied here. A direct prediction of $\x_0$ from an initial white random noise $\x_T$ will generate poor results. To reach $\x_0$, it is important to move gradually along the sequence of $\x_T, \x_{T-1}, \dots, \x_2, \x_1$ by sampling at every step from the distribution in \eref{eq: GM ddpm reverse_process_approximated}.

    The inference process can be summarized as follows:
    \begin{itemize}
        \item From the initial $\x_T$, evaluate $\hat{\x}_0 = \hat{\x_\theta}(\x_T)$. This is the first estimate and is generally not good enough.
        \item From $\hat{\x}_0$ and $\x_T$, sample from the distribution $p_\theta(\x_{t-1} | \x_t) = \N(\x_{t-1}; \bmu_\theta(\x_t), \bSigma_\theta(t))$ the data-point $\x_{T-1}$ as follows:
        \begin{equation}
            \x_{T-1} = \frac{(1-\bar{\alpha}_{t-1})\sqrt{\alpha_t}}{1-\bar{\alpha}_t}\x_t + \frac{\beta_t \sqrt{\bar{\alpha}_{t-1}}}{1-\bar{\alpha}_t}\hat{\x}_\theta(\x_t) + \sqrt{\tilde{\beta}_t} \n,
        \end{equation}
        with $\n \sim \N(0, \I)$ if $t\geq2$ else $\n=0$.
        \item Repeat this iterative process from the new $\x_{T-1}$ to generate $\x_{T-2}$ and over again until the final $\hat{\x}_0$ is obtained.
    \end{itemize}
    In this way, by going back and forth between $\x_t$ and an estimate of $\x_0$ and then back to better denoised version $\x_{t-1}$, it is possible to generate a good enough $\x_0$.

    Since in general, for a normal application $T=1000$, repeating the process 1000
    times is not efficient. For this reason in \cite{SongME2021DDIM} was proposed the so-called \gls{ddim}, a variation of the discussed \gls{ddpm}, where $\x_{t-1}$ is generated in a deterministic way rather than probabilistic. In doing so it is possible to skip some of the $T$ steps in the reverse process. With the \gls{ddim} instead of computing $T$ steps, only a small subset is required, speeding up the generation process. In general the subset is composed of a number in between $10$ and $100$ equidistant steps.

\end{itemize}

\subsection{Advanced Techniques}\label{sec: GM ddpm_advanced_techniques}
What has been discussed so far is the basic structure of a \gls{ddpm} where, given some white noise $\x_T$, the model can generate a good enough image $\x_0$. \\
The drawback with this formulation is that there is no control over the output. Suppose that the dataset used to train the model contains multiple images of different classes, like "car" and "people". In inference, when starting from a white noise $\x_T$, the model will generate an image similar to one used for the training process. Unfortunately, there is no way to know beforehand which image will be generated, if a car or a pedestrian, and no way to control the output. This is, a significant limitation of the model as described until now, and in this section this limitation will be addressed.

Before analyzing some different techniques, it is important to clarify that there are only two possible approaches to control the generation process: \textit{conditioning} and \textit{guidance}. These two approaches are often used in combination and can be applied using a different variety of techniques. 
\begin{itemize}[label={}]
    \item{\textbf{Conditioning}:} This is the first way to control the output of a model. This is how the model is told on "what" to consider during generation. In general, control is obtained by incorporation some additional input information to the model, rather than only $\x_t$ and $t$. This can be performed in many ways. For example, in \sref{sec: GM unet} it was discussed how the \gls{spade} technique is used to condition the generation process based on some \gls{ssm} $\s$. In that case, conditioning was enforced in the bottleneck and decoder of the \gls{unet} by replacing the normalization layer with the \gls{spade} layer.  
    Another possible conditioning example is the one mainly used for \gls{sr}, where the \gls{lr} image is up-scaled and concatenated to $\x_t$ at the beginning of the \gls{unet}. In this way, the model considers both the noisy image $\x_t$ and the \gls{lr} input to influence the reconstruction. 
    
    \item{\textbf{Guidance}:} If conditioning tells the model "what" to consider, guidance tells it "how" and "how strongly" the additional information should be enforced. The methods to guide the model generally fall into two broad categories: \textit{Classifier Guidance} and \gls{cfg}. Let's describe them in more detail \footnote{In the following discussion will be considered the use of the noise prediction approach with the network $\hat{\bepsilon}_\theta$. The same concepts can be applied to the direct prediction approach with $\hat{\x}_\theta$.}:
    \begin{itemize}
        \item \textbf{Classifier Guidance}: Introduced by Dhariwal and Nichol in \cite{Dhariwal2021DDPM_beat_GAN}, the first application was proposed to control the output based on a specific class $y$. This technique allows to control the level of enforcement and adjust it at inference time. 
        
        The core concept behind classifier guidance is to use a pre-trained classifier to guide the reverse diffusion process. Specifically, during inference the noise model $\hat{\bepsilon}_\theta$ is modified to steer the predicted noise $\hat{\bepsilon}_\theta(\x_t)$ toward a particular class $y$. This is achieved by combining the gradients of the classifier's output with the predicted noise.

        Formally, let $p_\phi(y|\x_t)$ be a classifier trained to predict the class label $y$ given a noisy input $\x_t$. The classifier guidance technique modifies the predicted noise $\hat{\bepsilon}_\theta(\x_t)$ by incorporating the gradient of the log-probability of the target class $y$ with respect to $\x_t$. This change generates a new estimate defined as:
        \begin{equation}
            \hat{\bepsilon}_\theta'(\x_t|y) = \hat{\bepsilon}_\theta(\x_t) - a \cdot \sigma_t \nabla_{\x_t} \log p_\phi(y|\x_t),
            \label{eq: GM ddpm classifier_guidance}
        \end{equation}
        where $a$ is a scaling factor that controls the strength of the guidance, and $\sigma_t=\sqrt{\tilde{\beta}_t}$ is the standard deviation of the noise at time-step $t$.

        The intuition behind this approach is that the gradient $\nabla_{\x_t} \log p_\phi(y|\x_t)$ points in the direction that increases the likelihood of the image $\x_t$ being classified as class $y$. By subtracting this gradient from the predicted noise $\hat{\bepsilon}_\theta(\x_t)$, the generative process is pushed toward producing images more likely to belong to the desired class.

        This method allows for flexible and controlled generation. If the dataset contains multiple classes like "car" and "pedestrian", classifier guidance can be used to steer the generation process towards producing either the image of a "car" or of a "pedestrian", based on the target class $y$.

        One limitation of classifier guidance is that it requires a pre-trained classifier. This extra model is adding computational overhead to the whole architecture. Additionally, the strength of the guidance parameter $a$ needs to be carefully tuned. If $a$ is too small, the guidance effect might be weak, resulting in generated images that are not strongly aligned with the target class. On the other hand, if $a$ is too large, the generated images might become distorted as the model overfit to the classifier's gradients.

        \item \textbf{Classifier-Free Guidance}: Proposed by Ho and Salimans in \cite{ho2021classifierfree}, the classifier-free guidance eliminates the need for an external classifier, as shown in \fref{fig: GM schema_cfg}.
        \begin{figure}[!t]
            \centering
            \includegraphics[width=0.7\textwidth]{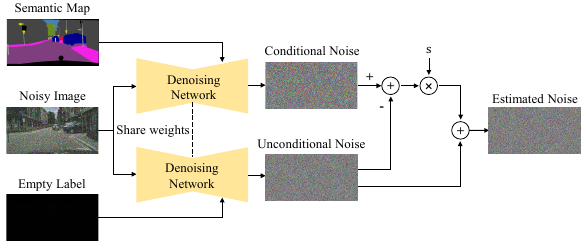}
            \captionsetup{width=.75\linewidth}
            \caption[\acrshort{cfg} graphical scheme]{The control-flow graphical scheme used for the \acrshort{cfg} mechanism as proposed in \cite{ho2021classifierfree}. The same \acrshort{unet} model is trained with and without conditioning to allow control over the output.}
            \label{fig: GM schema_cfg}
        \end{figure}
        Formally, let $\hat{\bepsilon}_\theta(\x_t | y)$ represent the noise prediction model conditioned on class $y$ and let $\hat{\bepsilon}_\theta(\x_t|\emptyset)$ represent the the same model without the conditioning. \footnote{The conditioning is always present but it is set to zero.} During training, the network learns to predict the noise both conditionally and unconditionally. This dual training approach allows the model to learn how to generate data in a way that is either influenced by a specific class label or entirely unconstrained by class information.

        During inference, the generation process can be controlled by adjusting the amount of conditioning applied to the noise prediction. Specifically, the noise prediction for a given time-step $t$ can be guided by the class label $y$ using the following equation:

        \begin{equation}
        \hat{\bepsilon}_\theta'(\x_t | y) = \hat{\bepsilon}_\theta(\x_t|\emptyset) + a \cdot \left(\hat{\bepsilon}_\theta(\x_t | y) - \hat{\bepsilon}_\theta(\x_t|\emptyset)\right),
        \label{eq: GM ddpm classifier_free_guidance_inference}
        \end{equation}
        where $\hat{\bepsilon}_\theta'(\x_t | y)$ is the modified noise prediction influenced by the conditioning, depending on the scaling factor.
        
        Essentially, thanks to the term $a \cdot(\hat{\bepsilon}_\theta(\x_t | y) - \hat{\bepsilon}_\theta(\x_t|\emptyset))$ it is possible to enforce more or less conditioning. This is replacing the gradient of the classifier in the classifier guidance in \eref{eq: GM ddpm classifier_guidance}. \\
        By doing so, some of the pitfalls associated with classifier guidance are eliminated. In fact potential biases introduced by the classifier and the need for a separate pre-trained model are removed. However, like classifier guidance, the choice of the guidance scale $a$ is still critical. Too small value of  may lead to weak conditioning, while too large values of $a$ might result in less natural or overly biased outputs.
    \end{itemize}
\end{itemize}

\section{Evaluation Metrics}\label{sec: GM evaluation metrics}

After a comprehensive introduction to the fundamental concepts in the field of generative models, it is important to introduce the evaluation metrics and losses that will be employed throughout this work. In fact, the choice of appropriate metrics and losses is crucial when designing, training and assessing the performances of a model.

This section is divided into two main subsections. The first subsection focuses on classic evaluation metrics commonly used in image compression and reconstruction tasks. These metrics primarily measure the quality of the reconstructed image $\hat{\x}$ in terms of pixel-wise differences compared to the original image $\x$. These metrics are agnostic of the overall semantic information contained in the image itself. The second subsection introduces \textit{semantic-relevant metrics}, which are designed to evaluate the semantic information preserved in the reconstructed image. These metrics are particularly important in a \gls{sc} framework as the one introduced in \sref{sec: SEMCOM sem}.

In the following discussion will be used the following notation:
\begin{itemize}
    \item the original image $\x$ and the reconstructed  $\hat{\x}$ have both a  shape of   $3 \times H \times W$ pixels.
    \item the original \gls{ssm} $\s$ and the reconstructed  $\hat{\s}$ have both a shape of  $n_c \times H \times W$, where $n_c$ is the number of semantic classes.
    \item The subscript $_{(h,w)}$ refers to the values of the vector located at coordinates $(h,w)$. For example in the case of $\x_{(h,w)}$ this refers to the vector with the RGB values of the pixel, while $\s_{(h,w)}$ refers to the one-hot encoded vector with $n_c$ components.
    \item $\s_i$ and $\hat{\s}_i$ represent the "slice" associated to the $i$-th semantic class of $\s$ and $\hat{\s}$, respectively. This means that $\s_i$ and $\hat{\s}_i$ can be considered as binary masks of shape $H \times W$ with 1 in correspondence of objects belonging to the $i$th the semantic class.
    \item $\phi$ refers to a generic pre-trained model used in the metric evaluation process. $\phi$ is a \gls{sota} model used to extract the latent features from the images $\x$ and $\hat{\x}$. The specific model used will be clarified in detail depending on the metric. 
    \item the arrows ($\uparrow$) and ($\downarrow$) indicate if the best performance is obtained with higher or lower values of the metric, respectively.
\end{itemize}

\subsection{Classic Evaluation Metrics}

In the domain of image compression and reconstruction, several classic metrics are widely used to quantify the quality of reconstructed images. These metrics typically focus on measuring the difference between the original image $\x$ and the reconstructed image $\hat{\x}$ on a pixel-by-pixel basis. The most common classic evaluation metrics used in this work are:

\begin{itemize}[label={}]
    \item {\textbf{Weighted \gls{l2} metric} ($\downarrow$):}
    The weighted \gls{l2} metric is one of the most fundamental metrics used to measure the average squared difference between the original and reconstructed images. Differently from the classic \gls{l2}, this weighted version is used in scenarios where certain regions of the image are more critical than others and it is defined as follows: 

    \begin{equation}
        \text{W}L_2(\mathbf{x}, \hat{\mathbf{x}}) = \frac{1}{H\times W}\sum_{(h,w)} \text{w}_{(h,w)} \left( \x_{(h,w)} - \hat{\x}_{(h,w)} \right)^2 \equiv \mathcal{L}_{\text{W}L_2}(\mathbf{x}, \hat{\mathbf{x}}), 
    \end{equation}

    where $\text{w}_{(h,w)}$ is the weight assigned to the pixel at position $(h, w)$. By adjusting these weights, the metric can prioritize and give more importance to certain parts of the image over others.

    \item {\textbf{Peak Signal-to-Noise Ratio (PSNR)} ($\uparrow$):}
    The \gls{psnr} is a widely used metric for evaluating the quality of reconstructed images, particularly in image compression tasks. It is derived from the \gls{l2} metric and expresses the ratio between the maximum possible value of a signal, the power, and the error in reconstruction, the noise.

    The \gls{psnr} is defined in decibels (dB) as:

    \begin{equation} 
        \text{PSNR}(\mathbf{x}, \hat{\mathbf{x}}) = 20 \cdot \log_{10} \left( \frac{255}{\sqrt{\text{W}L_2(\mathbf{x}, \hat{\mathbf{x}})}} \right), 
    \end{equation}

    where $255$ represent the maximum possible pixel value of the image data represented in 8-bit.

    Higher \gls{psnr} values indicate better image reconstruction quality on a pixel-by-pixel level. A higher \gls{psnr} implies that the ratio of the signal (original image) to the noise (error introduced by compression or reconstruction) is larger, meaning less distortion. 
\end{itemize}

\subsection{Semantic-Relevant Metrics}

In many applications where the preservation of semantic information is fundamental, classic metrics cannot be used. 

The reason is simple and can be explained with one example. Suppose to have an image $\x$ and the reconstructed $\hat{\x}$ is exactly like $\x$ but shifted one pixel on the right. In this case by evaluating the \gls{psnr} between the two images the value will be low, even though the images are basically the same. 

To overcome this limitation it is important to introduce metrics that can evaluate performances on a semantic level rather than a pixel level.

\begin{itemize}[label={}]
    \item {\textbf{Mean Intersection over Union (mIoU)} ($\uparrow$):}
    The  \gls{miou} is a standard metric for evaluating how much two \glspl{ssm} differ from each other. 

    The evaluation is performed by measuring the overlap between $\s$ and $\hat{\s}$, the higher the overlap the closer the two \glspl{ssm} are. 

    For each semantic class $i$, the process starts by evaluating the Intersection over Union (IoU) as:

    \begin{equation} 
        \text{IoU}(\s_i, \hat{\s}_i) = \frac{|\s_i \cap \hat{\s}_i|}{|\s_i \cup \hat{\s}_i|} , 
    \end{equation}
    where $|\s_i \cap \hat{\s}_i|$ represent the number of pixels where both $\s_i$ and $\hat{\s}_i$ have value 1, and $|\s_i \cup \hat{\s}_i|$ is the number of pixels were at leas one of the two has value of 1.

    The \gls{miou} is then computed as the mean over all the semantic classes of the IoU:

    \begin{equation} 
        \text{mIoU}(\s, \hat{\s}) = \frac{1}{n_c} \sum_{i=1}^{n_c} \text{IoU}(\s_i, \hat{\s}_i)
        \label{eq: mIoU_score} 
    \end{equation}

    Higher values of \gls{miou} indicate that the two \glspl{ssm} are closer to each other. 

    \item {\textbf{Weighted Cross-Entropy Loss} ($\downarrow$):}
    The Weighted Cross-Entropy loss is commonly used as a loss function for training classification models, including semantic segmentation networks. It measures the difference between the predicted class probabilities and the ground truth labels, and it is evaluated as follows:

    \begin{equation} 
        \Loss_{\text{WCE}}(\s, \hat{\s}) = - \sum_{(h,w)} \text{w}_{(h,w)} \s_{(h,w)} \log\left( \s_{(h,w)} \odot \hat{\s}_{(h,w)} \right),
    \end{equation}
    where $\text{w}_{(h,w)}$ is the weight associated with the semantic class of the pixel $\s_{(h,w)}$ and $\odot$ represent the element-wise product.

    Minimizing the Weighted Cross-Entropy loss encourages the model to produce \glspl{ssm} that are similar to the true $\s$. In the context of semantic segmentation, it helps the model to correctly classify each pixel into the appropriate semantic class. By increasing or decreasing the weights, the importance of different classes can be adjusted to reflect their relative significance in the task.

    \item {\textbf{Learned Perceptual Image Patch Similarity (LPIPS)} ($\downarrow$):}
    The \gls{lpips} metric \cite{Zhang2018LPIPS} is a metric used to evaluate the perceptual similarity between two images. As for other semantic metrics it is based on the idea that the preservation of fine details and texture matter more than reconstructing the exact pixel values. To extract these information the \gls{lpips} metric is evaluated with the help of a pre-trained VGG-16  model $\phi$ \cite{Simonyan2015VGG}. 

    The two images, $\x$ and $\hat{\x}$, are processed by the model to extract a series of latent representation tensors  at different levels of resolution. The idea is that the pre-trained model $\phi$ is able to capture details that are not captured by pixel-by-pixel comparisons. For example, one of the intermediate representations might have values close to 1 only if in the image is present a green traffic light and close to -1 if the traffic light is red. By evaluating the \gls{l2} distance between these latent representations of $\x$ and $\hat{\x}$ it is possible to capture all these information.

    The \gls{lpips} is defined as:
    \begin{equation}
        \text{LPIPS}(\x, \hat{\x}) = \sum_{l\in L} \frac{1}{H_l W_l} \sum_{(h_j,w_j)} \left\| \text{w}_l \odot  \left( \phi_i(\x) - \phi_i(\hat{\x})  \right) \right\|_2^2
    \end{equation}
    where $\phi(\cdot)$ is the output of the pre-trained model $\phi$ at the $l$-th layer, $L$ is the set of layers used to extract the features, $H_l$ and $W_l$ are the height and width of the $l$-th layer, respectively, and $\text{w}_l$ is a learnable weight associated with the $l$-th layer.

    \gls{lpips} is designed to correlate with human perceptual similarity judgments. Lower \gls{lpips} values indicate that the images are more perceptually similar.

    \item {\textbf{Perceptual Loss} ($\downarrow$):} The concept of Perceptual Loss refers to the use of one of the many perceptual metrics during training. By introducing this term in the loss function, the model is encouraged to generate images that are to the original images in terms of latent representation.

    This is particularly used in the context of generative models. In this thesis as Perceptual Loss it is employed the \gls{lpips} metric as described above. So the perceptual loss is intended as:
    \begin{equation} 
        \mathcal{L}_{\text{perc}} = \text{LPIPS}(\x, \hat{\x})
    \end{equation}

    \item {\textbf{Fréchet Inception Distance (FID)} ($\downarrow$):}
    The \gls{fid} \cite{Heusel2017FID}, differently form the previous semantic metrics is used to evaluate the overall quality of a set of images. It compares the distribution of real images to that of reconstructed images in the feature space of an Inception-v3 model \cite{Szegedy2015Inceptionv3} pre-trained for image classification on the Imagenet dataset \cite{Russakovsky2015ImageNet}.

    Differently from the \gls{lpips}, the features are extracted only from a single layer of the model $\phi$  and the \gls{fid} is evaluated as follows\footnote{More specifically the output is represented by the 1024 elements vector of the \textit{pool3} layer of the Inception-v3}:

    \begin{equation} 
        \text{FID}(\mathbf{X}, \hat{\mathbf{X}}) = \left\| \mu_{\phi}(\mathbf{X}) - \mu_{\phi}(\hat{\mathbf{X}}) \right\|_2^2 + \text{Tr}\left( \Sigma_{\phi}(\mathbf{X}) + \Sigma_{\phi}(\hat{\mathbf{X}}) - 2 \left( \Sigma_{\phi}(\mathbf{X}) \Sigma_{\phi}(\hat{\mathbf{X}}) \right)^{1/2} \right)    
    \end{equation}
    where $\mathbf{X}$ and $\hat{\mathbf{X}}$ are the sets of real and reconstructed images, respectively, $\mu_{\phi}(\mathbf{X})$ and $\mu_{\phi}(\hat{\mathbf{X}})$ are the mean of the features extracted from the Inception-v3 model, and $\Sigma_{\phi}(\mathbf{X})$ and $\Sigma_{\phi}(\hat{\mathbf{X}})$ are the covariance matrices of the features. The structure of the \gls{fid} metric is based on the Fréchet distance, with the assumption that the distributions are Gaussian \cite{DOWSON1982frechet}. 

    Lower \gls{fid} scores indicate that the generated images have feature distributions that are more similar to those of the real images, implying higher quality and diversity in the generated images. However, is not possible to compare directly two images with the \gls{fid} metric, but only sets of images.

    \item {\textbf{Traffic signs classification accuracy (ACC)} ($\uparrow$):}
    The last semantic relevant metric is introduced in this work and designed from scratch to evaluate the semantic quality of an image based on the classification accuracy of traffic signs. To perform the classification of traffic signs it has been chosen to adopt an architecture based on the MobileNetV2 model \cite{Sandler2018MobileNetV2} pre-trained on the ImageNet dataset \cite{Russakovsky2015ImageNet}. This model has been fine-tuned for traffic sign classification using the GTSRB dataset \cite{Stallkamp2011GTSRB}, which is a comprehensive collection of more than 50000 images of traffic signs spanning over 43 distinct classes. The result of the fine-tuning was a model able classify correctly up to 94\% of the traffic signs in validation. 

    With a pre-trained model is now possible to describe the process to evaluate the traffic signs classification accuracy. As a first step, this metric involves the identification of all the traffic signs in the frame. This task is performed by using the \gls{ssm} to identify all the different traffic signs via the two-pass binary connected-component labeling algorithm \cite{Shapiro2001connectedCOmp}.  With this algorithm is possible to create a binary mask $\m_j$ for every traffic sign by using the \gls{ssm} to produce different masks of every one of them. The different masks are then used to locate the traffic signs in the original image $\mathbf{x}$ and the reconstructed image $\hat{\mathbf{x}}$. These traffic signs are referred to as $\ob_j=\x \odot \m_j$ and $\hat{\ob}_j=\hat{\x} \odot \m_j$, respectively. After the identification, the  classification via the fine-tuned MobileNetV2 is performed on every pair $(\ob_j, \hat{\ob}_j)$ to produce the pair of resulting classes $(a_{\ob_j}, a_{\hat{\ob}_j})$. The traffic signs classification accuracy ACC is defined as the percentage of traffic signs that are classified identically in both the original and the reconstructed images:

    \begin{equation}
        \text{ACC} = \frac{1}{n} \sum_{i=1}^n \text{ACC}(\hat{\x}_i) = \frac{1}{n} \sum_{i=1}^n \frac{1}{N_i} \sum_{j=1}^{N_i} 1(a_{\ob_j} = a_{\hat{\ob}_j})
        \label{eq: app Accuracy_score}
    \end{equation}

    where $n$ is the total number of images that contain at least one traffic sign, $N_i$ is the total number of traffic signs detected by the \gls{ssm} in a single image, and $1(\cdot)$ is the indicator function, which equals 1 if the condition inside is true and 0 otherwise.

    The closer the value of ACC is to 1, the more the classification on the reconstructed traffic signs is closer to the classification on the original, thus the images are considered semantically close, with $\text{ACC}=1$ indicating the semantical equivalence $\x \longleftrightarrow \hat{\x}$. 

    This metric is particularly useful and will be adopted in evaluating the semantic preservation quality of the model proposed in the following chapter.
\end{itemize}

% %   CHAPTERS

%   CHAPTERS
\chapter{\textcolor{black}{Semantic-Preserving Image Coding based on Conditional Diffusion Models}}\label{ch: SPIC}

The content of this chapter is entirely based on the following publications:
\begin{quotation}
    \noindent \textit{\textbf{\large Semantic-Preserving Image Coding based on Conditional Diffusion Models}}\\
    \textit{Francesco Pezone, Osman Musa, Giuseppe Caire, Sergio Barbarossa}

    \vspace{0.1cm}
    \noindent \textit{\textbf{\large C-SPIC: Class-Specific Semantic-Preserving Image Coding with Residual Enhancement for Accurate Object Recovery}}\\
    \textit{Francesco Pezone, Osman Musa, Giuseppe Caire, Sergio Barbarossa}
\end{quotation}

\section{Introduction}
After having discussed the overall idea behind the concept of \gls{sc} and some of the most important generative architectures, it is possible to merge these two worlds. In this chapter the challenge of high-quality image reconstruction is addressed. This is performed via the development of a framework based on \glspl{ddpm} that can serve as a viable alternative to classical image compression algorithms. \\

Before continuing any further it is important to define what constitutes the semantically relevant information in this context. The assumption that will be made in this chapter is that the semantic relevant information consists of the following elements: (i) the lossless reconstruction of the \gls{ssm}; (ii) the reconstructed image should retain as much of the \gls{ssm} as possible; (iii) the overall characteristics of the objects such as colors and details should be preserved; and (iv), if required, certain semantic classes will be considered semantically relevant information, necessitating their high-quality reconstruction regardless of the quality of the rest of the image.

According to these assumptions, this chapter introduces the \gls{spic} framework, a modular semantic image communication scheme based on the generating power of \glspl{ddpm}. The \gls{spic} framework consist of a transmitter and receiver side. At the transmitter the image $\x$ is processed to generate the \gls{ssm} $\s$ and a low-quality (coarse) version $\co$ before sending them to the receiver. The \gls{ssm} is send losslessly while the coarse lossy. At the receiver they are used as input of the proposed  \gls{semcore} to reconstruct the full-resolution image $\hat{\x}$ similar to $\x$.\\

Prominent recent models have exploited the use of the  \gls{ssm} to guide the image reconstruction process. For example, Isola et al. \cite{isola2017image2image} introduced the pix2pix model, a conditional \gls{gan} that uses the \gls{ssm} as input to generate an image that is able to preserve the \gls{ssm} content.

Wang et al. \cite{Wang2018HighRes} addressed the challenge of image reconstruction by proposing a more efficient variant of a pix2pix model. Instead of generating the image solely from the \gls{ssm}, they introduced a multi-step approach that integrates features extracted from the original image. This method was able to achieve incredible reconstruction capabilities to produce an image $\hat{\x}$ more similar to $\x$ in term of colour and texture, while preserving the same level of \gls{ssm} as for the pix2pix model.

Recently, \gls{ddpm} \cite{Ho2020ddpm} have demonstrated remarkable results in image synthesis \cite{Dhariwal2021DDPM_beat_GAN, Nichol2021Improved_DDPM}. Building upon the works of Isola et al. and Wang et al., in \cite{Wang2022SISDM} it was introduced a \gls{ddpm} exploiting the conditioning on the \gls{ssm}. This conditioning was enforced by the extensive use of the \gls{spade} normalization layer \cite{Park2019SPADE}. While being able to produce images with a high level of \gls{ssm} preservation, even this model presents a limitation on other aspects of reconstruction.

A common limitation to the cited models is the oversight of the original image. By considering only the \gls{ssm} as input, the resulting image may be visually appealing but every time an image is produced it will look differently in colors and textures, even if the \gls{ssm} is the same. The sole constraint of these algorithms is the preservation of the \gls{ssm}, which is not enough to preserve all the other semantically relevant information discussed before.

Classical image compression algorithms, such as \gls{bpg} \cite{Bellard2017BPG} and \gls{jpeg2000}, approach the problem from a different perspective. These algorithms are designed to reduce the file size of images by exploiting redundancies in the data, employing techniques like discrete cosine transforms and wavelet transforms. While they are effective at maintaining a certain level of visual quality at lower bit rates, they primarily focus on optimizing classical metrics like \gls{psnr}  rather than the semantic content. Consequently, they may indiscriminately discard information that is crucial for semantic interpretation, such as fine details in specific regions or features important for tasks like object recognition or segmentation. This lack of semantic awareness can lead to a degradation of semantically relevant information, which is detrimental in applications where understanding the content of the image is as important as its visual appearance.

An alternative approach to enhancing image quality is through \gls{sr} models, which aim to reconstruct \gls{hr} images from their \gls{lr} counterparts. Techniques like \cite{Dong2015SRCNN, Lim2017EDSR, Bruna2016SuperResolution, Rombach2022SR_CVPR} utilize various architectures, from classical \gls{dnn} to more advanced \gls{ddpm}, to learn the mapping between \gls{lr} and \gls{hr} image pairs. While \gls{sr} models can effectively enhance details and improve visual quality, they often  operate under the assumption that the input is a uniformly down-scaled version of the original image, which may not hold true in practical scenarios involving complex degradation or compression artifacts. As a result, while they can improve the overall appearance of an image, they may not adequately restore semantically significant details required for tasks that rely on precise content understanding.\\

In this context is inserted the proposed \gls{spic} framework. Although slightly suboptimal compared to conventional approaches in terms of the overall rate-distortion curve, the proposed method allows for a significantly better reconstruction when considering the combination of the multiple criticalities of other models.

What is more, thanks to the modular approach and the doubly conditioned \gls{semcore}  this framework is able to be easily modified without the need of further fine-tuning or re-training. This chapter proposes additionally to the \gls{spic} also an improved version, referred to as \gls{cspic}. This variation is introduced to overcome some limitations of the \gls{spic} framework in reconstructing small and detailed objects.\\

This chapter is organized as follows: In \sref{sec: SPIC spic}, the basic modular structure of \gls{spic} is introduced, focusing on the transmitter and receiver architectures and the proposed \gls{semcore} model. In \sref{sec: SPIC residuo}, the \gls{cspic} is introduced as a modified version of the \gls{spic} to allow better reconstruction of specific details, without requiring fine-tuning or re-training of the \gls{semcore} model. In \sref{sec: SPIC results}, the results of the proposed architectures are compared with classical compression algorithms such as \gls{bpg} and \gls{jpeg2000}.

\section{Semantic-Preserving Image Coding (SPIC)}\label{sec: SPIC spic}

\begin{figure}
    \centering
    \includegraphics[width=\textwidth]{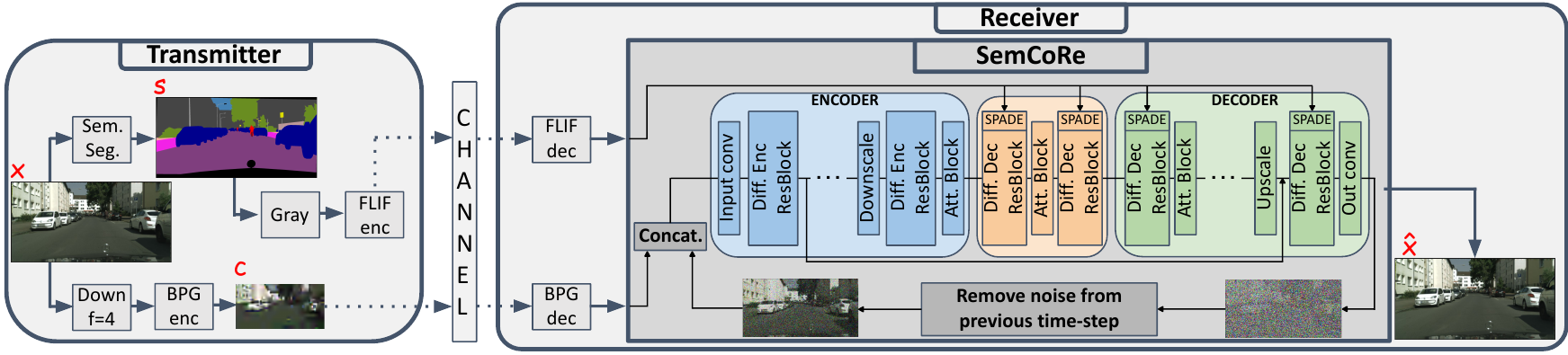}
    \caption[\acrshort{spic} architecture scheme]{Overview of the \acrshort{spic} architecture. At the transmitter side the \acrshort{ssm} $\s$ and the coarse image $\co$ are extracted from the original image and compressed with classical off-the-shelf compression algorithms. At the receiver it is employed the \acrshort{semcore} that leverages both $\s$ and $\co$ for high-fidelity semantic-relevant image recovery even at low BPP.}
    \label{fig: SPIC spic_scheme}
\end{figure}

In this section, the basic modular structure of the \gls{spic} framework is introduced, as illustrated in \fref{fig: SPIC spic_scheme}. The architecture comprises a transmitter and a receiver, which will be discussed separately.

\subsection{Transmitter Side}\label{sec: SPIC spic_transmitter}

As shown in \fref{fig: SPIC spic_scheme}, the transmitter consists of two separate pipelines designed to extract the \gls{ssm} $\mathbf{s}$ and the coarse image $\mathbf{c}$ from the input image $\mathbf{x}$.

The design aims to keep the system simple while allowing for future improvements and modifications. By adopting a modular approach, it is possible to integrate multiple components without the need for fine-tuning or re-training the model.

\begin{itemize}[label={}]
    \item \textbf{\gls{ssm} Pipeline:} As discussed at the beginning of the chapter, one of the fundamental semantic information that has to be available at the receiver is the \gls{ssm}, transmitted in a lossless way.

    In this work, the \gls{ssm} $\s$ is extracted directly from the original image $\x$ using an out-of-the-shelf pre-trained \gls{sota} \gls{ssmodel}. The INTERN-2.5 model \cite{Wang2022internimage} is chosen for this task due to its high performance in semantic segmentation. However, the modular structure of the \gls{spic} allows other choices to be made.

    After generating the \gls{ssm}, the next step is lossless compression. The output RGB-colored \gls{ssm} is mapped to grayscale by assigning each class a specific shade of gray. This grayscale \gls{ssm} is then compressed using the lossless compression algorithm \gls{flif} \cite{Sneyers2016FLIF}. This algorithm ensures efficient encoding of the \gls{ssm} with an average of 0.112 \gls{bpp}. \\

    \item \textbf{Coarse Image Pipeline:} This step is fundamental in compressing all other semantically relevant information about the original image that are not contained in the \gls{ssm}. Examples include object details and patterns, colors, and more.

    The process starts with an initial down-scaling. The image $\x$ is transformed from its original resolution of $256 \times 512$ pixels to $64 \times 128$ pixels. This operation is performed by averaging the pixel values in corresponding patches.

    Thanks to the down-scaling it is possible to preserve a general "overview" of the original image $\x$ with some of the most semantically relevant information about color, patterns, etc. by using only a fraction of the total number of pixels.

    Unlike the \gls{ssm}, perfect reconstruction of the coarse image at the receiver is not required. Therefore, a lossy compression algorithm can be employed. In this work, the \gls{bpg} algorithm is used.

    This algorithm can compress the coarse image $\co$ at different resolutions, measured in terms of \gls{bpp}, providing a lot of flexibility. In fact, the value at which $\co$ is compressed will affect the level of detail preservation and, consequently, the quality of the final reconstructed image $\hat{\x}$.
\end{itemize}

After this pre-processing, both the encoded $\s$ and $\co$ are sent to the receiver.

\subsection{Semantic-Conditioned Super-Resolution Diffusion Model (SemCoRe)}\label{sec: SPIC semcore}

At the receiver, the incoming encoded $\s$ and $\co$ are decoded using their corresponding algorithms.

Beyond this, the receiver consists solely of the proposed \gls{semcore} block. Its purpose is to reconstruct an image $\hat{\x}$ that preserves all the semantic information of $\x$. \\

The architecture of the proposed \gls{semcore} is based on the work of Wang et al. \cite{Wang2022SISDM}. They introduced an \gls{ssm}-conditioned \gls{ddpm} capable of generating visually appealing images with high levels of \gls{ssm} preservation but colors and texture completely different from the original image. In this thesis, thanks to the additional conditioning on the coarse image this problem has been mitigated. The architecture of the proposed \gls{semcore} is as follows:

\begin{itemize}[label={}]
    \item {\textbf{Encoder}:} The encoder is the part of the network responsible for the extraction of the most important features. As already introduced in \sref{sec: GM unet} and \sref{sec: GM ddpm_advanced_techniques}, the conditioning in a \gls{unet} is strongly dependent on the task and since the scope of the encoder is to extract as much information as possible from the input, it is useful to use the conditioning on the coarse $\co$ at this stage. Specifically, the input tensor has a shape of $6 \times 256 \times 512$ pixels. The first three channels correspond to the RGB values of the noisy image $\x_t$ at time-step $t$, and the last three channels correspond at every iteration to the same RGB values of the up-scaled coarse image $\co$\footnote{The up-scaling is performed via linear interpolation to restore the original shape of $256 \times 512$ pixels. This is done to reverse the down-scaling performed at the transmitter side.}. This concatenated input allows the encoder to process both images simultaneously and capture important details from the coarse image at every step of the diffusion process. 
    
    The structure of the encoder $E$ consist of an initial convolutional layer that map the input tensor of shape $6 \times 256 \times 512$ to a tensor of shape $512 \times 256 \times 512$. This is followed by a repeated alternating series of two time-conditioned \glspl{resblock}, depicted in \fref{fig: GM resblock with time}, a multi-head self-attention layer with 8 heads and a down-scaling layer. This sequence is repeated 2 times. After an additional time-conditioned \glspl{resblock} and a multi-head self-attention layer with 8 heads are employed to conclude the encoding phase. 
    
    At the end of the encoding phase the tensor $\z = E(\x_t, \co)$ has a shape of $256 \times 64 \times 128$.\\
    \item {\textbf{Bottleneck}:} The bottleneck is that part of the \gls{unet} at the beginning of the up-scaling phase. For this reason it is important to start the conditioning process on the \gls{ssm} to enforce its structure. This is performed by the use of the \gls{spade} layer introduced in the \gls{resblock}. In this case the \gls{resblock} is obtained by merging the time-conditioned version in \fref{fig: GM resblock with time} with the one that uses the \gls{ssm} conditioning in \fref{fig: GM resblock spade} \cite{Wang2022SISDM}. In this way the \gls{unet} will be able to influence the denoising on the current time-step and enforce the structure of the \gls{ssm}. This \gls{resblock} will be referred to as Diff-Dec \gls{resblock}.
    
    The structure of the bottleneck is composed by one Diff-Dec \gls{resblock} followed by a multi-head self-attention layer with 8 heads and another Diff-Dec \gls{resblock} and the output shape is preserved.
    \item {\textbf{Decoder}:} The scope of the decoder is to produce the denoised version of $\x_t$ in a way that is coherent with the \gls{ssm}. Its structure is composed of an initial Diff-Dec \gls{resblock} followed by a multi-head self-attention layer with 8 heads. Then there is a sequence of two Diff-Dec \gls{resblock}, a multi-head self-attention layer with 8 heads and a linear up-scaling. This sequence is repeated 2 times. The final block of the decoder consist in a  Diff-Dec \gls{resblock} followed by a convolutional layer. 
    
    At the end the output correspond to the prediction $\bepsilon_0$ of shape $3 \times 256 \times 512$ to be removed from $\x_t$ to obtain the estimate of $\x_0\equiv \x$.
\end{itemize}

After that the \gls{unet} structure of the \gls{semcore} has been defined it is possible to focus on the diffusion step. 
As introduced in \sref{sec: GM ddpm training_inference}, the training consist of $T=1000$ diffusion steps. At inference, to save computational resources, only a reduced number of diffusion steps is performed, $T=20$.

On top of that, to better enforce the preservation of the \gls{ssm} the training is performed by applying the \gls{cfg} technique discussed in \sref{sec: GM ddpm_advanced_techniques} illustrated in \fref{fig: GM schema_cfg}.

By combining the dual conditioning approach and advanced training procedures, the \gls{semcore} model is able to produce \gls{sr} images of the coarse $\co$, that are visually close to the original. In this way the relevant semantic information is better preserved.

\section{Class-Specific SPIC (C-SPIC)}\label{sec: SPIC residuo}

\begin{figure}[!t]
    \centering
    \includegraphics[width=\textwidth]{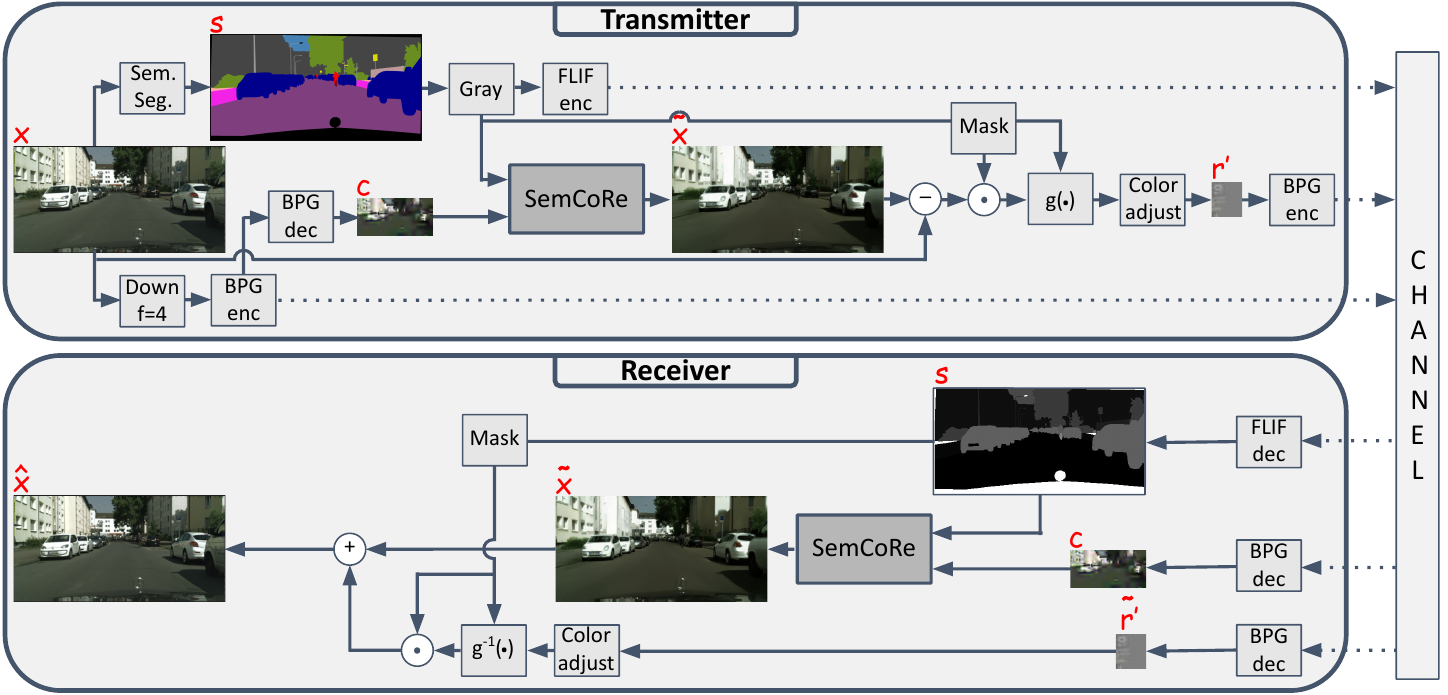}
    \caption[\acrshort{cspic}$^w$ architecture scheme]{Overview of the \acrshort{cspic}$^w$ architecture. At the transmitter after the generation and compression of $\s$ and  $\co$ the \acrshort{semcore} is used to evaluate the intermediate reconstructed image $\tilde{\x}$ and the relevant residual $\br$. This residual is processed and compacted to preserve only information about some relevant classes before being send to the receiver. At the receiver the $\tilde{\x}$ is evaluated starting from $\s$ and $\co$ and then the relevant residual is processed and de-compacted before adding it back to produce the final reconstructed image $\hat{\x}$.}
    %[C-SPIC scheme with residual]{Overview of the proposed C-SPIC architecture with residual.}
    \label{fig: SPIC cspic with residual scheme}
\end{figure}

Even if the \gls{spic} framework is already able to reconstruct images that are more semantically relevant and visually appealing compared to classical compression algorithms and other \gls{dnn}-based models, it still lacks one important detail.
In one of the first stages of \gls{spic}, the original image $\x$ is down-scaled to produce the coarse image $\co$. Although this down-scaling allows for a more efficient compression of the data, some specific small details are inevitably lost.

This phenomenon can be neglected in many cases but not in some other. For example in an urban scenario where knowing only that  the color of a traffic sign is blue is not always enough. By down-scaling the image to produce the coarse, some information about colors are preserved, but the small details  inside the traffic sign might be lost.

To address this limitation, the \gls{spic} framework is further modified to consider some specific object as semantically relevant, thus improving their reconstruction. This modified version is referred to as \gls{cspic} and is implemented without any need for fine-tuning or re-training of the \gls{semcore} model.\\

The idea is based on the \gls{dsslic} framework proposed in \cite{Akbari2019DSSLIC}. In their work, they introduced a framework to improve the reconstruction quality of images. Instead of evaluating the model only at the receiver, the model is evaluated at both ends to produce an intermediate reconstruction $\tilde{\x}$. At the transmitter side, the difference between the original image and the intermediate reconstruction is evaluated, compressed and sent to the receiver. This quantity is referred to as residual, $\br = \x - \tilde{\x}$. At this point at the receiver, after reconstructing the intermediate representation $\tilde{\x}$, the residual is added back. This will improve the quality of the final reconstructed image $\hat{\x}$.\\

A similar idea is implemented and proposed in this section, but with a substantial difference: the residual is considered only for the semantically relevant classes and not for the whole frame. This change allows for significant savings in terms of \gls{bpp} compared to the transmission of the full residual and this modification will not affect the reconstruction of the remaining semantic information.

This section is divided into two parts. In \sref{sec: SPIC residual dsslic}, a first version named \gls{cspic}$^w$ is introduced as a variation of the \gls{dsslic} framework. The apex $^w$ is used to specify that the residual is produced and transmitted. In \sref{sec: SPIC image dsslic}, another version referred to as \gls{cspic}$^{w/o}$ is presented. This variation is introduced to overcome the computational complexity of \gls{cspic}$^w$. In fact, to obtain the residual $\br$ it is fundamental to use the \gls{semcore} even at the transmitter side. This can be feasible in some applications but not doable in all those in which the transmitter has a low computational power. The \gls{cspic}$^{w/o}$ is designed to have the same computational complexity as for the classic \gls{spic}, and allow the same performances as \gls{cspic} with an increase of at most 10\% of the total \gls{bpp}.

\subsection{C-SPIC with Residual (C-SPIC$^w$)}\label{sec: SPIC residual dsslic}

The scheme of the proposed \gls{cspic}$^w$ framework is depicted in \fref{fig: SPIC cspic with residual scheme}.

Differently from the base version of \gls{spic} shown in \fref{fig: SPIC spic_scheme}, the structure of both the transmitter and receiver consist now of multiple elements. This is absolutely necessary  to improve the reconstruction capabilities of certain relevant semantic classes. 

The \gls{cspic}$^w$ is composed as follows:

\subsubsection{Transmitter Side}\label{sec: SPIC cspic_with transmitter}

The initial processing of the original image $\x$ is the same as discussed for \gls{spic} in \sref{sec: SPIC spic_transmitter}, the coarse image $\co$ and the \gls{ssm} $\s$ being produced and transmitted to the receiver using the same approach.

Differently from before, the coarse image and \gls{ssm} are now processed by the \gls{semcore} directly at the transmitter to output the intermediate reconstructed image $\tilde{\x}$. This image is then used to produce the general residual $\br = \x - \tilde{\x}$ of the whole image.

\begin{figure}[!t]
    \centering
    \begin{subfigure}[b]{0.49\textwidth}
        \centering
        \includegraphics[width=\textwidth]{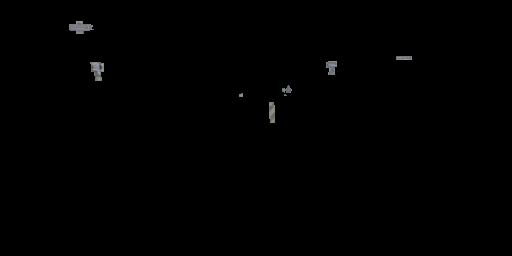}
        \caption*{(a) Residual $\br_m = \br \odot \m$ after masking}
    \end{subfigure}%
    \begin{subfigure}[b]{0.24\textwidth}
        \centering
        \raisebox{-\height}{\includegraphics[height=\heightof{\includegraphics[width=2.04\textwidth]{Figures/SPIC/improved/Residual_masked.png}}]{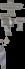}}
        \caption*{(b) Compacted $g(\br_m)$}
    \end{subfigure}
    \begin{subfigure}[b]{0.24\textwidth}
        \centering
        \raisebox{-\height}{\includegraphics[height=\heightof{\includegraphics[width=2.04\textwidth]{Figures/SPIC/improved/Residual_masked.png}}]{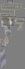}}
        \caption*{(c) Final $\br'$}
    \end{subfigure}%

    \caption[Residual processing pipeline in the \acrshort{cspic}]{Processing pipeline of the residual $\br$ in the \acrshort{cspic}: (a) the residual isis initially masked using the binary mask $\m$, (b) the function $g(\cdot)$ is applied to compact the relevant regions and reduce empty spaces, and (c) the pixel values are re-scaled to an 8-bit representation.}
    \label{fig: SPIC cspic residual pipeline}
\end{figure}

However, since only certain semantic classes are considered as relevant and require higher reconstruction performance, the residual is processed as depicted in \fref{fig: SPIC cspic residual pipeline} undergoing three different steps:

\begin{enumerate}
    \item \textbf{Masking:} First, a binary mask $\m$ is obtained from the \gls{ssm} $\s$. Only the pixels corresponding to the semantically relevant classes are set to 1, and the others to 0. This binary mask $\m$ is used to perform an element-wise multiplication with the residual $\br$, obtaining $\br_m = \br \odot \m$, as shown in \fref{fig: SPIC cspic residual pipeline}(a). This step effectively filters out the residual information for non-relevant classes.

    \item \textbf{Compacting:} This is the step that allows to transform the residual from the representation in \fref{fig: SPIC cspic residual pipeline}(a) to (b). 
    
    The invertible algorithm $g$ for compacting the residual, and removing the empty regions is described in detail in \aref{app: SPIC compacting} and only requires the \gls{ssm} to be performed. The advantage of using the compacted representation lies in the fact that the residual can be represented with fewer bits because of the smaller shape of compacted version. For instance, from its original shape of $256 \times 512$ pixels, the reshaped $g(\br_m)$ has an average shape of $40 \times 40$ pixels when considering traffic signs, $30 \times 40$ pixels when considering traffic lights, and $80 \times 40$ pixels when considering pedestrians. This represents a significant reduction in data size, allowing for lower amount of \gls{bpp} in compression.

    \item \textbf{Color Adjusting:} The final step involves re-scaling the pixel values. Since the residual is a difference between two images, the values of each RGB component are in the range $[-255, 255]$. To make these values compatible with the \gls{bpg} compression algorithm, they are re-scaled into the range $[0, 255]$. This results in the final processed residual $\br'$, as shown in \fref{fig: SPIC cspic residual pipeline}(c).
\end{enumerate}

The final processed residual $\br'$ can now be compressed in a lossy manner using the \gls{bpg} algorithm and sent to the receiver.

\subsubsection{Receiver Side}\label{sec: SPIC cspic_with receiver}

At the receiver, the \gls{ssm} $\s$ and coarse image $\co$ are processed as in the \gls{spic} framework: they are decoded and used as conditioning inputs to the \gls{semcore} model to produce the same intermediate reconstructed image $\tilde{\x}$ as at the transmitter.

The processing of the residual involves reversing the transformations applied at the transmitter:

\begin{enumerate}
    \item \textbf{Reverse Color Adjusting:} The received and decoded residual $\tilde{\br}'$ is first processed to adjust the values of the pixels back into the original range of $[-255, 255]$.

    \item \textbf{De-compacting:} Since the \gls{ssm} $\s$ is transmitted losslessly the compacting algorithm is reversed and the original shape of the residual is restored. This operation is referred to as $g^{-1}$ and described in \aref{app: SPIC compacting}.

    \item \textbf{Masking:} The residual is then multiplied element-wise with the mask $\m$ to ensure that only the relevant semantic classes are considered.
\end{enumerate}

Finally, the masked reconstructed residual is added back to the intermediate reconstructed image $\tilde{\x}$ to obtain the final reconstructed image $\hat{\x}$. In this way $\hat{\x}$ will contain enhanced details for the specific semantic classes considered as relevant. The quality of these details will be influenced by the compression level at which the residual has been encoded.

\subsection{C-SPIC without Residual (C-SPIC$^{w/o}$)}\label{sec: SPIC image dsslic}

In scenarios where the transmitter cannot perform the computationally intensive task of evaluating the \gls{semcore} model, an alternative approach is necessary.

\begin{figure}
    \centering
    \includegraphics[width=\textwidth]{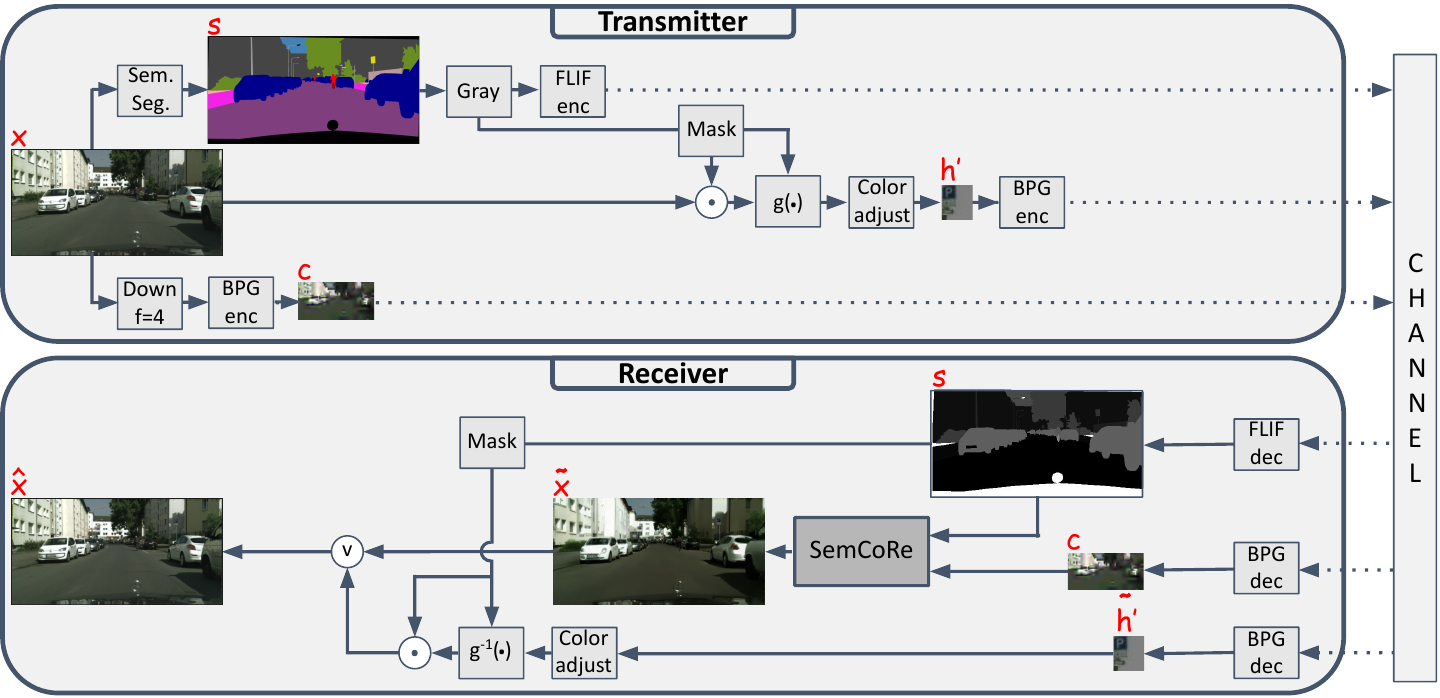}
    \caption[\acrshort{cspic}$^{w/o}$ architecture scheme]{Overview of the \acrshort{cspic}$^{w/o}$ architecture. At the transmitter after the generation and compression of $\s$ and  $\co$ the original image is directly used to extract the information about relevant classes. After masking and compacting, the transformation $\h'$ is send to the receiver. At the receiver the intermediate $\tilde{\x}$ is evaluated starting from $\s$ and $\co$ by using the \acrshort{semcore} model and then the semantically relevant parts are replaced with the original extracted from the received $\tilde{\h}'$.} 
    %[C-SPIC scheme without residual]{Overview of the proposed C-SPIC architecture without residual.}
    \label{fig: SPIC cspic without scheme}
\end{figure}

To address this, the \gls{cspic}$^{w/o}$ is introduced as a variant of the \gls{cspic}$^w$ and the scheme is depicted in \fref{fig: SPIC cspic without scheme}. This new implementation significantly reduces the computational requirements at the transmitter side while preserving similar reconstruction quality. The draw back is the increase of up to 10\% of the required \gls{bpp}, this may or may not represent a problem depending on the communication infrastructure.

\subsubsection{Transmitter Side}\label{sec: SPIC cspic_no transmitter}

In this variant, the transmitter's structure remains similar to that discussed in \sref{sec: SPIC cspic_with transmitter}, with one crucial difference: the residual is not computed but instead the original image $\x$ is used directly to extract the relevant details as follows:

\begin{enumerate}
    \item \textbf{Masking:} The original image $\x$ is multiplied element-wise with the binary mask $\m$ obtained from the \gls{ssm} $\s$. This results in an image where only the relevant semantic classes are preserved, and the rest of the pixels are zeroed out.

    \item \textbf{Compacting:} The function $g(\cdot)$ is applied to the masked image to compact the relevant regions.

    \item \textbf{Color Adjusting (Optional):} To reduce the \gls{bpp} a color correction step can be applied. This involves modifying the RGB values of the pixels that have been set to zero  to match the mean values of the relevant pixels. This can slightly improve the compression efficiency.
\end{enumerate}

The resulting compacted image $\h'$ is then compressed using a lossy compression algorithm like \gls{bpg} and transmitted to the receiver.

While this approach avoids the need for the transmitter to compute $\tilde{\x}$ using the \gls{semcore} model, the amount of \gls{bpp} required to compress $\h'$ are higher compared to $\br'$. This is a trade-off that has to be considered when choosing between \gls{cspic}$^w$ and \gls{cspic}$^{w/o}$.

\subsubsection{Receiver Side}

The receiver processes the \gls{ssm} $\s$ and coarse image $\co$ as before, using them to generate the intermediate reconstructed image $\tilde{\x}$ via the \gls{semcore} model.

The received compacted image $\h'$ is decoded and then processed as follows:

\begin{enumerate}
    \item \textbf{De-compacting:} The inverse function $g^{-1}(\cdot)$ is applied to restore the compacted image to its original dimensions, using the \gls{ssm}.

    \item \textbf{Masking:} The de-compacted masked image is multiplied element-wise with the mask $\m$ to ensure that only the relevant semantic classes are considered and $\h_m$ is obtained.
\end{enumerate}

Finally, the relevant regions from the de-compacted image are substituted into the intermediate reconstructed image $\tilde{\x}$ to obtain the final reconstructed image $\hat{\x}$:

\[\hat{\x} = \tilde{\x} \;\circv\; \h_m\]

Here, the symbol $\circv$ denotes the substitution of pixel values in $\tilde{\x}$ with those in $\h_m$ when the value of the mask $\m$ is 1.\\

\section{Results}\label{sec: SPIC results}

This section provides a comprehensive analysis of the performance of the proposed frameworks, \gls{spic} and \gls{cspic}. The evaluation is conducted on the Cityscapes dataset \cite{Cordts2016Cityscapes}, although the frameworks are applicable to any dataset from which a \gls{ssm} with fewer than 255 semantic classes can be extracted.

The Cityscapes dataset contains 5000 pairs of images recorded in 50 German cities from the interior of a moving vehicle. Each pair consists of an RGB image $\x$ of the street scene and the associated \gls{ssm} $\s$. The images have a resolution of $1024\times2048$ pixels and are divided into 2975 pairs for training, 500 for validation, and 1525 for testing. Both the images and \glspl{ssm} are considered RGB images.

The \gls{ssm} in the Cityscapes dataset comprises multiple classes. In most applications, including this one, only $C_\s=19+1$ classes are used. The first 19 classes refer to actual semantic categories like "car", "person", etc., while the last one is the "null class", representing all objects that do not belong to any other class (e.g. the silhouette of the recording vehicle). Before being input to the network, the images and \gls{ssm}s are down-scaled to $256\times512$ pixels. This re-scaling is performed by averaging the RGB values in the same patch for the image while considering the  nearest-neighbours interpolation for the \gls{ssm}.

The true \gls{ssm} provided with the dataset is used only during training. At inference time, the \gls{ssm} is generated by an out-of-the-shelf \gls{sota} \gls{ssmodel}. The INTERN-2.5 model \cite{Wang2022internimage} is chosen for its remarkable ability to extract \glspl{ssm} from images, though other options can be adopted without affecting the results, thanks to the modularity of the approach.

An important metric used in this section is the \gls{bpp}, evaluated in this context as follows:

\begin{equation} \text{BPP} = \frac{B_{\x} + B_{\s} + B_{\br'}}{256\times512}, \end{equation}

where $B_{\x}$ is the number of bits used to lossy encode the image, $B_{\s}$ is the number of bits used to losslessly encode the \gls{ssm}, $B_{\br'}$ is the number of bits used to lossy encode the residual and $256\times512$ is the total number of pixels in the original image. The first two terms are always present in both \gls{spic} and \gls{cspic}. The term associated to the residual is instead present only in the \gls{cspic} framework and can refer to either the compacted masked residual $\br'$ or the compacted masked image $\h'$.\\

This section is divided into two parts. In \sref{sec: SPIC results spic}, the performance of \gls{spic} is compared to classical compression algorithms like \gls{bpg} and \gls{jpeg2000}. In \sref{sec: SPIC results c-spic}, the results for \gls{cspic} are discussed, highlighting the visual improvement in reconstructing semantically relevant classes.

\subsection{SPIC}\label{sec: SPIC results spic}
\begin{figure}[t]
    \centering
    % Titles
    \begin{subfigure}{\textwidth}
        \centering
        \begin{minipage}{0.32\textwidth}
            \centering
            \textbf{BPG} ($0.176$ BPP)
        \end{minipage}
        \hfill
        \begin{minipage}{0.32\textwidth}
            \centering
            \textbf{Original}
        \end{minipage}
        \hfill
        \begin{minipage}{0.32\textwidth}
            \centering
            \textbf{SPIC} ($0.166$ BPP)
        \end{minipage}
    \end{subfigure}
    
    % Top row of images
    \begin{subfigure}{\textwidth}
        \centering
        \begin{tikzpicture}
            \node[anchor=south west,inner sep=0] (frame1) at (0,0) {\includegraphics[width=.32\textwidth]{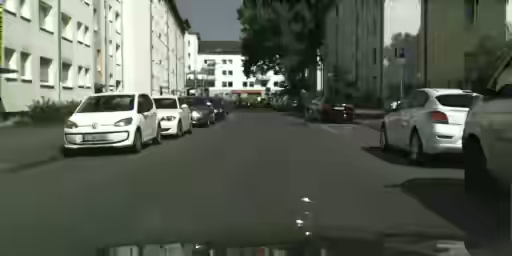}};
            \node[anchor=south west,inner sep=0] (frame2) at (frame1.south east) {\includegraphics[width=.32\textwidth]{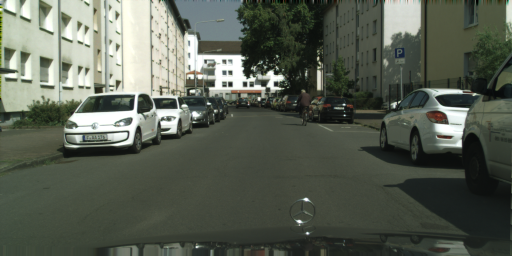}};
            \node[anchor=south west,inner sep=0] (frame3) at (frame2.south east) {\includegraphics[width=.32\textwidth]{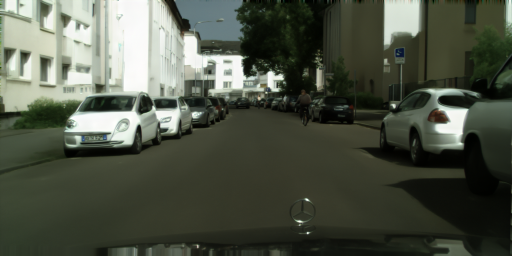}};
            
            % Add \gls{ssm} overlays
            \begin{scope}[x={(frame1.south east)},y={(frame1.north west)}]
                \node[anchor=south east,inner sep=0] at (1,0) {\includegraphics[width=.13\textwidth]{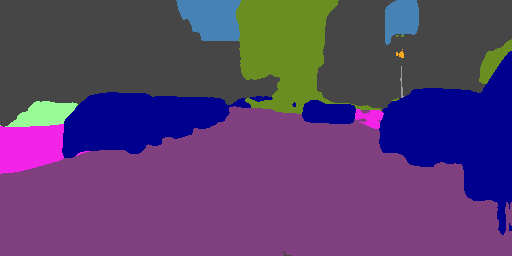}};
            \end{scope}
            \begin{scope}[x={(frame2.south east)},y={(frame2.north west)}]
                \node[anchor=south east,inner sep=0] at (1,0) {\includegraphics[width=.13\textwidth]{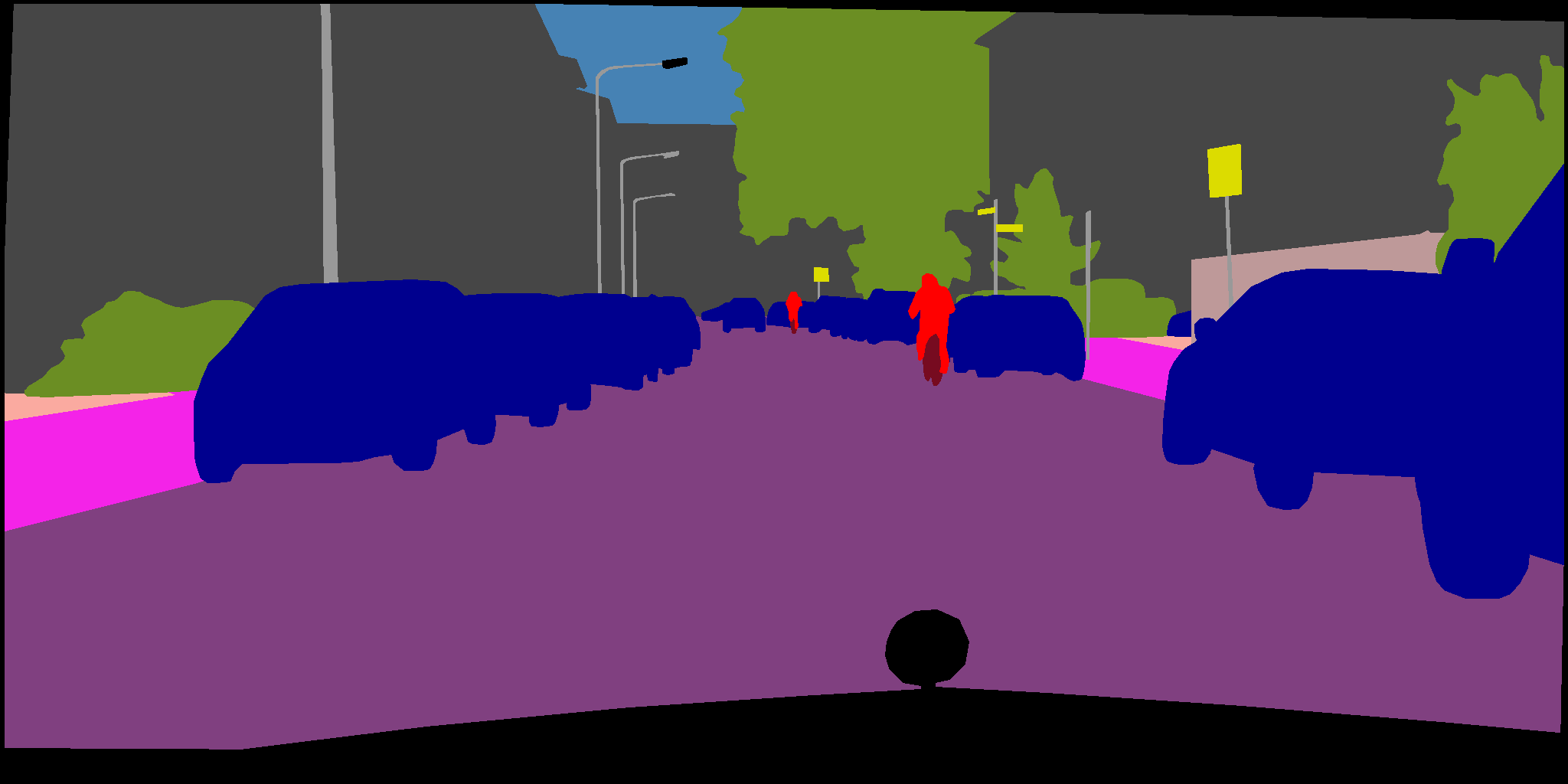}};
            \end{scope}
            \begin{scope}[x={(frame3.south east)},y={(frame3.north west)}]
                \node[anchor=south east,inner sep=0] at (1,0) {\includegraphics[width=.13\textwidth]{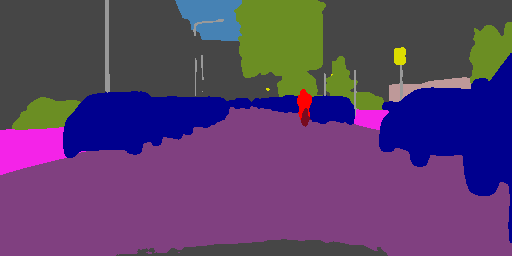}};
            \end{scope}
        \end{tikzpicture}
    \end{subfigure}
    
    % Bottom row of images with trimming
    \begin{subfigure}{\textwidth}
        \centering
        \begin{tikzpicture}
            \node[anchor=south west,inner sep=0] (frame4) at (0,0) {\includegraphics[width=.32\textwidth, trim=190 100 90 40, clip]{Figures/SPIC/Classico/bpg40.png}};
            \node[anchor=south west,inner sep=0] (frame5) at (frame4.south east) {\includegraphics[width=.32\textwidth, trim=190 100 90 40, clip]{Figures/SPIC/Classico/original_512.png}};
            \node[anchor=south west,inner sep=0] (frame6) at (frame5.south east) {\includegraphics[width=.32\textwidth, trim=190 100 90 40, clip]{Figures/SPIC/Classico/our_bpg35.png}};
            
            % Add \gls{ssm} overlays with different trimming
            \begin{scope}[x={(frame4.south east)},y={(frame4.north west)}]
                \node[anchor=south east,inner sep=0] at (1,0) {\includegraphics[width=.13\textwidth, trim=190 100 90 40, clip]{Figures/SPIC/Classico/bpg40_SSM.png}};
            \end{scope}
            \begin{scope}[x={(frame5.south east)},y={(frame5.north west)}]
                \node[anchor=south east,inner sep=0] at (1,0) {\includegraphics[width=.13\textwidth, trim=760 400 360 160, clip]{Figures/SPIC/Classico/original_512_SSM.png}};
            \end{scope}
            \begin{scope}[x={(frame6.south east)},y={(frame6.north west)}]
                \node[anchor=south east,inner sep=0] at (1,0) {\includegraphics[width=.13\textwidth, trim=190 100 90 40, clip]{Figures/SPIC/Classico/our_bpg35_SSM.png}};
            \end{scope}
        \end{tikzpicture}
    \end{subfigure}
    \caption[Visual comparison between \acrshort{bpg} and \acrshort{spic}]{Visual comparison between the original image and \acrshort{ssm} (CENTER), the image compressed with \acrshort{bpg} at $0.176$ \acrshort{bpp} with the associated generated \acrshort{ssm} (LEFT), and the image obtained with the proposed \acrshort{spic} framework at $0.166$ \acrshort{bpp} and the associated generated \acrshort{ssm} (RIGHT). At similar values of \acrshort{bpp} the proposed approach preserves semantically relevant details better than \acrshort{bpg}. The person riding the bicycle is clearly detected by the generated \acrshort{ssm}.}
    \label{fig: SPIC Visual_comparison_OUR_vs_BPG}
\end{figure}

An advantage of the proposed model is its capability to retain semantic information while providing a good trade-off between overall image quality and compression rate. Several existing compression algorithms and \gls{sr} models often reconstruct visually pleasing images. However, a closer inspection reveals a significant drawback: the degradation of semantic content. This effect is amplified and particularly evident as the size of semantically relevant objects within the image diminishes. For larger foreground objects the level of detail and \gls{ssm} retention is good enough. However, as the object size shrinks, conventional models falter, failing to accurately process the image and preserve details.

This aspect can be observed by examining \fref{fig: SPIC Visual_comparison_OUR_vs_BPG} where the proposed \gls{spic} is compared to the classical \gls{bpg} algorithm at similar \gls{bpp} values.

The upper row shows the reconstructed images $\hat{\x}$, while the lower row displays the associated \glspl{ssm}. For the original image is reported the original \gls{ssm}, while for the other two columns, the \glspl{ssm} shown are the ones generated form $\hat{\x}$ via the INTERN-2.5 semantic segmentation model.

As can be seen, the overall visual quality of the images is very similar. In some details, such as the shape of the windows, the \gls{bpg} algorithm performs even better. However, when attention is focused on the generated \gls{ssm}, the results are completely different. The image reconstructed using \gls{bpg} fails to retain detailed information about small objects. An example is the person on the bicycle, shown in detail in the bottom row. The \gls{bpg} algorithm fails to correctly compress and reconstruct the rider in $\hat{\x}$, causing the \gls{ssmodel} to not detect the rider. However, in the case of the \gls{spic} framework, by conditioning the reconstruction of the image on the original \gls{ssm}, the preservation of the semantic information is guaranteed. This can be visually appreciated in the generated \gls{ssm}.
\begin{figure}[!t]
    \centering
    % Titles Row
    \begin{subfigure}{\textwidth}
        \centering
        \begin{minipage}{0.333\textwidth}
            \centering
            \textbf{SOTA SR model} \cite{Rombach2022SR_CVPR}
        \end{minipage}%
        \begin{minipage}{0.333\textwidth}
            \centering
            \textbf{SPIC}
        \end{minipage}%
        \begin{minipage}{0.333\textwidth}
            \centering
            \textbf{\gls{ssm}-based model} \cite{Wang2022SISDM}
        \end{minipage}
    \end{subfigure}
    
    % Images Row
    \begin{subfigure}{\textwidth}
        \centering
        \begin{subfigure}{0.333\textwidth}
            \centering
            \includegraphics[width=\linewidth, trim=200 110 180 70, clip]{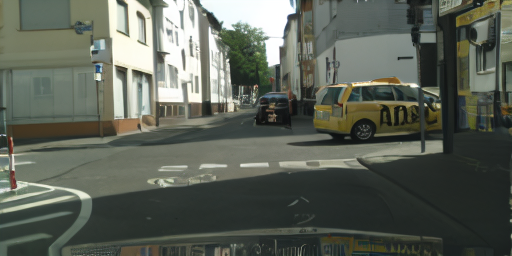}
            \caption*{}
        \end{subfigure}%
        \begin{subfigure}{0.333\textwidth}
            \centering
            \includegraphics[width=\linewidth, trim=200 110 180 70, clip]{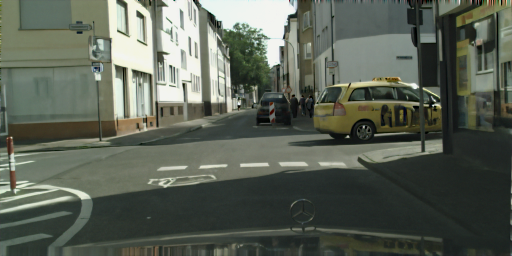}
            \caption*{}
        \end{subfigure}%
        \begin{subfigure}{0.333\textwidth}
            \centering
            \includegraphics[width=\linewidth, trim=200 110 180 70, clip]{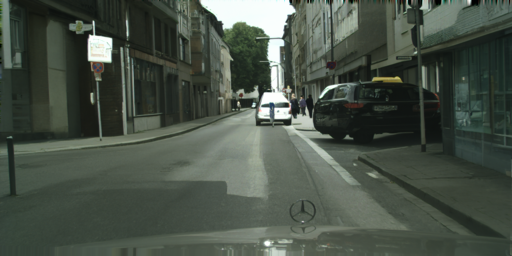}
            \caption*{}
        \end{subfigure}
    \end{subfigure}
    \vspace{-1cm}
    \caption[Visual comparison with Super-Resolution and SSM conditioned models]{Visual comparison between: the \acrshort{sota} \acrshort{sr} model \cite{Rombach2022SR_CVPR} (LEFT), the proposed \acrshort{spic} framework (CENTER), and the \acrshort{ssm} conditioned model \cite{Wang2022SISDM} (RIGHT).}
    \label{fig: SPIC super_res_comparison}
\end{figure}

Not only do classical compression approaches present these problems, but issues also arise when analyzing \gls{sr} models. All the \gls{sr} models work under a very strict condition: the input coarse image $\co$ must already contain sufficient details. To better explain this concept, \fref{fig: SPIC super_res_comparison} illustrates a comparison between the reconstruction capabilities of the \gls{sota} \gls{sr} model proposed in \cite{Rombach2022SR_CVPR} (left) and the proposed \gls{spic} framework (middle).

The main drawback of \gls{sr} models is that they are designed to enhance the quality of images that contain minor defects. Nonetheless, when the starting image lacks relevant details, the model is unable to infer what the output image should look like. The result is that the \gls{sota} \gls{sr} model fails to reconstruct the three pedestrians between the two cars. This failure is due to the very low-resolution input image, which does not contain enough details. Removing objects is not the only problem. The model begins adding pedestrians near the sidewalk or on the street when detecting another object that cannot correctly identify. This happens due to the low-resolution coarse image, where objects vaguely resembling pedestrians are wrongly considered as belonging to that class.

Conditioning only on the low-resolution image does not yield good results in terms of object reconstruction. The other extreme is conditioning solely on the \gls{ssm}. In \fref{fig: SPIC super_res_comparison}, the result of this conditioning is shown in the right image. This image has been reconstructed by conditioning only on the \gls{ssm} using the model proposed in \cite{Wang2022SISDM}. In this case, the pedestrians are placed in the correct positions, as are all the other elements in the scene. Still, the drawback is that the resulting image is completely different from the original: the colors and traffic signs have changed. This is because this family of models is agnostic to any other semantic information beside the \gls{ssm}.

In this scenario, \gls{spic} manages to balance all these factors. With this framework, it is possible to transmit the original \gls{ssm} losslessly to the receiver and at the same time reconstruct an overall good-looking image able to preserve all the most important semantic details, thanks to the conditioning on the \gls{ssm} itself.
\begin{figure}[!t]
    \centering
    \begin{subfigure}[t]{0.49\textwidth}
        \centering
        \includegraphics[width=\textwidth, trim=5 10 40 20, clip]{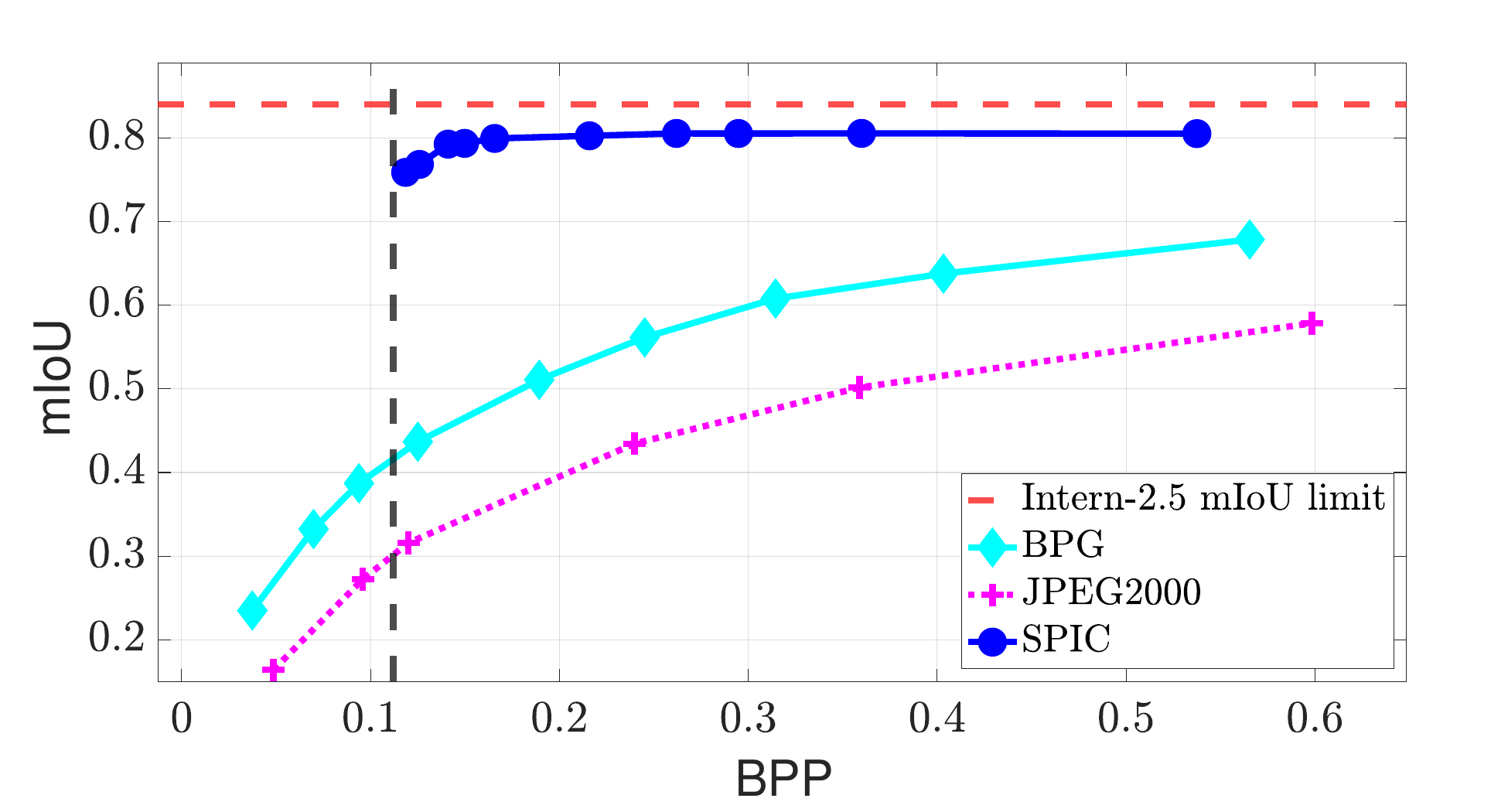}
        \caption*{(a)} % Caption under image without adding to list
    \end{subfigure}%
    \begin{subfigure}[t]{0.49\textwidth}
        \centering
        \includegraphics[width=\textwidth, trim=5 10 40 20, clip]{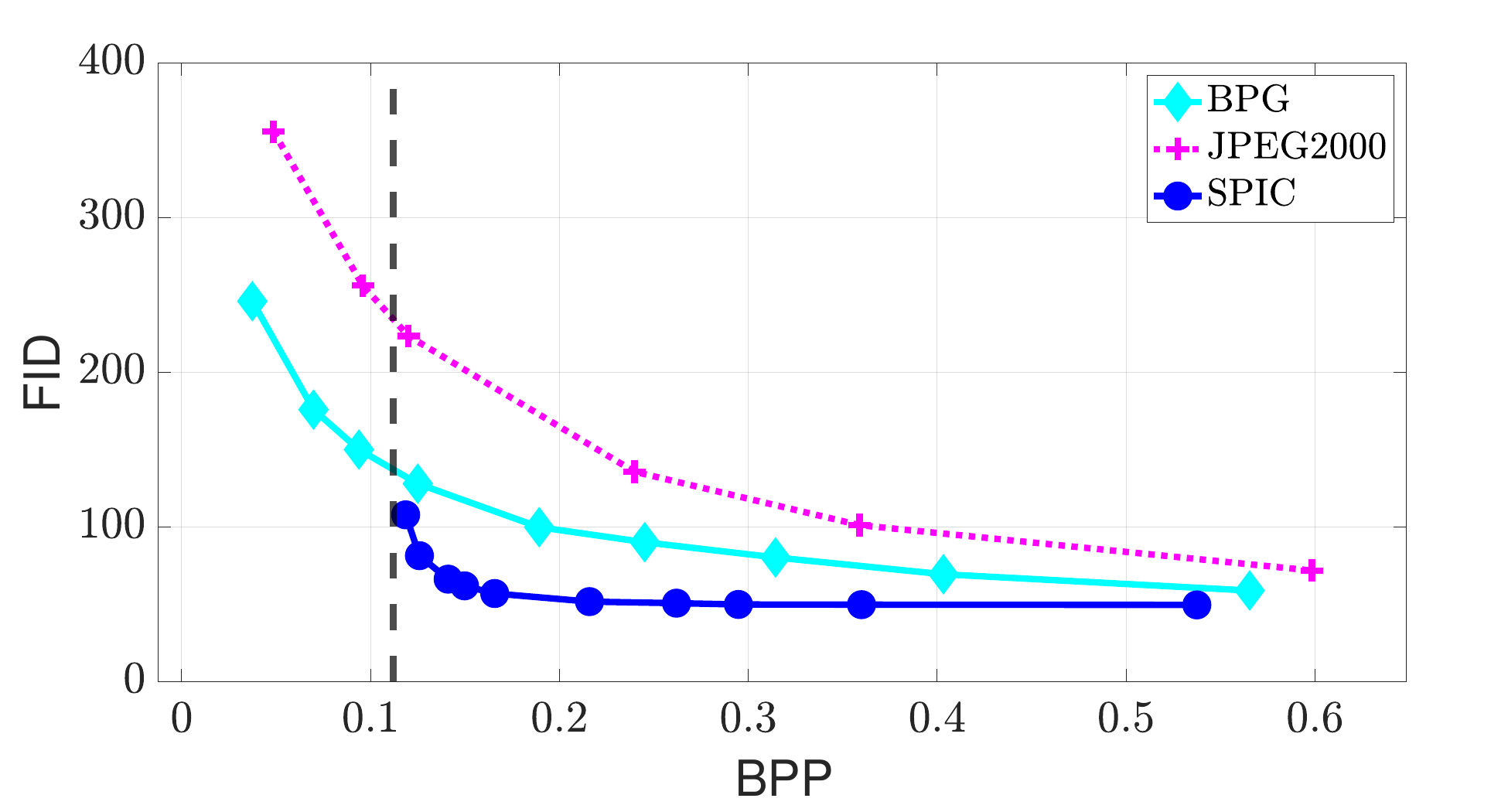}
        \caption*{(b)}
    \end{subfigure}
    \caption[Performance comparison between the \acrshort{spic} and classical compression algorithms]{Performance comparison between \acrshort{bpg}, \acrshort{jpeg2000} and \acrshort{spic} evaluated in terms of \acrshort{miou} and \acrshort{fid}.}
    \label{fig: SPIC spic metrics comparison}
\end{figure}

The performances of the \gls{spic} framework are shown in \fref{fig: SPIC spic metrics comparison}.\footnote{All the performances are evaluated by averaging the results on 500 images.} On the left is shown the plot associated with the concept of \gls{ssm} retention. This quantity is evaluated by the \gls{miou} between the original \gls{ssm} and the one generated from $\hat{\x}$ via the INTERN-2.5 \gls{ssmodel}.

The black dotted vertical line, positioned at $0.112$ \gls{bpp}, represents the \gls{bpp} required for the lossless compression of the \gls{ssm}. The blue point represents the \gls{ssm} retention of the proposed \gls{spic} framework, evaluated at a \gls{bpp} given by the sum of the \gls{bpp} necessary for the lossless encoding of the \gls{ssm} and the lossy encoding of the coarse image. The green and magenta curves represent the performances achieved with \gls{bpg} and \gls{jpeg2000} compression methods.

The red dotted horizontal line represents the upper bound achievable with the INTERN-2.5 \gls{ssmodel}. This means that this model is not able to generate \glspl{ssm} that perform better than an \gls{miou} of $0.84$.

The results are clear: the \gls{spic} framework outperforms the other methods. The \gls{bpg} can achieve \gls{miou} performances close to what the \gls{spic} can achieve at $0.17$\gls{bpp} only when the compression is in the order of $1$ \gls{bpp}.

The other comparison, on the right, is on the \gls{fid} metric. Even in this case, the proposed \gls{spic} is able to outperform the other two methods.

Overall, the reconstructed image is visually more appealing and can better preserve the \gls{ssm}, as visually shown in \fref{fig: SPIC Visual_comparison_OUR_vs_BPG}.

\subsection{C-SPIC}\label{sec: SPIC results c-spic}

While the proposed \gls{spic} is able to outperform classical compression algorithms both visually and on different semantic metrics, it can still be improved by introducing the \gls{cspic} framework.

\begin{figure}[!t]
    \centering
    % Titles Row
    \begin{subfigure}{\textwidth}
        \centering
        \begin{minipage}{0.333\textwidth}
            \centering
            \textbf{BPG} (0.176 BPP)
        \end{minipage}%
        \begin{minipage}{0.333\textwidth}
            \centering
            \textbf{SPIC} (0.166 BPP)
        \end{minipage}%
        \begin{minipage}{0.333\textwidth}
            \centering
            \textbf{C-SPIC$^w$} (0.171 BPP)
        \end{minipage}
    \end{subfigure}
    
    % Images Row
    \begin{subfigure}{\textwidth}
        \centering
        \begin{subfigure}{0.333\textwidth}
            \centering
            \includegraphics[width=\linewidth, trim=330 140 50 40, clip]{Figures/SPIC/Classico/bpg40.png}
            \caption*{}
        \end{subfigure}%
        \begin{subfigure}{0.333\textwidth}
            \centering
            \includegraphics[width=\linewidth, trim=330 140 50 40, clip]{Figures/SPIC/Classico/our_bpg35.png}
            \caption*{}
        \end{subfigure}%
        \begin{subfigure}{0.333\textwidth}
            \centering
            \includegraphics[width=\linewidth, trim=330 140 50 40, clip]{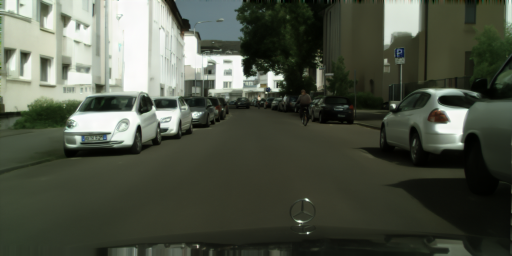}
            \caption*{}
        \end{subfigure}
    \end{subfigure}
    \vspace{-1cm}
    \caption[Visual comparison between \acrshort{bpg}, \acrshort{spic} and \acrshort{cspic}$^w$ on the "traffic sign" class]{Visual comparison of the traffic sign "P"  compressed with the \acrshort{bpg} algorithm at 0.176 \gls{bpp} (LEFT), the proposed \acrshort{spic} framework at 0.166 \acrshort{bpp}(CENTER), and the \gls{cspic}$^w$ at 0.171 \acrshort{bpp} (RIGHT).}
    %[Visual comparison for improved SPIC]{Visual comparison of the parking traffic sign detail between the image compressed with the \gls{bpg} algorithm at 0.176 \gls{bpp} (LEFT), the proposed \gls{spic} framework at 0.166 \gls{bpp}(CENTER), and the \gls{cspic}$^w$ with enhanced traffic signs at 0.171 \gls{bpp} (RIGHT).}
    \label{fig: SPIC detail traffic sign}
\end{figure}

In \fref{fig: SPIC detail traffic sign}, a detail of a "Parking" traffic sign is shown extracted from reconstructed images at similar \gls{bpp} values. The image on the right is the result of the classical \gls{bpg} compression algorithm at $0.176$ \gls{bpp}. It is possible to see that the traffic sign is blue and vaguely resembles a "P", although it is not very clear. In the middle, at $0.166$ \gls{bpp}, is the reconstruction obtained with the \gls{spic} framework. The traffic sign is blue, but the "P" has been completely removed.

Finally, the left image, obtained with the proposed \gls{cspic}$^w$ at $0.171$ \gls{bpp}, is clearly superior in terms of preservation of meaning. The sign is blue, and the "P" is clearly visible. By using only $0.004$ more \gls{bpp} than the \gls{spic} framework, the traffic sign is now perfectly reconstructed.
\footnote{The exact same visual result obtained with the \gls{cspic}$^w$ at $0.171$ \gls{bpp} can be achieved with the \gls{cspic}$^{w/o}$ at $0.173$ \gls{bpp}. Due to the small difference in terms of metrics between the two approaches, the comparison are evaluated only between the \gls{cspic}$^w$ and classical compression algorithms. The choice of \gls{cspic}$^{w/o}$ over \gls{cspic}$^w$ is only a matter of computational constraints at the transmitter and if the 10\% increase in \gls{bpp} can really make the difference.}
\begin{figure}[!t]
    \centering
    \begin{subfigure}[t]{0.49\textwidth}
        \centering
        \includegraphics[width=\textwidth, trim=5 10 40 20, clip]{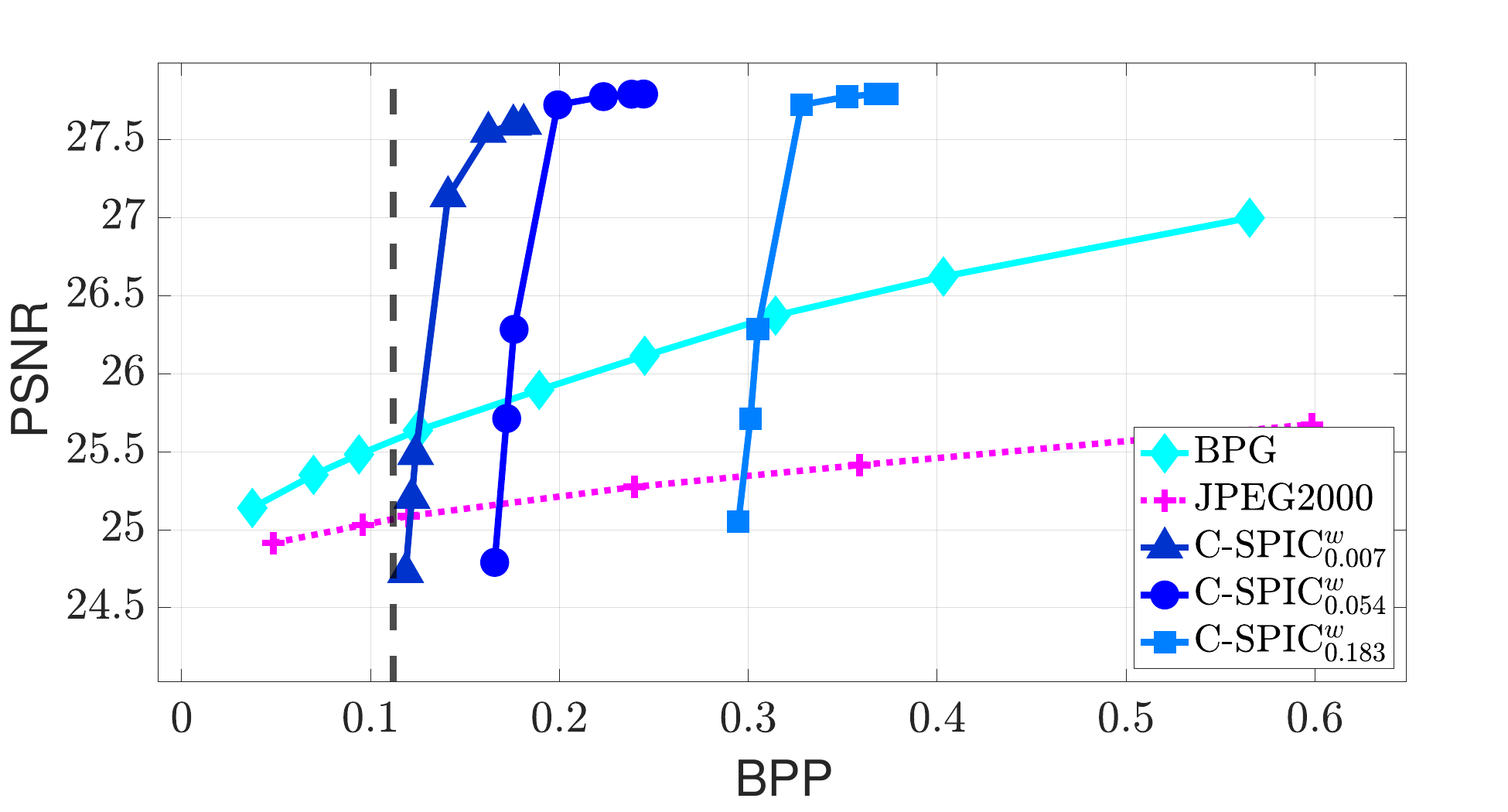}
        \caption*{(a)}
    \end{subfigure}%
    \begin{subfigure}[t]{0.49\textwidth}
        \centering
        \includegraphics[width=\textwidth, trim=5 10 40 20, clip]{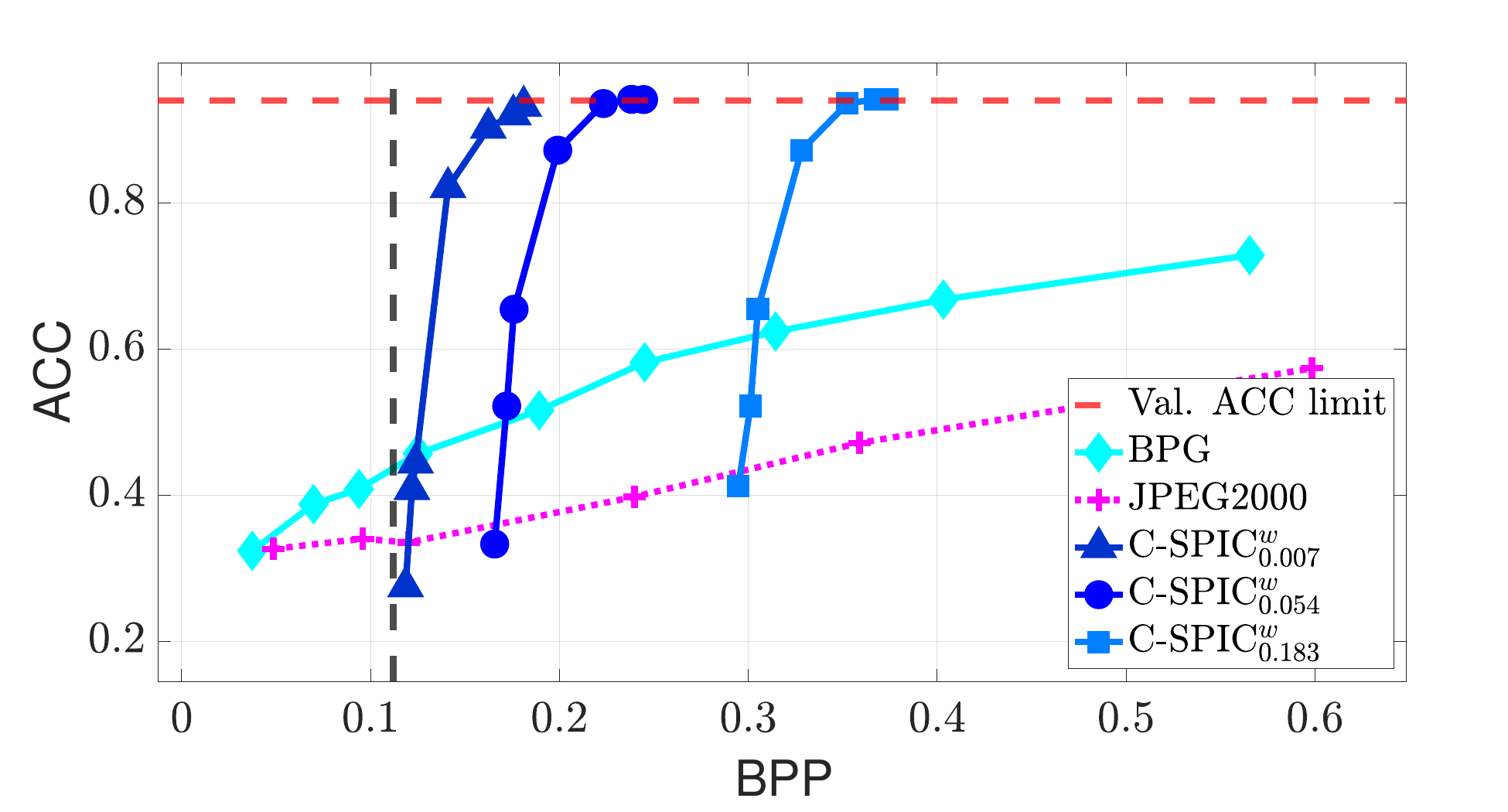}
        \caption*{(b)}
    \end{subfigure}

    \caption[Performance comparison between the \acrshort{cspic}$^w$ and classical compression algorithms]{Performance comparison between \acrshort{bpg}, \acrshort{jpeg2000} and \acrshort{cspic}$^w$ on the "traffic sign" class evaluated in term of masked \acrshort{psnr} and traffic signs classification accuracy.}
    \label{fig: SPIC cspic metrics comparison}
\end{figure}

In \fref{fig: SPIC cspic metrics comparison}, some numerical comparisons are presented. The color scheme is similar to that discussed in the previous section, with the red dotted horizontal line representing the upper bound for accuracy in traffic sign classification, as described in \sref{sec: GM evaluation metrics}. The black dotted vertical line again represents the amount of \gls{bpp} required to losslessly encode the \gls{ssm} and is placed at $0.112$ \gls{bpp}.

The performances of the proposed \gls{cspic}$^w$ are evaluated multiple times and reported with three different shades of blue. The difference between the three versions lies in the level of compression of the coarse image $\co$. In the \gls{cspic}$^w$, both the coarse image and the residual can be compressed at different levels. Since the effects of varying the quality of the coarse image were presented in \fref{fig: SPIC spic metrics comparison}, in this case, the quality of the coarse is fixed to three different values, and the quality of the residual is varied. The coarse image is compressed at these values of \gls{bpp}: $0.007$, $0.054$, and $0.183$. In the plot, the different lines are identified by different subscript, e.g. \gls{cspic}$_{0.007}^w$ indicates that the coarse has been compressed at $0.007$ \gls{bpp}.

The results show that, as soon as the artifacts of the lossy compression of the residual decreases, the proposed method outperforms the comparison approaches both on a classical metric like \gls{psnr}\footnote{The \gls{psnr} is evaluated only on the portion of the image containing the traffic signs.} and on the traffic sign classification accuracy.

These results are even more interesting when considering that the overall image still preserves the same qualities described in the \gls{spic} framework.\\

In conclusion, the \gls{spic} framework, along with its enhanced variant \gls{cspic}, represents a significant advancement in the field of \gls{sc} for image compression. By employing a modular architecture that distinctly handles the \gls{ssm}, the coarse image and the reconstruction, the \gls{spic} framework effectively balances compression efficiency with the preservation of essential semantic information. The introduction of the \gls{semcore} model at the receiver side enables high-fidelity image reconstruction, ensuring that semantically relevant details are maintained even at low \gls{bpp} levels.

The \gls{cspic} extension further refines this approach by incorporating a residual enhancement mechanism tailored for specific semantic classes. This enhancement ensures that smaller and more detailed objects, which are crucial for accurate semantic interpretation, are reconstructed with higher fidelity without imposing a significant additional \gls{bpp} overhead. Experimental results on the Cityscapes dataset clearly demonstrate that both \gls{spic} and \gls{cspic} outperform classical compression algorithms such as \gls{bpg} and \gls{jpeg2000} in terms of \gls{miou} and \gls{fid}, underscoring their superior capability in preserving semantic content.

These findings establish a robust foundation for further exploration and optimization in gls{sc}-based image compression algorithms and will be considered in the next chapter.

\chapter{Semantic Image Coding Using
Masked Vector Quantization}\label{ch: SQGAN}
The content of this chapter is entirely based on the following publication:
\begin{quotation}
    \noindent \textit{\textbf{\large SQ-GAN: Semantic Image Coding Using
Masked Vector Quantization}}\\
    \textit{Francesco Pezone, Sergio Barbarossa, Giuseppe Caire}
\end{quotation}

\section{Introduction}
In this chapter will be introduced the \gls{sqgan}, another \gls{ssm}-based image compression algorithms designed to address some limitations of the model proposed in the previous chapter. \\

Before continuing it is important to clearly define what constitutes the semantic relevant information in this context. The assumption that will be made is that the semantic relevant information consists of the following elements: (i) the reconstructed \gls{ssm} should preserve as much of the information of $\s$ as possible\footnote{In \cref{ch: SPIC} the lossless reconstruction of the \gls{ssm} was required. Now this constraint is relaxed allowing the lossy compression of the \gls{ssm}.}; (ii) the reconstructed image should retain as much of the \gls{ssm} as possible; (iii)  the overall
characteristics of the objects, such as colors and details, should be preserved; and (iv) some classes are more important than others and should be reconstructed better than the rest.\footnote{In \cref{ch: SPIC} this was optional and not required in the classic \gls{spic}, but only in the \gls{cspic}. In this chapter this is the default.}

The \gls{sqgan} introduces several innovations tailored to enhance semantic image compression. Firstly, the integration of semantic information into the compression process is achieved by modifying the \gls{maskvqvae} architecture. This modification involves the development of a \gls{samm}, which selectively prioritizes latent vectors associated with semantically relevant regions. By conditioning the masking process on the \gls{ssm}, the \gls{samm} ensures that critical classes such as "traffic signs" and "traffic lights" receive higher values of relevance score, thereby improving their reconstruction.

Secondly, to improve the efficiency of the image encoding, the \gls{sqgan} introduces the  \gls{spe}  that leverages the \gls{ssm} to assign different weights depending on the class of the objects. This facilitates the compression and preservation of details in significant regions of the image while reducing the focus on non-relevant regions.

Additionally, the \gls{sqgan} incorporates a multi-step adversarial training with a specifically designed Semantic-Aware Discriminator for image reconstruction. This discriminator is designed to reduce the focus of the discriminator on non-relevant regions to focus on the most important ones.

Finally, to address the challenge of underrepresented but crucial classes within the dataset, it is introduced the Semantic Relevant Classes Enhancement data augmentation technique. This novel augmentation method increases the prevalence of semantically important classes, such as "traffic signs" and "traffic lights", during the training phase. By augmenting the dataset with additional instances of these critical classes, the model exposure increases,  improving its ability to accurately reconstruct them, thereby enhancing the robustness and effectiveness of the compression process.\\

This chapter is structured as follows. In \sref{sec: SQGAN model description} will be introduced the architecture of the \gls{sqgan} and the details of the sub-networks $G_\s$ and $G_\x$ with the proposed \gls{spe} and the \gls{samm}. In \sref{sec: SQGAN training} will be discussed the multi-step training process of the \gls{sqgan} with the proposed Semantic Relevant Classes Enhancement data augmentation technique and the new Semantic-Aware Discriminator. In \sref{sec: SQGAN numerical results} will be shown the results and performances of the \gls{sqgan}.
\section{Model Architecture}\label{sec: SQGAN model description}
\begin{figure}[!t]
    \centering
    \includegraphics[width=0.9\textwidth]{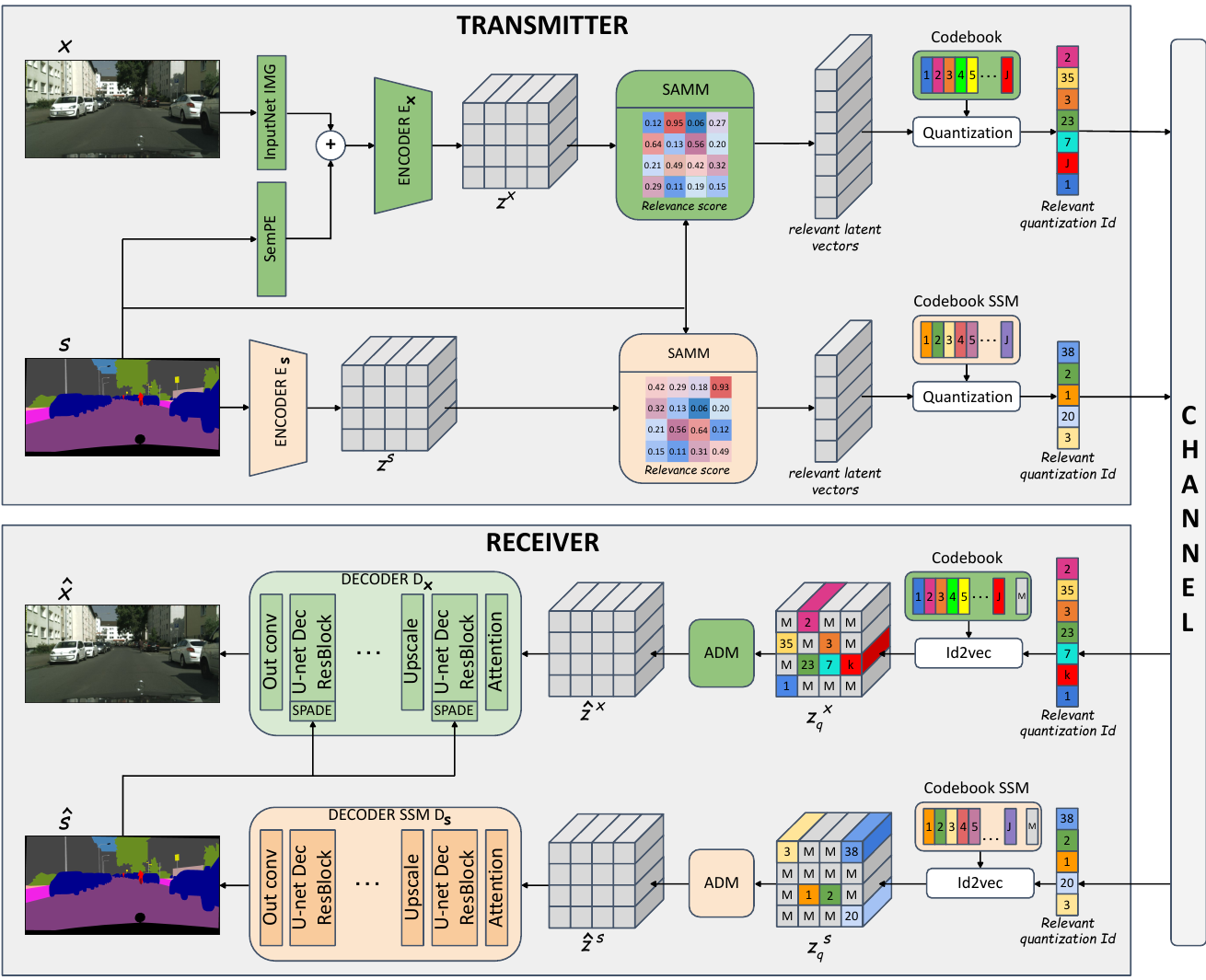}
    \caption[\acrshort{sqgan} architecture scheme]{Overview of the generator of the \acrshort{sqgan} architecture. The generator is composed of two sub-networks: $G_\s$ in beige and $G_\x$ in green. They are responsible for the compression and reconstruction of the \acrshort{ssm} and the image, respectively.}
    %[Scheme of the proposed SQ-GAN]{Sceme illustrating the architecture of the proposed SQ-GAN. It is composed of two sub-network ($G_\s$ and $G_\x$) responsible for the compression and reconstruction of the image and the SSM. The SAMM blocks are responsible for selecting only the most relevant latent vectors based on a relevance score.}
    \label{fig: SQGAN Scheme masked Sementic VQ-GAN}
\end{figure}
The \gls{sqgan} architecture depicted in \fref{fig: SQGAN Scheme masked Sementic VQ-GAN} consist of a main network $G$ composed of two sub-networks, $G_\x$ (green) and $G_\s$ (beige). They are responsible for the compression and reconstruction of the image $\x$ and the \gls{ssm} $\s$, respectively.\footnote{The presence of the apex/subscripts $\x$ and $\s$ will be used to refer to the specific pipeline.} Each sub-network consists of an encoder and a decoder pair: $E_\x$-$D_\x$ for the image and $E_\s$-$D_\s$ for the \gls{ssm} plus other additional blocks specific for the semantic vector quantization.

The encoder $E_\x$ processes the image $\x$ in conjunction with the \gls{ssm} $\s$  and the \gls{spe} producing a latent tensor $\z^\x$ with dimensions $C \times H_{16} \times W_{16}$. Similarly, the encoder $E_\s$ maps the \gls{ssm} $\s$ to a latent tensor $\z^\s$ of the same shape. Here, $C$ denotes the number of channels, while $H_{16} = \frac{H}{16}$ and $W_{16} = \frac{W}{16}$ represent the spatial dimensions reduced by a factor of 16.\footnote{In general the following notation if applied: $H_{n} = \frac{H}{n}$ and $W_{n} = \frac{W}{n}$.} 

To enhance compression efficiency while preserving semantic relevance, the \gls{samm} is integrated into both sub-networks. The \gls{samm} selectively prioritizes latent vectors based on their association with semantically important regions conditioned on the \gls{ssm}. As for the \gls{amm}, only $N_\x = m_\x \cdot K$ and $N_\s = m_\s \cdot K$ latent vectors are retained for the image and \gls{ssm}, respectively. The masking fractions $m_\x$ and $m_\s$ can be dynamically adjusted to control the compression level, allowing the model to operate at different compression rates.

This section will describe the architecture of the generator $G$ of the \gls{sqgan} in detail. Every block is discussed separately, starting from the encoder $E_\s$ and $E_\x$.

\subsection{Semantic Encoder}\label{sec: SQGAN semantic Encoder}
The first part of the sub-network $G_\s$ consists of the encoder block $E_\s$. The encoder is used to map the $n_c \times H \times W$ \gls{ssm} $\s$ to a latent tensor $\z^\s=E_\s(\s)$ with shape $C \times H_{16} \times W_{16}$, where $C$ is the number of channels and $H_{16}$ and $W_{16}$ are the final height and width.\\
$E_\s$ is formed by a repeated sequence of two \glspl{resblock} and one down-scaling layer, that will be referred to as \gls{resblockdown}. The average pooling down-scaling layer is used to halve the height $H$ and width $W$ of the \gls{ssm} after every iteration. After the first \gls{resblockdown}, the intermediate latent representation will have a shape of $\frac{H}{2}=H_{2}$ and $\frac{W}{2}=W_{2}$. By applying the \gls{resblockdown} for a total of 4 times the final latent tensor will have a spatial shape of $H_{16} \times W_{16}$. \\
The value of 4 consecutive \glspl{resblockdown} has been chosen to achieve a good balance between compression and detail retention. This is an important hyperparameter that strongly influences the results. Changing to 3 or 5, namely to $H_{8} \times W_{8}$ or $H_{32} \times W_{32}$, will change the performances of the whole architecture, causing poor compression in the first case and poor semantic preservation in the second. After an extensive ablation study it was concluded that $H_{16} \times W_{16}$ achieved the best performance.\\
%A total if 3 consecutive \glspl{resblockdown} will guarantee better reconstruction quality paying the price of higher \glspl{bpp}. On the other hand, the 5 \glspl{resblockdown} will compress more at the expense of the reconstruction quality. For this reason, after training the model with a different number of consecutive \glspl{resblockdown} it was concluded that 4 represent the best trade-off between compression and reconstruction.\\
The output of the last \gls{resblockdown} is used in the final multi-head self-attention layer of the encoder.\\
% This concludes the description of the semantic encoder $E_\s$. A block composed of four consecutive \gls{resblockdown} and one multi-head self-attention layer.  This block takes as input the \gls{ssm} $\s$ and outputs a latent tensor $\z^\s = E_\s(\s)$. The output shape of $\z^\s$ is $C \times H_{16} \times W_{16}$, where $C$ is a hyperparameter usually set to $C=256$.

\subsection{Image Encoder}\label{sec: SQGAN image Encoder}
\begin{figure}[!t]
    \centering
    \includegraphics[width=0.75\textwidth]{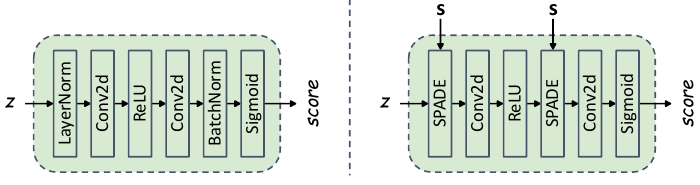}
    \caption[\acrshort{samm} architecture scheme]{Architectural diagram of the \acrshort{amm} as in \cite{Huang2023MaskedVQ-VAE} (LEFT), and the proposed \acrshort{samm} employing the \acrshort{spade} layer to introduce the \acrshort{ssm} conditioning (RIGHT).}
    %{Comparison between AMM on the left and the proposed SAMM on the right. The strength of the SAMM is the conditioning on the SSM $\s$ via the \gls{spade} layer.}
    \label{fig: SQGAN SemAdaptiveMask module vs old}
\end{figure}
At the same time, the image $\x$ is processed by the image encoder $E_\x$. This block is responsible for mapping the image $\x$ to the latent representation $\z^\x$ and it is conditioned by the \gls{ssm} $\s$. 
The structure of the encoder $E_\x$ is the same as the one of the encoder $E_\s$.
The main difference lies in the conditioning process. In fact, not all the parts of the image $\x$ have the same semantic meaning and relevance: for example  pedestrians are more important than sky. To help the encoder $E_\x$ in giving different importance to different parts, the image $\x$ is modified before being inserted in the encoder.\\
The proposed method draws inspiration from the version of the \gls{pe} introduced by Dosovitskiy et al. in transformer networks for vision tasks \cite{Dosovitskiy2021ViT}. By subdividing the images into $16 \times 16$ they made possible to assign a specific \gls{pe} vector to every patch. In this way the transformer network was able to  correctly interpret the relative positions of the different patches in the frame.

In this work, the implementation of the \gls{pe} as proposed in \cite{Dosovitskiy2021ViT} is not needed. Unlike transformer architectures, the spatial correlations are preserved in convolution-based architectures. However, the idea of giving different weights to different semantic regions of the frame is used to improve the overall performances.
For this reason in this work is proposed a variation of the \gls{pe}, referred to as \gls{spe}. The goal of the \gls{spe} is to provide the encoder $E_\x$ with a suitable transformation of $\s$ that can be used to influence the feature extraction process.\\
To obtain this \gls{spe}, the \gls{ssm} $\s$ is processed by a two layer \gls{cnn}. This network is designed to take into account the semantic classes of adjacent pixels and provide a meaningful, learnable transformation of $\s$. At the same time the image $\x$ is processed by a one layers \gls{cnn} called $InputNet_\x$. Similar to the classical \gls{pe}, the \gls{spe} is added to the output of the $InputNet_\x$ as:
\begin{equation}
    \h_{SemPE}(\x, \s) = InputNet_\x(\x) + SemPE(\s).
    \label{eq: SQGAN SemPE}
\end{equation}
This new augmented representation of the original image $\x$ can now be processed by the encoder $E_\x$ and transformed into the latent tensor $\z^\x = E_\x(\h_{SemPE}(\x, \s))$. 
The encoding process of $\x$ is now complete and the latent tensor $\z^\x$ has the same shape as $\z^\s$, $C \times H_{16} \times W_{16}$. 

\subsection{Semantic Conditioned Adaptive Mask Module}\label{sec: SQGAN SemAdaptiveMask}
After the encoding process, the image $\x$ and the \gls{ssm} $\s$ have been transformed to the respective latent tensors $\z^\x$ and $\z^\s$, of shape $C \times H_{16} \times W_{16}$. These tensors can be interpreted as being composed of $K=H_{16} \times W_{16}$ latent vectors in $C$ dimensions. The next step consist in selecting only the most relevant of these vectors. This selection is influenced by the relevance masking fraction $m_\x$ and $m_\s$. These values represent the fraction $N_\x=m_\x\cdot K$ and $N_\s=m_\s \cdot K$of the total $K$ latent vectors that will be selected by the proposed \gls{samm}.\\
The following discussion will consider only the latent tensor $\z^\x$ and the pipeline $G_\x$ to simplify the analysis since the process introduced in this section, in \sref{sec: SQGAN quantization} and in \sref{sec: SQGAN ADM} are exactly the same for both sub-networks.

The \gls{samm} is the proposed variation of the previously discussed \gls{amm}, see \sref{sec: GM mqvae}. The idea is to assign to everyone of the $K$ latent vectors a relevance score and then select only the fraction $m_\x$ of those with the highest relevance score. In the proposed \gls{samm} the relevance score is conditioned on the \gls{ssm} $\s$ and the masking fraction $m_\x$ can be adjusted dynamically. This allows the network to use the same weights to compress images at different levels of compression. The differences in architecture between \gls{amm} and \gls{samm} are shown in \fref{fig: SQGAN SemAdaptiveMask module vs old}.\\
While the classic \gls{amm}, on the left, is more suitable for general purpose applications, the \gls{samm} on the right is designed to take into account the semantic class of the different regions of the image. The new \gls{samm} is able to enforce this information thanks to the \gls{spade} normalization layer \cite{Park2019SPADE} used to replace all the classic normalization layers in the \gls{amm} to obtain a conditioned normalization based on the \gls{ssm}.\\
The \gls{samm} is a network that takes as input the latent tensor $\z^\x$ and the \gls{ssm} $\s$ and outputs a relevance score $s_k^\x \in [0,1]$ for each latent vector $\z_k^\x$. Of all the $\z_k^\x$ vectors composing $\z^\x$, only the $N_\x$ with the highest scores are selected. The final step involves the multiplication of the selected latent vectors by their respective relevance scores. This is done to allow the backpropagation to flow through the \gls{samm} and train its parameters, as discussed in \sref{sec: GM mqvae}.
%the process of selecting the elements with the highest relevance score is not differentiable and without the product it would be impossible to train the \gls{samm}.\\
%This completes the masking process where, starting from the output of the encoder, a list of the most relevant $N_\x$ latent vectors is obtained alongside with a list with the relative position of these vectors in the latent tensor $\z^\x$.\\

\subsection{Quantization and Compression}\label{sec: SQGAN quantization}
After the the most relevant $N_\x$ latent vectors have been selected the next step consist of the vector quantization and data compression. All the non-relevant $K-N_\x$ latent vectors are automatically dropped and never used again.\\
The vector quantization process is the same as described in \sref{sec: GM mqvae}. A learnable codebook $\C_\x$ is used to find the closest codeword $\e_j^\x$ to the selected score-scaled relevant vector $\z_k^{'\x}= \z_k^\x \times s_k^\x$ and replace it with the associated quantization index $e_j^\x$. In this application it was chosen to use $J=1024$ codewords of dimensionality $C=256$.\\
In the classical \gls{maskvqvae} this would have marked the end of this process. In fact, the only purpose of the vector quantization was to avoid redundancy and improve the final image quality. In this case however, this is additionally used to compress the data.\\
The compression of the quantization indices is performed by entropy coding. Every one of the $N_\x$ selected $e_j^\x$ is located in the respective location in a $H_{16} \times W_{16}$ shaped array. To each of the remaining empty $K-N_\x$ elements of the array is assigned an additional index $e_0^\x$ that is specific for the non-relevant vectors. This additional index brings the total possible amount of quantization indices to $J+1=1025$.\\
The index $e_0^\x$ appears with probability $1-\frac{N_\x}{K}$, while the other indices $e_j^\x$ appear with probability $\frac{N_\x}{K} \alpha_j$ for some $\alpha_j \geq 0$. Generally this probability is unknown and depends on the specific image $\x$ to be encoded. However, using the fact that entropy is maximized by the uniform probability, it is possible to upper-bound the entropy of the index sequence (in bits per index) by letting $\alpha_j = \frac{1}{J}$.
Therefore, the entropy coding rate for the index sequence is upper-bounded by:
\begin{equation}
    R_\x = h_2\!\left(\frac{N_\x}{K}\right) + log_2(J)\frac{N_\x}{K} ,
\end{equation}
where $h_2(p) = -(1-p)log_2(1 - p) - p log_2(p)$ is the binary entropy function. 
Since the index sequence length is $K$, the number of bits necessary to represent such a sequence is therefore upper-bounded by:
\begin{equation}
    B_\x = K R_\x = K\cdot h_2\!\left(\frac{N_\x}{K}\right) + log_2(J)\cdot N_\x . 
\end{equation}
However, in this work this value is further approximated. To maintain a linear relationship between the number of bits and the number of relevant latent vectors the condition $h_2(p) \leq 1$ is used. By substituting $h_2(p)=1$ the final amount of bits is expressed as follows:
\begin{equation}
    B_\x = K + log_2(J) \cdot N_\x = K(1+10\cdot m_\x),
\end{equation}
where $J=1024$. This corresponds to overestimating the bits required for compression and intentionally decreasing the performances of the proposed \gls{sqgan} only to simplify future analysis when linearity will be preferred (\sref{sec: EN_nn}). However, as will be pointed out in \sref{sec: SQGAN numerical results} the \gls{sqgan} is constantly outperforming classical image compression algorithms even by overestimating the require compression bits.\\   
By considering the same approach for the $N_s$ relevant vectors in $\z_\s$ and normalizing by the total number of pixels $H \times W$, it is possible to express the total amount of \gls{bpp} as a function of $m_\x$ and $m_\s$ as follows:
\begin{equation}
    BPP = \frac{B_\x + B_\s}{H \times W} = BPP_\x + BPP_\s = \frac{1}{256}[10(m_\x + m_\s) + 2].\footnotemark
    \label{eq: SQGAN BPP}
\end{equation}
\footnotetext{In the reminder of this chapter when referring to the \gls{bpp} it will be intended the total \gls{bpp} unless when differently specified.}
\subsection{Tensor Reconstruction and Adaptive De-Masking Module}\label{sec: SQGAN ADM}
At the decoder the quantization indices are rearranged in the $H_{16} \times W_{16}$ shaped array and copies of the codebooks $\C_\x$ and $\C_\s$ are stored at the receiver end. The following discussion will again consider only the sub-network $G_\x$, but the same identical approach is applied to $G_\s$.\\
The non-relevant elements encoded with index $e_0^\x$ are replaced with a learnable codeword $\bM_\x$ learned during training.\\
This new tensor composed of $\e_j^\x$'s and $\bM_\x$'s is now ready to be processed by the \gls{adm}. This step is the same as the one introduced in the \gls{maskvqvae} \cite{Huang2023MaskedVQ-VAE} and is used to gradually let the information flow from the relevant $\e_j^\x$ to the non-relevant placeholder $\bM_\x$. In fact, this additional codeword contains no information about $\x$. Being available only at the receiver, and being the same for all the non-relevant latent vectors will negatively affect the performances of the decoder in reconstructing $\x$. The \gls{adm} with the direction-constrained self-attention is actively reducing these unwanted effects.\\
The output $\hat{\z}^\x$ of the \gls{adm} is then used as input of the decoder $D_\x$, the same being true for $\hat{\z}^\s$.\\
From now on the flows for the two sub-networks $G_\x$ and $G_\s$ will diverge again. The next subsection will discuss the decoding process of the \gls{ssm} $\s$.

\subsection{Semantic Decoder}\label{sec: SQGAN semantic Decoder}
After the \gls{adm} has correctly processed the received relevant quantization indices it is time to decode the latent tensor to obtain the reconstruction $\hat{\s}$.\\
The structure of the decoder $D_\s$ is very similar to the mirrored version of the encoder $E_\s$. The first input layer is composed of the multi-head self-attention layer, after which the series of \glspl{resblockup} is placed. Every \gls{resblockup} is composed of two consecutive \glspl{resblock} and one up-scaling layer. The up-scaling is performed by copying the value of one element in the up-scaled $2\times2$ patch. The goal is to gradually transform the latent tensor $\hat{\z}^\s$ from a spatial shape of $H_{16} \times W_{16}$ back to the original spatial shape of $H \times W$, the same as the \gls{ssm} $\s$. This decoding and up-scaling process is completed by using four consecutive \glspl{resblockup}. \\
The final reconstructed \gls{ssm} $\hat{\s}$ is obtained by applying the argmax operator to assign any pixel to a specific semantic class.\\
The reconstructed \gls{ssm} $\hat{\s}$ can now be used to condition the reconstruction $\hat{\x}$ of the image $\x$. 

\subsection{Image Decoder}\label{sec: SQGAN image Decoder}
The reconstruction of $\x$ is obtained by applying the decoder $D_\x$ to $\hat{z}^\x$  and to the reconstructed \gls{ssm} $\hat{\s}$. This decoder has a structure similar to the mirrored version of the encoder $E_\x$, with some key conceptual differences. The input  multi-head self-attention layer remains the same. At the same time the sequence of four consecutive \gls{resblockup} is modified to incorporate the conditioning via the \gls{ssm}. The modifications focus on the normalization layers within the \glspl{resblock}, replacing every normalization layer with the \gls{spade} layer, as depicted in \fref{fig: GM resblock spade}. Without such conditioning, the decoder $D_\x$ will require much more computational power, longer training and bigger latent representation to have a similar level of \gls{ssm} retention. \\
%This ends the decoding process with the reconstruction of the image $\hat{\x}$. \\

%This end the discussion of the proposed \gls{sqgan} architecture, a model composed of different blocks that process the input $\x$ and $\s$ in two interconnected pipelines. The encoders $E_\x$ (\sref{sec: SQGAN image Encoder}) and $E_\s$ (\sref{sec: SQGAN semantic Encoder}) are the first blocks of these pipelines. They are responsible for the mapping of the input data in the respective latent tensors. From this point the two parallel pipeline are composed of the same blocks. The \gls{samm} are used to select the relevant latent vectors (\sref{sec: SQGAN SemAdaptiveMask}), then they are vector quantized, and the quantization indeces are compressed and send them to the receiver (\sref{sec: SQGAN quantization}). At the receiver the relevant quantization indices are restructured using a copy of the codebooks and the final latent tensor is processed (\sref{sec: SQGAN ADM}). After the information has flown from the relevant to the non-relevant latent vectors it is time of reconstructing $\hat{\x}$ (\sref{sec: SQGAN image Decoder}) and $\hat{\s}$ (\sref{sec: SQGAN semantic Decoder}).\\
%The interconnectivity in between parts in the \gls{sqgan} is one of the strength of this architecture but could also represent a problem if not controlled properly during training. Being composed of so many specialized parts could lead to some bad results if trained as a single mono-block model. In the next section the training process of the \gls{sqgan} will be discussed.

\section{Training and Inference}\label{sec: SQGAN training}
In the previous section, the architecture of the  \gls{sqgan} was presented as being composed of two interconnected sub-networks, $G_\x$ and $G_\s$, forming the single network $G$. This section focuses on the training process specifically designed for this architecture. 

The training of the entire network $G$ as a monolithic entity is not feasible due to the \gls{sqgan}'s structural complexity, which makes such an approach inefficient and impractical. The two sub-networks are in fact intended to reconstruct objects in distinct domains—the image domain and the \gls{ssm} domain—each requiring different loss functions. By using a single loss function for the whole process the performances will decrease and none of the sub-networks will perform at its best. Additionally, the masking fractions $m_\s$ and $m_\x$ are considered variable rather than fixed as in \cite{Huang2023MaskedVQ-VAE}. This allows  the \gls{sqgan} to compress images and \glspl{ssm} at various compression levels but also increases the training complexity. \\

To effectively train  the \gls{sqgan}, a multi-step approach consisting of three stages is employed: (i) train $G_\s$ using the original $\s$, (ii) train $G_\x$ with the original $\x$ and $\s$ and (iii) fine-tune the entire network $G$ using the original $\x$, original $\s$, and the reconstructed $\hat{\s}$ by freezing $G_\s$'s parameters and only fine-tuning $G_\x$'s parameters. 

The necessity of this multi-step approach arises from the distinct objectives and dependencies of the sub-networks. Initially training $G_\s$ ensures that the \gls{ssm} is accurately reconstructed independently of the rest. At the same time training $G_\x$ with the original $\x$ and $\s$ allows it to learn the conditional dependencies required for image reconstruction based on a reliable \gls{ssm}. However, since $G_\x$ is later required to be conditioned by the reconstructed $\hat{\s}$ rather than the original $\s$, the fine-tuning step is crucial to adapt $G_\x$ to mitigate the imperfections of $\hat{\s}$ that are not present in $\s$.

In addition to the multi-step training approach, another crucial challenge is the identification and preservation of semantically relevant information. To address this problem the training process has been further improved by proposing and incorporating the Semantic Relevant Classes Enhancement data augmentation technique and the Semantic-Aware discriminator network. The idea of these proposed approaches is to emphasize the importance of semantically relevant classes (i.e. "traffic signs", "traffic lights", "pedestrians" etc.), while preventing non-relevant classes (i.e. "sky", "vegetation" and "street") from being too dominant.

This section is divided as follows. \sref{sec: SQGAN Data Augmentation} will present the data augmentation process discussing commonly used techniques and the proposed Semantic Relevant Classes Enhancement method. \sref{sec: SQGAN training G_s} will focus on the training of $G_\s$ while \sref{sec: SQGAN training G_x} will focus on the training of $G_\x$ with the introduction of the proposed Semantic-Aware Discriminator. At the end \sref{sec: SQGAN training G} will focus on the final fine-tuning process.

\subsection{Data Augmentation}\label{sec: SQGAN Data Augmentation}
\begin{figure}[!t]
    \centering
    \adjustbox{valign=t}{\includegraphics[width=0.45\textwidth]{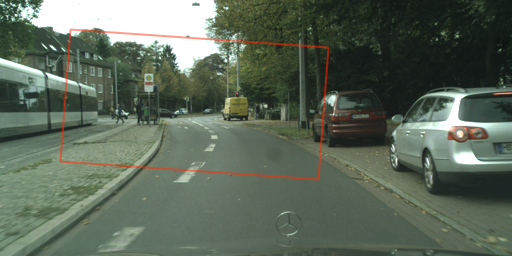}}%
    \hspace{5mm}
    \adjustbox{valign=t}{\includegraphics[width=0.45\textwidth]{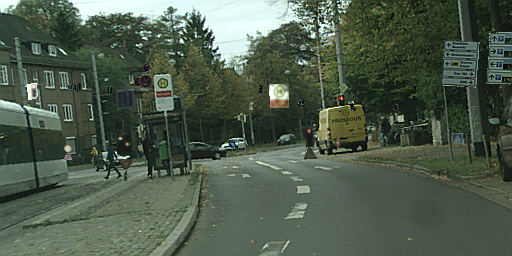}}\\[1mm]
    \adjustbox{valign=t}{\includegraphics[width=0.45\textwidth]{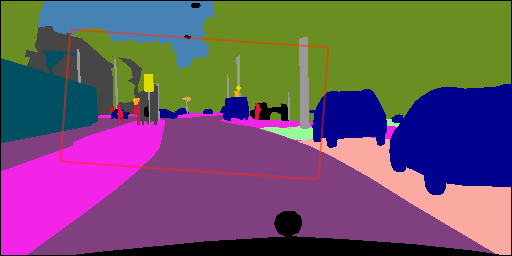}}%
    \hspace{5mm}
    \adjustbox{valign=t}{\includegraphics[width=0.45\textwidth]{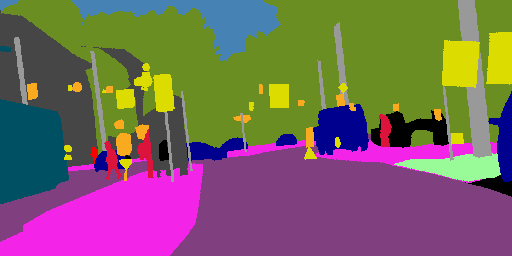}}
    \caption[Data Augmentation for \acrshort{sqgan}]{Effect of combined data augmentation techniques, including rotation, cropping and the proposed Semantic Relevant Classes Enhancement. On the left, the original image and \gls{ssm} and on the right the resulting augmented version.}
    \label{fig: SQGAN Data augmentation}
\end{figure}

This section introduces a novel data augmentation technique, the  Semantic Relevant Classes Enhancement, designed to solve the problem of underrepresented but semantically relevant classes in datasets. The core idea lies in the assumption that some specific classes, i.e. "traffic signs" and "traffic lights", are crucial in applications like autonomous driving but unfortunately underrepresented. In fact they often occupy a small portion of the frame and are not always present in every frame. Enhancing their representation is fundamental for improving the model's ability to efficiently reconstruct these critical classes.

This new data augmentation technique is fundamentally different from other proposed through the years \cite{Konushin2021dataAug1, Jockel2021dataAug2}. Prior methods often focus on swapping existing objects---for example, replacing a "turn left" sign with a "stop" sign---or employ complex \gls{nn} architectures to introduce new objects into images. These approaches can lead to increased computational costs or fail to adequately address the under-representation of critical classes. In contrast, the proposed technique is designed to be straightforward, efficient, and fast.

Below will be described the data augmentation pipeline used in this work and the effects are shown in \fref{fig: SQGAN Data augmentation}. At first the implementation of the commonly used techniques will be discussed, then the proposed data augmentation method.
\begin{itemize}[label={}]
    \item{\textbf{Random Rotation}:} With probability $p=0.5$, each pair $(\x,\s)$ is rotated by a random angle between $[-5^\circ,5^\circ]$. This small range prevents unrealistic scenarios, e.g. a pedestrian walking at a 45 degrees angle. This is performed to help the model generalize slight variations in camera orientation.
    
    \item{\textbf{Random Cropping}:} With probability $p=0.5$, this technique selects a random portion of the frame of the pair $(\x,\s)$. Unlike standard random cropping, the selected portion of the frame is constrained to avoid regions too close to the top or the bottom. This is because the top and bottom areas are often dominated by non-relevant classes such as "sky", "vegetation", or "street". By focusing on the central regions where relevant objects like "traffic signs" and "traffic lights" are more likely to appear, the model is encouraged to pay attention to smaller details in relevant classes. 
    
    \item{\textbf{Random Color Manipulation}:} With probability $p=0.3$, the brightness and saturation of the image $\x$ are altered. This simulates overexposed or underexposed images, such as those seen when a vehicle enters or exits a tunnel, helping the model deal with challenging lighting conditions.
    
    \item{\textbf{Random Semantic Relevant Classes Enhancement}:} This is the novel data augmentation technique specifically designed in this work to address the under-representation of "traffic signs" and "traffic lights". The process is based on the use of mini-batches of pairs $(\x,\s)$.

    For each image in a mini-batch, the \gls{ssm} is used to identify all instances of "traffic signs" and "traffic lights" present in that image. The main idea is to augment each image and its corresponding \gls{ssm} by adding more instances of these critical classes. This is achieved by copying "traffic signs" and "traffic lights" from other images within the same mini-batch and pasting them into the current image and \gls{ssm}. By increasing the presence of these objects, the model will be more exposed to them during training.

    The process begins by collecting all "traffic signs" and "traffic lights" from the images and \glspl{ssm} in the current mini-batch. For each image in the mini-batch, a random number $n$ between $0$ and $25$ is selected, representing the number of objects to add. Then, $n$ objects are randomly chosen from the collected set and are carefully placed into the image and its \gls{ssm}. Placement is done in such a way to avoid overlapping with existing instances of the same classes or with the other relevant class.
\end{itemize}

The final augmented pair $(\x,\s)$ is represented in \fref{fig: SQGAN Data augmentation} with the original pair on the left and the data augmented version on the right. Other than the cropping, rotation and color correction it is interesting to focus on the increase in "traffic signs" (yellow), and "traffic lights" (orange) in the augmented $\x$ and $\s$.

\subsection{Training of $G_\s$}\label{sec: SQGAN training G_s}
\begin{figure}[!t]
    \centering
    \includegraphics[width=1\textwidth]{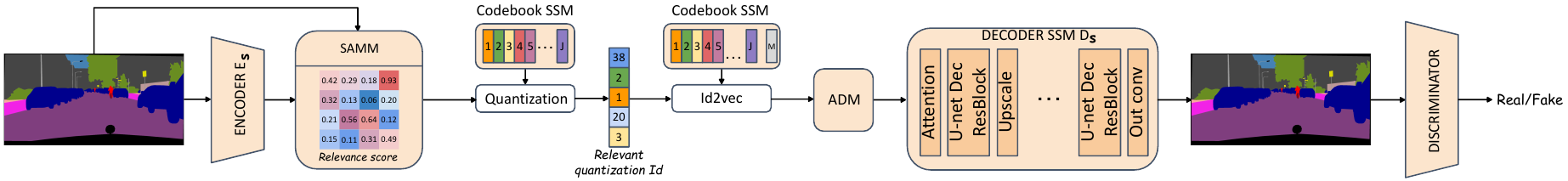}
    \caption[\acrshort{sqgan} training pipeline of the  \acrshort{ssm} generator]{Schematic representation of the semantic generator network $G_\s$ training pipeline.}
    \label{fig: SQGAN Gen_sem sqgan}
\end{figure}
After data augmentation, the training of the sub-network $G_\s$ can be implemented, as depicted in \fref{fig: SQGAN Gen_sem sqgan}. The training is based on an adversarial approach typical of \glspl{vqgan}. The input and output tensors of $G_\s$ have a shape of $n_c \times H \times W$, representing the one-hot encoded \glspl{ssm}. The loss function is designed as follows:
\begin{equation}
    \Loss_{\text{SQ-GAN}}^\s = \lambda_{\text{GAN}} \Loss_{\text{GAN}} + \lambda_{\text{WCE}} \Loss_{\text{WCE}}  + \lambda_{\text{vq}} \Loss_{\text{vq}} + \lambda_{\text{commit}} \Loss_{\text{commit}},
\end{equation}
In contrast to the classic \gls{vqgan} loss function discussed in \eref{eq: GM vq-gan total_loss}, the loss function adopted here uses the weighted cross-entropy loss $ \Loss_{\text{WCE}}$ as the reconstruction loss, rather than the \gls{l2} or the perceptual loss. This choice is motivated by the nature of the \gls{ssm}.

To emphasize the importance of semantically relevant classes over non-relevant ones, the weight factors $ \Loss_{\text{WCE}}$ are set as follows: 

\vspace{0.2cm}
\begin{tabular}{ll}
    \textbullet \;\; $\text{w} = 1$  & for the relevant classes "traffic signs" and "traffic lights". \\
    \textbullet \;\; $\text{w} = 0.85$  & for the classes "people" and "rider". \\
    \textbullet \;\; $\text{w} = 0.20$  & for the non-relevant classes "sky" and "vegetation". \\
    \textbullet \;\; $\text{w} = 0.50$  & for all the other classes.
\end{tabular}\\

These weight assignments encourage the model to focus more on reconstructing the relevant classes, thus guiding the \gls{samm} to prioritize the selection of latent vectors containing information about these classes.

The other important part involved in training is the discriminator network \cite{isola2017image2image}. It is composed of a convolutional layer, batch normalization layer and a leaky ReLU \cite{Bing2015Rectified} repeated 3 times and followed by the last convolutional layer that maps the output to a single number. This value is the output of the discriminator used to classify the \gls{ssm} as real or fake.

During training, the masking fraction $m_\s$ is varied randomly, selected from a set of values ranging from 5\% to 100\% with and expected value of 35\%. This approach allows the model to learn to compress the \gls{ssm} at various levels of compression.

The sub-network $G_\s$ is trained using the Adam optimizer \cite{Kingma2015Adam} with a learning rate of $10^{-4}$ and a batch size of 8. The training is conducted for 200 epochs with early stopping to prevent overfitting.

\subsection{Training of $G_\x$}\label{sec: SQGAN training G_x}
\begin{figure}[!t]
    \centering
    \includegraphics[width=1\textwidth]{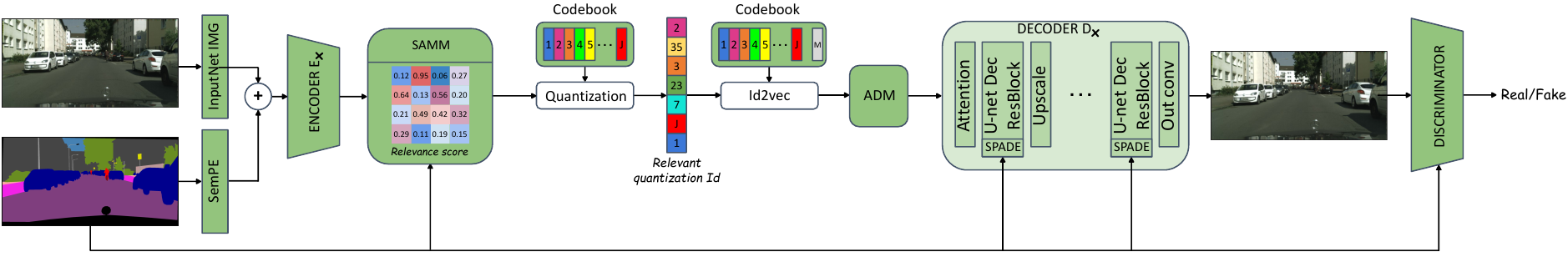}
    \caption[\acrshort{sqgan} training pipeline of the image generator]{Schematic representation of the image generator network $G_\x$ training pipeline.}
    \label{fig: SQGAN Gen_img sqgan}
\end{figure}
The training of the sub-network $G_\x$ follows a similar approach to that of $G_\s$. At this stage, $G_\x$ is trained using the original image $\x$ and the original \gls{ssm} $\s$. The training pipeline is illustrated in \fref{fig: SQGAN Gen_img sqgan}, highlighting that the true \gls{ssm} $\s$ is used to condition the decoder.\footnote{This is in contrast with the functioning of the network $G$ where the reconstructed \gls{ssm} $\hat{\s}$ is used to condition the discriminator.}

An adversarial approach typical of \glspl{vqgan} is employed for training, with the loss function defined as:
\begin{equation}
    \Loss_{\text{SQ-GAN}}^\x = \lambda_{\text{GAN}} \Loss_{\text{GAN}} + \lambda_{WL_2} \Loss_{WL_2}  + \lambda_{\text{perc}} \Loss_{\text{perc}} + \lambda_{\text{vq}} \Loss_{\text{vq}} + \lambda_{\text{commit}} \Loss_{\text{commit}},
\end{equation}
The weighted $l_2$ loss $\Loss_{WL_2}$ adjusts the importance of different semantic classes as follows:

\vspace{0.2cm}
\begin{tabular}{ll}
    \textbullet \;\; $\text{w} = 1$    & for the relevant classes "traffic signs" and "traffic lights". \\
    \textbullet \;\; $\text{w} = 0.55$ & for the classes "people" and "rider". \\
    \textbullet \;\; $\text{w} = 0$    & for the non-relevant classes "sky" and "vegetation". \\
    \textbullet \;\; $\text{w} = 0.15$ & for all other classes.
\end{tabular}\\

Among all the choices of weights the most interesting is the one concerning the weights for "sky" and "vegetation", set to zero. This choice has been made to force the network not to consider the pixel-by-pixel reconstruction of these classes. In fact, the real color of the sky and trees in the image are semantically non-relevant. 

The perceptual loss is instead the part that ensures that the reconstructed image maintains visual similarity to the original, preventing unrealistic alterations such as an unnatural sky color.

Another important improvement is in the adversarial loss $\mathcal{L}_{\text{GAN}}$ and involves the discriminator network. In this work a new version of the discriminator is proposed to avoid the focus on non-relevant regions of the image.

\subsubsection{Semantic-Aware Discriminator}
\begin{quote}
The discriminator $D_{disc}^\x$ plays a crucial role in adversarial training by determining whether an image reconstructed by $G_\x$ is real or fake. It achieves this by classifying images between real and fake images, and is trained on both $\x$ and $\hat{\x}$  to identify the characteristics that make an image appear authentic. However, a potential drawback is that the discriminator might focus on non-relevant parts of the image to influence the classification. For instance, it might prioritize vegetation details, and classify images as real only if the leaves on the trees have a certain level of detail. This will force the generator $G_\x$ to reconstruct images with better vegetation details to fool the discriminator. Unfortunately, this is not optimal for the structure of the \gls{sqgan}. Giving more importance to the vegetation will decrease the importance of other classes, thus causing the \gls{samm} module  to select the wrong latent vectors as relevant.

In recent years various articles have proposed ways to modify the discriminator to introduce various conditioning. For example, \cite{Oluwasanmi2020condDiscr} conditioned the discriminator on the \gls{ssm} to improve \gls{sseg} retention, while \cite{Chen2020ssd} enforced the discriminator to focus on high-frequency components. However, these methods do not adequately address the issue at hand. 

To achieve the desired performance, it is essential to adjust the discriminator to minimize its focus on non-relevant regions. To this scope it is possible to consider these key observations about the discriminator's behavior:
\begin{itemize}
    \item For a trained discriminator $D_{disc}$ and two images $\x$ and $\y$, $D_{disc}(\x)$ is likely similar to $D_{disc}(\y)$ if both images originate from the same data distribution, i.e., $p_{\x} = p_{\y}$. However, $D_{disc}(\x) = D_{disc}(\y)$ is not guaranteed unless the images are identical or indistinguishable by the discriminator.
    \item If $p_{\y}$ differs from $p_{\x}$, $D_{disc}(\y)$ will likely differ from $D_{disc}(\x)$. The output difference is primarily influenced by the aspects of $p_{\y}$ that deviate from $p_{\x}$.
    \item The greater the difference between $p_{\x}$ and $p_{\y}$, the higher the uncertainty in predicting $D_{disc}(\y)$ based on $D_{disc}(\x)$.
\end{itemize}
Based on these insights, a technique is proposed to reduce the discriminator's focus on non-relevant semantic classes.

The approach consist in artificially modifying the reconstructed image $\hat{\x}$ before it is evaluated by the discriminator. This modification aims to minimize the differences in non-relevant regions between $\x$ and $\hat{\x}$, bringing the data distribution of these regions of $\hat{\x}$ closer to the real distribution of $\x$. For example, if the generator reconstructs a tree with dark green leaves when in reality they are light green, the color in $\hat{\x}$ will be artificially shifted toward light green. 

This artificial editing is performed by considering the residual between the real and reconstructed images, defined as:
\begin{equation}
    \br = \x - \hat{\x}.
\end{equation}
By masking this pixel-wise difference between the two images and adding back a fraction of the residual to $\hat{\x}$ it is possible to obtain the new image:
\begin{equation}
    \hat{\x}_{rel} = \hat{\x} + \w_{rel} \odot \br,
\end{equation}
where $\w_{rel}$ is the re-scaling relevance tensor with the following values:

\vspace{0.2cm}
\begin{tabular}{ll}
    \textbullet \;\; $\w_{rel} = 0.90$    & for the class "sky". \\
    \textbullet \;\; $\w_{rel} = 0.80$ & for the class "vegetation". \\
    \textbullet \;\; $\w_{rel} = 0.40$    & for the class "street". \\
    \textbullet \;\; $\w_{rel} = 0$ & for the other classes.
\end{tabular}\\

This means that, for instance, 80\% of the difference between the shades of green leaves is removed before presenting the image to the discriminator. The modified vegetation in $\hat{\x}_{rel}$ will appear much closer to the real light green in $\x$, thus reducing the discriminator's focus on this non-relevant region.

It is important to acknowledge that this approach negatively impacts the generator's ability to accurately reproduce these specific non-relevant classes. Nevertheless, the perceptual loss still considers the entire image, guiding the generator to reconstruct the "sky," "vegetation," and "streets" to maintain overall realism.
\end{quote}
The proposed approach is a simple yet effective way to reduce the discriminator's focus on non-relevant regions. Alongside the weighted \gls{l2} loss these techniques allow the generator to focus on the relevant parts of the image.

Similarly to $G_\s$, the sub-network $G_\x$ is trained for different masking fractions $m_\x$. These values are selected from a finite set ranging from $5\%$ to $100\%$ with an expected value of $35\%$.
The structure of the discriminator $D_{disc}^\x$ is the same as the one used in the training of $G_\s$ and the training is performed using the Adam optimizer with a learning rate of $10^{-4}$ with batch size of $8$. The model is trained for $200$ epochs with early stopping.

\subsection{Fine-tuning of $G$}\label{sec: SQGAN training G}

The final step involves fine-tuning the entire network $G$ by freezing the parameters of $G_\s$ and updating only those of $G_\x$. This fine-tuning addresses the scenario where the original \gls{ssm} $\s$ is unavailable at the receiver, that is when the \gls{sqgan} is used for \gls{sc}.

This fine-tuning is essential since the quality of $\hat{\s}$ may not perfectly match that of the original $\s$, as it is influenced by the masking fraction $m_\s$.

Fine-tuning allows the model to adapt to variations in the quality of $\hat{\s}$ by training $G_\x$ with the reconstructed \gls{ssm}. This ensures that $G_\x$ can effectively reconstruct images based on $\hat{\s}$ despite its imperfections.

The loss function and the Semantic-Aware Discriminator $D_{disc}^\x$ remain identical to those used in the training of $G_\x$. Fine-tuning is conducted over 100 epochs with early stopping to prevent overfitting. During each iteration, both masking fractions $m_\x$ and $m_\s$ are randomly selected from the same distribution. The Adam optimizer is employed with a learning rate of $10^{-4}$ and a batch size of 8.

\section{Results} \label{sec: SQGAN numerical results}
This section will focus on the performances of the proposed \gls{sqgan} evaluated using the 500 pairs of images and \glspl{ssm} from the validation set of the Cityscape dataset. \footnote{More details on the dataset can e found in \sref{sec: SPIC results}}

This chapter is structured as follows. In \sref{sec: SQGAN result samm} the latent vector selection performed by the \gls{samm} will be discussed to show which are the parts of the $\x$ and $\s$ that are considered semantically relevant. Additionally, this section will show how varying the masking fractions $\m_\x$ and $\m_\s$ affects the final reconstruction. In \sref{sec: SQGAN result comparison} the performances of the proposed \gls{sqgan} will be compared with classical compression algorithms like \gls{bpg} and \gls{jpeg2000}.

\subsection{SAMM Performance}\label{sec: SQGAN result samm}
\begin{figure}[!t]
    \centering
    \adjustbox{valign=t}{\includegraphics[width=0.45\textwidth]{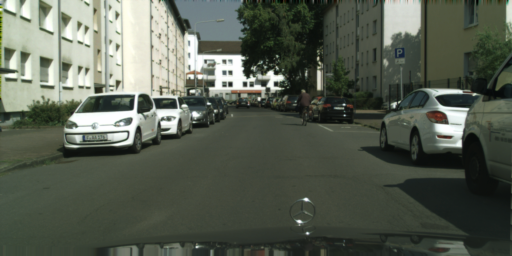}}%
    \hspace{5mm}
    \adjustbox{valign=t}{\includegraphics[width=0.45\textwidth]{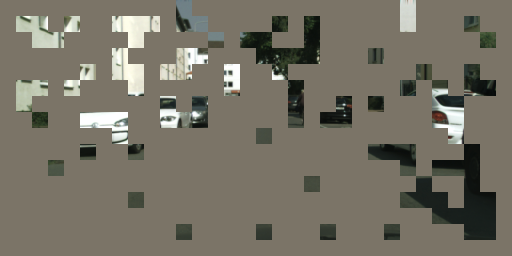}}\\[1mm]
    \adjustbox{valign=t}{\includegraphics[width=0.45\textwidth]{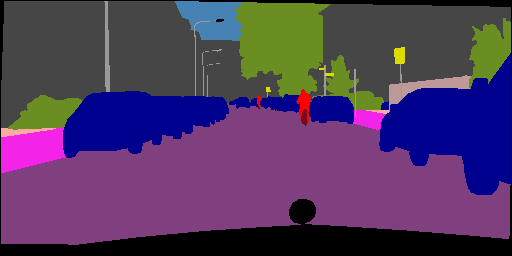}}%
    \hspace{5mm}
    \adjustbox{valign=t}{\includegraphics[width=0.45\textwidth]{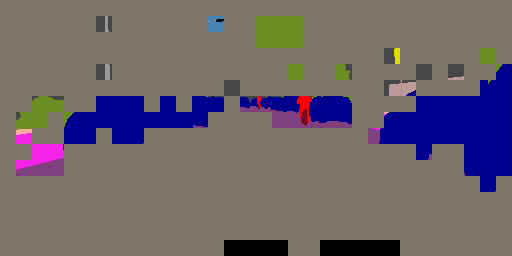}}
    \caption[Visual representation of the effect of the \acrshort{samm} module]{Visual representation of the latent tensor selection of the \acrshort{samm} projected in the image and \acrshort{ssm} domain. In both cases the masking has been fixed to $m_\x=m_\s=0.20$ and the region considered semantically relevant are shown on the right.}
    \label{fig: SQGAN masked images}
\end{figure}

\begin{figure}[!t]
    \centering
    % Text boxes above columns
    \begin{tabular}{>{\centering\arraybackslash}m{0.24\textwidth} 
                    @{\hspace{1mm}}>{\centering\arraybackslash}m{0.24\textwidth} 
                    @{\hspace{1mm}}>{\centering\arraybackslash}m{0.24\textwidth} 
                    @{\hspace{1mm}}>{\centering\arraybackslash}m{0.24\textwidth}}
        Original & 
        $m_\x=0.95, \;m_\s=0.15$ & 
        $m_\x=0.55, \;m_\s=0.55$ & 
        $m_\x=0.15, \;m_\s=0.95$ \\
    \end{tabular}
    \\% Adjust space between text and images

    % Images with corresponding SSM overlayed on the top row
    \begin{tabular}{>{\centering\arraybackslash}m{0.24\textwidth} 
                    @{\hspace{1mm}}>{\centering\arraybackslash}m{0.24\textwidth} 
                    @{\hspace{1mm}}>{\centering\arraybackslash}m{0.24\textwidth} 
                    @{\hspace{1mm}}>{\centering\arraybackslash}m{0.24\textwidth}}
        \begin{tikzpicture}
            \node[anchor=south west,inner sep=0] (image1) at (0,0) {\includegraphics[width=\linewidth]{Figures/SQ-GAN/real_image.png}};
            \node[anchor=south east,inner sep=0] at (image1.south east) {\includegraphics[width=0.45\linewidth]{Figures/SQ-GAN/real_ssm.png}};
        \end{tikzpicture} & 
        \begin{tikzpicture}
            \node[anchor=south west,inner sep=0] (image4) at (0,0) {\includegraphics[width=\linewidth]{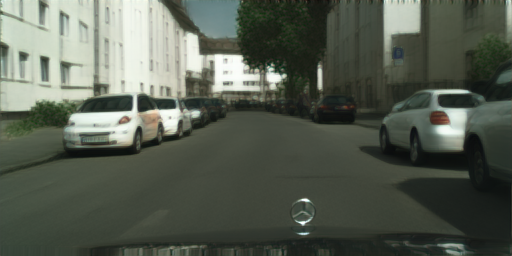}};
            \node[anchor=south east,inner sep=0] at (image4.south east) {\includegraphics[width=0.45\linewidth]{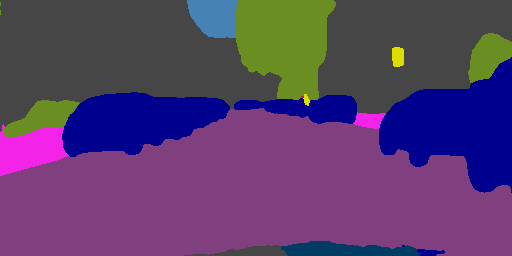}};
        \end{tikzpicture} & 
        \begin{tikzpicture}
            \node[anchor=south west,inner sep=0] (image3) at (0,0) {\includegraphics[width=\linewidth]{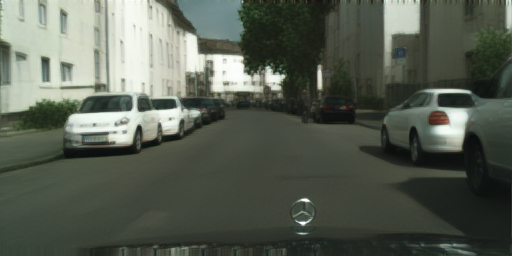}};
            \node[anchor=south east,inner sep=0] at (image3.south east) {\includegraphics[width=0.45\linewidth]{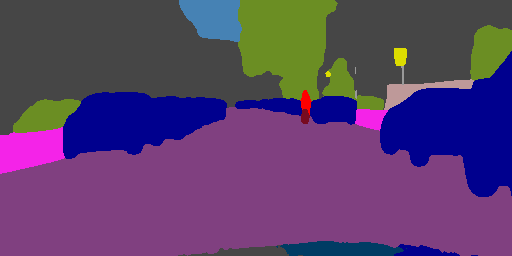}};
        \end{tikzpicture} & 
        \begin{tikzpicture}
            \node[anchor=south west,inner sep=0] (image2) at (0,0) {\includegraphics[width=\linewidth]{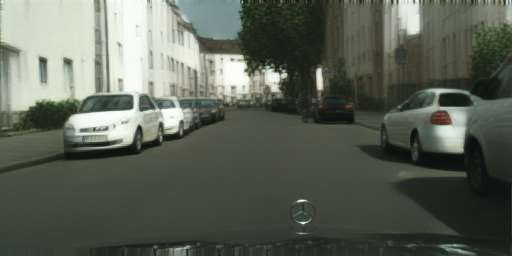}};
            \node[anchor=south east,inner sep=0] at (image2.south east) {\includegraphics[width=0.45\linewidth]{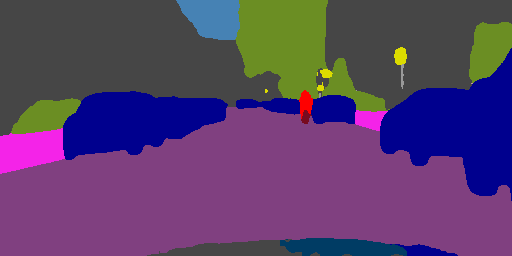}};
        \end{tikzpicture} \\

        % new row for SSM
        \includegraphics[width=\linewidth]{Figures/SQ-GAN/real_ssm.png} & 
        \includegraphics[width=\linewidth]{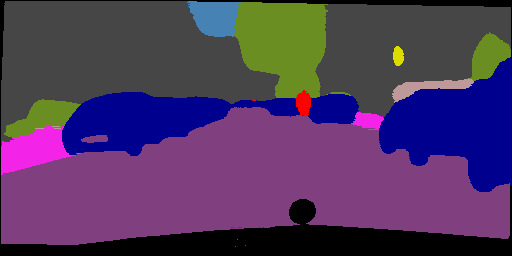} & 
        \includegraphics[width=\linewidth]{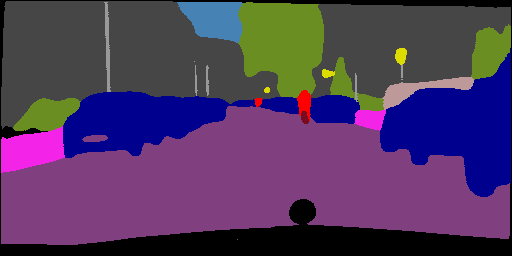} & 
        \includegraphics[width=\linewidth]{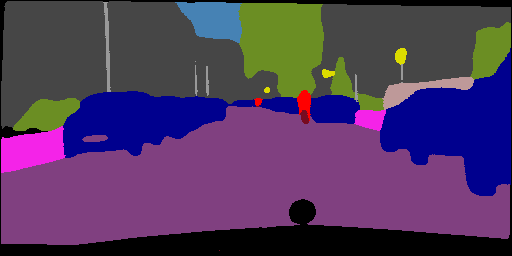} \\
    \end{tabular}
    
    \caption[Visual comparison between different results at different masking fractions]{Visual comparison between the same image and \acrshort{ssm} at different masking reactions $m_\x$ and $m_\s$. The original image and \acrshort{ssm} are shown on the left. The upper row shows the reconstructed $\hat{\x}$ and the generated \gls{ssm} using the \gls{sota} \gls{ssmodel} INTERN-2.5 \cite{Wang2022internimage}. The bottom row shows the reconstructed \gls{ssm} $\hat{\s}$. All pairs $(\hat{\x}, \hat{\s})$ are obtained at $0.05$\gls{bpp}.}
    \label{fig: SQGAN visual result changing masking}
\end{figure}
The \gls{sqgan} adopt various techniques to force the model to correctly reconstruct the relevant regions of the image. The weighted loss function based on the semantic classes, the Semantic-Aware Discriminator $D_{disc}^\x$, and the Semantic Relevant Classes Enhancement data augmentation have been designed with this idea in mind. Their scope is to guide the \gls{samm} in identifying and selecting the relevant latent vectors $\z_k^\x$ and $\z_k^\s$. By prioritizing certain regions over other the \gls{samm} learns to assign the truly relevant latent vectors a higher score. 

Since the selection is performed on the latent tensor, the effect of the \gls{samm} has to be projected from the latent tensor to the image and \gls{ssm} domain. These projections are  illustrated in \fref{fig: SQGAN masked images}.\footnote{The projection showed in figure \ref{fig: SQGAN masked images} has been realized by pivoting on the convolutional nature of the two encoders that allow the preservation of the spatial correlation between the latent tensor and the original frame.} 
In the left column the original $\x$ and $\s$ are represented, while the right column shows which are the regions associated to the latent vectors that the \gls{samm} consider more relevant. In this analysis the masking fractions have been fixed to $m_\x=m_\s=0.20$.

It is immediately evident that $G_\x$ and $G_\s$ consider different regions as relevant. The \gls{samm} in $G_\s$ focuses on regions with the most change in semantic classes. The street, building and sky requires very few dedicated latent vectors to be represented. On the contrary the parts of the \gls{ssm} containing relevant classes are strongly  preferred. 

In a similar way the \gls{samm} in $G_\x$ shows a strong preference for relevant classes like cars and people. However, it also focuses on areas previously ignored, such as the street and buildings. In fact, thanks to the conditioning on the \gls{ssm}, the sub-network $G_\x$ knows the location of every object and their shape and can focus on different aspects like colors and textures. For this reason the streets and sky will require some latent vectors to reconstruct the colors correctly. However, most of the latent vectors are selected from the truly relevant regions.

In both cases the \gls{samm} tends to prefer regions that contain more semantically relevant objects. This is a direct effect of the various techniques adopted to train the model as described in \sref{sec: SQGAN training}. \\

Another important analysis examines the effects of different masking fractions on the output. In fact increasing $m_\x$ and $m_\s$ is expected to increase the overall quality of $\hat{\x}$ and $\hat{\s}$ respectively, while decreasing is expected to do the opposite. \fref{fig: SQGAN visual result changing masking} shows visually how masking fractions influences the reconstructed outputs.

The column on the left shows the original $\x$ and $\s$. The other columns show on top the reconstructed $\hat{\x}$ and the generated \gls{ssm} obtained form $\hat{\x}$ via the INTERN-2.5 \gls{ssmodel} \cite{Wang2022internimage}. The bottom row shows the reconstructed $\hat{\s}$. All the pairs of $\hat{\x}$ and $\hat{\s}$ have been reconstructed, fixing the compression level to a total amount of $0.05$\gls{bpp} and by letting the masking fractions $m_\x$ and $m_\s$ vary.

The first consideration is the reconstructed $\hat{\x}$. It is evident that the presence of an object in $\hat{\x}$ is influenced more by the quality of $\hat{\s}$ and thus on $m_\s$  than on $m_\x$. The lower the amount of object and detail in $\hat{\s}$ the lower will the amount that $\hat{\x}$ is able to retain be. This is clearly be seen by comparing the $\hat{\s}$ on the bottom row and the generated \gls{ssm} in the upper row. If $\hat{\s}$ does not contain some detail, that same detail is not present in the generated \gls{ssm}.\\
The other aspect is the influence of $m_\x$ on the quality. As expected the increase of $m_\x$ tends to improve the overall image quality especially on the non-relevant details. The relevant details are reconstructed and prioritized even for low values of $m_\x$, while the windows of the building are better reconstructed when $m_\x=0.95$. This is important since it shows that the model is able to first reconstruct relevant classes and then improve the non-relevant.

\begin{figure}[!t]
    \centering
    \begin{minipage}[b]{0.49\textwidth}
        \centering
        \includegraphics[width=\textwidth]{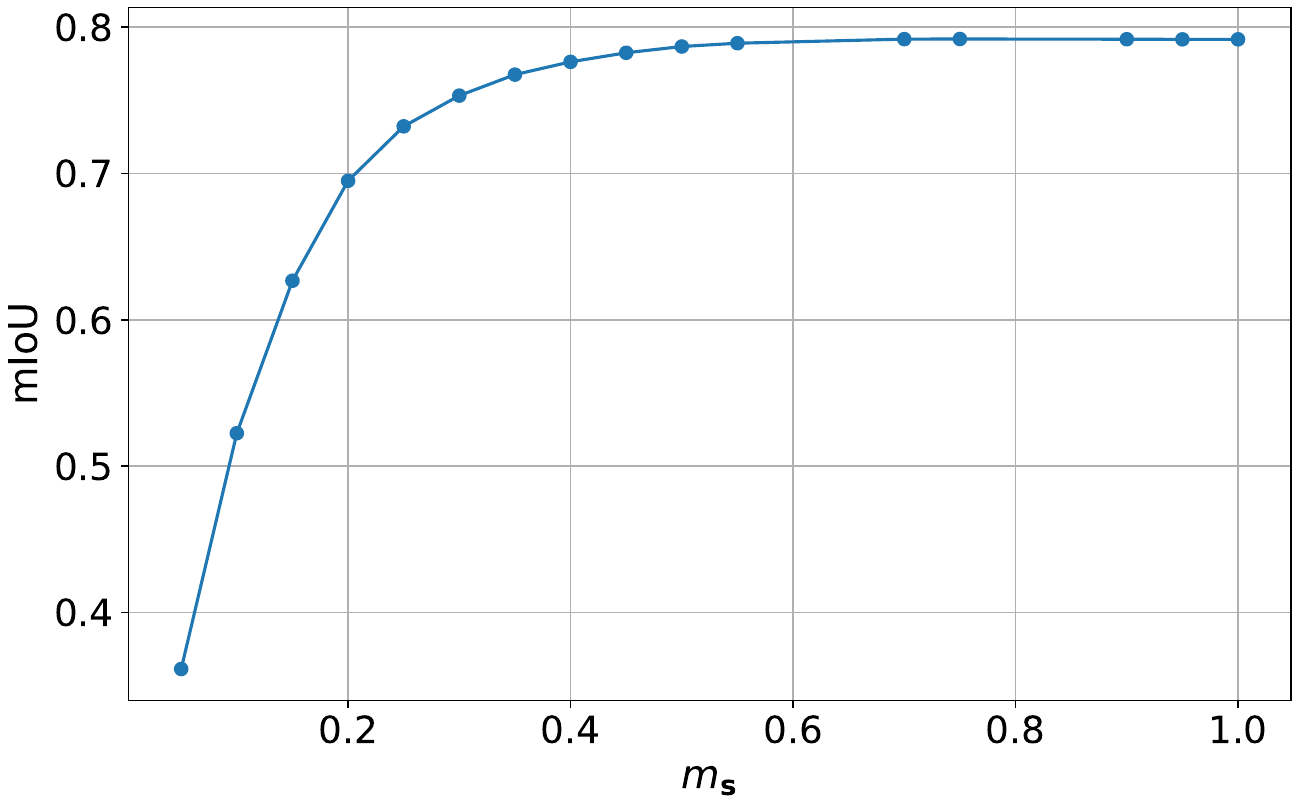}
        \caption[\acrshort{ssm} retention as a function of the masking fraction]{\acrshort{ssm} retention  evaluated between the true $\s$ and the reconstructed $\hat{\s}$ with the \gls{miou} metric as a function of the masking fraction $m_\s$. As the masking fraction $m_\s$ increases the network $G_\s$ is able to better reconstruct the \gls{ssm}. However, the increase of performances reaches a plateau from $m_\s \geq 0.2$ ($BPP_\s=0.011$\gls{bpp}).}
        \label{fig: SQGAN miou vs m_s}
    \end{minipage}
    \hfill
    \begin{minipage}[b]{0.49\textwidth}
        \centering
        \includegraphics[width=\textwidth]{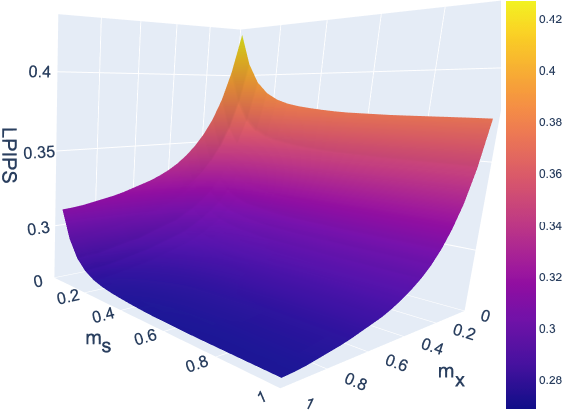}
        \caption[\acrshort{lpips} as a function of the masking fractions]{\acrshort{lpips} evaluated between $\x$ and $\hat{\x}$ as the masking fractions $m_\s$ e $m_\x$ vary.}
        \label{fig: SQGAN lpips 3d plot}
    \end{minipage}
\end{figure}

The other consideration is on the reconstructed $\hat{\s}$. It is possible to see that the change in masking fractions $m_\s$ influences the reconstruction of the \gls{ssm} up to a certain value. Both the reconstructed $\hat{\s}$ obtained at $m_\s=0.55$ and $m_\s=0.95$ look in fact almost identical.\\
This behavior can be seen in \fref{fig: SQGAN miou vs m_s} where the \gls{miou} between $\s$ and $\hat{\s}$ is reported as a function of $m_\s$. For very low values of $m_\s$ the performances are not optimal. As soon as $m_\s \geq 0.20$, the model is able to reconstruct $\hat{\s}$ with a decent level of semantic retention. To give a term of comparison the value of $m_\s =0.20$  correspond to a compression of $BPP_\s=0.011$\gls{bpp}. This means that at this low value of \gls{bpp} the model is already able to preserve a lot of details in the \gls{ssm}. After the value of $m_\s=0.55$ (equivalent to $BPP_\s=0.025$\gls{bpp}), there is no further advantage in increasing the masking fraction. After this value, all the relevant latent vectors characterizing the \gls{ssm} have been used. Adding other vectors will only increase the amount of redundant information, thus not improving the reconstructed \gls{ssm}. This is the opposite of what was discussed about $m_\x$, where even for values close to 1 the improvements were still visible on the final output.

Similar to the result in \fref{fig: SQGAN miou vs m_s} it is interesting to see how both $m_\x$ and $m_\s$ influence $\hat{\x}$. For this scope in \fref{fig: SQGAN lpips 3d plot} the \gls{lpips} evaluated on $\hat{\x}$ is reported. The effects of both masking fractions are clear. As expected, the performances along the $m_\s$ axis are influencing the output only for values of $m_\s \leq 0.55$. The influence on $m_\x$ is instead along the whole range from 0 to 1. This is perfectly explaining the visual results analysed in \fref{fig: SQGAN visual result changing masking}.

The results discussed in this section help to understand how the model performs as a function of the masking fractions $m_\x$ and $m_\s$. However, the most important part is to understand how the model is performing in comparison with other data compression techniques. This will be discussed in the next section.

\subsection{Visual Results and Comparisons with Classical Approaches}\label{sec: SQGAN result comparison}
\begin{figure}[!t]
    \centering
    % Text boxes above columns
    \begin{tabular}{>{\centering\arraybackslash}m{0.24\textwidth} 
                    @{\hspace{1mm}}>{\centering\arraybackslash}m{0.24\textwidth} 
                    @{\hspace{1mm}}>{\centering\arraybackslash}m{0.24\textwidth} 
                    @{\hspace{1mm}}>{\centering\arraybackslash}m{0.24\textwidth}}
        Original & 
        $0.019$\textit{BPP} & 
        $0.038$\textit{BPP} & 
        $0.078$\textit{BPP} \\
    \end{tabular}
    \\% Adjust space between text and images

    % Images with corresponding SSM overlayed
    \begin{tabular}{>{\centering\arraybackslash}m{0.24\textwidth} 
                    @{\hspace{1mm}}>{\centering\arraybackslash}m{0.24\textwidth} 
                    @{\hspace{1mm}}>{\centering\arraybackslash}m{0.24\textwidth} 
                    @{\hspace{1mm}}>{\centering\arraybackslash}m{0.24\textwidth}}
        \begin{tikzpicture}
            \node[anchor=south west,inner sep=0] (image1) at (0,0) {\includegraphics[width=\linewidth]{Figures/SQ-GAN/real_image.png}};
            \node[anchor=south east,inner sep=0] at (image1.south east) {\includegraphics[width=0.45\linewidth]{Figures/SQ-GAN/real_ssm.png}};
        \end{tikzpicture} & 
        \begin{tikzpicture}
            \node[anchor=south west,inner sep=0] (image2) at (0,0) {\includegraphics[width=\linewidth]{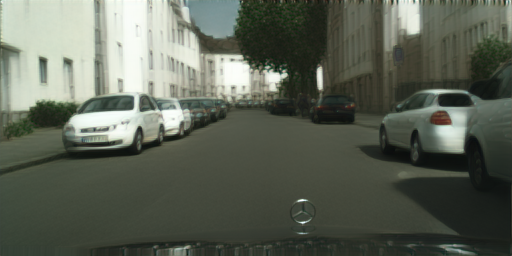}};
            \node[anchor=south east,inner sep=0] at (image2.south east) {\includegraphics[width=0.45\linewidth]{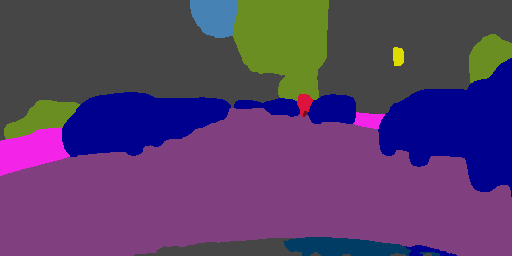}};
        \end{tikzpicture} & 
        \begin{tikzpicture}
            \node[anchor=south west,inner sep=0] (image3) at (0,0) {\includegraphics[width=\linewidth]{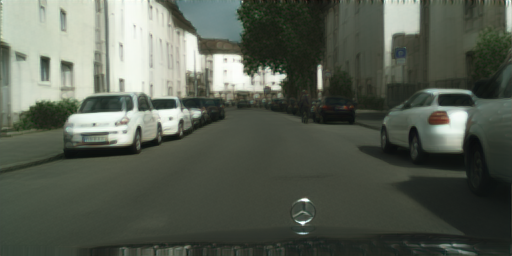}};
            \node[anchor=south east,inner sep=0] at (image3.south east) {\includegraphics[width=0.45\linewidth]{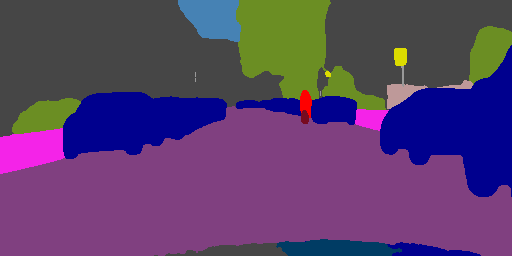}};
        \end{tikzpicture} & 
        \begin{tikzpicture}
            \node[anchor=south west,inner sep=0] (image4) at (0,0) {\includegraphics[width=\linewidth]{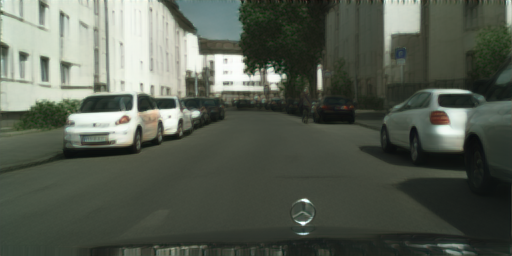}};
            \node[anchor=south east,inner sep=0] at (image4.south east) {\includegraphics[width=0.45\linewidth]{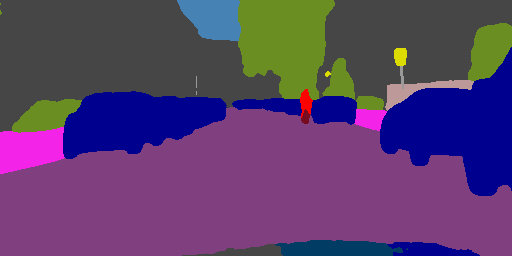}};
        \end{tikzpicture} \\
        % new row for BPG
        \begin{tikzpicture}
            \node[anchor=south west,inner sep=0] (image5) at (0,0) {\includegraphics[width=\linewidth]{Figures/SQ-GAN/real_image.png}};
            \node[anchor=south east,inner sep=0] at (image5.south east) {\includegraphics[width=0.45\linewidth]{Figures/SQ-GAN/real_ssm.png}};
        \end{tikzpicture} & 
        % Empty slot for (row 2, col 2) 
        & 
        \begin{tikzpicture}
            \node[anchor=south west,inner sep=0] (image7) at (0,0) {\includegraphics[width=\linewidth]{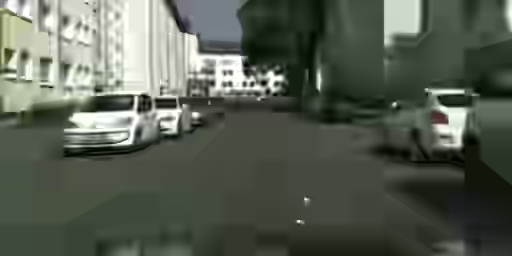}};
            \node[anchor=south east,inner sep=0] at (image7.south east) {\includegraphics[width=0.45\linewidth]{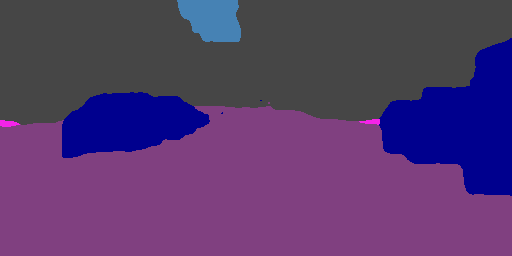}};
        \end{tikzpicture} & 
        \begin{tikzpicture}
            \node[anchor=south west,inner sep=0] (image8) at (0,0) {\includegraphics[width=\linewidth]{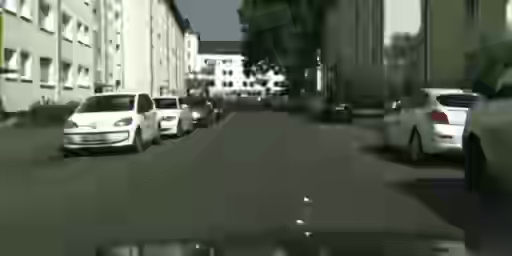}};
            \node[anchor=south east,inner sep=0] at (image8.south east) {\includegraphics[width=0.45\linewidth]{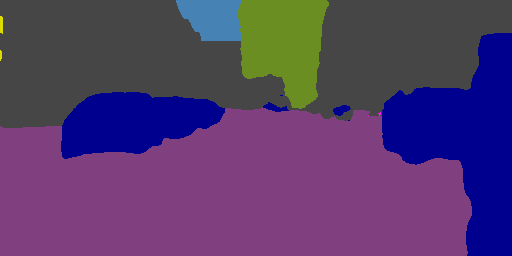}};
        \end{tikzpicture} \\
    \end{tabular}
    
    \caption[Visual comparison between \acrshort{bpg} and \acrshort{sqgan}]{Visual comparison at different compression rates between the proposed \acrshort{sqgan} (TOP) and the classical \acrshort{bpg} (BOTTOM). The \glspl{ssm} shown are generated from $\hat{\x}$ via the \acrshort{sota} \acrshort{ssmodel} INTERN-2.5 \cite{Wang2022internimage}. The proposed model is able to reconstruct images with higher semantic retention and lower values of \acrshort{bpp} compared with \acrshort{bpg}. The \acrshort{bpg} algorithm is not able to compress images at lower values than $0.038$ \acrshort{bpp}, thus the comparison is limited to $0.038$ and $0.078$ \acrshort{bpp}.}
    \label{fig: SQGAN visual comparison sqgan bpg}
\end{figure}
In this section the results of the proposed \gls{sqgan} are compared with the classical compression algorithm \gls{bpg} and \gls{jpeg2000}. Before numerically presenting the performances in term of semantic and classical metrics, it is important to visually see the differences. The visual comparison between \gls{bpg} and \gls{sqgan} is shown in \fref{fig: SQGAN visual comparison sqgan bpg}. \\

The top row represent the reconstructed image $\hat{\x}$ obtained with the proposed \gls{sqgan} and the associated generated \gls{ssm} via the INTERN-2.5 \gls{ssmodel}. The bottom row is showing the reconstructed image obtained by using the \gls{bpg} algorithms and the associated generated \gls{ssm}. 

As a first examination it is noticable that the \gls{sqgan} can reach lower values of \gls{bpp}. The classic \gls{bpg} cannot compress images at \gls{bpp} lower than $0.038$.

The other observation is that while the classical \gls{bpg} is using precious resources to reconstruct the windows of the buildings, the \gls{sqgan} is focusing on the relevant parts. This is shown in the amount of semantic retention of the generated \gls{ssm}. The person riding the bicycle is still visible and correctly identified by the \gls{ssmodel} even if the windows of the building are reconstructed with less detailed. On the contrary the \gls{bpg} is trying to reconstruct the window and the person riding the bicycle with the same level of detail.\\
As a matter of comparison, to obtain a level of semantic retention similar to the one obtained by the \gls{sqgan} at $0.038$\gls{bpp} for the \gls{bpg} algorithm almost $0.280$\gls{bpp} are required.\\

\begin{figure}[!t]
    \centering
    \begin{subfigure}[t]{0.45\textwidth}
        \centering
        \includegraphics[width=\textwidth]{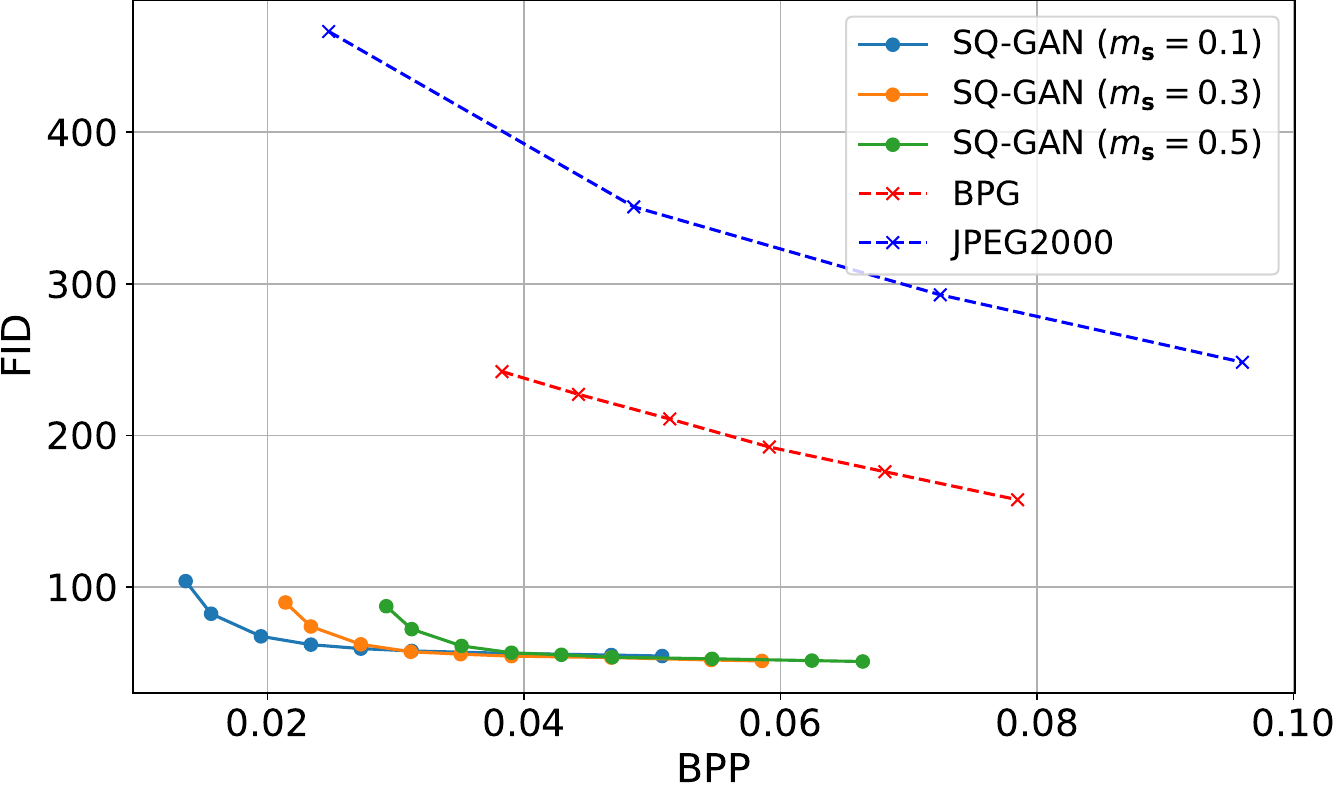}
        \caption*{(a)} % Caption under image without adding to list
    \end{subfigure}%
    \hspace{5mm}
    \begin{subfigure}[t]{0.45\textwidth}
        \centering
        \includegraphics[width=\textwidth]{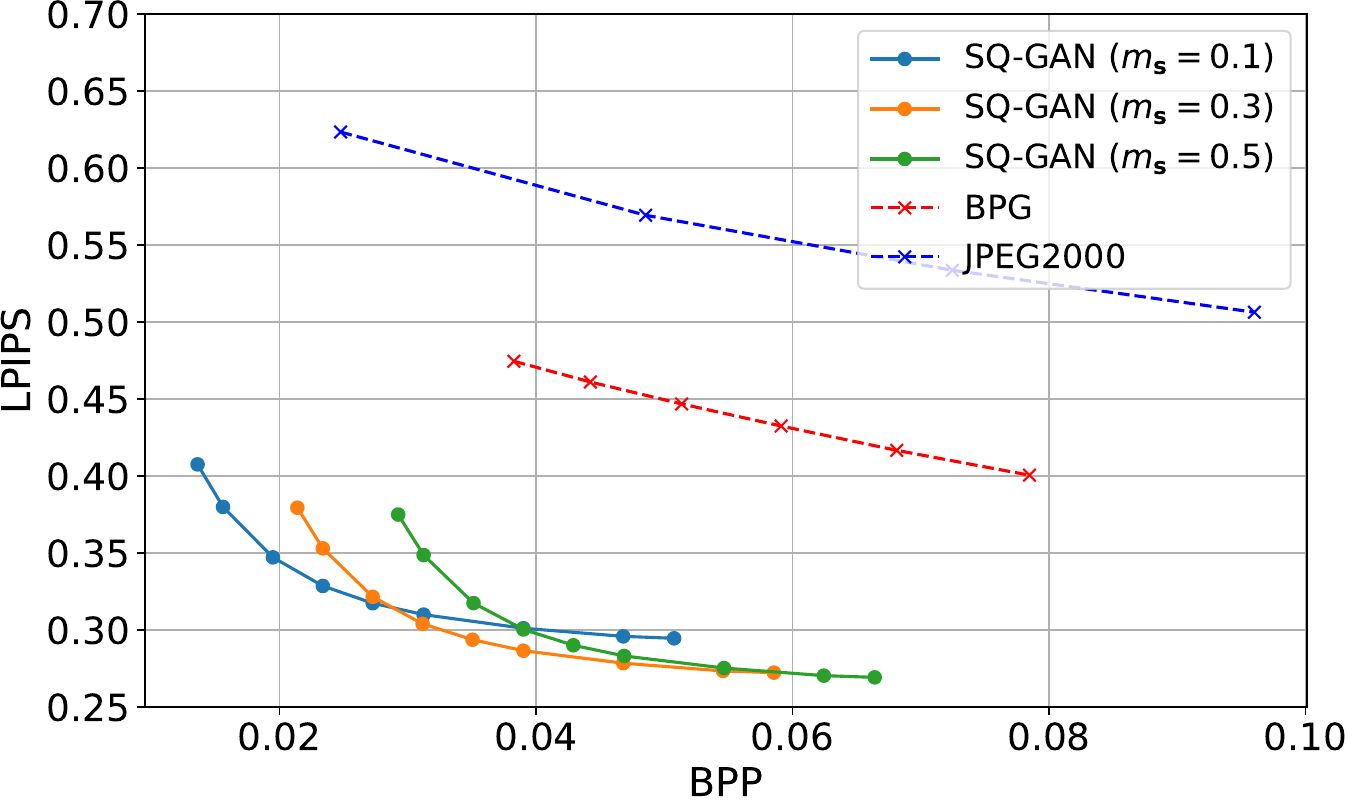}
        \caption*{(b)}
    \end{subfigure}
    
    \vspace{5mm} % Add space between rows

    \begin{subfigure}[t]{0.45\textwidth}
        \centering
        \includegraphics[width=\textwidth]{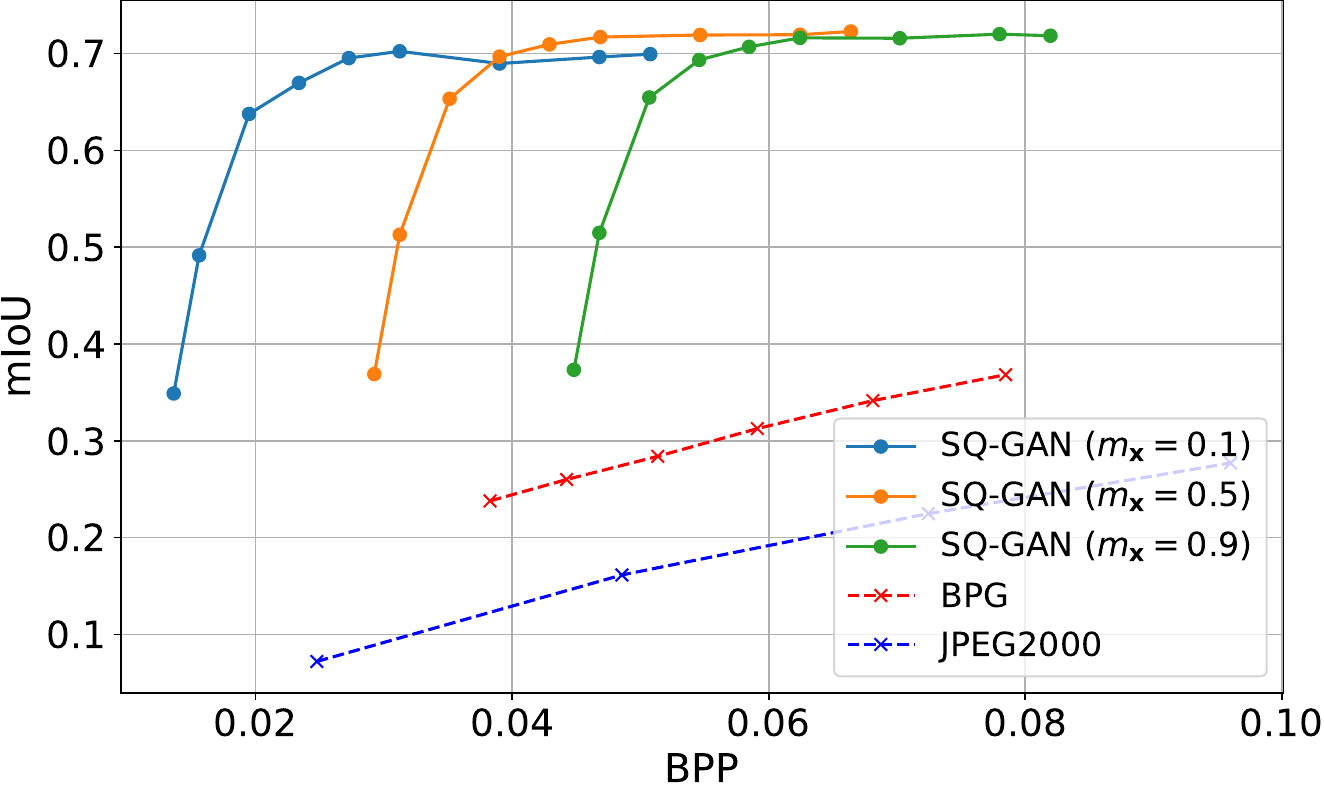}
        \caption*{(c)}
    \end{subfigure}%
    \hspace{5mm}
    \begin{subfigure}[t]{0.45\textwidth}
        \centering
        \includegraphics[width=\textwidth]{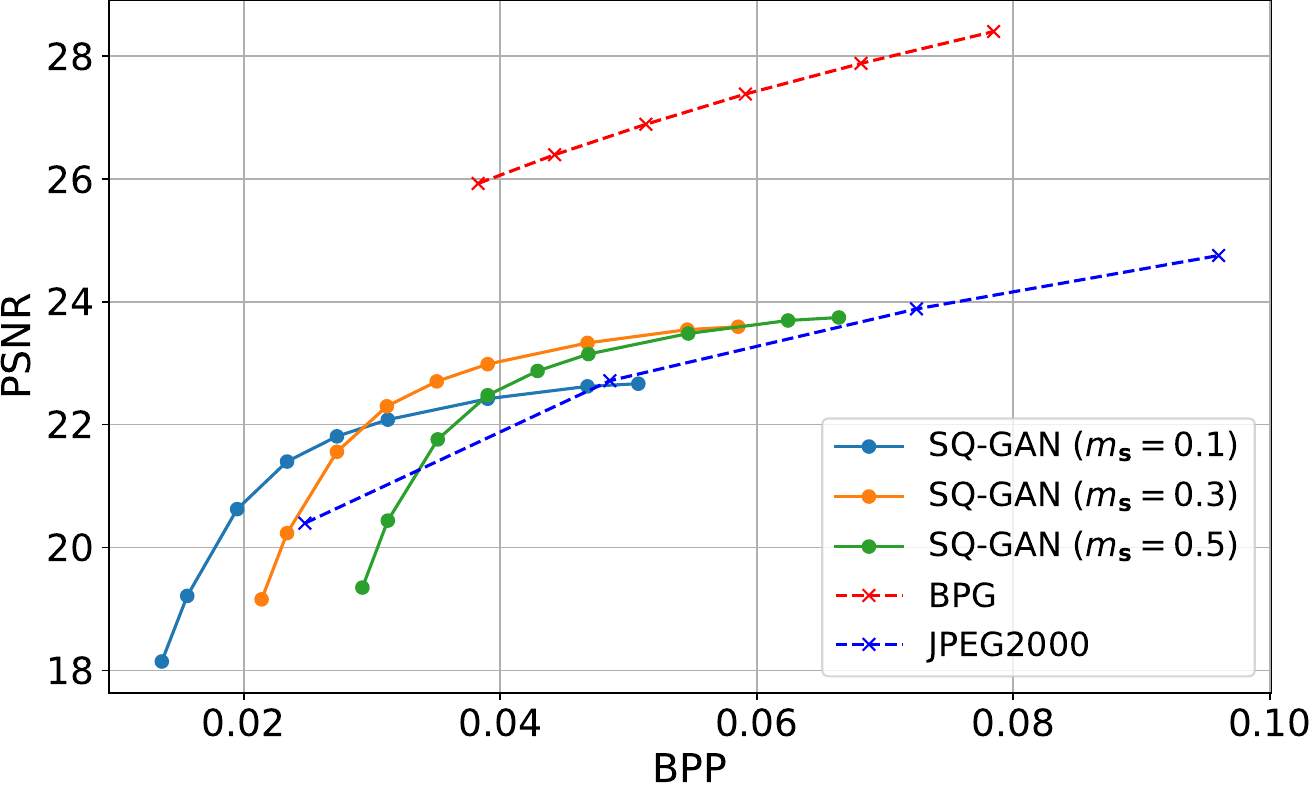}
        \caption*{(d)}
        \label{fig: SQGAN psnr vs bpp}
    \end{subfigure}

    \caption[Performance comparison between the \acrshort{sqgan} and classical compression algorithms]{Performance comparison between \acrshort{bpg}, \acrshort{jpeg2000} and \acrshort{sqgan} in term of semantic metrics and the classic pixel-by-pixel \acrshort{psnr}.}
    %[Comparison performances between \acrshort{sqgan} and classical compression algorithms]{Performance comparisons between \gls{sqgan} and classical compression algorithms. The non-semantic metric \gls{psnr} (d) is the only metric where \gls{bpg} performs better. In all the semantic-related metrics assessing image quality, such as \gls{fid} (a), \gls{lpips} (b), and \gls{miou} (c), the proposed model outperforms classical algorithms.}
    \label{fig: SQGAN all metrics comparison}
\end{figure}
These improvements can additionally be evaluated via comparison metrics as in \fref{fig: SQGAN all metrics comparison}. The figure shows the comparison between \gls{sqgan} and classical image compression algorithms in terms of \gls{fid}, \gls{lpips}, \gls{psnr} and \gls{miou}. The first 3 metrics are evaluated between the original $\x$ and the reconstructed $\hat{\x}$, while the \gls{miou} is evaluated between $\s$ and the \gls{ssm} generated via the \gls{sota} INTERN-2.5 \gls{ssmodel}.

The first important advantage is that the proposed \gls{sqgan} is able to compress at a lower level of \gls{bpp}. Moreover, on semantic metrics like \gls{fid}, \gls{lpips}, and \gls{miou}  the \gls{sqgan} consistently outperforms classical algorithms at a fraction of the \gls{bpp}. The only metric where \gls{bpg} performs better is the \gls{psnr}. This is not a surprise since this is a classical pixel-by-pixel metric that does not consider the overall visual quality but only the distance in the pixel domain. However, in a \gls{sc} framework, not all pixels have the same importance. Performing better on pixel-by-pixel metrics and not on semantic relevant metrics is not an advantage. For example, at $0.038$\gls{bpp} the \gls{bpg} algorithm can reconstruct images with a \gls{psnr} of 26, but it fails to preserve the \gls{ssm}. As shown in \fref{fig: SQGAN visual comparison sqgan bpg}, while \gls{bpg} reconstructs the windows of buildings better, only the first two cars are detected and the rest of the objects are lost.

Overall, the proposed \gls{sqgan} demonstrates exceptional performances. The model compresses images at very low \gls{bpp} while preserving their semantic content. It outperforms classical algorithms by learning to select relevant parts of the image focusing on key elements and considering less relevant parts only when asked to.\\

In summary, the proposed \gls{sqgan} achieves remarkable \gls{sc} image compression by effectively balancing low \gls{bpp} and high semantic retention. The integration of the \gls{samm} and \gls{spe} allows for selective prioritization of semantically relevant regions, ensuring that critical classes such as "traffic signs" and "traffic lights" are accurately reconstructed. The introduction of the Semantic-Aware Discriminator further enhances the model's ability to focus on important semantic details while minimizing attention on less relevant areas. Additionally, the Semantic Relevant Classes Enhancement data augmentation technique plays a crucial role in addressing the challenge of underrepresented classes, thereby improving the overall robustness and effectiveness of the compression process.

Experimental results on the Cityscapes dataset demonstrate that \gls{sqgan} consistently outperforms classical compression algorithms like \gls{bpg} and \gls{jpeg2000} across various metrics, including \gls{miou}, \gls{fid} and \gls{lpips}. These findings highlight \gls{sqgan}'s superior capability in preserving semantic content while maintaining efficient compression rates. The visual comparisons further corroborate the quantitative results, showcasing \gls{sqgan}'s ability to retain essential semantic details even at lower \gls{bpp} levels.

These advancements establish \gls{sqgan} as a potent tool for applications requiring both high compression efficiency and meticulous preservation of semantic information. Building on the \gls{sqgan}, the following chapter explores its application in \gls{goc} Resource Allocation in Edge Networks.

\chapter{\textcolor{black}{Goal-Oriented Resource Allocation in Edge Networks}}
\label{ch: Goal_oriented}
The content of this chapter is entirely based on the following publications:
\begin{quotation}
    \noindent \textit{\textbf{\large Goal-Oriented Communication for Edge Learning based on the Information Bottleneck}}\\
    \textit{Francesco Pezone, Sergio Barbarossa, Paolo Di Lorenzo}

    \vspace{0.1cm}
    \noindent \textit{\textbf{\large SQ-GAN: Semantic Image Coding Using
Masked Vector Quantization}}\\
    \textit{Francesco Pezone, Sergio Barbarossa, Giuseppe Caire}
\end{quotation}

\section{Introduction}
At this point, it should be clear that \gls{sc} is achievable thanks to generative models. The models presented in previous chapters have been specifically designed to enable the compression and reconstruction of images at very low values of \gls{bpp} while maintaining high levels of semantic preservation. By tweaking a limited number of parameters\footnote{For the \gls{spic} and \gls{cspic} it is possible to tweak the compression level of the coarse $\co$ and the compacted residual $\br'$. For the \gls{sqgan} it is possible to tweak the masking fractions $m_\x$ and $m_\x$. In both cases the effects have been discussed in the respective chapters in the Result sections.}, it is possible to easily explore the trade-off between compression and semantic preservation. However, identifying this optimal trade-off goes beyond the scope of \gls{sc} and is deeply rooted in the concept of \gls{goc}, as introduced in \sref{sec: SEMCOM go}.

This framework enables the communication system to be tailored to specific application objectives, accounting for external factors such as resource constraints and network conditions \cite{Strinati20216G}. In particular, the dynamic adaptation of compression parameters based on feedback mechanisms becomes essential for optimizing performance in varying environments.

This chapter introduces a methodology for achieving optimal Goal-Oriented resource allocation within \glspl{en} by leveraging stochastic optimization techniques. The primary objective is to ensure that the \gls{en} minimizes the average power consumption over time. Concurrently, the system aims to constrain the processing and transmission delays, as well as the level of performance over time.\footnote{Performance is quantified using task-specific metrics, which are defined subsequently.}

In contrast to existing research that predominantly focuses on \gls{en} optimization in isolation \cite{Mohammad2019Adaptive, Wang2018whenedge, Skatchkovsky2019optimizing, Merluzzi2020dynamic}, this chapter integrates \gls{goc} principles into the optimization process. Two distinct approaches are proposed: pure \gls{goc} via the \gls{gib} problem and \gls{sgoc} through the \gls{sqgan} model.

The \gls{gib} framework is particularly well-suited for \gls{goc}, as detailed in \sref{sec: SEMCOM ib}. By identifying the trade-off between complexity and relevance through the parameter $\beta$, the \gls{gib} method allows the system to reduce resource consumption while maintaining task performance \cite{Chechik2004GIB}. In \sref{sec: EN_ib} a mechanism based on a dynamic feedback will be proposed to adjust $\beta$ based on real-time network conditions and resource availability. This will enable the \gls{en} to autonomously optimize its operations, thereby ensuring efficient resource utilization in a \gls{goc} fashion.

Alternatively, when preserving semantic information is of paramount importance, the \gls{sqgan} model offers a robust solution for \gls{sgoc}. As introduced in \cref{ch: SQGAN}, the \gls{sqgan} model provides semantic image compression capabilities. By applying stochastic optimization techniques to dynamically adjust the masking fractions $m_\x$ and $m_\s$, it becomes possible to exert precise control over the performance metrics of the \gls{en}.

The integration of stochastic optimization and feedback mechanisms, informed by real-time monitoring of resource usage, network conditions, and performance indicators, allows the system to dynamically balance resource consumption and performance. This comprehensive framework is designed to orchestrate resource allocation within the \gls{en}, ensuring that computational and communication resources are utilized efficiently while satisfying the specific objectives of the application.

\begin{figure}[!t] 
    \centering 
    \includegraphics[width=0.9\textwidth]{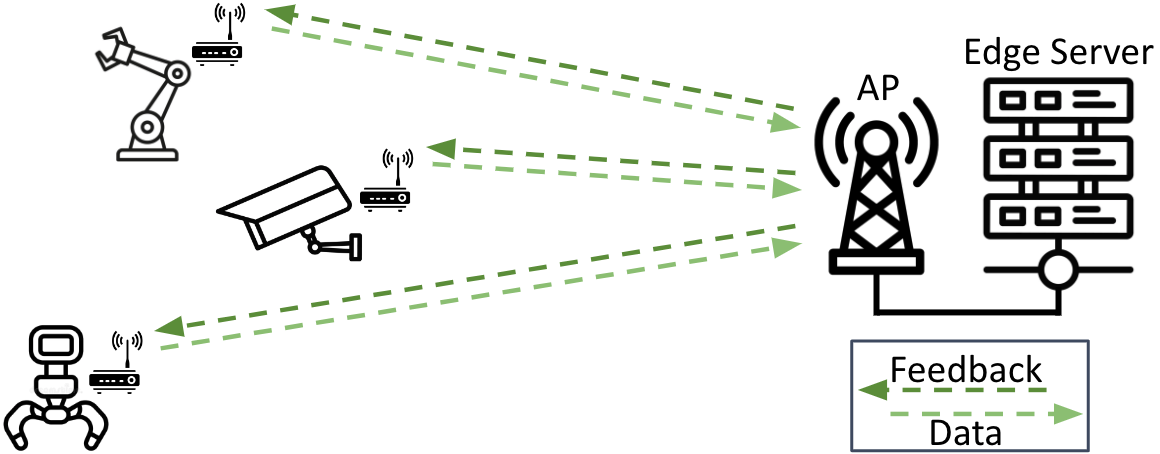} 
    \caption[Edge Network Scheme]{\acrshort{en} scheme with connected \acrshort{ed}s transmitting signals and receiving feedback from the \acrshort{es}.}
    %[Schema of the Edge Network]{Edge Network scheme.} 
    \label{fig: EN en_scheme} 
\end{figure} 

By adopting these methodologies, the system is capable of transmitting only the most relevant information necessary for the task at hand, thereby optimizing overall performance within the constraints of available resources. The adaptive mechanisms ensure that the communication process is closely aligned with application-specific goals, resulting in enhanced efficiency and effectiveness.

The structure of this chapter is as follows. \sref{sec: EN_en desciption} provides a comprehensive overview of the \gls{en}architecture. \sref{sec: EN_ib} discusses the implementation of \gls{goc} within the \gls{en} using the \gls{gib} framework, presenting the mathematical formulations and optimization strategies for resource allocation. This section also explores how stochastic optimization facilitates the adjustment of the compression parameter $\beta$. \sref{sec: EN_nn} focuses on the application of the \gls{sqgan}-based \gls{sgoc} paradigm, highlighting how this model can be integrated into the \gls{en} and how to dynamically adjust the masking fractions $m_\x$ and $m_\s$.

\section{Edge Network}\label{sec: EN_en desciption}
Edge computing represents a decentralized computing architecture that brings data processing, storage, and networking capabilities closer to the physical location where data is generated \cite{Shi2016edge,Satyanarayanan2017edge,Liu2019edge}. This approach contrasts with traditional cloud computing, where data is transmitted to centralized data centers via the internet for processing. By moving computational tasks closer to the data source—the \textit{edge} of the network—latency is minimized, and real-time applications become more feasible. This is particularly beneficial for applications requiring immediate processing and response, such as autonomous vehicles, industrial automation, and real-time analytics.

An Edge Network (\gls{en}) typically consists of multiple Edge Devices (\glspl{ed}) connected via wireless communication to an Edge Server (\gls{es}). The \glspl{ed} are capable of data collection and preliminary processing, while the \gls{es} handles more intensive computational tasks and orchestrates the network resources. In this work, the \gls{en} is composed of $K$ \glspl{ed}, each connected to the \gls{es} through an Access Point operating at a predefined carrier frequency $f_0$. The total available bandwidth $B$ is equally divided among the \glspl{ed} and the wireless communication is characterized by flat-fading channels and additive white Gaussian noise.

For the purpose of focusing on the most significant factors affecting performance, several simplifications are made in the network model:

\begin{itemize} \item \textbf{Edge Devices (\glspl{ed}):} Each \gls{ed} is equipped with: \begin{itemize} \item \textbf{Central Processing Unit (CPU):} Responsible for pre-processing data, the CPU is characterized by a maximum power consumption $p_k^{cpu}$, a working clock frequency $f_k$, a maximum clock frequency $f_k^{max}$, and an effective switch capacitance $\eta_k$ \cite{Burd1996EffSwithc}. \item \textbf{Wireless Network Interface:} Facilitates wireless communication, allowing the \gls{ed} to transmit data at a rate $R_k$, with a maximum value $R_k^{max}$. This rate depends on the allocated bandwidth $B_k$, the flat-fading channel coefficient $h_k$, the noise power spectral density $N_0$, and the maximum transmission power $p_k^{tr}$. \end{itemize} \item \textbf{Edge Server (\gls{es}):} Acts as the central processing unit for the network and is responsible for final data processing and resource orchestration. The \gls{es} includes: \begin{itemize} \item \textbf{Central Processing Unit (CPU):} More powerful than those in the \glspl{ed}, the \gls{es} CPU is characterized by maximum power consumption $p^{cpu}$, working clock frequency $f_c$, maximum clock frequency $f_c^{max}$, and effective switch capacitance $\eta$. The \gls{es} allocates a portion of its computing time, denoted by $\frac{\tau_k}{\tau}$ or equivalently $f_k^{es}$, to each \gls{ed}, ensuring that $\sum_{k=1}^K f_k^{es} \leq f_c$. \end{itemize} \end{itemize}

Other components present in a real-world \gls{en} are omitted for simplicity, as their impact on the resource optimization problem is either limited or constant.

Resource optimization in an \gls{en} refers to the efficient management and allocation of computational and networking resources to achieve specific performance objectives \cite{Mach2017mobile, Huang2019multi, Mach2017mobile, Binucci2023goaloriented}. In this context, a dynamic resource allocation strategy is proposed, aiming to minimize the average power expenditure while satisfying constraints on the average service delay and the desired performance metrics. The service delay encompasses the time required for data pre-processing at the \gls{ed}, data transmission to the \gls{es}, and post-processing at the \gls{es}. The performance metrics are task-specific and could include measures such as reconstruction quality, classification accuracy, or other relevant indicators.

The optimization problem involves several controllable variables:

\begin{itemize} \item \textbf{CPU Clock Frequencies ($f_k$ and $f_k^{es}$):} Adjusting the clock frequencies of the CPUs at the \glspl{ed} and the \gls{es} affects both power consumption and processing delay. \item \textbf{Data Transmission Rate ($R_k$):} Modulating the transmission rate influences the transmission power consumption and the communication delay. \item \textbf{Model Hyperparameters:} Parameters of the encoding and decoding models, such as the trade-off parameter $\beta$ in the Information Bottleneck method or the masking fractions $m_\x$ and $m_\s$ in the \gls{sqgan}, impact the quality of the reconstructed data and the amount of information transmitted. \end{itemize}

Time is considered to be slotted, indexed by $t$, and tasks starting in one time-slot are completed within the same slot to prevent queue formation. The subsequent sections delve into the specifics of the optimization problems formulated for both the Information Bottleneck approach and the \gls{sqgan}-based \gls{sgoc} paradigm.

\section{Goal-Oriented communication based on the Information Bottleneck}\label{sec: EN_ib}
In this section the scenario composed of $K$ \glspl{ed} sending data to an \gls{ed} using a \gls{gib}-based encoder is discussed.

The solution of the \gls{gib} problem described in \sref{sec: SEMCOM ib} was obtained starting from two joint random variables $\x\sim \mathcal{N}(\mathbf{0}, \Sigma_{X})$ and $\y\sim \mathcal{N}(\mathbf{0}, \Sigma_{Y})$ of dimension $d_\x$ and $d_\y$, respectively, with the cross-covariance matrix denoted by $\Sigma_{XY}$. In this specific case, Chechik et al. were able to express in a closed-form solution the best transformation $z= \Phi(\x)$ able to optimize the \gls{ib} problem in \eref{eq: SEMCOM ib_problem} as a function of the trade-off parameter $\beta$  \cite{Chechik2004GIB}. This solution was represented by the linear projection $\z = \mathbf{A} \x + \boldsymbol{\xi}$, where $\boldsymbol{\xi}$ represent the added white Gaussian noise, while $\mathbf{A}$ is the projection matrix described in \eref{eq: EN_ib Matrice_A} and reported below:

\begin{equation}
    \mathbf{A} = \left\{\begin{matrix}
                [\mathbf{0}^T;...;\mathbf{0}^T] & 0 \leq \beta \leq \beta_1^c\\
                [\alpha_1\mathbf{v}_1^T; \mathbf{0}^T;...;\mathbf{0}^T] & \beta_1^c < \beta \leq \beta_2^c\\
                [\alpha_1 \mathbf{v}_1^T;\alpha_2\mathbf{v}_2^T;\mathbf{0}^T;\ldots;\mathbf{0}^T] & \beta_2^c < \beta \leq \beta_3^c\\
                \vdots\\
                [\alpha_1 \mathbf{v}_1^T;\alpha_2\mathbf{v}_2^T;\ldots;\alpha_{n_{\beta}}\mathbf{v}_{n_{\beta}}^T] & \beta_{n_{\beta}}^c < \beta 
                \end{matrix}\right.
    \tag{\ref{eq: SEMCOM Matrice_A}}
    \label{eq: EN_ib Matrice_A}
\end{equation}

The structure of the matrix $\mathbf{A}$ is strongly influenced by the parameter $\beta$. By changing $\beta$, the number of rows in $\mathbf{A}$ will change accordingly, up to a certain value $n_{\beta}$. The value $n_{\beta}$ can be computed starting from the covariance matrices $\Sigma_{X}$, $\Sigma_{Y}$, and $\Sigma_{XY}$.

Due to the complicated structure of the matrix $\mathbf{A}$, in this section, only some specific values of  $\beta$ will be considered. Instead of letting $\beta$ vary from $0$ to infinity, a finite subset composed as follows will be taken into account: 
\begin{equation} 
    \mathcal{B}_k = \{ \beta_{2_k}^c, \beta_{3_k}^c, \ldots, \beta_{n_{\beta_k}}^c, 10*\beta_{n_{\beta_k}}^c\}. \label{eq: EN_ib beta_set} 
\end{equation}

These values correspond to the last values of $\beta$ before the number of rows in $\mathbf{A}_k$ is increased. They are selected in a way that at least one row is considered, thus avoiding cases where no data is transmitted, and the maximum value of $\beta_k= 10*\beta_{n_{\beta_k}}^c$ is sufficiently high to be considered as infinite.\

At the \gls{es}, the goal is to reconstruct an estimate $\hat{\y}_k=\mathbf{M}_k\z_k$ of the original data $\y_k$. The specifications of the matrix $\mathbf{M}_k$ will be described in the following when discussing the evaluation metric.\\

After this introductory review and clarification about the \gls{gib} problem, it is possible to discuss how the \gls{en} is structured and what the role of the optimization variables is. This will be done by discussing how the power consumption, the delays and the evaluation metric are represented in this framework.\

\begin{itemize}[label={}] 
    \item {\textbf{Power consumption:}} The power consumption represents the objective function of the optimization problem. This term is composed of the contributions of the \glspl{ed} and the \gls{es}. After the observation of the data, the first task to be performed is the encoding based on the \gls{gib}. This task will require the CPU of the \gls{ed} to be active and work at a certain clock frequency $f_k(t)$. The power consumption will then be a function of the frequency and expressed as: \begin{equation} P_k^{ib}(f_k(t)) = \eta_k \cdot (f_k(t))^3 \equiv P_k^{ib}(t), \end{equation} where $\eta_k$ is the effective switch capacitance of processor $k$ \cite{Burd1996EffSwithc}.

    For simplicity of notation, the power consumption will be written only as a function of time as $P_k^{ib}(f_k(t)) \equiv P_k^{ib}(t)$. This notation will be used for all the other terms in the following sections.

    After the encoding step, the data is ready to be transmitted. The transmission power can be expressed as a function of the transmission rate $R_k(t)$ using Shannon's formula:
    \begin{equation}
        P_k^{tr}(R_k(t)) =  \frac{B_k N_0}{h_k(t)} \left[  {\rm exp} \left(\frac{R_k(t) \ln(2)}{B_k} \right)   -1 \right] \equiv P_k^{tr}(t).
    \end{equation}

    After the data has been transmitted to the \gls{es}, the final processing has to be performed. The power consumption of the CPU at the \gls{es} is evaluated in the same way as the one at the \gls{ed}:
    \begin{equation}
        P^{es}(f_c(t)) = \eta \cdot (f_c(t))^3 \equiv P^{es}(t).
    \end{equation}
    These terms can be combined to obtain the total power consumption of the \gls{en}:

    \begin{equation}
        P^{tot}(t) = P^{es}(t) + \sum_{k=1}^K \Gamma_k P_k^{ib}(t) + P_k^{tr}(t),
        \label{eq: EN_ib total power}
    \end{equation}
    where $\Gamma_k$ is the tunable term that weights the power consumption of the CPU of the \gls{ed}.

    \item {\textbf{Delay:}}
    Every stage of the processing pipeline will be associated with a specific delay. After the data is collected, the encoding is performed. The first important term influencing the computational delay at the \gls{ed} is given by the number of operations to perform the matrix multiplication $\z_k = \mathbf{A}_k \x + \boldsymbol{\xi}$. This value is expressed as:
    \begin{equation}
        W_k(\beta_k(t)) = d_x\cdot n_{\beta_k}(t) \equiv W_k(t).
    \end{equation}
    The other term influencing the encoding delay is the number of \gls{flops} that the CPU can perform at a given clock frequency $f_k(t)$. This value depends on the product $\rho_k$ between the number of cores and the number of \gls{flopc} of the CPU. After the CPU has been selected, this value is fixed and cannot change. By considering all these terms, the delay of the encoding process can be expressed as:
    \begin{equation}
        D_k^{ib}(f_k(t)) = \frac{W_k(t)}{FLOPS(f_k(t))} = \frac{W_k(t)}{f_k(t) \cdot \rho_k} \equiv D_k^{ib}(t).
    \end{equation}
    After the encoding, the transformation $\z$ has to be transmitted. The delay involved with the transmission is dependent on the transmission rate $R_k(t)$ and on the number of bits to be transmitted $N_k(t)$. This number of bits corresponds to the entropy of the random Gaussian variable $\z_k(\beta_k(t))$ and can be expressed as:
    \begin{equation}
        N_k(\beta_k(t)) = h(\z_k(\beta_k(t)))  \equiv N_k(t),
    \end{equation}
    The delay associated with the transmission step can be expressed as:
    \begin{equation}
        D_k^{tr}(\beta_k(t), R_k(t)) = \frac{N_k(t)}{R_k(t)} \equiv D_k^{tr}(t).
    \end{equation}
    At the receiver, every sub-process associated with a device $k$ is performed. The estimate $\hat{\y}_k=\mathbf{M}_k\z_k$ has to be evaluated, and this matrix multiplication requires a number of operations equal to:
    \begin{equation}
        W_k^{es}(\beta_k(t)) = d_y\cdot n_{\beta_k}(t) \leq  d_y \cdot d_{min} \equiv W_k^{es},
    \end{equation}
    with $n_{\beta_k}(t)\leq d_{min}= \min\{ d_x, d_y\}$.
    
    Overall, this processing at the \gls{es} is associated with a delay per \gls{ed} equal to:
    \begin{equation}
        D_k^{es}(f(t)) = \frac{W_k^{es}}{FLOPS(f_k^{es}(t))} = \frac{W_k^{es}}{f_k^{es}(t) \cdot \rho_k^{es}} \equiv D_k^{es}(t),
    \end{equation}
    where $\rho_k^{es}$ is the product between the number of cores and the number of Floating Point Operations Per Cycle of the CPU at the \gls{es}.

    By considering all the three components, it is possible to express the total delay associated with the process of every \gls{ed} as:
    \begin{equation}
        D_k^{tot}(t) = D_k^{es}(t) + D_k^{ib}(t) + D_k^{tr}(t).
    \end{equation}

    \item {\textbf{Evaluation metric:}}
    The last term to consider is the evaluation metric.

    After the transformation reaches the \gls{es}, it has to be processed to estimate $\hat{\y}_k=\mathbf{M}_k\z_k$. This task is performed via a matrix $\mathbf{M}_k$ designed to minimize the \gls{mse} between $\y_k$ and $\hat{\y}_k$.

    This process is associated the \gls{nmse} evaluation metric that can be evaluated in closed-form solution as:
    \begin{equation}
        G_k(\beta_k(t)) = NMSE_{\beta_k(t)}(\y_k, \hat{\y}_k) =  1 - \frac{\mathrm{tr} \left( \bSigma_{Y_kZ_{\beta_k(t)}} \bSigma_{Z_{\beta_k(t)}}^{-1} \bSigma_{Y_kZ_{\beta_k(t)}}^{T}\right)}{\mathrm{tr}(\bSigma_{Y_k})} \equiv G_k(t).
    \end{equation}

\end{itemize}

Now that all the terms have been introduced, it is possible to discuss and present the optimization problem. As already mentioned at the beginning of this chapter, the goal is to minimize the average power consumption over time while respecting the average constraints on the delay and the evaluation metric.

The optimization problem can then be cast as follows:
\begin{mini}|s|[0]
    {\mathbf{\Psi}(t)}{\lim_{T \to +\infty}\; \frac{1}{T} \sum_{t=1}^T  \mathbb{E}[P^{tot}(t)] }
    {}{}
    \addConstraint{\lim_{T \to +\infty}\; \frac{1}{T} \sum_{t=1}^T  \mathbb{E}[D_k^{tot}(t)] \leq D_k^{avg}\qquad \forall k }{}
    \addConstraint{ \lim_{T \to +\infty}\; \frac{1}{T} \sum_{t=1}^T  \mathbb{E}[G_k(t)] \leq G_k^{avg}\qquad \forall k }{}
    \addConstraint{0 \leq f_k(t) \leq f_k^{max} \qquad \forall k,t }{}
    \addConstraint{0 \leq R_k(t) \leq R_k^{max}(t) \qquad \forall k,t }{}
    \addConstraint{\beta_k(t) \in \mathcal{B}_k  \qquad \forall k,t}{}
    \addConstraint{0 \leq f_c(t) \leq f_c^{max} \qquad \forall t}{}
    \addConstraint{f_k^{es}(t) \geq 0 \quad \forall k,t}, \qquad {\sum_{k=1}^K f_k^{es}(t) \leq f_c(t)  \quad \forall t,}{}
    \label{eq: EN_ib initial opt problem}
\end{mini}
where $\mathbf{\Psi}(t) = [\{f_k(t)\}_k,\{ R_k(t)\}_k, \{\beta_k(t)\}_k, \{f_k^{es}(t)\}_k, f_c(t)]$ is the vector of the optimization variables at time $t$.
The first two constraints refer to the average delay $D_k^{avg}$ and the average evaluation metric $G_k^{avg}$ constraints. All the others are instead used to define the feasible space of the optimization variables in $\mathbf{\Psi}(t)$ and the value of the maximum transmission rate $R_k^{max}(t)$ at time $t$ is evaluated as follows:
\begin{equation}
    R_k^{max}(t) = B_k\; log \left[ 1 + \frac{p_k^{tr}\; h_k(t)}{N_0 B_k} \right]
\label{eq: EN_ib maximum rate}
\end{equation}

This optimization problem is complex to solve. However, in the next subsection will be shown how to handle it resorting to stochastic optimization \cite{Neely2010Lyapunov}.
\subsection{The Edge Network Optimization}
The first step to handle problem \eref{eq: EN_ib initial opt problem} is to introduce two \textit{virtual queues} for each \gls{ed}. These virtual queues will be associated to the long-term delay and evaluation metric constraints, $T_k(t)$ and $U_k(t)$ respectively. Proceeding similarly as in \cite{Merluzzi2021EN}, these two virtual queues evolve as follows:
\begin{align}
    T_k(t+1) &= \max [0,T_k(t) + \varepsilon_k(D_k^{tot}(t) - D_k^{avg})] \label{eq: EN_ib virtulQueue T_true}\\
    U_k(t+1) &= \max [0,U_k(t) + \nu_k(G_k(t) - G_k^{avg})],  \label{eq: EN_ib virtulQueue U_true}
\end{align}
where $\epsilon_k$ and $ \nu_k $ are the learning rate for the update of the virtual queues. 

Based on these virtual queues is possible to define the \textit{Lyapunov function} $L(\mathbf{\Theta}(t))$ as:
\begin{equation}
    L(\mathbf{\Theta}(t)) = \frac{1}{2} \sum_{k=1}^K T_k^2(t) + U_k^2(t),
\label{eq: EN_ib Lyapunov function}
\end{equation}
where $\mathbf{\Theta}(t) = [\{T_k(t)\}_k, \{U_k(t)\}_k]$ is the vector composed by all the virtual queues at time $t$. The idea is to use this Lyapunov function to satisfy the constraints on $D_k^{avg}$ and $G_k^{avg}$ by enforcing the stability of $L(\mathbf{\Theta}(t))$. To this scope it is introduced the so called \textit{drift-plus-penalty} function
\begin{equation}
    \Delta_p(\Theta(t)) = \mathbb{E}\left[L({\Theta}(t+1))-L({\Theta}(t))+V\cdot P^{tot}(t)  \;\Big|\; \Theta(t)\right],
    \label{eq: EN_ib drift_plus_penalty}
\end{equation}
whose minimization aims to stabilize the virtual queues in \eref{eq: EN_ib virtulQueue T_true} and \eqref{eq: EN_ib virtulQueue U_true}, while promoting low-power solutions for large values of the parameter $V$.

Using stochastic approximation arguments \cite{Neely2010Lyapunov} in \aref{app: EN_ib}, it is possible to maximize the drift-plus-penalty function with the suitable upper-bound and obtain the following optimization problem:
\begin{mini}|s|[0]
    {\mathbf{\Psi}(t)}{\sum_{k=1}^K \bigg[ \epsilon_k T_k(t)D_k^{tot}(t) + \nu_k U_k(t) G_k(t)\bigg]  + V P^{tot}(t) }{}{}
    \addConstraint{\mathbf{\Psi}(t) \in \mathcal{T}(t),}{}
    \label{eq: EN_ib per-slot opt problem structure}
\end{mini}
where $\mathcal{T}(t)$ indicates the space of possible solutions given by the constraints on the optimization variables. 

This per-slot deterministic optimization problem can now be solved at every time slot $t$ and the constant update of the virtual queues will ensure that, over time, the two constrains on the average delay and the average evaluation metric are satisfied.

The optimization of \eref{eq: EN_ib per-slot opt problem structure}, even if much simpler than \eref{eq: EN_ib initial opt problem}, is still non-trivial. This is a direct cause of its mixed-integer nature.

However, the time-slot optimization problem can be decoupled into two sub-problems, one associated with the \gls{ed} parameters, i.e., $[\{f_k(t)\}_k,\{ R_k(t)\}_k,\{\beta_k(t)\}_k]$, and the other associated with the \gls{es} parameters, i.e., $[\{f_k^{es}(t)\}_k, f_c(t)]$.

\begin{itemize}[label={}]
    \item {\textbf{ED sub-problem:}}
    The optimization problem associated with the $k$-th \gls{ed} can be expressed as follows:

    \begin{mini}|s|[0]
        {\mathbf{\Psi}^{ed}(t)}{g_k : g_k=\frac{\epsilon_kT_k(t)N_k(t)}{R_k(t)} + \frac{\epsilon_kT_k(t)W_k(t)}{f_k(t)\rho_k } + \nu_k U_k(t)G_k(t) +}{}{} \breakObjective{\qquad  \qquad+ V \frac{B_k N_0}{h_k(t)} {\rm exp} \left(\frac{R_k(t) ln(2)}{B_k} \right) + V \Gamma_k \eta_k (f_k(t))^3 }{}{}
        \addConstraint{\mathbf{\Psi}^{ed}(t) \in \mathcal{T}^{ed}(t),}{}
        \label{eq: EN_ib per-slot opt ed}
    \end{mini}
    where $\mathbf{\Psi}^{ed}(t) = [f_k(t), R_k(t), \beta_k(t)]$ is the vector of the optimization variables at time $t$ for the \gls{ed} sub-problem and $\mathcal{T}^{ed}(t)$ is the space of possible solutions given by the constraints on the optimization variables at the \gls{ed}.

    For a fixed value of the parameters $\beta_k(t)$ it is possible to evaluate a close form solution of both the optimal values of the clock frequency $f_k(t)$ and the transmission rate $R_k(t)$, \aref{app: EN_ib ed opt} for more details. These solutions are given by:
    \begin{equation}
        R_k^*(t) = \frac{2 B_k}{ln(2)}\; W\! \!\left(\sqrt{\frac{\epsilon_k T_k(t)\; ln(2)\; h_k(t)N_k(t)\; }{4 B_k^2\;V \;N_0}}\right)\; \Biggr|_0^{R_k^{max}(t)}
    \label{eq: EN_ib optimal rate}
    \end{equation}
    \begin{equation}
        f_k^* (t) = \sqrt[4]{\frac{\epsilon_k T_k(t) W_k(t)}{3 V \Gamma_k \eta_k \rho_k} }\; \Biggr|_0^{f_k^{max}},
    \label{eq: EN_ib optimal freq device}    
    \end{equation}
    where $W(\cdot)$ in \eref{eq: EN_ib optimal rate} denotes the principal branch of the Lambert function.
    
    At this point the optimal value $\beta_k^*(t)$ can be found by simply searching the value in $\mathcal{B}_k$ that, together with \eref{eq: EN_ib optimal rate} and \eqref{eq: EN_ib optimal freq device}, minimize the objective of the sub-problem associated with device k in \eref{eq: EN_ib per-slot opt ed}. 
    \item {\textbf{ES sub-problem:}}
    The optimization problem associated with the \gls{es} can be expressed as follows:
    \begin{mini}|s|[0]
        {\mathbf{\Psi}^{es}(t)}{g : g= \sum_{k=1}^K \frac{\epsilon_kT_k(t)W_{max}^{es}}{f_k^{es}(t)\rho_k^{es}} + V \eta (f_c(t))^3 }{}{}
        \addConstraint{\mathbf{\Psi}^{es}(t) \in \mathcal{T}^{es}(t),}{}
        \label{eq: EN_ib per-slot opt ed}
    \end{mini}
    where $\mathbf{\Psi}^{es}(t) = [\{f_k^{es}(t)\}_k, f_c(t)]$ is the vector of the optimization variables at time $t$ for the \gls{es} sub-problem, $\mathcal{T}^{es}(t)$ is the space of possible solutions given by the constraints on the optimization variables at the \gls{es} and $W_{max}^{es} = \displaystyle\max_{k} W_k^{es}$.
    
    The choice of substituting $W_k^{es}$ with $W_{max}^{es}$ is to avoid a combinatorial optimization problem. By maximizing the number of operations in such a way it is in fact possible to identify the optimal value the clock frequencies $f_k^{es}(t)$ and $f_c(t)$ as:
    \begin{equation}
        f_c^{*} (t) = \frac{\sqrt{\sum_{k=1}^K \sqrt{\frac{\epsilon_k T_k(t)W_{max}^{es}}{\rho_k^{es}}}}}{\sqrt[4]{3V\eta}} \; \Biggr|_0^{f_{c}^{max}},
    \label{eq: EN_ib optimal freq device ES}    
    \end{equation} 
    \begin{equation}
        f_k^{es*}(t) = \frac{\sqrt{\frac{\epsilon_k T_k(t)W_{max}^{es}}{\rho_k^{es}}}}{\sqrt{\sum_{k=1}^K \sqrt{\frac{\epsilon_k T_k(t)W_{max}^{es}}{\rho_k^{es}}}}\sqrt[4]{3V\eta} },  \qquad \forall k.
    \label{eq: EN_ib optimal freq ES}    
    \end{equation}
    More details in \aref{app: EN_ib ed opt}.

\end{itemize}

The final algorithm to solve the resource allocation optimization problem in \eref{eq: EN_ib initial opt problem} can be summarized as the algorithm in \tref{tab: EN_ib algorithm}.

\begin{table}[ht]
    \centering
    \rule{\textwidth}{0.4pt} % Line at the top
    \vspace{-22pt} % Adjust the space between the top line and the caption
    \caption{Edge Network Resource Allocation Algorithm}
    \vspace{-10pt} % Adjust the space between the caption and the line below it
    \rule{\textwidth}{0.4pt} % Line below the caption
    \vspace{-15pt} % Adjust the space between the line and the algorithmic content
        \begin{algorithmic}[1]
            \item Set the Lyapunov parameters $V$, $T_k(0)$, $U_k(0)$, $\epsilon_k$ and $\nu_k \;\;\forall k$
            \item Set the desired values for delay $D_k^{avg}$ and evaluation metric $G_k^{avg}$
            \For{$t$ at least until $T_k$ and $U_k$ converges}
                \State Compute the optimal $R_k^*(t)$ and $f_k^*(t)$ with \eref{eq: EN_ib optimal rate} and \eqref{eq: EN_ib optimal freq device} as a function of $\beta_k(t)$ $\forall k$.
                \State Select the optimal $\{R_k^*(t)\}_k$,  $\{f_k^*(t)\}_k$ and $\{\beta_k^*(t)\}_k$ that minimizes \eref{eq: EN_ib per-slot opt ed} $\forall k$.
                \State Evaluate the optimal $f_c^*(t)$ with \eref{eq: EN_ib optimal freq device ES} and then $f_k^{es*}(t)$ with \eref{eq: EN_ib optimal freq ES} $\forall k$.
                \State Run the online inference task
                \State Update  the virtual queues $T_k$ and $U_k$ $\forall k$ via \eref{eq: EN_ib virtulQueue T_true} and \eqref{eq: EN_ib virtulQueue U_true}
                
            \EndFor
        \end{algorithmic}
    \vspace{-10pt} % Adjust the space between the content and the bottom line
    \rule{\textwidth}{0.4pt} % Line at the bottom
    \label{tab: EN_ib algorithm}
\end{table}

\subsection{Results}
In this section the results obtained by solving the optimization problem in \eref{eq: EN_ib initial opt problem} will be presented using the algorithm in \tref{tab: EN_ib algorithm}. 

To this scope the \gls{en} is designed to be composed of $K=100$ \glspl{ed} that sends independent tasks to a common \gls{es}. These \glspl{ed} are placed at a random distance, from 5 to 150 meters, from the \gls{es}. The maximum transmit power is $p_k^{tr} = 100mW$ . The access point operates with a carrier frequency $f_0 = 1GHz$ and the wireless channels are generated using the Alpha-Beta-Gamma model from \cite{MacCartney2016AlphaBetaGamma}. The total available bandwidth per \gls{ed} is set to $ B_k = 1kHz $ and the noise spectral density at the receiver is set to $ N_0 = -174  dBm/Hz $.

Both \gls{es} and all the \glspl{ed} are equipped with a $1.8 \; GHz $ CPU (Intel$^{\circledR}$ Celeron$^{\circledR}$  6305E Processor 4M Cache) and the product between the numbers of cores and the number of \gls{flopc} is $\rho_k=4$. With this designing choice the parameters for CPUs are set to $ f_c^{max} = f_k^{max} = 1.8GHz $, and $  \eta = \eta_k = 2.57*10 ^ {-27} $, for all $k$. 

The input data $\x_k$ has a dimension of $d_x=750$, whereas the output $\y_k$ has $d_y=8$ and this is the same for all the \glspl{ed}. \\
\begin{figure}[!t]
    \centering
    \begin{subfigure}[t]{0.49\textwidth}
        \centering
        \includegraphics[width=\textwidth]{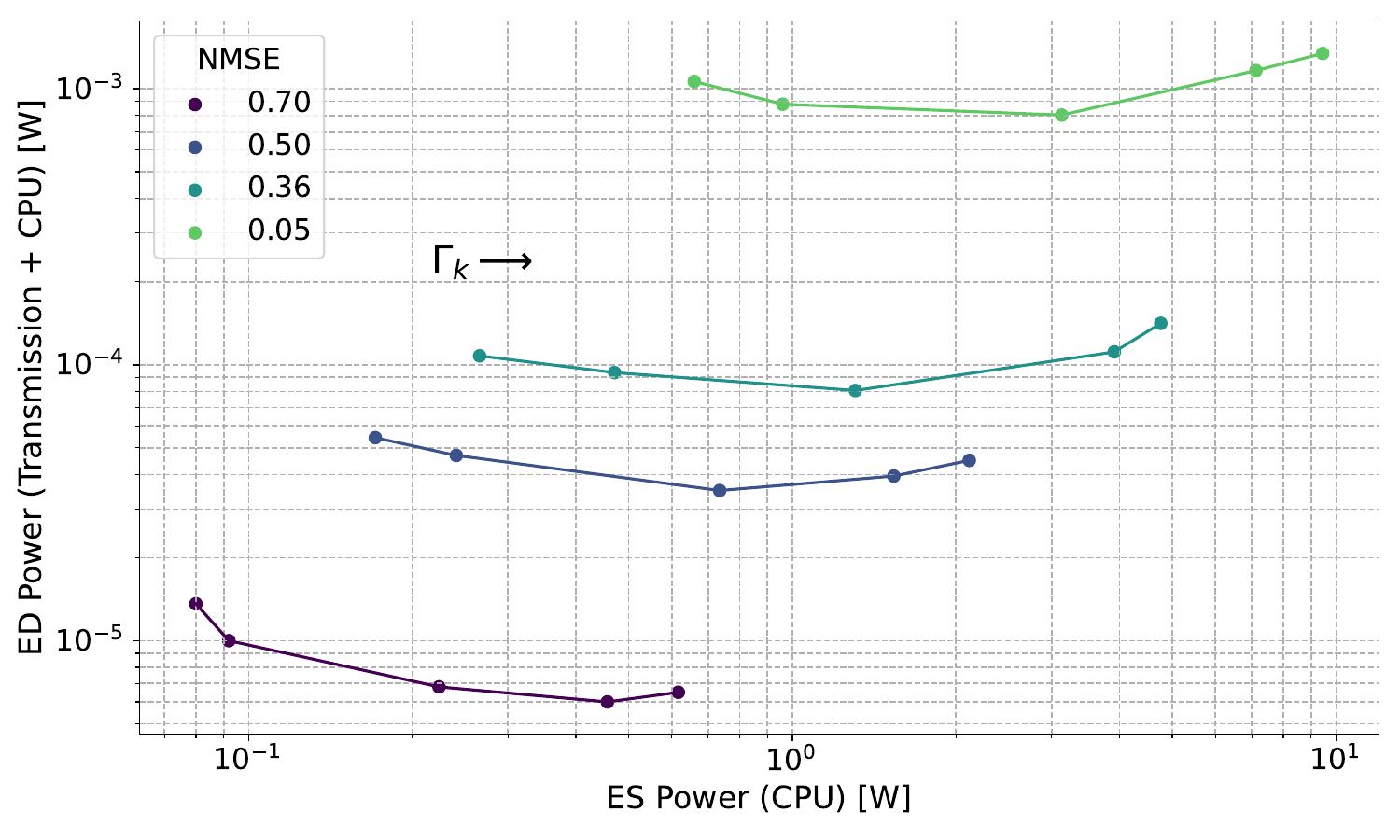}
        \caption*{(a)} % Caption under image without adding to list
    \end{subfigure}%
    \begin{subfigure}[t]{0.49\textwidth}
        \centering
        \includegraphics[width=\textwidth]{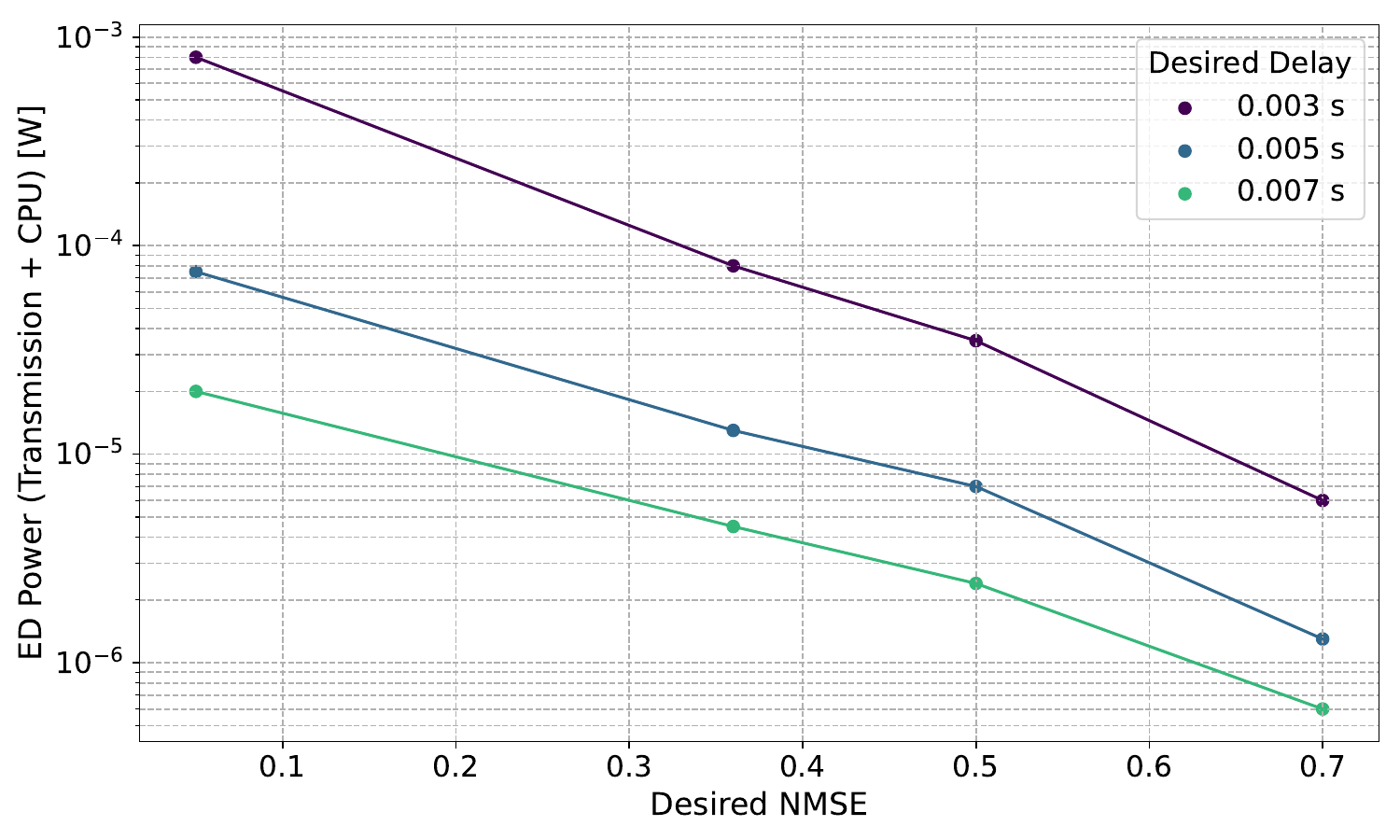}
        \caption*{(b)}
    \end{subfigure}
    \caption[Goal-Oriented communication Edge Network trade-off plots ]{(a) The effects of $\Gamma_k\in \{0.5, 1, 8, 30, 50\}$ on the power consumption of the \gls{ed} and the \gls{es}. In this plot the value of $G_k^{avg}$ changes as seen in the legend while the delay is fixed to $D_k^{avg}= 0.003s$. (b) The trade-off between device power consumption as a function of the \gls{nmse}. The value of the delay changes as in the legend and $\Gamma_k=8$.}
    \label{fig: EN_ib trade-off}
\end{figure}

After setting these architectural parameters, it is possible to execute the algorithm in \tref{tab: EN_ib algorithm} multiple times for different values of $D_k^{avg}$ and $G_k^{avg}$ to evaluate the performance of the \gls{en} in different scenarios. Every time, the algorithm is evaluated for the time necessary to let the virtual queues converge. After convergence, the average power consumption, the average delay and the average evaluation metric are obtained by averaging their values over the following $1000$ time-slots. The results are shown in \fref{fig: EN_ib trade-off}.

In \fref{fig: EN_ib trade-off}~(a), the effects of $\Gamma_k$ on the power consumption are shown. This parameter was introduced in \eref{eq: EN_ib total power} to adjust the importance of the CPU and transmission power consumption in the \gls{ed}. In the plot its effect are showed, where the total power consumption at the \gls{ed} versus the power consumption of the \gls{es} is reported. The different lines represent different values of $G_k^{avg}$ at a predefined $D_k^{avg}=0.003,\text{s}$.

By letting $\Gamma_k$ vary, it is possible to deliberately decide if the CPU or the transmission should consume more or less power. It was chosen to set $\Gamma_k \in {0.5, 1, 8, 30, 50}$.

As this value increase the points move toward the right showing that the power consumption at the \gls{es} increases. This is due to the fact that by increasing the weight of the CPU power consumption of the \gls{ed}, the optimization algorithm will prefer to process the data at lower frequencies. This will increase the associated delay that will require the \gls{es} to process the data faster, resulting in a higher power consumption at the \gls{es}.

Moreover, it is possible to identify a value at which the total power consumption of the \gls{ed} is minimized. This represents a very important point since, in general, \glspl{ed} are not connected to the main power supply like the \gls{es}. Minimizing their power consumption might be very helpful.

For almost all the different lines the minimum in the power consumption of the \gls{ed} is reached at $\Gamma_k=8$.  For this reason this value is used for the next plot.

In \fref{fig: EN_ib trade-off}~(b), the trade-off between the power consumption of the \gls{ed} and the evaluation metric is shown as the delay requirements change. This is a very fundamental trade-off that is used to fully characterize the network. In fact, it allows for a desired value of power consumption, an idea of the value of the associated delay and \gls{nmse}.

\section{Semantic-Goal-Oriented communication based on the SQ-GAN}\label{sec: EN_nn}
A more interesting approach is obtained by considering a less theoretical scenario.

In a real-world scenario, the \gls{en} is composed of a set of \glspl{ed} that are required to deal with non-simple data domains. Considering the data to follow a Gaussian distribution to apply the \gls{gib} might not always be possible.

Additionally, there are cases in which the goal of the communication lies in the transmission of the semantic information.

In this section, instead of encoding the data via \gls{gib}, the \gls{sqgan} is employed. In such a scenario, the orchestration of the resources is oriented to the transmission of the semantic information that the image $\x$ and the \gls{ssm} $\s$ contain.

The general setup of the \gls{en} discussed in the previous section will hold also in this specific application. However, some changes will be applied due to the design and nature of the \gls{sqgan}.

While in the \gls{gib} case the quality of the encoding was conditioned on the parameter $\beta$, in the \gls{sqgan} case the quality of the encoding is conditioned on the two masking fractions $m^{\x}$ and $m^{\s}$. Moreover, the number of operations performed by the model is mostly independent of the value of the masking fractions.

For this reason, the following modifications are applied to the \gls{en}. 
\begin{itemize}[label={}] 
    \item \textbf{ED:} For the \gls{ed}, some minor modifications are introduced: 
    \begin{itemize}[label={}] 
        \item \textbf{CPU:} The CPU is composed as before. Each \gls{ed} is equipped with a CPU with maximum power consumption $p_k^{cpu}$, working clock frequency $f_k$, maximum clock frequency $f_k^{max}$, and the effective switch capacitance of the processor $\eta_k$~\cite{Burd1996EffSwithc}.
        
        The difference is that, in this case, the CPU will be used only to perform the part of operations that are dependent on the masking fractions. This means that it will be employed only to perform the vector quantization.

        All the other constant computations will be implemented by a dedicated analog chip.

        \item \textbf{Analog chip:} The analog chip is a specifically designed chip that can perform the encoding and the \gls{samm}. Being independent of the masking fraction, its contribution will not be taken into account in the optimization. The only assumption is that it is powerful enough not to represent a bottleneck in the computations.

        \item \textbf{Wireless Network Interfaces:} The network interface remains unchanged. Each \gls{ed} will be able to transmit data at a rate $R_k$ with maximum value of $R_k^{max}$. This value will depend on the available bandwidth $B_k$, the flat-fading coefficient $h_k$, the noise power spectral density at the receiver $N_0$, and the maximum transmission power $p_k^{tr}$.
    \end{itemize}
    \item \textbf{ES:} For the same reasons of the independence from the masking fractions, the \gls{es} will not be considered in the optimization. All the operations at the receiver will not be taken into account.
\end{itemize}

After these necessary considerations, it is possible to start discussing all the optimization problem terms. As before, the power consumption, the delay, and the evaluation metric will be considered separately.

\begin{itemize}[label={}] 
    \item \textbf{Power Consumption:} The task of the device, after the analog chip has correctly encoded the data and evaluated the relevance score via the \gls{samm}, is to perform the vector quantization. The power consumption of the CPU associated with this task can be expressed as: 
    \begin{equation} 
        P_k^{vq}(f_k(t)) = \eta_k \cdot (f_k(t))^3 \equiv P_k^{vq}(t), 
    \end{equation} where $ \eta_k $ is the effective switch capacitance of processor $k$.

    After the vector quantization, the data is ready to be transmitted, and the power consumption of the transmission is evaluated as a function of the transmission rate $R_k(t)$ as:
    \begin{equation}
        P_k^{tr}(R_k(t)) =  \frac{B_k N_0}{h_k(t)} \left[  {\rm exp} \left(\frac{R_k(t) \ln(2)}{B_k} \right)   -1 \right] \equiv P_k^{tr}(t).
    \end{equation}

    These two terms are then summed to obtain the total power consumption of the \gls{ed} as:
    \begin{equation}
        P_k^{tot}(t) = P_k^{vq}(t) + P_k^{tr}(t).
    \end{equation}

    \item \textbf{Delay:} The first part of the delay is associated with the time required to perform the vector quantization. This term is proportional to the number of operations required to perform the vector quantization. By considering the same hyperparameters described in \cref{ch: SQGAN}, this value assumes the following form:
    \begin{equation}
        W_k(m_\x^k(t), m_\s^k(t)) = 769 \cdot (|\C| - 1) \cdot 512  \cdot [m_\x^k(t) + m_\s^k(t)] = 402{,}783{,}744 \cdot [m_\x^k(t) + m_\s^k(t)] \equiv W_k(t),
    \end{equation}
    where $769 = 256 \cdot 3 + 1$ is the number of operations to evaluate the distance between two vectors in $256$ dimensions, $|\C|=1024$ is the number of codewords in the codebooks,  and $512 \cdot [m_\x^k(t) + m_\s^k(t)]$ is the number of selected vectors after the \gls{samm}.

    The associated delay is expressed as:
    \begin{equation}
        D_k^{vq}(m_\x^k(t), m_\s^k(t), f_k(t)) = \frac{W_k(t)}{\mathrm{FLOPS}(f_k(t))} = \frac{W_k(t)}{f_k(t) \cdot \rho_k} \equiv D_k^{vq}(t),
    \end{equation}
    where $\rho_k$ is the product between the number of cores and the number of Floating Points Operations Per Cycle of the CPU.

    After the vector quantization, the data is ready to be sent. The number of bits to be sent can be expressed as a function of the two masking fractions by following the steps in \eref{eq: SQGAN BPP} as:
    \begin{equation}
        N_k(m_\x^k(t), m_\s^k(t)) = 512 \cdot [10(m_\x^k(t) + m_\s^k(t)) + 2] \equiv N_k(t),
    \end{equation}
    where $512$ is the maximum number of elements to be transmitted. The delay associated with the transmission is then expressed as:
    \begin{equation}
        D_k^{tr}(m_\x^k(t), m_\s^k(t), R_k(t)) = \frac{N_k(t)}{R_k(t)} \equiv D_k^{tr}(t).
    \end{equation}

    The total delay is then expressed as:
    \begin{equation}
        D_k^{tot}(t) = D_k^{vq}(t) + D_k^{tr}(t).
    \end{equation}
    \item \textbf{Evaluation Metric:} In \sref{sec: SQGAN numerical results}, it was shown that the evaluation metrics are in general functions of both $m_\x^k(t)$ and $m_\s^k(t)$. The \gls{lpips} metric shown in \fref{fig: SQGAN lpips 3d plot} is a clear example of this relationship. Both the masking fractions influence the result. For this reason, in this \gls{en} application, the evaluation metric considered is the \gls{lpips}.

    Working directly on the \gls{lpips} is not the best option. The case discussed in the previous section with the \gls{gib} approach is an example. The best value for the optimization variable $\beta_k^*$ was selected via a grid search approach. This is not an elegant approach and can be improved.

    The idea is to approximate the true metric $G_k(m_\x^k(t), m_\s^k(t))$ with a function $G_k^{approx}(m_\x^k(t), m_\s^k(t))$. This function will be used to replace the original one in the optimization problem of the \gls{en}, removing the grid search approach in two variables.

    The smooth structure of the \gls{lpips}, shown in \fref{fig: SQGAN lpips 3d plot}, makes this approximation even easier. The \gls{lpips} can, in fact, be approximated with the following separable function:
    \begin{equation}
        G_k^{approx}(m_\x^k(t), m_\s^k(t)) = \frac{a}{(m_\x^k(t))^b} + \frac{c}{m_\s^k(t)} \equiv G_k^{approx}(t).
    \end{equation}
    By fitting this function to the true \gls{lpips}, the $l_2$ error is on the order of $1\mathrm{e}{-5}$. The optimal values for the parameters are selected to be $a=2.58\mathrm{e}{-1}$, $b=1.20\mathrm{e}{-1}$, and $c=2.95\mathrm{e}{-3}$.
\end{itemize}

By knowing all the components of the \gls{en}, the optimization problem can be formulated as follows:
\begin{mini}|s|[0]
    {\mathbf{\Psi}(t)}{\lim_{T \to +\infty}\; \frac{1}{T} \sum_{t=1}^T  \mathbb{E}[P^{tot}(t)] }
    {}{}
    \addConstraint{\lim_{T \to +\infty}\; \frac{1}{T} \sum_{t=1}^T  \mathbb{E}[D_k^{tot}(t)] \leq D_k^{avg}\qquad \forall k }{}
    \addConstraint{ \lim_{T \to +\infty}\; \frac{1}{T} \sum_{t=1}^T  \mathbb{E}[G_k(t)] \leq G_k^{avg}\qquad \forall k }{}
    \addConstraint{0 \leq f_k(t) \leq f_k^{max} \qquad \forall k,t }{}
    \addConstraint{0 \leq R_k(t) \leq R_k^{max}(t) \qquad \forall k,t }{}
    \addConstraint{0 \leq m_\x^k(t) \leq 1  \qquad \forall k,t}{}
    \addConstraint{0 \leq m_\s^k(t) \leq 1  \qquad \forall k,t}{}
    \label{eq: EN_nn initial opt problem}
\end{mini}
with  $\mathbf{\Psi}(t) = [\{f_k(t)\}_k,\{ R_k(t)\}_k, \{m_\x^k(t)\}_k, \{m_\s^k(t)\}_k]$ the vector of the optimization variables at time $t$. The only constraint on the optimization variable that is time dependent is $R_k^{max}(t)$, evaluated at every time-step as in \eref{eq: EN_ib maximum rate}.

\subsection{The Edge Network Optimization}
After the formulation of the minimization problem, it is possible to proceed with the optimization. The first step is to introduce two \textit{virtual queues} for each \gls{ed}, one for the long-term constraint on delays, $T_k(t)$, and the other one for the long-term constraint on the metric, $U_k(t)$.

The update of the virtual queues is expressed as: 
\begin{align}
    T_k(t+1) &= \max [0,T_k(t) + \varepsilon_k(D_k^{tot}(t) - D_k^{avg})] \label{eq: EN_nn virtulQueue T_true}\\
    U_k(t+1) &= \max [0,U_k(t) + \nu_k(G_k(t) - G_k^{avg})],  \label{eq: EN_nn virtulQueue U_true}
\end{align}
where $\epsilon_k$ and $ \nu_k $ are the learning rate for the update of the virtual queues. 

While the virtual queue $T_k(t)$ can be used directly, for $U_k(t)$ a clarification is needed. The virtual queue $U_k(t)$ will always be updated as shown in \eref{eq: EN_nn virtulQueue U_true}. However, for defining the Lyapunov function $L(\Theta(t))$, the virtual queue will be written as: 
\begin{equation} 
    U_k(t+1) = \max [0,U_k(t) + \nu_k(G_k^{approx}(t) - G_k^{avg})]. \label{eq: EN_nn virtulQueue U_approx} 
\end{equation} 
By using the approximation of the \gls{lpips} instead of the true value, it is possible to work with a smooth continuous function. This will help in the optimization process and avoid the need for a grid search approach.

In the remainder of this section, the virtual queue $U_k(t)$ will be considered as in \eref{eq: EN_nn virtulQueue U_approx}, unless differently specified.

The Lyapunov function $L(\Theta(t))$ and the drift-plus-penalty function $\Delta(\Theta(t))$ are defined and obtained as in \eref{eq: EN_ib Lyapunov function} and \eref{eq: EN_ib drift_plus_penalty}, respectively. The final per-slot optimization problem has the same form as the one in \eref{eq: EN_ib per-slot opt problem structure}. By substituting the correct expressions of delay, power, and evaluation metric specific to this application, the per-slot problem can be written as: 
\begin{mini}|s|[0]
    {\mathbf{\Psi}(t)}{\sum_{k=1}^K \bigg[ \frac{\epsilon_kT_k(t)N_k(t)}{R_k(t)} + \frac{\epsilon_kT_k(t)W_k(t)}{f_k(t)\rho_k } + \nu_k U_k(t) \left(\frac{a}{(m_\x^k(t))^b} + \frac{c}{m_\s^k(t)}\right)+}{}{} \breakObjective{\qquad+ V \frac{B_k N_0}{h_k(t)} {\rm exp} \left(\frac{R_k(t) ln(2)}{B_k} \right) + V \eta_k (f_k(t))^3} \bigg] 
    \addConstraint{\mathbf{\Psi}(t) \in \mathcal{T}(t).}{}
    \label{eq: EN_nn per-slot opt problem}
\end{mini}  
The solution of this problem is not simple. It is, however, possible to see the similarities with the problem discussed in the previous section. The solutions for the transmission rate $R_k(t)$ and the CPU clock frequency $f_k(t)$ are, in fact, the same as in \eref{eq: EN_ib optimal rate} and \eref{eq: EN_ib optimal freq device}. The only difference is that these values can now be expressed as functions of the two masking fractions $m_\x^k(t)$ and $m_\s^k(t)$ as follows:

\begin{equation}
    R_k^*(m_\x^k(t), m_\s^k(t)) = \frac{2 B_k}{ln(2)}\; W\! \!\left(\sqrt{\frac{512\epsilon_k T_k(t)\; ln(2)\; h_k(t)}{4B_k^2\;V \;N_0} [2+ 10(m_\x^k(t) + m_\s^k(t))] }\right)\; \Biggr|_0^{R_k^{max}(t)}
\label{eq: EN_nn optimal rate masks_func}
\end{equation}
\begin{equation}
    f_k^* (m_\x^k(t), m_\s^k(t)) = 107.64\sqrt[4]{\frac{\epsilon_k T_k(t)}{V \eta_k \rho_k} [m_\x^k(t) + m_\s^k(t)]}\; \Biggr|_0^{f_k^{max}}.
\label{eq: EN_nn optimal freq device masks_func}    
\end{equation}
Due to the very intricate nature of these equations, an exact analytical solution is not possible. However, the solution can be found by using numerical approaches. The first step involves the representation of all the optimization variables as functions of $m_\s^k(t)$.

To achieve this result, the first step is to derive the Lagrangian associated with the optimization problem in \eref{eq: EN_nn per-slot opt problem} with respect to $m_\x^k(t)$ and $m_\s^k(t)$. Then, by setting the derivative to zero, it is possible to drop all the constant terms and write $m_\x^k(t)$ as follows
\begin{equation}
    m_\x^{k*}(m_\s^k(t)) = \left(\frac{a}{c}\right)^{\frac{1}{b}} (m_\s^k(t))^{\frac{2}{b}}\; \Biggr|_0^{1}.
    \label{eq: EN_nn optimal m_x masks_func}
\end{equation}
The last step involves the substitution of \eref{eq: EN_nn optimal m_x masks_func} in the solutions of the transmission rate, \eref{eq: EN_nn optimal rate masks_func}, and the CPU clock frequency, \eref{eq: EN_nn optimal freq device masks_func}. The final system of equations is:
\begin{align}
    R_k^*(m_\s^k(t)) &= \frac{2 B_k}{ln(2)}\; W\! \!\left(\sqrt{\frac{512\epsilon_k T_k(t)\; ln(2)\; h_k(t)}{4B_k^2\;V \;N_0} \left[2+ 10\left( \left(\frac{a}{c}\right)^{\frac{1}{b}} (m_\s^k(t))^{\frac{2}{b}} + m_\s^k(t)\right)\right] }\right)\; \Biggr|_0^{R_k^{max}(t)}, \label{eq: EN_nn optimal rate}\\
    f_k^* (m_\s^k(t)) &= 45.26\sqrt[4]{\frac{\epsilon_k T_k(t)}{V \eta_k} \left[\left(\frac{a}{c}\right)^{\frac{1}{b}} (m_\s^k(t))^{\frac{2}{b}} + m_\s^k(t)\right]}\; \Biggr|_0^{f_k^{max}}, \label{eq: EN_nn optimal freq device}\\
    m_\x^{k*}(m_\s^k(t)) &= \left(\frac{a}{c}\right)^{\frac{1}{b}} (m_\s^k(t))^{\frac{2}{b}}\; \Biggr|_0^{1}. \label{eq: EN_nn optimal m_x}
\end{align}

By substituting the solutions at \eref{eq: EN_nn optimal rate}, \eqref{eq: EN_nn optimal freq device}, and \eqref{eq: EN_nn optimal m_x} inside the per-slot objective function in \eqref{eq: EN_nn per-slot opt problem}, the resulting problem can be optimized as a single-variable optimization problem. The final solution can be found by using a numerical approach.

Once the optimal solutions for all the optimization variables are found, the virtual queues can be updated as in \eref{eq: EN_nn virtulQueue T_true} and \eqref{eq: EN_nn virtulQueue U_approx}.

The overall procedure is summarized in the algorithm in \tref{tab: EN_nn algorithm}.

\begin{table}[ht]
    \centering
    \rule{\textwidth}{0.4pt} % Line at the top
    \vspace{-22pt} % Adjust the space between the top line and the caption
    \caption{Edge Device Resource Allocation Algorithm}
    \vspace{-10pt} % Adjust the space between the caption and the line below it
    \rule{\textwidth}{0.4pt} % Line below the caption
    \vspace{-15pt} % Adjust the space between the line and the algorithmic content
        \begin{algorithmic}[1]
            \item Set the Lyapunov parameters $V$, $T_k(0)$, $U_k(0)$, $\epsilon_k$ and $\nu_k \;\;\forall k$
            \item Set the desired values for delay $D_k^{avg}$ and metric $G_k^{avg}$
            \For{$t$ at least until $T_k$ and $U_k$ converges $\forall k$}
                \State Find the minimum of \eref{eq: EN_nn per-slot opt problem} as a function of $m_\s^k(t)$ via numerical optimization.
                \State Select the optimal $m_\s^k(t)$.
                \State Evaluate the optimal $R_k^*(t)$, $f_k^*(t)$ and $m_\x^{k*}(t)$ with \eref{eq: EN_nn optimal rate}, \eqref{eq: EN_nn optimal freq device} and \eqref{eq: EN_nn optimal m_x} respectively.
                \State Run the online inference task with the \gls{sqgan} model.
                \State Update  the virtual queues $T_k$ and $U_k$ $\forall k$ via \eref{eq: EN_nn virtulQueue T_true} and \eqref{eq: EN_nn virtulQueue U_approx}
            \EndFor
        \end{algorithmic}
    \vspace{-10pt} % Adjust the space between the content and the bottom line
    \rule{\textwidth}{0.4pt} % Line at the bottom
    \label{tab: EN_nn algorithm}
\end{table}
\subsection{Results}
\begin{figure}
    \centering
    \includegraphics[width=0.7\textwidth]{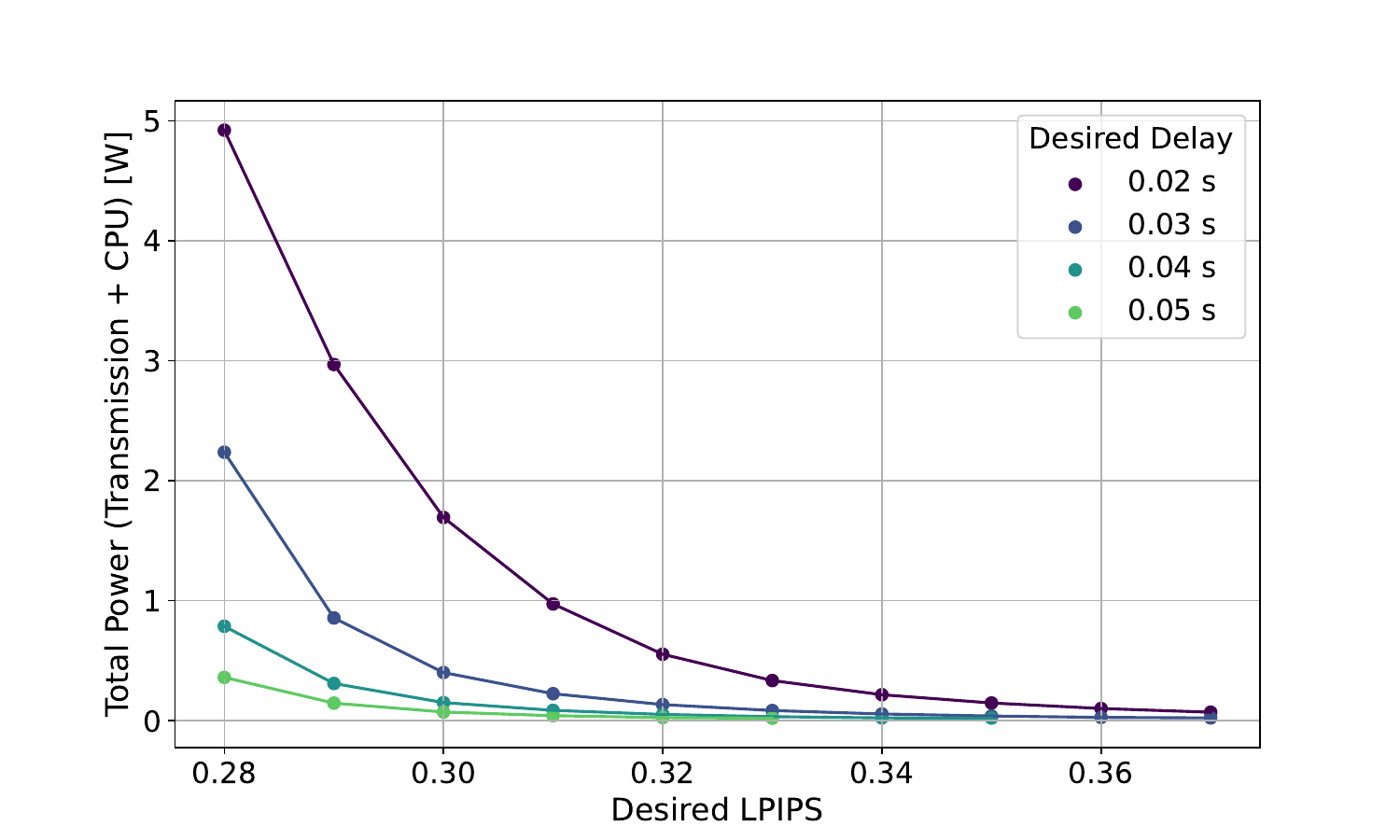}
    \caption[Semantic-Goal-Oriented communication Edge Network trade-off plot]{The trade-off between average power consumption, delay, and \gls{lpips} metric.}
    \label{fig: EN_nn power-delay-metric trade-off}
\end{figure}
In the previous subsection, the algorithm in \tref{tab: EN_nn algorithm} was introduced. By using this algorithm, it is possible to dynamically allocate the resources of the \gls{en} to guarantee target levels of delay and \gls{lpips}.

In this section, we will discuss the trade-off between power consumption, delay and metric. The \gls{en} will be composed of $K=10$ \glspl{ed} placed at a random distance from the \gls{es} between $5$ and $150$ meters. The specifics of the CPU are: (i) $f_k^{max}=1GHz$, (ii) the effective switched capacitance $\eta_k = 10^{-26}$, (iii) the product between the number of cores and the number of \gls{flopc} $\rho_k=16$. The antenna can work at a maximum power of $p_k^{tr}=500mW$.

The access point operates with a carrier frequency $f_0 = 1GHz$ and the wireless channels are generated using the Alpha-Beta-Gamma model from~\cite{MacCartney2016AlphaBetaGamma}. The total available bandwidth is set to $ B = 1MHz $ equally distributed among \glspl{ed}. The noise spectral density at the receiver is set to $ N_0 = -174dBm/Hz $.

Now that the \gls{en} is defined, it is possible to see how it will perform for different values of required delay $D_k^{avg}$ and metric $G_k^{avg}$. The performances of the \gls{en} can be evaluated by using the algorithm in \tref{tab: EN_nn algorithm} until convergence of the virtual queues. After convergence, the value of $P^{tot}$ is obtained by averaging the results over the following $1000$ time-steps. The results are shown in \fref{fig: EN_nn power-delay-metric trade-off}.

In summary, the results demonstrate a clear trade-off between power consumption, delay, and the \gls{lpips} metric within the \gls{sgoc} framework based on the \gls{sqgan}. Specifically, as the desired \gls{lpips} increases, the overall power consumption of the \gls{en} decreases, indicating more efficient resource utilization when lower semantic fidelity is acceptable. Conversely, achieving lower \gls{lpips} values necessitates higher power consumption, especially under stringent delay constraints. This is primarily due to the increased computational load on the \gls{ed} CPUs to meet the tight delay requirements, thereby highlighting the inherent balance between maintaining high semantic accuracy and conserving energy in edge networks.\\

This chapter has thoroughly explored the integration of \gls{goc} principles into \gls{en} for optimal resource allocation. Initially, it introduced the foundational concepts of \gls{goc} and its significance in tailoring communication systems to specific application objectives. Building on this, two distinct methodologies were presented: one leveraging the \gls{ib} framework and the other utilizing the \gls{sqgan} model for \gls{sc}-based image compression. Both approaches were formulated as stochastic optimization problems aimed to minimize average power consumption while adhering to constraints on processing delays and performance metrics. The implementation details, including the decomposition of the optimization problem and the application of Lyapunov-based techniques, were meticulously discussed allowing real-time adaptive resource management.

The empirical evaluations substantiated the efficacy of the proposed frameworks, illustrating how \gls{goc} and \gls{sgoc} paradigms can significantly enhance the performance and efficiency of edge networks. By dynamically adjusting key parameters such as compression levels, data transmission rates and CPU frequencies the systems were able to achieve a balanced trade-off between energy consumption, latency, and semantic fidelity based on real-time network conditions and resource availability. These findings underscore the potential of Goal-Oriented resource allocation strategies in meeting the demanding requirements of modern edge computing applications.

\chapter{\textcolor{black}{Conclusion}}
\label{ch: Conclusion}
\thispagestyle{plain}

In this thesis, the potential of \gls{sc} and \gls{goc} paradigms within modern digital networks has been explored and exploited. The rapid proliferation of data driven technologies such as the \gls{iot}, autonomous vehicles and smart cities has underscored the limitations of traditional bit-centric communication systems. These systems, grounded in Shannon's information theory, focus primarily on the accurate transmission of raw data without considering the contextual significance of the information being conveyed. This fundamental mismatch between data production and communication infrastructure capabilities has necessitated the exploration of more efficient and intelligent communication frameworks.

\cref{ch: SEMCOM} discussed how the core of this thesis focused on integrating of \gls{sc} principles with generative models and their potential applications in the context of edge computing. By focusing on the conveyance of relevant meaning rather than exact data reproduction, \gls{sc} reduces unnecessary bandwidth consumption and inefficiencies. In all those cases where it is possible and reasonable to discuss the semantics, then the faithful representation of the original data is unnecessary as long as the meaning has been conveyed. This paradigm also aligns with \gls{goc}, where the transmitted data is tailored to meet specific objectives, further reducing the communication overhead. The goal of the communication can either be the classical syntactic data transmission or the semantic preservation of the data. By focusing on the goal of the communication, it is possible to transmit only the most pertinent information, thereby reducing the load in communication networks and optimizing resource utilization.

In \cref{ch: SPIC}, the \gls{spic} framework was introduced as a novel method for semantic-aware image compression. The framework demonstrated the potential for high-fidelity image reconstruction from compressed semantic representations. The proposed modular transmitter-receiver architecture is based on a doubly conditioned \gls{ddpm} model, the \gls{semcore}, specifically designed to perform \gls{sr} under the conditioning of the \gls{ssm}. By doing so the reconstructed images preserve their semantic features at a fraction of the \gls{bpp} compared to classical methods such as \gls{bpg} and \gls{jpeg2000}.

Furthermore, the enhancement introduced by \gls{cspic} addressed a critical aspect in image reconstruction: the accurate representation of small and detailed objects. Without requiring extensive retraining of the underlying \gls{semcore} model, \gls{cspic} improved the preservation of important semantic classes, such as traffic signs.  The modular design at the core of the \gls{spic} and \gls{cspic} showcased the flexibility and adaptability of the system in different contexts.

The integration of \gls{sc} principles continued in \cref{ch: SQGAN}, where the \gls{sqgan} model was proposed. This architecture employed vector quantization in tandem with a semantic-aware masking mechanism, enabling selective transmission of semantically important regions of the image and the \gls{ssm}. By prioritizing critical semantic classes and utilizing techniques such as Semantic Relevant Classes Enhancement or the Semantic-Aware discriminator, the model excelled at maintaining high reconstruction quality even at very low bit rates, further emphasizing the efficiency gains of the proposed approach.

Finally, in \cref{ch: Goal_oriented}, the thesis was extended to include the \gls{goc} for resource allocation in \glspl{en}. By adopting the \gls{ib} principle to perform \gls{goc} was developed a framework to dynamically adjust compression and transmission parameters based on network conditions and resource constraints. This dynamic adaptation was crucial in balancing compression efficiency with semantic preservation, optimizing the use of computational and communication resources in edge networks.

By leveraging the \gls{sqgan} within the \gls{en}, the research demonstrated the synergy between \gls{sc} and \gls{goc}. Real-time network conditions informed adjustments to the masking process, enabling the edge network to operate autonomously and efficiently. This approach validated the potential of \gls{sgoc} to enhance resource utilization in modern network infrastructures.

\cleardoublepage
\addcontentsline{toc}{chapter}{Bibliography}
\bibliography{Bibliography}

%   APPENDIX
\cleardoublepage
\appendix % to tell LaTeX that the following chapters are appendices
\chapter{\textcolor{black}{Compacting algorithm}}\label{app: SPIC compacting}

The compacting algorithm $g$ is a crucial component of the proposed \gls{spic} framework, enabling the removal of non-relevant parts of the residual for class-specific applications. Operating only with a mask $\m$, the algorithm eliminates redundant spaces, resulting in a series of bounding box transformations that can be applied to any image.

The process begins by extracting a binary mask $\m$ from the \gls{ssm} $\s$. For example, if the relevant class is "traffic sign," the mask $\m$ will contain 1s at positions corresponding to traffic signs and 0s elsewhere.

This mask is then processed using the standard two-pass binary connected-component labeling algorithm \cite{Shapiro2001connectedCOmp}, which identifies all distinct connected components in the frame and assigns each a unique label. For every connected component $c_i$, a bounding box $b_i$ is defined. Each bounding box is the smallest rectangle that entirely contains its connected component, specified by the coordinates of its top-left and bottom-right corners.

With the bounding boxes identified the iterative process of removing empty regions can start. The algorithm repositions the bounding boxes toward the top-left corner of the frame while preventing any overlap between them. It does this through iterative upward and leftward shifts, continuing until no further movement is possible.

During the upward shift phase, bounding boxes move toward the top edge of the frame until they either contact the top border or another bounding box. Importantly, a bounding box is moved only after all bounding boxes positioned above it have been moved. The leftward shift operates similarly, moving bounding boxes toward the left edge of the frame. By alternating between these two shifts, the algorithm incrementally relocates all bounding boxes to the top-left corner, ensuring that no overlaps occur.

Throughout the process, the algorithm records the original and new coordinates of each bounding box. This information is essential for applying the transformations to other images. After completion, the algorithm outputs pairs of original and final coordinates for every bounding box. These pairs coordinates enable the identification and repositioning of corresponding objects in the residual or original image.

Moreover, since the algorithm relies solely on the \gls{ssm} to generate the mask $\m$ and reposition the bounding boxes, it can be executed at the receiver's end as well. In this context, the coordinate pairs are used to restore the compacted bounding boxes to their original positions. This inverse operation is referred to as $g^{-1}$, the inverse of the compacting algorithm.
\chapter{\textcolor{black}{Edge Network optimization}}\label{app: EN_ib}

In this section the mathematical solution of the optimization problem \eref{eq: EN_ib initial opt problem} in \sref{sec: EN_ib} reported below:

\begin{mini}|s|[0]
    {\mathbf{\Psi}(t)}{\lim_{T \to +\infty}\; \frac{1}{T} \sum_{t=1}^T  \mathbb{E}[P^{tot}(t)] }
    {}{}
    \addConstraint{\lim_{T \to +\infty}\; \frac{1}{T} \sum_{t=1}^T  \mathbb{E}[D_k^{tot}(t)] \leq D_k^{avg}\qquad \forall k }{}
    \addConstraint{ \lim_{T \to +\infty}\; \frac{1}{T} \sum_{t=1}^T  \mathbb{E}[G_k(t)] \leq G_k^{avg}\qquad \forall k }{}
    \addConstraint{0 \leq f_k(t) \leq f_k^{max} \qquad \forall k,t }{}
    \addConstraint{0 \leq R_k(t) \leq R_k^{max}(t) \qquad \forall k,t }{}
    \addConstraint{\beta_k(t) \in \mathcal{B}_k  \qquad \forall k,t}{}
    \addConstraint{0 \leq f^{es}(t) \leq f_{es}^{max} \qquad \forall t}{}
    \addConstraint{f_k^{es}(t) \geq 0 \quad \forall k,t}, \qquad {\sum_{k=1}^K f_k^{es}(t) \leq f_c(t)  \quad \forall t,}{}
\end{mini}

These virtual queues associated to the long-term delay and evaluation metric constraints, $T_k(t)$ and $U_k(t)$ respectively are introduced as follows \cite{Neely2010Lyapunov}:
\begin{align}
    T_k(t+1) &= \max [0,T_k(t) + \varepsilon_k(D_k^{tot}(t) - D_k^{avg})] \\
    U_k(t+1) &= \max [0,U_k(t) + \nu_k(G_k(t) - G_k^{avg})],  
\end{align}
where $\epsilon_k$ and $ \nu_k $ are the learning rate for the update of the virtual queues. 

Based on these virtual queues is possible to define the \textit{Lyapunov function} $L(\mathbf{\Theta}(t))$ as:
\begin{equation}
    L(\mathbf{\Theta}(t)) = \frac{1}{2} \sum_{k=1}^K T_k^2(t) + U_k^2(t),
    \tag{\ref{eq: EN_ib Lyapunov function}}
    \label{app: EN_ib Lyapunov function}
\end{equation}
where $\mathbf{\Theta}(t) = [\{T_k(t)\}_k, \{U_k(t)\}_k]$ is the vector composed by all the virtual queues at time $t$. The idea is to use this Lyapunov function to satisfy the constraints on $D_k^{avg}$ and $G_k^{avg}$ by enforcing the stability of $L(\mathbf{\Theta}(t))$. 

To this scope it is introduced the so called \textit{drift-plus-penalty function}:
\begin{align}
    \Delta(\Theta(t)) &= \mathbb{E}\left[L({\Theta}(t+1))-L({\Theta}(t))+V\cdot P^{tot}(t)  \;\Big|\; \Theta(t)\right] \\
    &=\mathbb{E}\left[\;\sum_{k=1}^K \frac{T_k^2(t+1)-T_k^2(t)}{2} +  \frac{U_k^2(t+1)-U_k^2(t)}{2} +V\cdot P^{tot}(t)\;\; \Big|\;\; \Theta(t)\right]\\
    &= \mathbb{E}\left[\;\sum_{k=1}^K \Delta_{T_k} +  \Delta_{U_k} +V\cdot P^{tot}(t) \;\; \Big|\;\; \Theta(t)\right],
    \label{app: EN_ib drift plus penalty}
\end{align}
where, starting from a generic virtual queue evolving as 
$H(t+1) = \max [0,H(t) +h(t) - \Bar{h}]$ the quantity $\Delta_H$ is defined as follows:
\begin{align*}
    \Delta_H &= \frac{H^2(t+1)-H^2(t)}{2} = \frac{\max [0,(H(t) +h(t) - \Bar{h})^2]-H^2(t)}{2} \\
   &\leq   \frac{(h(t) - \Bar{h})^2}{2} + H(t)[h(t)-\Bar{h}].
\end{align*} 

By applying the same upper bound to $\Delta_{T_k}$ it is possible to obtain:
\begin{align}
    \Delta_{T_k} &= \frac{T_k^2(t+1)-T_k^2(t)}{2} = \frac{\max [0,(T_k(t) + \nu_k(D_k^{tot}(t) - D_k^{avg}))^2]-T_k^2(t)}{2} \\
    &\leq   \nu_k^2\frac{(D_k^{tot}(t) - D_k^{avg})^2}{2} + \nu_k T_k(t)[D_k^{tot}(t) - D_k^{avg}] \\
    &\leq \nu_k^2\frac{(D_k^{max} - D_k^{avg})^2}{2}  + \nu_k T_k(t)[D_k(t) - D_k^{avg}],
    \label{app: EN_ib delta U_k}
\end{align}
where $D_k^{max}(t)$ is the maximum delay allowed for the $k$-th \gls{ed}.

By applying the same reasoning to $\Delta_{U_k}$ it is possible to obtain:
\begin{equation}
    \Delta_{U_k} \leq \nu_k^2\frac{(G_k^{max} - G_k^{avg})^2}{2}  + \nu_k U_k(t)[G_k(t) - G_k^{avg}],
    \label{app: EN_ib delta U_k}
\end{equation}
where $G_k^{max}(t)$ is the maximum value allowed for the evaluation metric for the $k$-th \gls{ed}.

Substituting now \eref{app: EN_ib delta U_k} and \eqref{app: EN_ib delta U_k} inside \eref{app: EN_ib drift plus penalty} and rearranging the terms it is possible to obtain:

\begin{align}
    \Delta_p(\Theta(t)) &\leq
    \sum_{k=1}^K \Bigg{[} \nu_k^2\frac{(D_k^{max} - D_k^{avg})^2}{2} + \nu_k^2\frac{(G_k^{max}(t) - G_k^{avg})^2}{2}  \Bigg{]}  \\ &\;\;\;
    + \mathbb{E} \Bigg{[}\;\sum_{k=1}^K \Big{[} - \varepsilon_k Z_k(t)Q_k^{avg} - \nu_k S_k(t)G_k^{avg}   + \Big|\;\; \Theta(t) \Bigg{]} \\ &\;\;\; + \mathbb{E} \Bigg{[}\;\sum_{k=1}^K \Big{[} \varepsilon_k Z_k(t)Q_k^{tot}(t)  +  \nu_k S_k(t)G_k(t) \Big{]} + V\cdot P^{tot}\;\; \Big|\;\; \Theta(t) \Bigg{]}, 
\end{align}
where some constants that have been taken out of the expected value (first line), while others even if within the expected value do not depend on the optimization parameters (second line).

Pivoting therefore on the Lyapunov optimization it is possible to neglect all these terms. Moreover, it is possible to remove the expected value to obtain the following per-slot optimization:

\begin{mini}|s|[0]
    {\mathbf{\Psi}(t)}{\sum_{k=1}^K \bigg[ \frac{\epsilon_kT_k(t)N_k(t)}{R_k(t)} + \frac{\epsilon_kT_k(t)W_k(t)}{f_k(t)\rho_k } + \frac{\epsilon_kT_k(t)W_{max}^{es}}{f_k^{es}(t) \rho_k^{es}}+}{}{} \breakObjective{\qquad +  \frac{B_k N_0}{h_k(t)} {\rm exp} \left(\frac{R_k(t) ln(2)}{B_k} \right) + V \Gamma_k \eta_k (f_k(t))^3 +}{}{} \breakObjective{\;+  V \eta (f_c(t))^3 + \nu_k U_k(t)G_k(t)\bigg]}{}{}
    \addConstraint{\mathbf{\Psi}(t) \in \mathcal{T}(t),}{}
    \label{eq: EN_ib per-slot opt problem structure}
\end{mini}
where $\mathcal{T}(t)$ indicates the space of possible solutions given by the constraints on the optimization variables. 

at this point it is possible to split the problem for the resource allocation at the \gls{ed} and at the \gls{es}.

\section{Edge Device optimization}\label{app: EN_ib ed opt}
The sub-problem for the \gls{ed} as defined in \eref{eq: EN_ib per-slot opt ed} can be split in two further sub-problems for the transmission rate $R_k(t)$ and the clock frequency $f_k(t)$.

\subsection*{Transmission rate optimal solution}
The sub-problem associated to the transmission rate $R_k(t)$ can be defined as follows:
\begin{mini}|s|[0]
    {R_k(t)}{\frac{\epsilon_kT_k(t)N_k(t)}{R_k(t)} +  V \frac{B_k N_0}{h_k(t)} {\rm exp} \left(\frac{R_k(t) ln(2)}{B_k} \right) }{}{}
    \addConstraint{0 \leq R_k(t) \leq R_k^{max}(t)}{} 
\end{mini}

To simplify the notation, define:
\[
A = \epsilon_k T_k(t) N_k(t), \quad B = V \dfrac{B_k N_0}{h_k(t)}, \quad C = \dfrac{\ln(2)}{B_k}.
\]

Computing the derivative of the objective function $J(R_k(t))$ with respect to $R_k(t)$and set it to zero it is possible to obtain:
\[
\frac{dJ}{dR_k(t)} = -\dfrac{A}{[R_k(t)]^2} + B C \exp\left( C R_k(t) \right) = 0.
\]

By defining Let $x = C R_k(t)$ and $d = \dfrac{A C}{B}$ the derivative can be rearranged as:
\[
x e^{\frac{x}{2}} = \sqrt{d}.
\]

Fortunately, there is an exact solution to this problem and it is based on the \textit{Lambert W function}. By applying the definition and substituting back all the terms it is possible to obtain the final solution:
\begin{equation}
    R_k^*(t) = \frac{2 B_k}{ln(2)}\; W\! \!\left(\sqrt{\frac{\epsilon_k T_k(t)\; ln(2)\; h_k(t)N_k(t)\; }{4 B_k^2\;V \;N_0}}\right)\; \Biggr|_0^{R_k^{max}(t)}
\end{equation}

\subsection*{Clock frequency optimal solution}
The sub-problem associated to the transmission rate $R_k(t)$ can be defined as follows:
\begin{mini}|s|[0]
    {f_k(t)}{\frac{\epsilon_k T_k(t)W_k(t)}{f_k(t)\rho_k } +  V \Gamma_k \eta_k (f_k(t))^3 }{}{}
    \addConstraint{0 \leq f_k(t) \leq f_k^{max}}{} 
\end{mini}

To simplify the notation define:
\[
A = \dfrac{\epsilon_k T_k(t) W_k(t)}{\rho_k}, \quad B = V \Gamma_k \eta_k
\]

Computing the derivative of the objective function $J(f_k(t))$ with respect to $f_k(t)$ and set it to zero it is possible to obtain:
\[
\frac{dJ}{df_k(t)} = -\dfrac{A}{[f_k(t)]^2} + 3B [f_k(t)]^2 = 0
\]

After multiply both sides by $[f_k(t)]^2$, rearranging the terms and substituting back  $A$ and $B$ the final solution is:
\[
f_k(t) = \left( \dfrac{A}{3B} \right)^{1/4} \; \Biggr|_0^{f_k^{max}} \implies f_k^* (t) = \sqrt[4]{\frac{\epsilon_k T_k(t) W_k(t)}{3 V \Gamma_k \eta_k \rho_k} }\; \Biggr|_0^{f_k^{max}},
\]

\section{Edge Server optimization}\label{app: EN_ib es opt}

\begin{mini}|s|[0]
    {\{f_f^{es}(t)\}_k, f_c(t)}{\sum_{k=1}^K \frac{\epsilon_k T_k(t)W_{max}^{es}}{f_k^{es}(t)\rho_k^{es}} + V \eta (f_c(t))^3 }{}{}
    \addConstraint{0 \leq f_c(t) \leq f_c^{max} }{}
    \addConstraint{f_k^{es}(t) \geq 0 \quad \forall k}, \qquad {\sum_{k=1}^K f_k^{es}(t) \leq f_c(t)}{}
\end{mini}

Define:
\[
A_k = \dfrac{ \epsilon_k T_k(t) W_{\text{max}}^{es} }{ \rho_k^{es} }, \quad B = V \eta, \quad S = \sum_{k=1}^K \sqrt{ A_k }
\]

The objective function becomes:
\[
J(\{f_k^{es}(t)\}_k,\ f_c(t)) = \sum_{k=1}^K \dfrac{A_k}{f_k^{es}(t)} + B [f_c(t)]^3
\]

As a first step it is possible to define the associated Lagrangian $L$ of the sub-problem with respect to  $f_k^{es}(t)$ given $f_c(t)$ as:
\[
L = \sum_{k=1}^K \dfrac{A_k}{f_k^{es}(t)} + \lambda \left( \sum_{k=1}^K f_k^{es}(t) - f_c(t) \right)
\]

By deriving it and isolating with respect to $f_k^{es}(t)$ it is possible to obtain:

Solve for $f_k^{es}(t)$:
\[
    \frac{\partial L}{\partial f_k^{es}(t)} = -\dfrac{A_k}{[f_k^{es}(t)]^2} + \lambda = 0  \implies [f_k^{es}(t)]^2 = \dfrac{A_k}{\lambda} \implies f_k^{es}(t) = \sqrt{ \dfrac{A_k}{\lambda} }
\]

Apply the coupling constraint on $f_c(t)$ and by solving for $\lambda$ it is possible to identify:
\[
\sum_{k=1}^K f_k^{es}(t) = \dfrac{1}{\sqrt{\lambda}} \sum_{k=1}^K \sqrt{ A_k } = f_c(t) \implies \sqrt{\lambda} = \dfrac{ S }{ f_c(t) } \implies \lambda = \left( \dfrac{ S }{ f_c(t) } \right)^2
\]

Therefore:
\[
f_k^{es}(t) = \dfrac{ \sqrt{ A_k } }{ S } f_c(t)
\]

This term can now be substituted back into the objective function that is then derived with respect to $f_c(t)$ and set to zero as:

\[
J(f_c(t)) = \dfrac{ S^2 }{ f_c(t) } + B [f_c(t)]^3 \implies \frac{dJ}{df_c(t)} = - \dfrac{ S^2 }{ [f_c(t)]^2 } + 3 B [f_c(t)]^2 = 0
\]

By solving for $f_c(t)$, substituting back the expressions of $A$, $B$ and $S$ and applying the constraints it is possible to obtain the final solution:
\[
    f_c^*(t) = \left[ \left( \dfrac{ S^2 }{ 3 B } \right)^{1/4} \right]_0^{f_c^{\text{max}}}  = \frac{\sqrt{\sum_{k=1}^K \sqrt{\frac{\epsilon_k T_k(t)W_{max}^{es}}{\rho_k^{es}}}}}{\sqrt[4]{3V\eta}} \; \Biggr|_0^{f_{c}^{max}}
\]

Therefore, for every $k$:
\[
f_k^{es}(t) = \dfrac{ \sqrt{ A_k } }{ S } f_c^*(t) = f_k^{es*}(t) = \frac{\sqrt{\frac{\epsilon_k T_k(t)W_{max}^{es}}{\rho_k^{es}}}}{\sqrt{\sum_{k=1}^K \sqrt{\frac{\epsilon_k T_k(t)W_{max}^{es}}{\rho_k^{es}}}}\sqrt[4]{3V\eta} }
\]

%   ACKNOWLEDGEMENTS
\cleardoublepage
\pagenumbering{gobble}

\end{document}